\documentclass[11pt]{article}
\usepackage{amsmath}\begin{normalsize}
\end{normalsize}
\usepackage{algorithm}
\usepackage[noend]{algpseudocode}
\usepackage{color}
\usepackage{bm,verbatim}
\usepackage[authoryear]{natbib}
\usepackage[colorlinks,citecolor=blue,linkcolor=blue,urlcolor=blue]{hyperref}
\usepackage{cleveref}
\usepackage{latexsym, epsfig, amssymb, amsmath, amsthm, graphicx, mathrsfs, booktabs}
\usepackage{rotating, tabularx, setspace,paralist}
\usepackage{color}
\usepackage{graphicx}
\usepackage[OT1]{fontenc}
\usepackage{lscape}
\usepackage{multirow,multicol,mathtools}
\usepackage{subcaption}
\usepackage{anysize}
\renewcommand{\baselinestretch}{1.2}
\marginsize{1.1in}{1.1in}{0.55in}{0.8in} %

\usepackage{ifthen,amsmath,amssymb}

\makeatletter
\def\singlespace{\def\baselinestretch{1}\@normalsize}

\newtheorem{assumption}{Assumption}
\newtheorem{lemma}{Lemma}
\newtheorem{proposition}{Proposition}
\newtheorem{theorem}{Theorem}
\newtheorem{definition}{Definition}
\newtheorem{corollary}{Corollary}
\newtheorem{condition}{Condition}

\theoremstyle{remark}
\newtheorem{remark}{Remark}
\newtheorem{example}{Example}

\renewcommand{\theequation}{
\arabic{equation}%
}

\makeatother

\newcommand{\beq}{\begin{equation}}
\newcommand{\eeq}{\end{equation}}

\newcommand{\e}{{\varepsilon}}

\newcommand{\cE}{{\cal E}}

\newcommand{\cP}{{\cal P}}

\newcommand{\cC}{{\cal C}}

\renewcommand{\cal}{\mathcal}

\newcommand{\dd}{\mathrm{d}}

\renewcommand{\le}{\leq}
\renewcommand{\ge}{\geq}

\renewcommand{\P}{\mathbb{P}}
\newcommand{\E}{\mathbb{E}}
\newcommand{\R}{\mathbb{R}}
\newcommand{\C}{\mathbb{C}}
\newcommand{\N}{\mathbb{N}}

\DeclareMathOperator{\im}{Im}

\newcommand{\OO}{O}
\newcommand{\oo}{o}

\renewcommand{\Im}{{\rm{Im}}}

\newcommand{\cor}{\color{red}}
\newcommand{\cob}{\color{blue}}

\newcommand{\cM}{{\cal M}}

\newcommand{\ba}{\mbox{\bf a}}

\newcommand{\be}{\mbox{\bf e}}

\newcommand{\br}{\mbox{\bf r}}

\newcommand{\bu}{{\mathbf u}}
\newcommand{\bv}{{\mathbf v}}
\newcommand{\bw}{\mbox{\bf w}}
\newcommand{\bx}{\mbox{\bf x}}
\newcommand{\by}{\mbox{\bf y}}

\newcommand{\bA}{\mbox{\bf A}}
\newcommand{\bB}{\mbox{\bf B}}

\newcommand{\bF}{\mbox{\bf F}}
\newcommand{\bG}{\mbox{\bf G}}
\newcommand{\bH}{\mbox{\bf H}}
\newcommand{\bI}{\mbox{\bf I}}
\newcommand{\bL}{\mbox{\bf L}}
\newcommand{\bQ}{\mbox{\bf Q}}
\newcommand{\bR}{\mbox{\bf R}}

\newcommand{\bP}{\mbox{\bf P}}
\newcommand{\bU}{\mbox{\bf U}}
\newcommand{\bM}{\mbox{\bf M}}
\newcommand{\bV}{\mbox{\bf V}}
\newcommand{\bW}{\mbox{\bf W}}
\newcommand{\bcW}{\overline{\mathbf W}}
\newcommand{\ocW}{\overline{W}}
\newcommand{\bX}{\mbox{\bf X}}

\newcommand{\bY}{\mbox{\bf Y}}

\newcommand{\bpi}{\mbox{\boldmath $\pi$}}

\newcommand{\bDelta}{\mbox{\boldmath $\Delta$}}
\newcommand{\bhDelta}{\widehat{\mbox{\boldmath $\Delta$}}}
\newcommand{\btheta}{\mbox{\boldmath $\theta$}}
\newcommand{\bTheta}{\mbox{\boldmath $\Theta$}}

\newcommand{\bLam}{\mbox{\boldmath $\Lambda$}}

\newcommand{\bSig}{\mbox{\boldmath $\Sigma$}}

\newcommand{\bPi}{\mbox{\boldmath $\Pi$}}
\newcommand{\bcE}{\mbox{\boldmath $\mathcal E$}}

\newcommand{\pn}{\psi_n}

\newcommand{\pnk}{\psi_n(\delta_k)}
\newcommand{\tpnk}{\widetilde{\psi}_n(\delta_k)}

\newcommand{\fA}{\mathfrak{A}}
\newcommand{\fB}{\mathfrak{B}}
\newcommand{\fC}{\mathfrak{C}}
\newcommand{\fD}{\mathfrak{D}}
\newcommand{\fE}{\mathfrak{E}}

\newcommand{\fM}{\mathfrak{M}}

\newcommand{\fP}{\mathfrak{P}}

\newcommand{\fS}{\mathfrak{S}}

\newcommand{\hs}{\widehat{s}}

\newcommand{\hv}{\widehat{v}}

\newcommand{\cA}{\mathcal{A}}
\newcommand{\cB}{\mathcal{B}}
\newcommand{\cF}{\mathcal{F}}
\newcommand{\cI}{\mathcal{I}}

\newcommand{\cN}{\mathcal{N}}

\newcommand{\cY}{\mathcal{Y}}

\newcommand{\tM}{\widetilde{M}}

\newcommand{\tX}{\widetilde{X}}

\newcommand{\bhv}{\widehat{\mathbf{v}}}

\newcommand{\btM}{\widetilde{\mathbf{M}}}

\newcommand{\btX}{\widetilde{\mathbf{X}}}

\newcommand{\hW}{\widehat{W}}
\newcommand{\bhW}{\widehat{\mathbf{W}}}
\newcommand{\bhV}{\widehat{\mathbf{V}}}

\newcommand{\bLambda}{\boldsymbol{\Lambda}}
\newcommand{\bUpsilon}{\boldsymbol{\Upsilon}}
\newcommand{\btUpsilon}{\widetilde{\boldsymbol{\Upsilon}}}
\newcommand{\hdelta}{\widehat{\delta}}
\newcommand{\tdelta}{\widetilde{\delta}}

\newcommand{\bcY}{\boldsymbol{\cY}}
\newcommand{\sft}{\mathsf{t}}

\newcommand{\wt}{\widetilde}
\newcommand{\wh}{\widehat}

\newcommand{\hK}{\widehat{K}}

\newcommand{\hSig}{\widehat\Sig}

\newcommand{\sdeg}{\beta_n}

\newcommand{\var}{\mathrm{var}}

\newcommand{\Sig}{\mathbf{\Sigma}}

\newcommand{\tr}{\mathrm{tr}}
\newcommand{\diag}{\mathrm{diag}}

\newcommand{\rank}{\mathrm{rank}}

\newcommand{\supp}{\mathrm{supp}}

\def\independenT#1#2{\mathrel{\setbox0\hbox{$#1#2$}%
\copy0\kern-\wd0\mkern4mu\box0}}

\def\beginn{\begin{eqnarray*}}
\def\endn{\end{eqnarray*}}
\def\beginy{\begin{eqnarray}}
\def\endy{\end{eqnarray}}
\def\begine{\begin{enumerate}}
\def\ende{\end{enumerate}}
\definecolor{bittersweet}{rgb}{0.8, 0.5, 0.2}

\newcommand{\bbA}{{\bf A}}

\newcommand{\bbW}{{\bf W}}

 \newcommand{\La}{\ensuremath{\Lambda}}

\newcommand{\al}{\ensuremath{\alpha}}

\allowdisplaybreaks

\usepackage[draft,inline,nomargin,index]{fixme}
\fxsetup{theme=color,mode=multiuser}
\FXRegisterAuthor{yf}{ayf}{\color{magenta} YF}
\renewcommand{\top}{T}

\renewcommand{\hat}{\widehat}
\renewcommand{\ldots}{\cdots}
\renewcommand{\sft}{t}

\begin{document}

\title{Asymptotic Theory of Eigenvectors for Latent Embeddings with Generalized Laplacian Matrices%
\thanks{
Jianqing Fan is Frederick L. Moore '18 Professor of Finance and Professor of Operations Research and Financial Engineering, Department of Operations Research and Financial Engineering, Princeton University, Princeton, NJ 08544, USA (E-mail: jqfan@princeton.edu). %
Yingying Fan is Centennial Chair in Business Administration and Professor, Data Sciences and Operations Department, Marshall School of Business, University of Southern California, Los Angeles, CA 90089 (E-mail: \textit{fanyingy@marshall.usc.edu}). %
Jinchi Lv is Kenneth King Stonier Chair in Business Administration and Professor, Data Sciences and Operations Department, Marshall School of Business, University of Southern California, Los Angeles, CA 90089 (E-mail: \textit{jinchilv@marshall.usc.edu}). %
Fan Yang is Associate Professor, Yau Mathematical Sciences Center, Tsinghua University, China (E-mail: \textit{fyangmath@mail.tsinghua.edu.cn}). %
Diwen Yu is Ph.D. candidate, Yau Mathematical Sciences Center, Tsinghua University, China (E-mail: \textit{ydw23@mails.tsinghua.edu.cn}). %
This work was supported in part by NSF grants EF-2125142 and DMS-2324490, and National Key R\&D Program of China (No. 2023YFA1010400).
}
\date{March 1, 2025}
\author{Jianqing Fan$^1$, Yingying Fan$^2$, Jinchi Lv$^2$, Fan Yang$^3$, and Diwen Yu$^3$
\medskip\\
Princeton University$^1$, University of Southern California$^2$, and Tsinghua University$^3$
\\
} %
}

\maketitle

\begin{abstract} 
Laplacian matrices are commonly employed in many real applications, encoding the underlying latent structural information such as graphs and manifolds. The use of the normalization terms naturally gives rise to random matrices with dependency. It is well-known that dependency is a major bottleneck of new random matrix theory (RMT) developments. To this end, in this paper, we formally introduce a class of generalized (and regularized) Laplacian matrices, which contains the Laplacian matrix and the random adjacency matrix as a specific case, and suggest the new framework of the asymptotic theory of eigenvectors for latent embeddings with generalized Laplacian matrices (ATE-GL). Our new theory is empowered by the tool of generalized quadratic vector equation for dealing with RMT under dependency, and delicate high-order asymptotic expansions of the empirical spiked eigenvectors and eigenvalues based on local laws. The asymptotic normalities established for both spiked eigenvectors and eigenvalues will enable us to conduct precise inference and uncertainty quantification for applications involving the generalized Laplacian matrices with flexibility. We discuss some applications of the suggested ATE-GL framework and showcase its validity through some numerical examples.
\end{abstract}

\textit{Running title}: ATE-GL


\section{Introduction} \label{Sec.new1}

Graphs and manifolds are commonly associated with sequence data such as texts. To enable text modeling and token generation, one may first construct Word2Vec embeddings of individual words and then build a graph of short sequences, where each short sequence can be viewed as a node of the graph and also be viewed as a point in a latent low-dimensional manifold. The link strengths between each pair of nodes can be calculated using a certain similarity measure of the embedding vectors, giving rise to a high-dimensional random matrix representing the graph data. For network applications, an important question is how to uncover the latent structural information underlying the graphs via low-dimensional manifold representations, often much lower than the ambient embedding dimensionality of each node. The Laplacian matrices for network data have been widely used to construct latent embeddings of graphs, where the nodes of the graph are represented in a latent subspace spanned by the corresponding leading eigenvectors of the Laplacian matrix. A natural question is how to characterize the asymptotic distributions of the leading eigenvectors and eigenvalues of the Laplacian matrix. The existing results in random matrix theory (RMT) have focused almost always on the setting of independent entries modulo symmetry, which is a major bottleneck of new RMT developments. Due to the use of the normalization terms, the Laplacian matrix is an example of a random matrix with dependency. To enable more flexible latent embeddings of graphs, we will extend the concept of the Laplacian matrix to that of the generalized (regularized) Laplacian matrix with index $\alpha \in [0, \infty)$. A key question we aim to address in this paper is how to characterize the asymptotic distributions of the leading eigenvectors and eigenvalues of the generalized (regularized) Laplacian matrices, a \textit{new} class of high-dimensional random matrices with \textit{dependency} representing the network data.

In the realm of spectral graph theory and network analysis, the Laplacian matrix is a fundamental object of study and connects to a multitude of valuable graph properties; 
see, e.g., \cite{chung1997spectral,mohar1991laplacian,merris1994laplacian,godsil2001algebraic} for an overview. Given a network with \textit{adjacency matrix} $\btX$, its (symmetric normalized) Laplacian is defined as 
\begin{equation}\label{eq:Laplace}
\bI-\bL^{-1/2}\btX \bL^{-1/2},
\end{equation}
where $\bL$ is a diagonal matrix consisting of the row sums of $\btX$, i.e., the node degrees. The Laplacian matrix finds applications in various domains such as 
information theory, communication, 
and Ramanujan graphs \citep{sipser1996expander,lubotzky1988ramanujan,donetti2006optimal,hoory2006expander}; quantum graphs and quantum chaos \citep{smilansky2007quantum,braunstein2006laplacian,kurasov2008graph,kook2011combinatorial}; and mathematical biology and chemistry \citep{trinajstic1994laplacian,klein2002resistance,xiao2003resistance,estrada2010topological,freschi2011improved}. 
Furthermore, the importance of the Laplacian matrix extends to other domains, such as manifold learning, where a similar and related concept is the ``transition matrix" derived from the affinity matrix constructed based on a noisy point cloud of the manifold \citep{hardoon2004canonical, michaeli2016nonparametric, lederman2018learning, 
ding2020spectral}.

Statistical inference of network data involves matrices beyond the adjacency and Laplacian matrices. To give an important example, we consider the degree-corrected mixed membership (DCMM) model introduced in \cite{jin2017estimating}.
The adjacency matrix $\btX$ of this model is a random matrix, whose entries are independent Bernoulli random variables (up to symmetry $\btX=\btX^\top$). The expectations of the entries of $\btX$ take the form 
\begin{equation}
    \bH:=\E[\btX]=\bTheta\bPi\bP\bPi^\top\bTheta,\label{intro_DCMM}
\end{equation} 
where $\bTheta$ denotes a diagonal matrix reflecting the degree heterogeneity of nodes in the random graph, $\bPi$ is the matrix of community membership probability vectors, and $\bP$ is a matrix representing the connection probabilities between communities; see \Cref{example:DCMM} in Section \ref{modelsetting} for more details. Compared to network models without significant degree heterogeneity, statistical inference for the DCMM model encounters additional complexities due to the presence of matrix $\bTheta$, whose entries can vary wildly in magnitude; see e.g., the related discussions in \cite{fan2022simple,SIMPLERC, bhattacharya2023inferences}. To deal with such an issue, notice that under certain normalization, $\bLambda:=\E[\bL]$ is proportional to $\bTheta$. On the other hand, by the law of large numbers (LLN), $\bLambda$ can be well-approximated by $\bL$ for large networks when the node degrees diverge. Thus, dividing $\btX$ by $\bL$ on both sides largely removes the intrinsic degree heterogeneity of the DCMM model. This motivates the exploration of the following random matrix 
\begin{equation}\label{eq:Laplace2}
    \bL^{-1}\btX \bL^{-1},
\end{equation}
which is more suitable for certain applications. 


Motivated by the above applications, in this paper, we consider the \textit{generalized (regularized) Laplacian matrices} of large random networks, which incorporate both the Laplacian matrix \eqref{eq:Laplace} and random matrix \eqref{eq:Laplace2} as special cases. 
Specifically, given an $n\times n$ adjacency matrix $\btX$ representing a network with $n$ nodes, we define the \textit{generalized (regularized) Laplacian matrices} as
\begin{equation}\label{GLM}
    \bX:=\bL^{-\alpha}\btX \bL^{-\alpha}
\end{equation}
for an arbitrary constant $\alpha\in [0,\infty)$. When $\al=0$, $\bX$ reduces to the adjacency matrix; when $\al=1/2$, $\bX$ becomes the Laplacian matrix \eqref{eq:Laplace} (up to a trivial transformation $\bI-\bX$); when $\al=1$, we obtain random matrix \eqref{eq:Laplace2}. 
In defining $\bL$, we can also add some commonly used regularization terms to its entries; see the definition in \eqref{eq:defbL} later. 
To facilitate the theoretical analysis, we further assume that $\btX$ can be decomposed into a low-rank signal plus a random noise matrix
\begin{equation}\label{signal+noise}
     \btX= \bH + \bW \ \text{ with } \bH:= \E[\btX],
\end{equation}
where the signal matrix $\bH$ has 
rank $K$ and large signal eigenvalues in magnitude, and $\bW$ is a random noise matrix with independent (up to symmetry) centered entries. For network models, this assumption just means that the model contains $K$ communities. Take the DCMM model \eqref{intro_DCMM} as an example, this amounts to assuming that $\bP$ is a $K\times K$ matrix. For many applications, it is also desired to consider the generalized (regularized) Laplacian random matrices beyond networks, whose entries may take non-zero-one values. For example, one may take a signal-plus-noise model \eqref{signal+noise}, where $\bW$ is a Gaussian random matrix. Then, we get a model that behaves similarly to dense networks. We can further introduce sparsity to the model by considering the case of random matrices with missing entries. One way to model such missingness is to take the Hadamard product of a signal-plus-noise model with a symmetric Bernoulli random matrix whose upper triangular entries are independent and identically distributed (i.i.d.) Bernoulli$(p)$ random variables. The resulting random matrix model will behave like a sparse random network model with sparsity $p$. To have a unified model to cover all these important settings, instead of assuming Bernoulli distributions for the entries of $\btX$, we will impose certain general moment conditions on them following \cite{erdHos2013spectral}. The formal definitions of our model setting and the generalized (regularized) Laplacian matrix with index $\alpha \in [0, \infty)$ will be presented in \Cref{modelsetting}.


The primary objective of this paper is to investigate the asymptotic behaviors of the empirical spiked eigenvalues and eigenvectors of the generalized (regularized) Laplacian matrix \eqref{GLM} (with some commonly used regularization terms) for the signal-plus-noise model \eqref{signal+noise} when the signals are above a certain threshold. In particular, we will derive both the law of large numbers (LLN) and central limit theorems (CLTs) for the spiked sample eigenvalues and eigenvector components. Our results extend significantly the previous works \cite{fan2020asymptotic,SIMPLERC} to the context of the generalized Laplacian matrix framework. These prior studies established the LLN and CLTs for spiked sample eigenvalues and eigenvector components of the adjacency matrices of large networks, which can be viewed as a special case of our results when $\al=0$.  
Our results also compensate for the results of a recent work \cite{ke2024optimal}, where entrywise large-deviation bounds for the eigenvectors associated with the largest eigenvalues of the Laplacian matrix for the DCMM model were established through the leave-one-out strategy. Additionally, in \cite{tang2018limit}, the CLTs for the components of eigenvectors pertaining to the adjacency matrix and the Laplacian matrix of a random dot product graph were established, under the assumption of a prior distribution on the mean adjacency matrix.

Our results can be of independent theoretical interest due to the important role played by Laplacian matrices in the spectral graph theory. On the other hand, they can also serve as crucial ingredients for statistical inference concerning large networks and more general models. For example, they may 
enhance the characterization of the community membership probability matrix $\bPi$ through spectral clustering methods for community detection, a widely used and scalable tool in the literature, as demonstrated in \cite{von2007tutorial,abbe2017community,jin2015fast,le2016optimization,lei2015consistency,rohe2011spectral}, 
or may enable hypothesis testing with network data, a prevalent technique utilized in various contexts such as \cite{arias2014community,verzelen2015community,bickel2016hypothesis,lei2016goodness,wang2017likelihood,fan2022simple,SIMPLERC}. 
Due to the length constraint, we leave the investigation of various important applications of our theoretical results obtained in this paper to future work. 


Compared to previous works \cite{fan2020asymptotic,SIMPLERC,bhattacharya2023inferences}, our paper introduces important theoretical and technical \textit{innovations}. First, to the best of our knowledge, this paper is the \textit{first} in the literature to establish the limiting distributions of the empirical eigenvalues and eigenvectors of the generalized Laplacian matrix for such a general model described in \eqref{signal+noise}, featuring general distributions for the entries of $\bW$ and severe sparsity. In particular, it is worth noting that our theoretical framework accommodates sparse networks with an average node degree exceeding $(\log n)^8/n$, encompassing both dense and sparse scenarios outlined in \cite{arias2014community} and \cite{verzelen2015community}, respectively. Consequently, our setting may also apply to matrix completion problems involving sparse random matrices with missing values that are beyond network models.

Second, similar to \cite{fan2020asymptotic,SIMPLERC}, our results are also based on some delicate precise estimates of the resolvent (or Green's function) of matrix $\bX$, defined as $(\bX-z\bI)^{-1}$ for $z\in \mathbb C$, called the \emph{entrywise and anisotropic local laws}. Then, we derive asymptotic expansions for the spiked eigenvalues, as well as components and general projections of spiked eigenvectors of the generalized Laplacian matrix using these local laws. The desired LLN and CLTs are the consequence of these new expansions. In establishing such local laws and deriving the limiting distributions, one of the most significant challenges is to handle the correlations between $\bL$ and $\btX$. In our technical proof, we exploit a similar decorrelation technique outlined in \cite{ke2024optimal} within the resolvents and asymptotic expansions framework. However, to get more precise results such as the CLT of each single eigenvector component, we need to introduce \textit{new} decorrelation techniques \textit{beyond} the leave-one-out strategy. One such key tool is to introduce an intermediate random matrix denoted as $\bL_{[i]}$, which is constructed by excluding all entries in the $i$th row and column of $\btX$ from the entries of $\bL$ \textit{except for $L_{ii}$}. This operation effectively diminishes the correlation between $\bL_{[i]}$ and $\btX$, enabling us to establish the \textit{almost sharp} entrywise local laws for the resolvent of $\bX$ by first deriving an entrywise local law for an intermediate resolvent of the matrix $\bL_{[i]}^{-\alpha}\btX\bL_{[i]}^{-\alpha}$. 

Third, we offer estimates for the asymptotic variances of the related statistics (i.e., empirical eigenvalues, and components and general projections of the empirical eigenvectors) by leveraging rank inference and bias correction techniques inspired by the methodologies presented in \cite{fan2020asymptotic,fan2022simple}. These approaches enable a more applicable analysis of the asymptotic behaviors of the spiked eigenvalues and eigenvectors of the generalized Laplacian matrix. Our findings further contribute to a deeper understanding of the principal components of the generalized Laplacian matrix, unveiling insightful properties and characteristics. For instance, we confirm the intriguing phase transition phenomenon identified in \cite{fan2020asymptotic} within the context of the generalized Laplacian matrix: the variances of the projections of spiked eigenvectors can exhibit distinct orders based on whether the direction of the projection operation aligns with the spiked eigenvector or not.

The theoretical study in this paper is based on some advanced probabilistic tools developed recently in the random matrix theory (RMT) literature. For a comprehensive overview of recent developments in RMT, see, e.g., \cite{anderson2010introduction,erdHos2017dynamical, tao2012topics}. 
The asymptotic behavior of spiked empirical eigenvalues and eigenvectors of Wigner and sample covariance matrices has been extensively studied \citep{furedi1981eigenvalues,baik2005phase,baik2006eigenvalues,knowles2013isotropic,pizzo2013finite,renfrew2013finite,knowles2014outliers,wang2017asymptotics,bao2021singular,bao2022statistical,abbe2020entrywise, yan2024inference, bhattacharya2023inferences, fan2024can,bloemendal2016principal,ding2021spiked,nadler2008finite,paul2007asymptotics,koltchinskii2016asymptotics,capitaine2018non}. 
Here, we have only mentioned some existing works that are most related to our paper, and the list is far from comprehensive. However, none of these works have considered the (generalized) Laplacian matrices. It is also worth mentioning that our generalized Laplacian matrix \eqref{GLM} with $\al=1$ has a very similar structure to the celebrated sample correlation matrix model. 


The rest of the paper is organized as follows. Section \ref{modelsetting} introduces the model setting. We suggest the new framework of the asymptotic theory of eigenvectors for latent embeddings with generalized Laplacian matrices (ATE-GL) and present the main results in Section \ref{sec:mainresult}. Section \ref{techinno} details the technical innovations of our new theoretical work at a high level. We showcase some applications of our new asymptotic theory in Section \ref{new.Sec.appl} and provide several simulation examples verifying the theoretical results in Section \ref{new.Sec.simu}. Section \ref{new.Sec.disc} discusses some implications and extensions of our work. All the proofs and technical details are provided in the Supplementary Material.

\section{Model setting} \label{modelsetting}

The central object of interest in this paper is the generalized regularized Laplacian matrix (called the ``generalized Laplacian" for short), which provides a family of random matrices containing both the Laplacian matrix and the adjacency matrix as specific cases. To formally introduce such a concept, we will use the network language based on graphs as a concrete example. Let us consider an undirected graph $\cN=(V,E)$, where $V=[n]:=\{1,\cdots,n\}$ denotes a set of $n$ nodes and $E$ represents the set of all the network edges. 
For the setting of a network, the network edge set $E$ is given by a symmetric random adjacency matrix $\btX=(\tX_{ij})\in\mathbb R^{n\times n}$ satisfying that $\tX_{ij}=\tX_{ji}$ with $1 \leq i \neq j \leq n$. In particular, the values of $\tX_{ij}=1$ or $\tX_{ij}=0$ correspond to the cases when network nodes $i$ and $j$ are connected or not connected, respectively. The random graph literature commonly assumes that $\btX$ is a Bernoulli random matrix with independent entries modulo the symmetry and heterogeneous variances. The mean matrix of $\btX$ encodes the interesting community structure of the underlying graph through the low-rank representation.

Given the $n \times n$ random adjacency matrix $\btX$, we can introduce a diagonal matrix $\bL:=\diag(d_1,\cdots,d_n)$, where $d_i:=\sum_{j\in[n]} \tX_{ij}$ denotes the degree of the $i$th node with $1 \leq i \leq n$. For each $\alpha\ge 0$, we can define the generalized Laplacian matrix as $\bX:=\bL^{-\alpha}\btX\bL^{-\alpha}$. To ensure that matrix $\bL$ is nonsingular in practice, we will also incorporate some regularization parameters formally as in \eqref{eq:defbL} later. Observe that for the case of $\alpha=1/2$, the random matrix $\bX$ introduced above corresponds to the symmetric normalized Laplacian matrix of the graph $\cN$. For the case of $\alpha=1$, the random matrix $\bX$ has natural applications in the network inference for the DCMM model, as discussed in the Introduction. For the case of $\alpha=0$, random matrix $\bX$ above reduces to the original random adjacency matrix $\btX$, and the asymptotic behavior of its spiked eigenvalues and eigenvectors has been investigated extensively in previous works such as \cite{erdHos2013spectral}; \cite{fan2020asymptotic,fan2022simple}; \cite{SIMPLERC}. In contrast, we will concentrate on the more challenging case with an arbitrary $\alpha>0$ in this paper. In particular, we will consider a more general class of sparse random matrices that go beyond the network models studied in \cite{fan2020asymptotic, fan2022simple} and \cite{SIMPLERC}.

We now provide the rigorous definitions of the aforementioned random matrices that can be \textit{beyond} binary or bounded. Assume that the $n\times n$ symmetric random ``adjacency" matrix $\btX$ admits a signal-plus-noise decomposition 
\begin{equation} \label{eq:model}
\btX=\bH+\bW,
\end{equation}
where $\bH=\E \btX = (H_{ij})_{1 \leq i, j \leq n}$ is a symmetric deterministic signal matrix and $\bW = (W_{ij})_{1 \leq i, j \leq n}$ is a symmetric random noise matrix with centered and independent upper triangular entries. Further, assume that the signal part $\bH$ is of low rank $K \geq 1$. In particular,  we allow $K$ to \textit{diverge} slowly in this work, as described in \eqref{eq:condK unrescaled} later. 
Denote by 
\begin{equation} \label{new.eq.thetapara}
\theta
:=n^{-2}\sum_{1 \leq i,j \leq n}\E |W_{ij}|^2
\end{equation}
a parameter representing the ``sparsity" level of random adjacency matrix $\btX$ in (\ref{eq:model}). Let us introduce a key parameter 
\begin{equation} \label{new.eq.qpara}
q:=\sqrt{n\theta}
\end{equation}
that plays an important role in our technical analysis, where $\theta$ is given in (\ref{new.eq.thetapara}). 
To facilitate the presentation of our technical assumptions and the main results, we introduce a diverging parameter $\xi$ that is much larger than $\log n$ but much smaller than $n^\e$ for any constant $\e>0$ as network size $n\to \infty$. Specifically, we set $\xi$ as 
\begin{equation}\label{eq:xi condition}
\xi = (\log n)^{1+a_0}
\end{equation}
with $a_0>0$ a constant (which can be chosen to be arbitrarily small). 

We are ready to formally state the setting of our random matrix model for the generalized Laplacian matrix with regularization below. Denote by 
\begin{equation} \label{eq:defbL}
\bL\equiv \bL_{\tau,\lambda}:=\diag(L_1,\cdots,L_n)=\diag\left(d_i+\tau_i\bar d+\lambda_i: i \in [n]\right) 
\end{equation}
the regularized node degree matrix, where $d_i:=\sum_{j=1}^n\tX_{ij}$ with $\btX=(\tX_{ij})$ given in the random matrix model \eqref{eq:model}, $\bar d:=n^{-1}\sum_{j=1}^nd_j$, and $\tau_i,\lambda_i\ge 0$ are regularization parameters that are introduced to ensure that $\bL$ is nonsingular almost surely with $1 \leq i \leq n$.

\begin{definition} \label{def: model setting}
For each fixed $\alpha\in(0,\infty)$, we define the generalized regularized Laplacian matrix (named the generalized Laplacian hereafter) as
\begin{equation} \label{new.eq.FL.gLap}
	\bX:=\bL^{-\alpha}\btX\bL^{-\alpha},
\end{equation}
where $\btX$ and $\bL$ are given in (\ref{eq:model}) and (\ref{eq:defbL}), respectively. 
To streamline the technical presentation, assume some basic regularity conditions with a constant $C_0>0$ that 
\begin{enumerate}
\item The sparsity parameter $q$ in (\ref{new.eq.qpara}) satisfies that 
    \begin{equation}\label{eq: q condition unrescaled}
   \xi^3\leq q \leq C_0n^{1/2}
\end{equation}
with $\xi$ given in (\ref{eq:xi condition}).

\item The regularization parameters satisfy that  $\tau_i\le C_0$ and $\lambda_i\le C_0q^2$ (allowing them to be zero or depend on $n$). 

\item The entries of $\bW$ satisfy that
\begin{equation}\label{eq:moment conditions unrescaled}
    \mathbb E W_{ij}=0,\quad s_{ij}:=\mathbb E|W_{ij}|^2\leq C_0\theta,\quad \mathbb E|W_{ij}|^p\leq C_0^p\theta 
\end{equation}
for all $ i,j\in [n]$ and $3\leq p\leq \xi $. 

\item The entries of $\bH$ are nonnegative and 
assume that  
\begin{equation}\label{eq:signal conditions}
\max_{i \in [n]} \theta_i \le C_0 
\end{equation}
with $\theta_i:=q^{-2}\sum_{1 \leq j \leq n} H_{ij}$.

\item The matrix $\bL$ is positive definite almost surely. 
\end{enumerate}
Moreover, we introduce another key rescaling parameter $\sdeg:=\min_{i \in [n]}(\theta_i+\tau_i\bar\theta+\lambda_i/q^2)$ with $\bar\theta:=n^{-1}\sum_{1 \leq i \leq n} \theta_i$, which is crucial in our technical analysis.
\end{definition}

\begin{remark}
The assumption that $\btX$ is real symmetric can be extended to the complex Hermitian case, and all our conclusions and proofs would still apply almost verbatim. For the definiteness of notation, we will focus on the real symmetric case in the current paper.
\end{remark}

Definition \ref{def: model setting} above provides a natural extension of the typical network models in the sense that instead of assuming the Bernoulli distributions, we only impose certain general mean, variance, and moment assumptions on the entries of $\btX$ in our random matrix model. Such a setting can accommodate the scenarios when the entries of $\btX$ may \textit{not} be binary or bounded. In particular, Conditions (ii) and (iii) in Definition \ref{def: model setting} above are motivated by the random network setting. Specifically, let $\btX$ be the adjacency matrix of an undirected random graph. The entries of $\btX$ are independent (modulo the symmetry) Bernoulli random variables, and we can write $\btX$ as in \eqref{eq:model}.
Consider the sparse network setting where there is an edge between each pair of nodes with probability $\sim \theta \le 1$, where $\sim$ stands for the asymptotic order. Then the means and variances for the entries of $\btX$ are typically of order $\theta$, 
which leads to the assumptions in \eqref{eq:moment conditions unrescaled} and \eqref{eq:signal conditions}. 
For such setting, the last bound in \eqref{eq:moment conditions unrescaled} follows from the second bound in \eqref{eq:moment conditions unrescaled} and the fact that when $|W_{ij}|\le C$, 
\begin{equation}\label{eq:network moment} 
\E |W_{ij}|^p \le C^{p-2} \E |W_{ij}|^2 \le C^{p-1}\theta
\end{equation}
for each $p \geq 3$.
Observe that condition \eqref{eq:network moment} for network models is slightly stronger than condition \eqref{eq:moment conditions unrescaled} in the sense that the former holds for all $p\ge 3$ without imposing any upper bound $p\le \xi$. 

In the context of network models, parameters $\theta_i$'s introduced in Condition (iv) above quantify the degree of heterogeneity among the nodes. Assumption \eqref{eq:signal conditions} essentially implies that $q^2=n\theta$ is of the same order as the maximum degree of nodes in the network. As a consequence, our results will be more adapted to networks with a non-negligible portion of nodes having large degrees. For networks with few nodes of large degrees, we can rescale the adjacency matrix with a different $q$, and all results of this paper can be developed in parallel for that setting as well. For the definiteness of notation, we have opted to work under assumption \eqref{eq:signal conditions}.  
Finally, Condition (v) above holds trivially for network models since the node degrees $d_i$'s and the averaged degree $\bar d$ must be nonnegative. To make our discussions more concrete, let us examine the specific example of the degree-corrected mixed membership (DCMM) model \citep{jin2017estimating}.



\begin{example}[DCMM model]\label{example:DCMM}
Assume that random graph $\mathcal N$ has some underlying network structure in that there exist $K$ disjoint subsets $C_1,\cdots, C_K$ called the latent communities of the network, and each network node $i \in [n]$ has an associated $K$-dimensional community membership probability vector $\bpi_i:=(\pi_i(1),\cdots,\pi_i(K))^\top$ with 
\begin{equation}
    \mathbb P[i\in C_k]=\pi_i(k)
\end{equation}
for each $1 \leq k \leq K$, which means that each node generally has mixed membership among the $K$ latent communities. Denote by $\bPi:=(\bpi_1,\cdots,\bpi_n)^\top$ the $n\times K$ matrix of community membership probability vectors. 
Further, assume that the connection probability of any two nodes $  i\neq j\in [n]$ is given by 
\begin{equation}\label{eq:DCMM connection}
    H_{ij}=\mathbb P[\tX_{ij}=1]=\vartheta_i\vartheta_j\sum_{k, \,l\in[K]} \pi_i(k)\pi_j(l)p_{kl},
\end{equation}
where parameter $\vartheta_i>0$ represents the degree heterogeneity of each node $i$, 
and parameter $p_{kl}$ can be understood as the probability that two nodes in communities $C_k$ and $C_l$ connect to each other with $1\leq k,l \leq K$. Rewriting (\ref{eq:DCMM connection}) in the matrix form, we have the representation 
\begin{equation} \label{new.eq.DCMM.H}
    \bH =\bTheta\bPi\bP\bPi^\top\bTheta,
\end{equation}
where $\bH = (H_{ij})_{1 \leq i, j \leq n}$,  $\bTheta:=\diag(\vartheta_1,\cdots,\vartheta_n)$, and $\bP=(p_{kl})\in\mathbb R^{K\times K}$. The DCMM model corresponds to model (\ref{eq:model}) with the mean matrix $\bH$ in (\ref{new.eq.DCMM.H}). Under the DCMM model setting, the network sparsity parameter $\theta$ in (\ref{new.eq.thetapara}) is given by 
$$\theta:=n^{-2}\sum_{i,j\in [n]}\vartheta_i\vartheta_j\sum_{k, \,l\in[K]} \pi_i(k)\pi_j(l)p_{kl}.$$
In particular, when $\sum_{k,l \in [K]}\pi_i(k)\pi_j(l)p_{kl}$ are all of order $1$, we have that $\theta\sim (n^{-1}\sum_{i\in [n]}\vartheta_i)^2$. 
\end{example}

By the classical law of large numbers (LLN) and central limit theorem (CLT), the entries of the regularized node degree matrix $\bL$ given in (\ref{eq:defbL}) would concentrate around the deterministic diagonal matrix 
\begin{equation} \label{new.eq.LamMat}
\bLambda=\diag(\Lambda_1,\cdots,\Lambda_n):=\mathbb E[\bL]. 
\end{equation}
In view of (\ref{eq:model}), (\ref{new.eq.FL.gLap}), and (\ref{new.eq.LamMat}), we will consider the spectral decompositions of the generalized Laplacian matrix $\bX$ introduced in Definition \ref{def: model setting} and its population counterpart $\bLambda^{-\alpha}\bH\bLambda^{-\alpha}$ given by 
\begin{equation} \label{spectral decompositions}
    \bX=\sum_{i\in[n]}\hdelta_i\bhv_i\bhv_i^\top \  \text{ and } \  \bLambda^{-\alpha}\bH\bLambda^{-\alpha}=\sum_{i\in[K]}\delta_i\bv_i\bv_i^\top,
\end{equation}
where we arrange the eigenvalues according to the descending order in magnitude with $|\hdelta_1|\geq\cdots\geq|\hdelta_n|$ and $|\delta_1|\geq\cdots\geq|\delta_K|>0$, and $\bhv_i$'s and $\bv_i$'s are the corresponding eigenvectors. Given the empirical and population eigen-decompositions in (\ref{spectral decompositions}) above, let us define the diagonal matrices of spiked eigenvalues 
\begin{equation} \label{new.eq.spikedeigvals}
\bhDelta:=\diag(\hdelta_1,\ldots, \hdelta_K) \ \text{ and } \ \bDelta:=\diag(\delta_1,\ldots, \delta_K), 
\end{equation}
as well as the corresponding spiked eigenvector matrices 
\begin{equation} \label{new.eq.spikedeigvecs}
\bhV=(\bhv_1,\cdots,\bhv_K) \ \text{ and } \ \bV=(\bv_1,\cdots,\bv_K). 
\end{equation}
The major goal of this work is to study the asymptotic behavior of the empirical spiked eigenvalues and eigenvectors $\hdelta_k$ and $\bhv_k$ with $1 \leq k \leq K$ and in particular, identify their dependence on the population spiked eigenvalues and eigenvectors $\delta_k$'s and $\bv_k$'s. 

To facilitate the technical presentation, let us introduce additional necessary notation. We focus on the asymptotic regime of network size $n\to \infty$ and refer to a constant whenever it does not depend on parameter $n$. We will use $C$ to denote a generic large positive constant whose value may change from line to line. Similarly, we will use notations such as $\epsilon$, $c$, and $\delta$ to represent generic small positive constants. For any two sequences $a_n$ and $b_n$, $a_n = \OO(b_n)$ (or $b_n = \Omega(a_n)$) means that $|a_n| \le C|b_n|$ for some constant $C>0$, whereas $a_n=\oo(b_n)$ or $|a_n|\ll |b_n|$ means that $|a_n| /|b_n| \to 0$ as $n\to \infty$. We say that $a_n \lesssim b_n$ if $a_n = \OO(b_n)$ and that $a_n \sim b_n$ if $a_n = \OO(b_n)$ and $b_n = \OO(a_n)$. Given a vector $\mathbf v=(v_i)_{i=1}^n$, $|\mathbf v|\equiv \|\mathbf v\|\equiv \|\mathbf v\|_2$ denotes the Euclidean norm and $\|\mathbf v\|_p$ denotes the $L_p$-norm. Given a matrix $\bbA = (A_{ij})$, denote by $\|\bbA\|$, $\|\bbA\|_F $, and $\|\bbA\|_{\max}:=\max_{i,j}|A_{ij}|$ the matrix operator norm, Frobenius norm, and entrywise maximum norm, respectively. For notational simplicity, we write $\bbA=\OO(a_n)$ and $\bbA=\oo(a_n)$ to mean that $\|\bbA\|=\OO(a_n)$ and $\|\bbA\|=\oo(a_n)$, respectively. Moreover, we will use $A_{ij}$ and $\bbA(k)$ to denote the $(i,j)$th entry and $k$th \textit{row vector} of a given matrix $\bbA$, respectively, and use $v(k)$ to denote the $k$th component of a given vector $\bv$. 
We will often write an identity matrix of appropriate size as $\bI$ without specifying the size in the subscript. Denote by $\be_i$ the unit vector with the $i$th component being $1$ and others being $0$. Given any $n\times n$ matrix $\bA$ and vectors $\bu,\bv\in \C^n$, we define
\begin{equation}\label{eq:gen_entries}
    A_{i\bv}:=\be_i^\top\bA\bv, \ A_{\bu i}:=\bu^\top\bA\be_i, \ A_{\bu\bv}:=\bu^\top\bA\bv,
\end{equation}
where $\C$ stands for the complex plane. 

Throughout the paper, we will use the notion of high probability events as defined below.

\begin{definition}\label{def:high}
Given an ($n$-dependent) event $\Omega=\Omega_n$, a constant $c>0$, and a sequence $\xi=\xi_n$ of parameters satisfying \eqref{eq:xi condition}, we say that event $\Omega$ holds with $(c,\xi)$-high probability if
 $  \mathbb P(\Omega^c)\leq e^{-c\xi} $
for all large enough $n$. 
Moreover, we say that event $\Omega$ holds \textit{with high probability (w.h.p.)} if for any large constant $D>0$, 
 $   \mathbb P(\Omega^c)\leq n^{-D} $
for all large enough $n$. 
\end{definition}

\section{ATE-GL for latent embeddings} \label{sec:mainresult}

In this section, we formally introduce the framework of the asymptotic theory of eigenvectors for latent embeddings with generalized Laplacian matrices (ATE-GL). 

\subsection{Technical conditions and preparation} \label{new.sec.condprep}

To facilitate the technical analysis, we will make some regularity conditions below in addition to the basic ones assumed in Definition \ref{def: model setting}.

\begin{assumption}\label{main_assm} 
For a fixed $\alpha\in(0,\infty)$, assume that the following conditions hold for some $1\le K_0\le K$. 

\begin{enumerate}
\item  (Network sparsity) The sparsity parameter $q$ defined in (\ref{new.eq.qpara}) satisfies that $q \gg(\log n)^4$.

\item  (Spiked eigenvalues) It holds that $|\delta_k|\gg q^{1-4\alpha}\sdeg^{-2\alpha}$ for all $1\leq k\leq K_0$ with $\sdeg$ defined in Definition \ref{def: model setting}.

\item (Eigengap) There exists some constant $\epsilon_0>0$ such that
\begin{equation}\label{eq:eigengap unrescaled}
    \min_{1\leq k\leq K_0}\frac{|\delta_k|}{|\delta_{k+1}|}>1+\epsilon_0,
\end{equation}
where we do not require eigengaps for smaller eigenvalues $|\delta_k|$ with $K_0+1\leq k\leq K$.
\item (Low-rankness of signals)  
The rank $K$ of $\bH$ satisfies that 
\begin{equation}\label{eq:condK unrescaled}
K\xi\left(\frac{q^{1-4\alpha}}{|\delta_{K_0}|\sdeg^{1+2\alpha}}+\frac{\xi}{q\sdeg^{2}}+\|\mathbf V\|_{\max}\right)\ll q
\end{equation}
with $\bV$ given in (\ref{new.eq.spikedeigvecs}). 
\end{enumerate}
\end{assumption}

The lower bound on $q$ in Condition (i) of Assumption \ref{main_assm} above places a restriction on the sparsity level of the network, specifically $\theta\gg (\log n)^8/n$ in light of (\ref{new.eq.qpara}). 
Condition (ii) of Assumption \ref{main_assm} puts a constraint on the signal-to-noise ratio. We will see (cf.~\Cref{prop:Lambda and E} and Proposition \ref{prop:W,R,G} below) that $\|\bW\|\lesssim q$ and the smallest eigenvalue of $\bL$ is of order $\Omega(\sdeg q^2)$, which implies that the noise eigenvalue is of order $\OO(q^{1-4\alpha}\sdeg^{-2\alpha})$.
Hence, Condition (ii) indicates that the spiked eigenvalues $\delta_k$ with $1\le k \le K_0$ are considered as ``true spikes." In other words, parameter $K_0$ represents the number of strong signals in our random matrix model, while the model may also include some weak signals that are not subject to any specific assumptions.  
Condition (iii) of Assumption \ref{main_assm} above implies that the strong signals are nondegenerate. Such condition is adopted primarily for the sake of convenience in our technical presentation and proofs, and has been commonly utilized in the literature; see e.g., \cite{abbe2020entrywise,fan2022simple}; \cite{SIMPLERC,jin2017estimating}. However, we believe that our results can be readily extended to the general case with degenerate signals. For such cases, we would need to analyze the eigenspace spanned by the near-degenerate empirical spiked eigenvectors instead of considering each individual eigenvector separately.
Condition (iv) of Assumption \ref{main_assm} imposes a rather weak assumption on $K$. For example, when $\sdeg\gtrsim 1$, $|\delta_{K_0}|\gtrsim q^{2-4\alpha}$ (which holds when the $K_0$th eigenvalue of $\bH$ is at least of order $\Omega(n\theta)$), and $\|\bV\|_{\max}\le q^{-1}$, we see that  \eqref{eq:condK unrescaled} requires 
$ K\ll q^2/\xi^2=n\theta/\xi^2$. In the setting of network models, this means that the number of latent communities $K$ is assumed to be ``slightly" below the typical order of the node degrees $n\theta$. 

We first provide some necessary technical preparation related to the tool of the so-called \textit{generalized quadratic vector equation (QVE)} before presenting the main results of the paper. Such a tool plays a crucial role in characterizing the asymptotic limit $\sft_k$ of the empirical spiked eigenvalue $\hdelta_k$. To define the population quantity $\sft_k$, let us introduce a complex-valued vector $\btM\equiv \btM_n(z)=(\tM_1(z),\cdots,\tM_n(z))^\top \equiv (\tM_1,\cdots,\tM_n)^\top \in \C^n$ that is the $z$-dependent solution to the \textit{generalized QVE} 
given by 
\begin{equation}\label{eq:tM definition}
   \frac{1}{\tM_i}=-z-\Lambda_i^{-2\alpha}\sum_{j\in[n]}\Lambda_j^{-2\alpha}s_{ij}\tM_j,
\end{equation}
where $\im \tM_{i}(z)\ge 0$ for all $i\in [n]$ and $z\in \C_+$ (with $\C_+$ standing for the upper half complex plane), and $s_{ij}$'s and $\Lambda_i$'s are given in (\ref{eq:moment conditions unrescaled}) and (\ref{new.eq.LamMat}), respectively. 
We next define an $n \times n$ complex-valued deterministic diagonal matrix
\begin{equation}\label{eq:tUpsilon definition}
    \btUpsilon(z):=\diag(\tM_1(z),\cdots,\tM_n(z)).
\end{equation}
Denote by $\bV_{-k}$ an $n\times (K-1)$ matrix obtained by removing the $k$th column of matrix $\bV$, and $\bDelta_{-k}$ a $(K-1)\times (K-1)$ matrix obtained by removing the $k$th row and $k$th column of matrix $\bDelta$, where $\bDelta$ and $\bV$ are given in (\ref{new.eq.spikedeigvals}) and (\ref{new.eq.spikedeigvecs}), respectively. 

For each $1\le k\le K_0$, we introduce the union of two intervals on the real line 
\beq\label{eq:defnIk} 
    \widetilde{\cI}_k:=\biggl\{x\in\mathbb R:\frac{|\delta_k|}{1+\epsilon_0/2}\leq|x|\leq (1+\epsilon_0/2){|\delta_k|}\biggr\}
\eeq
with $\epsilon_0$ given in (\ref{eq:eigengap unrescaled}). Then, we define $\sft_k \in \mathbb{R}$ as the solution to the nonlinear equation
\begin{equation}\label{eq:sft_k def}
    1+\delta_k\bv_k^\top\btUpsilon(x)\bv_k-\delta_k\bv_k^\top\btUpsilon(x)\bV_{-k}\frac{1}{\bDelta_{-k}^{-1}+\bV_{-k}^\top\btUpsilon(x)\bV_{-k}}\bV_{-k}^\top\btUpsilon(x)\bv_k=0
\end{equation}
over $x\in \widetilde{\cI}_k$, where $\btUpsilon(\cdot)$ is defined in (\ref{eq:tUpsilon definition}).
Using similar arguments as in the proof of Lemma 3 in \cite{fan2020asymptotic} and Section A.2 of \cite{SIMPLERC}, we can establish the following lemma, which provides the existence, uniqueness, and asymptotic properties of the population quantity $\sft_k$ introduced in (\ref{eq:sft_k def}).

\begin{lemma}\label{lemma:tk}
Under parts (ii) and (iii) of Assumption \ref{main_assm}, for each $1 \leq k \leq K_0$, there exists a unique solution $x=\sft_k$ to equation \eqref{eq:sft_k def} in the subset $\widetilde{\cI}_k$, and it holds that 
$$\sft_k=\delta_k+O\left(q^{2-8\alpha}\sdeg^{-4\alpha}/|\delta_k|\right).$$
\end{lemma}

From Lemma \ref{lemma:tk} above, we see that the population quantities $\sft_k$'s based on the generalized QVE are indeed well-defined. For the implementation of the numerical examples in Section \ref{new.Sec.simu}, we now give a computational algorithm for the calculation of the population quantities $\sft_k$'s and $\btUpsilon(z)$. In light of (\ref{eq:tM definition})--(\ref{eq:tUpsilon definition}), $\btUpsilon(z)$ with $z \in \C$ is an analytic function and $z=0$ is the only pole of $\btUpsilon(z)$ in the complex plane. Hence, we can consider the Laurent series expansion of the complex-valued matrix $\btUpsilon(z)$ given by 
\begin{equation}\label{eq:Upsilon Laurent}
    \btUpsilon(z)=\sum_{l=0}^\infty \frac{1}{z^{l}}\bY_l,
\end{equation}
where $\bY_l$'s are $n \times n$ deterministic diagonal matrices that does not depend on $z$. Let us define the covariance matrix $\bSig = (\Sigma_{ij})_{1 \leq i, j \leq n}$ defined as 
\begin{equation} \label{new.eq.Sig.mat}
\Sigma_{ij}:=\var \left(\be_i^\top\bLambda^{-\alpha}\btX\bLambda^{-\alpha}\be_j\right) = \Lambda_i^{-2\alpha}\Lambda_j^{-2\alpha}s_{ij},
\end{equation} 
where $\be_i$ denotes the $i$th basis vector of $\mathbb{R}^n$ with $1 \leq i \leq n$, and $\btX$ and $\bLambda$ are given in (\ref{eq:model}) and (\ref{new.eq.LamMat}), respectively. 
With the definition in (\ref{new.eq.Sig.mat}), we can express the generalized QVE \eqref{eq:tM definition} that defines $\tM_i(z)$'s in the matrix form 
\begin{equation} \label{new.eq.gqve.mat}
    z\btUpsilon(z)\be_{[n]}=-(\bI+\btUpsilon(z)\bSig\btUpsilon(z))\be_{[n]} 
\end{equation}
with $\be_{[n]}:=\sum_{i\in[n]}\be_i$. 

In view of (\ref{eq:Upsilon Laurent}) and (\ref{new.eq.gqve.mat}), through comparing the degrees of $z$, we can compute the values of diagonal matrices $\bY_l \in \mathbb{R}^{n \times n}$ in a recursive fashion 
\begin{equation}\label{eq:Y_l recursive formula}
    \bY_0=0, \ \ \bY_1=-\bI, \ \text{ and } \ \bY_{l+1}\be_{[n]}=-\sum_{m=0}^l\bY_m\bSig\bY_{l-m}\be_{[n]}
\end{equation}
for $l\geq 1$, where $\bSig$ is given in (\ref{new.eq.Sig.mat}). The recursive formula in (\ref{eq:Y_l recursive formula}) allows us to determine the values of diagonal matrices $\bY_l$ for all $l \geq 0$. Based on the theoretical representations in (\ref{eq:Upsilon Laurent}) and (\ref{eq:Y_l recursive formula}), and the technical analyses in \cite{fan2020asymptotic,fan2022simple}; \cite{SIMPLERC}, we choose to apply the quadratic approximations of $\tM_i(z)$'s and $\btUpsilon(z)$ that are given by 
\begin{equation}\label{eq:bcY def}
    \cM_i(z):=-z^{-1}-z^{-3}\Lambda_i^{-2\alpha}\sum_{j\in[n]}\Lambda_j^{-2\alpha}s_{ij} \ \text{ and } \ \bcY(z):=\diag\{\cM_1(z),\cdots,\cM_n(z)\},
\end{equation}
respectively, with $1 \leq i \leq n$. To compute the value of the population quantity $\sft_k$, we observe that in the nonlinear equation \eqref{eq:sft_k def}, the fractional term is asymptotically negligible compared to the leading term, with an error of order $O(|\delta_k||\sft_k|^{-4})$. Consequently, we can ignore such fractional term and replace $\btUpsilon(z)$ with its quadratic approximation $\bcY(z)$ introduced in (\ref{eq:bcY def}), giving rise to the simplified equation
\begin{equation}\label{eq:t_k approximation calculate}
    1+\delta_k\bv_k^\top\bcY(x)\bv_k=0.
\end{equation}
We can then employ the Newton--Raphson method to calculate the value of $t_k$ by iteratively solving the approximate equation \eqref{eq:t_k approximation calculate} above with respect to $x \in \mathbb{R}$ for each $1 \leq k \leq K_0$.

\subsection{Main results} \label{new.sec.mainresu}

We first characterize the fundamental role played by the population quantity $\sft_k$ introduced earlier in (\ref{eq:sft_k def}) based on the generalized QVE. Let us define 
\begin{equation}\label{eq:Apsink unrescaled}
\tpnk:=\frac{q^{1-4\alpha}}{|\delta_k|\sdeg^{1+2\alpha}}+\frac{\xi}{q\sdeg^{2}}+\|\bV\|_{\max},
 \end{equation}
which is another crucial population quantity throughout our technical analysis.

\begin{theorem}\label{thm:eigenvalue}
 Under \Cref{def: model setting} and Assumption \ref{main_assm}, it holds w.h.p.~that
\begin{equation}\label{eqn:t_k-gamma_k unrescaled}
    |\hdelta_k-\sft_k|=O\biggl\{ |\delta_k| \frac{\xi\tpnk}{q}\left(1+\frac{Kq^{4-16\alpha}}{ |\delta_k|^4\sdeg^{8\alpha}}\right)\biggr\}
\end{equation}
for each $1\leq k\leq K_0$, where $\tpnk$ is given in (\ref{eq:Apsink unrescaled}). 
\end{theorem}

Theorem \ref{thm:eigenvalue} above reveals that the population quantity $\sft_k$ is indeed the first-order asymptotic limit of the empirical spiked eigenvalue $\hdelta_k$. In view of \eqref{eq:tM definition}, we have that for $x\in \widetilde{\cI}_k$, $ \tM_{i}(x) = -x^{-1} +\OO(|x|^{-3})$ with $1 \leq i \leq n$; see Lemma \ref{lemma:Upsilon} in Section \ref{Preliminary Estimates} of the Supplementary Material for more details. Combining this fact with \eqref{eq:sft_k def}, we see that $\sft_k$ is close to the population spiked eigenvalue $\delta_k$ asymptotically as shown in Lemma \ref{lemma:tk}. Note that the error bound in \eqref{eqn:t_k-gamma_k unrescaled} above is much smaller than $|\delta_k|$ by \eqref{eq:condK unrescaled}. Thus, it follows from Theorem \ref{thm:eigenvalue} that w.h.p.,
\begin{equation}\label{eq:eigenvalue coro}
|\hdelta_k - \delta_k|=o(|\delta_k|) 
\end{equation}
for each $1\leq k\leq K_0$. From (\ref{eq:defnIk}) and (\ref{eq:eigenvalue coro}), we see that $\hdelta_k$ also lies within set $\widetilde{\cI}_k$ asymptotically. 

Based on \eqref{eq:eigenvalue coro} and eigengap condition \eqref{eq:eigengap unrescaled}, we can define a closed contour $\cC_k$ in the complex plane $\mathbb{C}$ such that w.h.p., $\cC_k$ encloses only $\hdelta_k$ and no other eigenvalues of the generalized Laplacian matrix $\bX$. Such property allows us to extract information about the empirical spiked eigenvector $\bhv_k$ through utilizing the contour integral of the Green's function (i.e., the resolvent) of the random matrix $\bX$, denoted as $(\bX-z\bI)^{-1}$ with $z \in \mathbb{C}$, and applying Cauchy's integral formula. This explains why we will begin with investigating the delicate asymptotic expansions of the empirical spiked eigenvectors $\bhv_k$'s. With such a technical tool, we can unveil the asymptotic behavior of the projection of the empirical spiked eigenvector $\bhv_k$ onto any deterministic unit vector $\bu \in \mathbb{R}^n$. 
To this end, denote by 
\begin{equation}\label{eq:tUpsilonk def}
	\btUpsilon_k(z):=\btUpsilon(z)-\btUpsilon(z)\bV_{-k}\frac{1}{\bDelta_{-k}^{-1}+\bV_{-k}^\top\btUpsilon(z)\bV_{-k}}\bV_{-k}^\top\btUpsilon(z)
\end{equation}
an $n \times n$ 
deterministic matrix with $\btUpsilon(z)$ given in (\ref{eq:tUpsilon definition}).

\begin{theorem}\label{thm:projection}
Under \Cref{def: model setting} and Assumption \ref{main_assm}, it holds w.h.p. that 
\begin{equation}\label{eq:projectionvk unrescaled}
\begin{split}
   \left|\bv_k^\top(\bL/\bLambda)^{-\alpha}\bhv_k- \frac{1}{\sqrt{\delta_k^2\bv_k^\top\btUpsilon_k'(\sft_k)\bv_k}}\right| \lesssim \frac{\xi\tpnk}{q}\left(1+\frac{Kq^{4-16\alpha}}{ |\delta_k|^4\sdeg^{8\alpha}}\right)   
\end{split}
\end{equation}
for each $1\le k\le K_0$, where we choose the direction of $\bhv_k$ such that $\bhv_k^\top\bv_k>0$ and $\btUpsilon_k(\cdot)$ is given in (\ref{eq:tUpsilonk def}).
Moreover, for any deterministic unit vector $\bu \in \mathbb{R}^n$, it holds w.h.p. that 
\begin{equation}\label{eq:projection unrescaled}
\begin{split}
    &\left|\bu^\top(\bL/\bLambda)^{-\alpha}\bhv_k+\frac{\delta_k\bu^\top\btUpsilon_k(\sft_k)\bv_k}{\sqrt{\delta_k^2\bv_k^\top\btUpsilon_k'(\sft_k)\bv_k}}\right| \\
    &\lesssim \frac{\xi\tpnk}{q}\left[1+\frac{Kq^{4-16\alpha}}{ |\delta_k|^4\sdeg^{8\alpha}} + {\|\bu^\top \bV_{-k}\|} \left(\sqrt K+\frac{Kq^{2-8\alpha}}{|\delta_k|^2\sdeg^{4\alpha}}\right)\right].
\end{split}    
\end{equation}
Further, for the second terms on the left-hand side (LHS) of \eqref{eq:projectionvk unrescaled} and \eqref{eq:projection unrescaled}, we have that 
\begin{equation}\label{eq:projectionvk2 unrescaled}
\delta_k^2\bv_k^\top\btUpsilon_k'(\sft_k)\bv_k= 1+O\left(\frac{q^{2-8\alpha}}{|\delta_k|^{2}\sdeg^{4\alpha}}\right),\quad \delta_k \bu^\top\btUpsilon_k(\sft_k)\bv_k = -  \bu^\top \bv_k +O\left(\frac{q^{2-8\alpha}}{|\delta_k|^{2}\sdeg^{4\alpha}}\right).
\end{equation}
\end{theorem}

Theorem \ref{thm:projection} above provides the first-order asymptotic limits of linear projections of the empirical spiked eigenvector $\bhv_k$ under different weight vectors. Observe that due to the concentration of node degrees $d_i$'s and average node degree $\bar d$, the random diagonal matrix $\bL/\bLambda$ is approximately equal to the identity matrix plus a small random error in light of (\ref{eq:defbL}) and (\ref{new.eq.LamMat}). Alternatively, by utilizing \Cref{prop:Lambda and E} in Section \ref{Preliminary Estimates} and low-rankness of signals in \eqref{eq:condK unrescaled}, one can easily derive an estimate for $\bu^\top\bhv_k$ using \eqref{eq:projection unrescaled} and \eqref{eq:projectionvk2 unrescaled} given by 
\begin{equation}\label{eq:projection_cor}
    \bu^\top \bhv_k= \bu^\top\bv_k+O\bigg( \frac{\xi}{q}+\frac{q^{2-8\alpha}}{|\delta_k|^{2}\sdeg^{4\alpha}}+\frac{\sqrt K \xi\tpnk}{q}{\|\bu^\top \bV_{-k}\|} \bigg).
\end{equation}
The major reason why we have chosen to examine $\bu^\top(\bL/\bLambda)^{-\alpha}\bhv_k$ instead of $\bu^\top \bhv_k$ in Theorem \ref{thm:projection} is that the higher-order fluctuations of $\bu^\top(\bL/\bLambda)^{-\alpha}\bhv_k$ have a much cleaner form (as demonstrated in our \Cref{thm:projection2} later). 
On the other hand, the higher-order fluctuations of $\bu^\top \bhv_k$ are generally more complex and may not be optimal in certain scenarios. 


For the special (but significant) case of $\bu=\be_i$, \eqref{eq:projection unrescaled} above provides the first-order asymptotic limits for individual components of the empirical spiked eigenvector $\hv_k(i)$. In fact, we can obtain a much more precise estimate for $\hv_k(i)$ in the theorem below, which will allow us to derive the central limit theorem 
as $n\to \infty$.
 
\begin{theorem}\label{thm:main thm}
Assume that \Cref{def: model setting} and Assumption \ref{main_assm} are satisfied, and 
\begin{equation}\label{eq:main thm assumption unrescaled}
 K\tpnk\sdeg\lesssim 1,\quad \|\bV\|_{\max}\ll \frac{q^{1-4\alpha}}{|\delta_k|\sdeg^{1+2\alpha}}+\frac{\xi}{q\sdeg^{2}}
\end{equation}
for each $1\le k\le K_0$. Then for each $1\le k\le K_0$ and $1 \leq i \leq n$, it holds w.h.p.~that
\begin{align}\label{eq:main thm unrescaled}
	\begin{split}
	 \hv_k(i)&=(\Lambda_i/L_i)^\alpha v_k(i)+\frac{1}{\sft_kL_i^{\alpha}}\sum_{j\in[n]}{W}_{ij}{\Lambda}_j^{-\alpha}v_k(j)\\
	&+O\left(\|\bV\|_{\max}\left(\frac{\sqrt K q^{1-4\alpha}}{|\delta_k|\sdeg^{2\alpha}}+\frac{ K\xi}{q}\right)\left(\frac{q^{1-4\alpha}}{|\delta_k|\sdeg^{1+2\alpha}}+\frac{ \xi}{q\sdeg^{2}}\right)\right)\\
 &+O\left(\frac{\xi q^{1-4\alpha}}{\sqrt n |\delta_k|\sdeg^{2\alpha}}\left(\frac{q^{1-4\alpha}}{|\delta_k|\sdeg^{2\alpha}}+\frac{\xi}{q\sdeg}\right)\right),
	\end{split}
\end{align}
where we choose the direction of $\bhv_k$ such that $\bhv_k^\top\bv_k>0$. Consequently, we have that w.h.p., 
\begin{align}\label{eq:main thm expand unrescaled}
\begin{split}
   	 \hv_k(i)&= v_k(i)- \frac{\al}{\Lambda_i}\bigg(\sum_{j\in[n]} W_{ij} +\frac{\tau_i}{n} \sum_{j,\,l\in[n]} W_{jl}\bigg)v_k(i) +\frac{1}{\sft_k}\sum_{j\in[n]}\Lambda_i^{-\al}{W}_{ij}{\Lambda}_j^{-\alpha}v_k(j)\\
	&+O\left(\|\bV\|_{\max}\left(\frac{\sqrt K q^{1-4\alpha}}{|\delta_k|\sdeg^{2\alpha}}+\frac{ K\xi}{q}\right)\left(\frac{q^{1-4\alpha}}{|\delta_k|\sdeg^{1+2\alpha}}+\frac{ \xi}{q\sdeg^{2}}\right)\right)\\
 &+O\left(\frac{\xi q^{1-4\alpha}}{\sqrt n |\delta_k|\sdeg^{2\alpha}}\left(\frac{q^{1-4\alpha}}{|\delta_k|\sdeg^{2\alpha}}+\frac{\xi}{q\sdeg}\right)\right)
\end{split}
\end{align}
for each $1\le k\le K_0$ and $1 \leq i \leq n$. 
\end{theorem}
\begin{remark}
By \Cref{def:high}, the componentwise asymptotic 
expansions for the empirical spiked eigenvectors $\bhv_k$'s established in \eqref{eq:main thm unrescaled} and \eqref{eq:main thm expand unrescaled} of Theorem \ref{thm:main thm} above hold with very high probability $1-\OO(n^{-D})$ for any large constant $D>0$. Then by applying a union bound, we can conclude that the asymptotic expansions in \eqref{eq:main thm unrescaled} and \eqref{eq:main thm expand unrescaled} hold simultaneously for all $1\leq k\leq K_0$ and $i\in [n]$. If we require only a weaker probability $1-\oo(1)$, it is indeed possible to improve the error term by dropping some $\xi$ factors. However, we refrain from doing so because, in many applications, a uniform estimate in $k$ and $i$ is often necessary. Thus, we opt to keep the $\xi$ factors in order to provide a more general, applicable result.
\end{remark}
\begin{remark}
The additional assumption (\ref{eq:main thm assumption unrescaled}) in Theorem \ref{thm:main thm} is introduced solely for the purpose of simplifying the error term, making its order more apparent to the reader. By imposing such a condition, we can provide a clearer and more concise expression for the error term in our results. Indeed, in network applications, condition (\ref{eq:main thm assumption unrescaled}) is typically considered to be weak. In these applications, it is often assumed that the number of communities $K$ is fixed or slowly diverging, and the spiked eigenvectors are assumed to be delocalized in the sense of satisfying that 
\begin{equation}\label{eq:condK_strong}
\|\bV\|_{\max}^2\lesssim K/n.
\end{equation}
Such condition has been utilized in works such as \cite{erdHos2013spectral,fan2022simple}; \cite{SIMPLERC}. For the interested reader, we provide in \Cref{prop:main_strong} in Section \ref{Sec.proof.thm3} of the Supplementary Material the asymptotic expansion of $L_i^\alpha\hv_k(i)$ without assuming \eqref{eq:main thm assumption unrescaled}. The key difference is that the error term, in this case, is slightly more intricate.

\end{remark}

It is natural to expect the asymptotic distributions based on the asymptotic expansions given in Theorem \ref{thm:main thm}. Specifically, from \eqref{eq:main thm unrescaled} we see that the error term is roughly of order $(q^{2-8\alpha}|\delta_k|^{-2}\sdeg^{-4\alpha}+q^{-2})\|\bV\|_{\max}$ (noting that we always have $\|\bV\|_{\max}\ge n^{-1/2}$) up to some $K$, $\xi$, and $\sdeg$ factors. 
On the other hand, the leading fluctuation term $\sft_k^{-1}\sum_{l\in[n]}\Lambda_i^{-\alpha}{W}_{il}{\Lambda}_l^{-\alpha}v_k(l)$ converges in law to a Gaussian distribution with variance given by
\begin{equation}\label{eq:entry_CLT_variance}
    \sigma_{k,i}^2:=\var\bigg\{\frac{1}{\sft_k}\sum_{l\in[n]}\Lambda_i^{-\alpha}{W}_{il}{\Lambda}_l^{-\alpha}v_k(l)\bigg\}=\frac{\Lambda_i^{-2\alpha}}{\sft_k^2}\sum_{l\in[n]}s_{il}\Lambda_l^{-2\alpha}|v_k(l)|^2,
\end{equation}
which is typically of order $n^{-1}q^{2-8\alpha}|\delta_k|^{-2}\sdeg^{-4\alpha}$. 
Hence, if the error term is much smaller than $\sigma_{k,i}$, we can derive a CLT for $L_i^\alpha\hv_k(i)$ as presented in the corollary below. Using \eqref{eq:main thm expand unrescaled}, we can also derive a similar CLT for $\hv_k(i)$. However, we omit the details for the latter here for simplicity.

\begin{corollary}\label{example:CLT entry}
Under the conditions of \Cref{thm:main thm}, if $\|\bv_k\|_\infty\to 0$ and 
\begin{equation}\label{eq:entry_cond_CLT}
\begin{split}
&\|\bV\|_{\max}\left(\frac{\sqrt K q^{1-4\alpha}}{|\delta_k|\sdeg^{2\alpha}}+\frac{ K\xi}{q}\right)\left(\frac{q^{1-4\alpha}}{|\delta_k|\sdeg^{1+2\alpha}}+\frac{ \xi}{q\sdeg^{2}}\right)\\
 &+\frac{\xi q^{1-4\alpha}}{\sqrt n |\delta_k|\sdeg^{2\alpha}}\left(\frac{q^{1-4\alpha}}{|\delta_k|\sdeg^{2\alpha}}+\frac{\xi}{q\sdeg}\right)\ll \sdeg^\al\sigma_{k,i},
\end{split}
\end{equation}
we have $(L_i^\al\hv_k(i)-\La_i^\al v_k(i))/{\sigma_{k,i}} \overset{d}{\longrightarrow} \mathcal N(0,1)$ 
as $n\rightarrow \infty$ for each $1\leq k\leq K_0$. In particular, \eqref{eq:entry_cond_CLT} holds provided that 
\begin{equation}\label{eq:entry_cond_CLT2}
\sigma_{k,i}\gtrsim \frac{q^{1-4\alpha}}{\sqrt{n}|\delta_k|\sdeg^{2\alpha}},\quad \xi \ll \frac{|\delta_k|\sdeg^{3\al}}{q^{1-4\al}}, \quad \xi^2\ll q\sdeg^{\al+1},\quad \|\bV\|_{\max} \le { \frac{a}{\sqrt{n}}},
\end{equation}
\begin{equation}\label{eq:entry_cond_CLT3}
 K\ll \frac{q\sdeg^{\al+1}}{a\xi}\wedge \frac{q^{3-4\al}\sdeg^{2-\al}}{a|\delta_k|\xi^2}\wedge \frac{|\delta_k|^2\sdeg^{6\al+2}}{a^2q^{2-8\al}}\wedge\frac{q^2\sdeg^{2\al+4}}{a^2\xi^2}
\end{equation}
for some parameter $a\ge 1$ (that may depend on $n$).
\end{corollary}


Let us gain some insights into the assumptions given in Corollary \ref{example:CLT entry} above. For the network setting, assume that $\sum_{l\in[n]}s_{il}|v_k(l)|^2\gtrsim \theta$, $H_{ij}=\OO(\theta)$, and condition \eqref{eq:condK_strong} is satisfied. Then these assumptions entail that $\Lambda_i^{-2\alpha}\sum_{l \in[n]}s_{il}\Lambda_l^{-2\alpha}|v_k(l)|^2\gtrsim \theta q^{8\alpha}\sdeg^{4\alpha}$ and $|\delta_k|^2 \le \|\bH\|^2 \lesssim q^{3-8\alpha}\sdeg^{-4\alpha}$. Under these assumptions, we see that \eqref{eq:entry_cond_CLT2} holds with the choice of $a=\sqrt{K}$, as long as  $q^{-1+4\alpha}|\delta_k|\sdeg^{3\al}\gg \xi$ and $K$ is not too large, specifically 
$$K\ll \left({q}/{\xi^2}\right)^{2/3}\wedge  \frac{|\delta_k|\sdeg^{2\al}}{q^{1-4\al}}.$$ 


We next turn our attention to investigating the delicate asymptotic expansions for the empirical spiked eigenvalues $\hdelta_k$'s. We will present higher-order asymptotic expansions for both $\hdelta_k-\sft_k$ and $\bu^\top(\bL/\bLambda)^{-\alpha}\bhv_k$ below, which improve the results in Theorems \ref{thm:eigenvalue} and \ref{thm:projection}, respectively. In other words, we will extract the leading order random fluctuations from the error terms in \eqref{eqn:t_k-gamma_k unrescaled} and \eqref{eq:projection unrescaled}, respectively. 

\begin{theorem}\label{thm:eigenvalue2}
Under Definition \ref{def: model setting} and Assumption \ref{main_assm}, for each $1\leq k\leq K_0$ it holds w.h.p. that
\begin{align}\label{eqn:t_k-gamma_k2 unrescaled}
\begin{split}
    &\hdelta_k-\sft_k - A_k= -2\alpha \sft_k\bv_k^\top\frac{\bL-\bLambda}{\bLambda}\bv_k+\bv_k^\top\bcW\bv_k\\
    &+B_k+O\left(\frac{q^{3-12\al}}{|\delta_k|^2\sdeg^{6\al}}+\frac{\xi^3|\delta_k|}{q^3\sdeg^3}+\frac{\sqrt K\xi|\delta_k|\tpnk}{q}\left(\frac{q^{2-8\al}}{|\delta_k|^2\sdeg^{4\al}}+\frac{\sqrt K\xi\tpnk}{q}\right)\right),
\end{split}
\end{align}
where $\bcW:=\bLambda^{-\alpha}{\bW}\bLambda^{-\alpha}$, $A_k$ is a deterministic term given by 
\begin{align}\label{eq:CLT_evalue mean}
\begin{split}
A_k & =\alpha(2\alpha+1)\sft_k\E\bv_k^\top\frac{(\bL-\bLambda)^2}{\bLambda^2}\bv_k-4\alpha\E\bv_k^\top\frac{\bL-\bLambda}{\bLambda}\bcW\bv_k,
\end{split}
\end{align}
and $B_k$ is a centered random error satisfying
\[
    \var (B_k)\lesssim \frac{|\delta_k|^2\|\bv_k\|_\infty^2}{q^4\sdeg^4}+\frac{|\delta_k|^2}{q^4 n^2 \sdeg^4}+\frac{\|\bv_k\|_\infty^2}{q^{8\al}\sdeg^{2+4\al}}+\frac{1}{q^{8\al}n\sdeg^{2+4\al}}+\frac{q^{3-16\al}}{\sqrt n|\delta_k|^2\sdeg^{8\al}}.
\]
\end{theorem}

Roughly speaking, the asymptotic expansion in \Cref{thm:eigenvalue2} above states that the fluctuation of the empirical spiked eigenvalue $ \hdelta_k$ is dominated by the random variable
$$-2\alpha \sft_k\bv_k^\top\frac{\bL-\bLambda}{\bLambda}\bv_k+\bv_k^\top\bcW\bv_k.$$
Through direct calculations, we can obtain its variance as 
\begin{align}\label{eq:CLT_evalue var}
\begin{split}
    \varsigma_k^2:=\sum_{1 \leq i\leq j \leq n}\left(\frac{\sft_k\fS_{ij}^{\bv_k\bv_k}}{1+\delta_i^j}\right)^{2}s_{ij},
\end{split}
\end{align}
where for any vectors $\bx = (x(i))_{i \in [n]},\,\by = (y(i))_{i \in [n]} \in \mathbb{R}^n$ and $i,\,j\in[n]$, 
\begin{align}\label{eq:variancesigma}
    \fS^{\bx\by}_{ij}:=&-{2\alpha}\left(\frac{x(i)y(i)}{\Lambda_i}+\frac{x(j)y(j)}{\Lambda_j}+\frac{2}{n}\sum_{l\in[n]}\frac{\tau_lx(l)y(l)}{\Lambda_l}\right)\\
    &+\sft_k^{-1}\frac{x(i)y(j)+x(j)y(i)}{(\Lambda_i\Lambda_j)^{\alpha}}, \nonumber
    \end{align}
and $\delta_i^j$ represents the Kronecker delta. As long as the variance of the error terms is asymptotically negligible, we can derive a CLT for the empirical spiked eigenvalue $\hdelta_k$ as presented in the corollary below.

\begin{corollary}\label{cor:CLT_evalue}
Under Definition \ref{def: model setting} and Assumption \ref{main_assm}, if $\|\bv_k\|_\infty\to 0$ and 
\begin{align}\label{eq:CLT_evalue var assumption}
\begin{split}
    \varsigma_k&\gg\frac{q^{3-12\al}}{|\delta_k|^2\sdeg^{6\al}}+\frac{\xi^3|\delta_k|}{q^3\sdeg^3}+\frac{\sqrt K\xi|\delta_k|\tpnk}{q}\left(\frac{q^{2-8\al}}{|\delta_k|^2\sdeg^{4\al}}+\frac{\sqrt K\xi\tpnk}{q}\right)\\
    &+\frac{|\delta_k|\|\bv_k\|_{\infty}}{q^2\sdeg^2}+\frac{\|\bv_k\|_{\infty}}{q^{4\al}\sdeg^{1+2\al}}+\frac{1}{\sqrt n q^{4\al}\sdeg^{1+2\al}},
\end{split}
\end{align}
we have 
\begin{align}\label{eq:CLT_evalue}
\begin{split}
    \frac{\hdelta_k-\sft_k-A_k}{\varsigma_k}\overset{d}{\longrightarrow}\cN(0,1)
\end{split}
\end{align}
as $n \rightarrow \infty$ for each $1\leq k\leq K_0$, where $\varsigma_k$ is given in (\ref{eq:CLT_evalue var}). 
\end{corollary}

\begin{remark}
Let us gain some insights into the assumptions in Corollary \ref{cor:CLT_evalue} above associated with the CLT established for the empirical spiked eigenvalue $\hdelta_k$. Notice that its asymptotic standard deviation $\varsigma_k$ is typically of order $$ \frac{q^{1-4\al}}{\sqrt{n}\sdeg^{2\al}}+\frac{|\delta_k|}{q} \|\bv_k\|_{4}^2 $$
in the generic case (when there are no ``essential cancellations" in the expression of $\varsigma_k^2$). Further, we have $\|\bv_k\|_{4}^2\gtrsim n^{-1/2}$. 
Thus, if $\delta_k^2\gg \xi\sqrt{n}q^{2-8\al}\sdeg^{-4\al}$ and $q \gg K\xi^2 \sqrt{n}\|\bV\|_{4}^2 + (K\xi^4 \sqrt{n})^{1/3}$, condition \eqref{eq:CLT_evalue var assumption} holds. From (\ref{eq:CLT_evalue}) above, we see that the asymptotic bias is given by the population quantity $A_k$, which takes the following form: 
\begin{align}\label{eq:A_k formula}
\begin{split}
    A_k=&~ \al(1+2\al)t_k\sum_{i\in[n]}\left(\left(1+\frac{4\tau_i}{n}\right)\sum_{j\in[n]}s_{ij}-\frac{2\tau_i}{n}s_{ii}+\frac{2\tau_i^2}{n^2}\Sigma_a-\frac{\tau_i^2}{n^2}\tr(\Sigma)\right)\frac{v_k(i)^2}{\Lambda_{ii}^2}\\
    &~ -4\al\sum_{i,j\in[n]}\left(\left(1+\frac{2\tau_i}{n}\right)s_{ij}-\frac{\tau_i}{n}\delta_i^js_{ii}\right)\frac{v_k(i)v_k(j)}{\Lambda_i^{1+\al}\Lambda_j^{\al}},
\end{split}
\end{align}
where $\Sigma_{a}:=\sum_{i,j\in[n]}s_{ij}$ and $\tr(\Sigma)=\sum_{i\in[n]}s_{ii}$. In practice, $A_k$ can be estimated as $\widehat{A}_k$, by replacing all parameters by their counterparts, see the bias correction idea and  \eqref{eq:biascorrectedAk} at the end of this section for more details. 
\end{remark}


We now examine the higher-order asymptotic expansions for the empirical spiked eigenvectors $\bhv_k$'s that will enable us to derive the associated CLT results. In particular, to simplify the results of Theorem \ref{thm:projection2} below, we will decompose vector $\bu$ into two parts that are perpendicular to or parallel to $\bv_k$, respectively.

\begin{theorem}\label{thm:projection2}
Assume that Definition \ref{def: model setting} and Assumption \ref{main_assm} are satisfied. Then we have that 

1) For each $1\leq k\leq K_0$ and any deterministic unit vector $\bu \in \mathbb{R}^n$ such that $\bu^\top\bv_k=0$, 
it holds w.h.p. that
\begin{align}\label{eq:projection2 unrescaled}
\begin{split}
    &\bu^\top(\bL/\bLambda)^{-\alpha}\bhv_k -\cA_k= \sft_k\bu^\top\bV_{-k}\frac{1}{\sft_k-\bDelta_{-k}}\bV_{-k}^\top\left(-2\alpha\frac{\bL-\bLambda}{\bLambda}+\sft_k^{-1}\bcW\right)\bv_k\\
    +&\bw^\top\left(-2\alpha\frac{\bL-\bLambda}{\bLambda}+t_k^{-1}\bcW\right)\bv_k+\sum_{l\in[K]\setminus\{k\}}\frac{\sft_k\bu^\top\bv_l}{\sft_k-\delta_l}\cB_{k,l}+\cB_k^{\bw}\\
    +&O\left(K\left(\frac{q^{2-8\al}}{|\delta_k|^2\sdeg^{4\al}}+\frac{\xi\tpnk}{q}\right)\left(\frac{q^{1-4\al}}{|\delta_k|\sdeg^{2\al}}+\frac{\xi}{q\beta_n}\right)
    +\frac{K^{3/2}\xi\tpnk}{q}\left(\frac{\xi\tpnk}{q}+\frac{q^{2-8\al}}{|\delta_k|^2\sdeg^{4\al}}\right)\right),
\end{split}
\end{align}
where we choose the sign of $\bhv_k$ such that $\bhv_k^\top\bv_k>0$
, $\bw=(\bI-\bV\bV^\top)\bu$, $\cA_k$ is a deterministic term given by 
\begin{align*}
\cA_k & =\E\bw^\top\left(\alpha(2\alpha+1)\frac{(\bL-\bLambda)^2}{\bLambda^2}- \frac{2\alpha}{ t_k}\left(\frac{\bL-\bLambda}{\bLambda}\bcW+\bcW\frac{\bL-\bLambda}{\bLambda}\right)+\frac{\bcW^2}{t_k^{2}}\right)\bv_k\\
&+\quad\sft_k\bu^\top\bV_{-k}\frac{1}{\sft_k-\bDelta_{-k}}\E \bV_{-k}^\top\left(\alpha(2\alpha+1)\frac{(\bL-\bLambda)^2}{\bLambda^2}\right.\\
& \quad \left.-\frac{2\alpha}{\sft_k}\left(\frac{\bL-\bLambda}{\bLambda}\bcW+\bcW\frac{\bL-\bLambda}{\bLambda}\right)+\frac{\bcW^2}{\sft_k^{2}}\right)\bv_k,
\end{align*}
$\cB_{k}^{\bw}$ is a centered random variable satisfying
\begin{align*}
\var(\cB_{k}^{\bw}) & \lesssim\frac{\|\bv_k\|_\infty\|\bw\|_\infty}{q^4\sdeg^4}+\frac{|w|}{q^4n^2\sdeg^4}+\frac{1}{q^{8\al}|\delta_k|^2\sdeg^{2+4\al}}\Big(\|\bv_k\|_\infty\|\bw\|_\infty \\
&\quad + \frac{|w|}{n} \Big)+\frac{q^{3-16\al}}{\sqrt n|\delta_k|^4\sdeg^{8\al}}|w|,
\end{align*}
and for each $l\in[K]\setminus\{k\}$, $\cB_{k,l}$ is a centered random variable satisfying
\begin{align*}
\var(\cB_{k,l}) & \lesssim\frac{\|\bv_k\|_\infty\|\bv_l\|_\infty}{q^4\sdeg^4}+\frac{1}{q^4n^2\sdeg^4}+\frac{1}{q^{8\al}|\delta_k|^2\sdeg^{2+4\al}}\Big(\|\bv_k\|_\infty\|\bv_l\|_\infty \\
&\quad + \frac{1}{n} \Big)+\frac{q^{3-16\al}}{\sqrt n|\delta_k|^4\sdeg^{8\al}}.
\end{align*}

2) For the case of $\bu=\bv_k$ and each $1\leq k\leq K_0$, it holds w.h.p. that
\begin{align}\label{eq:projection3unrescaled}
\begin{split}
     \bv_k^\top(\bL/\bLambda)^{-\alpha}\bhv_k & -\bv_k^\top(\bL/\bLambda)^{-\alpha}\bv_k -\fA_k =\frac{\alpha^2}{2}\bv_k^\top\left(\frac{\bL-\bLambda}{\bLambda}\right)^2\bv_k-\frac{1}{2\sft_k^{2}}\bv_k^\top\bcW^2\bv_k\\
    &+\fB_k+O\left(\frac{Kq^{4-16\al}}{|\delta_k|^4\sdeg^{8\al}}+\frac{K\xi^2\tpnk^2}{q^2}\right),
\end{split}
\end{align}
where $\fA_k$ is a deterministic term given by 
\[
\fA_k:=(\delta_k^2\bv_k^\top\btUpsilon'_k(\sft_k)\bv_k)^{-1/2}-1+\frac{1}{2}\bv_k^\top(\sft_k^2\btUpsilon'(\sft_k)+2\sft_k\btUpsilon(\sft_k)+\bI)\bv_k\]
and $\fB_k$ is a random variable satisfying
\[
\E\fB_k^2\lesssim\frac{n^2\|\bv_k\|_\infty^4}{q^8\sdeg^6}+\frac{n^2q^{4-24\al}\|\bv_k\|_\infty^4}{|\delta_k|^6\sdeg^{12\al}}.
\]
\end{theorem}

Using the higher-order asymptotic expansions established in Theorem \ref{thm:projection2} above, we are ready to present more general CLT results for the empirical spiked eigenvectors $\bhv_k$'s (than the one obtained in Corollary \ref{example:CLT entry} before) under certain conditions on $q$ and $|\delta_k|$ in the corollary below.


\begin{corollary}\label{cor:CLT_evector}
Assume that Definition \ref{def: model setting} and Assumption \ref{main_assm} are satisfied, and $\|\bv_k\|_\infty\to 0$, $\sqrt n\ll q^2$, $\sqrt n q^{2-8\al}\sdeg^{-4\al}\ll |\delta_k|^2$. Then we have that 

1) For each $1\leq k\leq K_0$ and any deterministic unit vector $\bu \in \mathbb{R}^n$ such that $|\bu^\top\bv_k|\neq 1$, 
if 
\begin{align}\label{eq:CLT_evector1 var assumption}
\begin{split}
    \frac{\mathfrak s_{\bu,k}}{\sqrt{1-|\bu^\top\bv_k|^2}} & \gtrsim K\left(\frac{q^{2-8\al}}{|\delta_k|^2\sdeg^{4\al}}+\frac{\xi\tpnk}{q}\right)\left(\frac{q^{1-4\al}}{|\delta_k|\sdeg^{2\al}}+\frac{\xi}{q\beta_n}\right)\\
    &+\frac{K^{3/2}\xi\tpnk}{q}\left(\frac{\xi\tpnk}{q}+\frac{q^{2-8\al}}{|\delta_k|^2\sdeg^{4\al}}\right)\\
    &+\frac{\sqrt K\|\bV\|_{\max}}{q\sdeg}\left(\frac{1}{q\sdeg}+\frac{q^{1-4\al}}{|\delta_k|\sdeg^{2\al}}\right)+\sqrt{\frac{K q^{1-8\al}}{\sqrt n|\delta_k|^2\sdeg^{4\al}}}\\
    &+\left(\frac{1}{q^2\sdeg^2}+\frac{1}{q^{4\al}|\delta_k|\sdeg^{1+2\al}}\right)\sqrt{\|\bv_k\|_\infty\|\bw\|_\infty}\\
    &+\frac{|\bu^\top\bv_k|}{\sqrt{1-|\bu^\top\bv_k|^2}}\left(\frac{q^{2-8\al}}{\sqrt n|\delta_k|^2\sdeg^{4\al}}+\frac{K q^{4-16\al}}{|\delta_k|^4\sdeg^{8\al}}+\frac{K\xi^2\tpnk^2}{q^2}\right.\\
    &\left.+\frac{\sqrt n\|\bv_k\|^2_\infty}{q^2\sdeg^2}+\frac{n q^{2-12\al}\|\bv_k\|^2_\infty}{|\delta_k|^3\sdeg^{6\al}}\right)
\end{split}
\end{align}
with 
\begin{align}\label{eq:CLT_evector1 var}
\mathfrak s^2_{\bu,k}:=\sum_{i\le j\in[n]}(1+\delta_i^j)^{-2}\left(\sum_{l\in[K]}^{(k)}\bu^\top\bv_l\frac{\sft_k}{\sft_k-\delta_l}\fS_{ij}^{\bv_l\bv_k}+\fS_{ij}^{\mathbf{w}\bv_k}\right)^2s_{ij},
\end{align}
it holds that 
\begin{align}\label{eq:CLT_evector1}
    \frac{\bu^\top(\bL/\bLambda)^{-\alpha}\bhv_k-\bu^\top\bv_k\bv_k^\top(\bL/\bLambda)^{-\alpha}\bv_k-\fD_{\bu,k}}{\mathfrak s_{\bu,k}}\overset{d}{\longrightarrow}\cN(0,1)
\end{align}
as $n \rightarrow \infty$, where we choose the direction of $\bhv_k$ such that $\bhv_k^\top\bv_k>0$, $\bw=(\bI-\bV\bV^\top)\bu$, and
\begin{align*}
    \fD_{\bu,k}:=&\E\bw^\top\left(\alpha(2\alpha+1)\frac{(\bL-\bLambda)^2}{\bLambda^2}- \frac{2\alpha}{ t_k}\left(\frac{\bL-\bLambda}{\bLambda}\bcW+\bcW\frac{\bL-\bLambda}{\bLambda}\right)+\frac{\bcW^2}{t_k^{2}}\right)\bv_k\\
    &+\sft_k\bu^\top\bV_{-k}\frac{1}{\sft_k-\bDelta_{-k}}\E \bV_{-k}^\top\left(\alpha(2\alpha+1)\frac{(\bL-\bLambda)^2}{\bLambda^2}\right.\\
    & \left.-2\alpha \sft_k^{-1}\left(\frac{\bL-\bLambda}{\bLambda}\bcW+\bcW\frac{\bL-\bLambda}{\bLambda}\right)+\sft_k^{-2}\bcW^2\right)\bv_k\\
    &+\frac{\alpha^2}{2}\bu^\top\bv_k\E\bv_k^\top\left(\frac{\bL-\bLambda}{\bLambda}\right)^2\bv_k-\frac{1}{2\sft_k^{2}}\bu^\top\bv_k\E\bv_k^\top\bcW^2\bv_k\\
    &+\bu^\top\bv_k\left((\delta_k^2\bv_k^\top\btUpsilon'_k(\sft_k)\bv_k)^{-1/2}-1+\frac{1}{2}\bv_k^\top(\sft_k^2\btUpsilon'(\sft_k)+2\sft_k\btUpsilon(\sft_k)+\bI)\bv_k\right) 
\end{align*}
with $\bcW:=\bLambda^{-\alpha}{\bW}\bLambda^{-\alpha}$.

2) For the case of $\bu=\bv_k$ and each $1\leq k\leq K_0$, if 
\begin{equation}\label{eq:CLT_evector2 var assumption}
    \kappa_{\bv_k}^{1/2}\ll \mathfrak s_{\bv_k,k}^2,\quad 
    \frac{\sqrt n}{q}\left(\frac{1}{q^4\sdeg^4}+\frac{q^{4-16\al}}{|\delta_k|^4\sdeg^{8\al}}\right)\|\bv_k\|_\infty^4+\frac{n^2\|\bv_k\|_\infty^4}{q^8\sdeg^6}+\frac{n^2q^{4-24\al}\|\bv_k\|_\infty^4}{|\delta_k|^6\sdeg^{12\al}}\ll \mathfrak s_{\bv_k,k}^2
\end{equation}
with $s_{\bv_k,k}^2$ and $\kappa_{\bv_k}$ given in  \eqref{eq:variance para def} and \eqref{eq:variance Pk def}, respectively (see Section \ref{Sec.proof.cor3} of the Supplementary Material), it holds that 
\begin{align}\label{eq:CLT_evector2}
    \frac{\bv_k^\top(\bL/\bLambda)^{-\alpha}\bhv_k-\bv_k^\top(\bL/\bLambda)^{-\alpha}\bv_k-\fE_{\bv_k,k}}{\mathfrak s_{\bv_k,k}}\overset{d}{\longrightarrow}\cN(0,1) 
\end{align}
as $n \rightarrow \infty$, where we choose the direction of $\bhv_k$ such that $\bhv_k^\top\bv_k>0$ and
\begin{align*}
    \fE_{\bv_k,k}:=&\frac{\alpha^2}{2}\E\bv_k^\top\left(\frac{\bL-\bLambda}{\bLambda}\right)^2\bv_k-\frac{1}{2\sft_k^{2}}\E\bv_k^\top\bcW^2\bv_k+(\delta_k^2\bv_k^\top\btUpsilon'_k(\sft_k)\bv_k)^{-1/2}\\
    &-1+\frac{1}{2}\bv_k^\top(\sft_k^2\btUpsilon'(\sft_k)+2\sft_k\btUpsilon(\sft_k)+\bI)\bv_k 
\end{align*}
with $\bcW:=\bLambda^{-\alpha}{\bW}\bLambda^{-\alpha}$.
\end{corollary}

\begin{remark}
It is worth mentioning that we can, in fact, derive even higher-order asymptotic expansions than those in \eqref{eqn:t_k-gamma_k2 unrescaled} and \eqref{eq:projection2 unrescaled}, where higher-order fluctuations are extracted from the error terms. This will allow us to derive the limiting distributions of $\hdelta_k-\sft_k$ and $\bu^\top(\bL/\bLambda)^{-\alpha}\bhv_k$ under weaker assumptions on $q$ and $|\delta_k|$, specifically for smaller values of $q$ and $|\delta_k|$. In principle, our technical analysis indeed allows us to derive arbitrarily high-order series of asymptotic expansions for $\hdelta_k-\sft_k$ and $\bu^\top(\bL/\bLambda)^{-\alpha}\bhv_k$ for $q\ge n^\e$ and $|\delta_k| \ge n^\epsilon q^{1-4\al}\sdeg^{-2\al}$ with $\epsilon$ some small positive constant, as shown in \cite{fan2020asymptotic}. However, unlike in \cite{fan2020asymptotic}, it is very challenging to determine the limiting distributions of the high-order terms in these asymptotic expansions due to the intrinsic correlation between random matrices $\bL$ and $\bW$. Due to the length constraint, we leave the study of this problem to future work.
\end{remark}

For the random network model setting with the entries of $\btX$ having Bernoulli distributions, we can obtain slightly sharper results. For the convenience of the reader, we include such refined results in Section \ref{app:network} of the Supplementary Material. 
For a complete theory, we finally consider the practical problem of estimating the latent embedding dimensionality $K_0$ (i.e., the number of strong spikes). 

\begin{theorem}\label{thm:K_0}
Assume that Definition \ref{def: model setting} and Assumption \ref{main_assm} are satisfied, 
\begin{equation}\label{eq:K_0+1 assumption unrescaled}
|\delta_{K_0+1}|\gg q^{1-4\al}\sdeg^{-2\al}, \ \  \left|\frac{\delta_{K_0+1}}{\delta_{K_0+2}}\right|\geq 1+\epsilon_0, \ \ K\xi\tpnk(\delta_{K_0+1})\ll q, 
\end{equation}
$K_0$ can be represented as
\begin{equation}
    K_0=\max\left\{k\in[K]:|\delta_k|\geq a_n\right\}
\end{equation}
with some deterministic sequence $a_n\gg q^{1-4\al}\sdeg^{-2\al}$, and there exists some deterministic sequence $a_n'$ such that
\begin{equation}\label{eq:a_n assumption unrescaled}
    \limsup_{n\to\infty}\left|\frac{a_n'}{a_n}\right|<1,\quad \limsup_{n\to\infty}\frac{|\delta_{K_0+1}|}{a_n'}<1.
\end{equation}
Then the estimate of the latent embedding dimensionality defined as 
\begin{equation} \label{new.eq.K_0estimate}
    \hK_0:=\max\{k\in[K]:|\hdelta_k|\geq a_n'\}
\end{equation}
is a consistent estimator of $K_0$, i.e., $\P\{\hK_0 = K_0\} \rightarrow 1$ as $n \rightarrow \infty$.
\end{theorem}

Theorem \ref{thm:K_0} above justifies the practical utility of the latent embedding dimensionality estimate $\hK_0$ constructed in (\ref{new.eq.K_0estimate}). For suggestions on the choices of $a_n'$, if $a_n$ in \eqref{eq:a_n assumption unrescaled} can be taken as $a_n=q^{1-4\al}\sdeg^{-2\al}(\log n)^c$ for some $c>0$, and condition (iv) of \Cref{def: model setting} can be strengthened as
\begin{equation}\label{eq:rank etimation sparsity condition}
    c_0\leq\max_{i\in[n]}\theta_i\leq C_0,
\end{equation}
we propose to use
\begin{equation}
    a_n'=\frac{\check{q}}{(\min_{j\in[n]} L_j)^{2\al}}(\log n)^c\log\log n,
\end{equation}
where $\check q>0$ and  $\check{q}^2:=\max_{j\in[n]}\sum_{l\in[n]}X_{lj}$, representing the maximum node degree of the network. Note that from a simple concentration inequality and \eqref{eq:rank etimation sparsity condition}, we have that with probability $1 - o(1)$,
\begin{equation}
    \check{q}^2=(1+o(1))q^2\max_{j\in[n]}\theta_j\sim q^2,\quad \min_{j\in[n]} L_j=(1+o(1))q^2\min_j(\theta_j+\tau_j\bar{\theta}+\lambda_j)\sim q^2\sdeg.
\end{equation}
For example, as suggested in \cite{SIMPLERC}, when considering the SIMPLE-RC test in the DCMM model (\Cref{example:DCMM}), we can choose $c=1/2$ for testing a given pair of nodes and $c=3/2$ for the group test. For more information on the rank inference in the network setting, see \cite{SIMPLERC, han2023universal}.


The above estimations from \Cref{example:CLT entry} to \Cref{cor:CLT_evector} suggest that the spiked eigenvector $\bv_k$ can be estimated by $\bhv_k$, and $t_k$ can be estimated by $\hdelta_k$. The estimation of $s_{ij}=\E|W_{ij}|^2$ is provided by the bias correction idea from \cite{fan2022simple}, as we discuss below.

A naive estimator of $s_{ij}$ is $\hW_{0,ij}^2$, with $\bhW_0=(\hW_{0,ij}):=\btX-\bL^\al\left(\sum_{k\in[\hK]}\hdelta_k\bhv_k\bhv_k^\top\right)\bL^\al$ and $\hK$ given by \eqref{new.eq.K_0estimate}. However, this estimator is not accurate enough in practice, as it is well-known that $\hdelta_k$ is biased upward. Thus, we exploit the following one-step refinement procedure, which is motivated by the higher-order asymptotic expansion of $t_k$ (as presented in \eqref{eq:sft_k def}, \eqref{eq:bcY def}, and \eqref{eq:t_k approximation calculate}). This refinement procedure aims to shrink $\hdelta_k$ and reduce the bias to a more reasonable level:
\begin{enumerate}
\item Compute the initial estimator $\bhW_0=\btX-\bL^\al\left(\sum_{k\in[\hK]}\hdelta_k\bhv_k\bhv_k^\top\right)\bL^\al$ and the estimate of $s_{ij}$ as $\hs_{ij,0}=\hW_{0,ij}^2$, and denote by $\hSig_0:=(\hs_{ij,0})_{i,j\in[n]}$. We also estimate $t_k$ with $\widehat{t}_{k,0}=\hdelta_k$ initially.
\item Calculate an estimate of the theoretical bias term $A_k$ as
\begin{align}
\widehat{A}_{k,0}=&\al(1+2\al)\widehat{t}_{k,0}\sum_{i\in[n]}\left(\left(1+\frac{4\tau_i}{n}\right)\sum_{j\in[n]}\hs_{ij,0}-\frac{2\tau_i}{n}\hs_{ii,0}+\frac{2\tau_i^2}{n^2}\hSig_{a,0}-\frac{\tau_i^2}{n^2}\tr(\hSig_{0})\right)\frac{\hv_k(i)^2}{L_{ii}^2} \nonumber\\
    &-4\al\sum_{i,j\in[n]}\left(\left(1+\frac{2\tau_i}{n}\right)\hs_{ij,0}-\frac{\tau_i}{n}\delta_i^j\hs_{ii,0}\right)\frac{\hv_k(i)\hv_k(j)}{L_i^{1+\al}L_j^{\al}} \label{eq:biascorrectedAk}
\end{align}
with $\hSig_{a,0}:=\sum_{i,j\in[n]}\hs_{ij,0}$.
\item Update the estimator of $t_k$ according to
\[
    \widehat{t}_{k,1}=\hdelta_k - \widehat{A}_{k,0}.
\]
\item Using the initial estimator $\bhW_0$, update the estimator of $\delta_k$ according to
\[
    \tdelta_{k}:=\left[\frac{1}{\widehat{t}_{k,1}}+\frac{\bhv_k\diag[(\bL^{-\al}\bhW_0\bL^{-\al})^2]\bhv_k^\top}{\widehat{t}_{k,1}^3}\right]^{-1}.
\]
\item Update the estimator of $\bW$ as $\bhW=(\hW_{ij}):=\btX-\bL^\al\left(\sum_{k\in[\hK]}\tdelta_k\bhv_k\bhv_k^\top\right)\bL^\al$ and calculate the estimate of $s_{ij}$ as $\hs_{ij}:=\hW_{ij}^2$.
\end{enumerate}
We propose estimating the asymptotic variances of the eigenvector components $\sigma_{k,i}^2$ and of the eigenvalue $\varsigma_k^2$ by substituting $t_k$, $\bv_k$, $s_{ij}$, and $\bLambda$ with $\widehat{t}_k$, $\bhv_k$, $\hs_{ij}$, and $\bL$, respectively, in \eqref{eq:entry_CLT_variance}, \eqref{eq:CLT_evalue var}, and \eqref{eq:variancesigma}. In particular, to estimate the population quantity $\delta_k$ (as opposed to the population quantity $t_k$), we can apply the idea of correction by estimating $\widehat{A}_k$ coupled with the empirical bias correction suggested above for the empirical spiked eigenvalues $\hdelta_k$.

\section{Technical innovations of our theory} \label{techinno}


As mentioned in Section \ref{sec:mainresult}, the main results of our paper are the high-order asymptotic expansions for the empirical spiked eigenvalues, and the components and projections of the empirical spiked eigenvectors of the generalized Laplacian matrix $\bX$, which are presented in Theorems \ref{thm:eigenvalue2}, \ref{thm:main thm}, and \ref{thm:projection2}, respectively. These results have practical implications and can be exploited to establish the CLTs for enabling valid inference of both spiked eigenvalues and spiked eigenvectors, as stated in Corollaries \ref{cor:CLT_evalue}, \ref{example:CLT entry}, and \ref{cor:CLT_evector}, respectively. We further provide the theoretical results under the specific case of network model setting when the underlying random matrix $\btX$ is generated as the adjacency matrix of a random graph. Such a setting introduces a stronger assumption in \eqref{eq:network case}, leading to enhanced technical results presented in Section \ref{app:network} of the Supplementary Material.

To provide a better picture of our technical innovations, we offer a detailed description of the structure of our proofs of the main results, as well as the additional mathematical challenges encountered in our setting. The complete proofs can be found in Sections \ref{sec.mainresu.rescaledmodel}--\ref{addtech.details} of the Supplementary Material. We follow a similar approach as in previous works on the empirical spiked eigenvalues and eigenvectors in RMT, such as \cite{fan2020asymptotic,SIMPLERC,ke2024optimal}. We begin by considering the master equation for the spiked eigenvalues. Our proofs are mainly based on the ``resolvents" (i.e., the Green functions) of relevant random matrices defined as     
\begin{equation}\label{eq:defGR unrescaled}
	\bG(z):=\left({\bcW}-z(\bL/\bLambda)^{2\alpha}\right)^{-1} \ \text{ and } \ \bR(z):=\left({\bcW}-z\bI\right)^{-1}
\end{equation}
where $\bcW:=\bLambda^{-\alpha}{\bW}\bLambda^{-\alpha}$ and $z \in \mathbb{C}$.  
We next focus on the equation governing the behavior of the empirical spiked eigenvalue $\hdelta_k$, observing that 
\begin{align}\label{eq:tech inno eigenvalue}
\begin{split}
    \det(\bX-\hdelta_k\bI)=0
    	\iff&\det(\bLambda^{-\alpha}\btX\bLambda^{-\alpha}-\hdelta_k(\bL/\bLambda)^{2\alpha})=0\\
    	\iff&\det(\bG^{-1}(\hdelta_k)+\bV\bDelta\bV^\top)=0\\
    	\iff&\det(\bDelta^{-1}+\bV^\top\bG(\hdelta_k)\bV)=0.
\end{split}
\end{align} 
To analyze the asymptotic behavior of the empirical spiked eigenvalue $\hdelta_k$, we introduce the asymptotic limit of the resolvent $\bG(z)$, denoted as $\btUpsilon(z)$. We then replace $\bG(z)$ in  \eqref{eq:tech inno eigenvalue} with $\btUpsilon(z)$ and obtain a deterministic equation 
\begin{equation}\label{eq:tech inno tk}
    \det(\bDelta^{-1}+\bV^\top\btUpsilon(t_k)\bV)=0,
\end{equation}
which characterizes the asymptotic limit of $\hdelta_k$, denoted as $\sft_k$. To establish the relationship between $\hdelta_k$ and $\sft_k$ and derive the asymptotic expansion of $\hdelta_k$, we subtract the expressions in \eqref{eq:tech inno eigenvalue} and \eqref{eq:tech inno tk}, and control the error term $\bV(\bG(z)-\btUpsilon(z))\bV$. This enables us to analyze the asymptotic behavior of the empirical spiked eigenvalues.

Moving on to the empirical spiked eigenvectors, we employ the Cauchy integral formula to extract a specific spiked eigenvector $\bhv_k$ from the random generalized Laplacian matrix $\bX$ using the formula
\begin{equation}
(\bL/\bLambda)^{-\alpha}\bhv_k\bhv_k^\top(\bL/\bLambda)^{-\alpha}=-\frac{1}{2\pi i}\oint_{\cC_k}(\bL/\bLambda)^{-\alpha}(\bX-z)^{-1}(\bL/\bLambda)^{-\alpha} \dd z,
\end{equation}
where $\cC_k$ represents a contour in the complex plane $\C$ that encloses only the eigenvalue $\hdelta_k$ and no other eigenvalues of random matrix $\bX$, and $i = (-1)^{1/2}$ denotes the imaginary unit. By leveraging the Woodbury matrix identity, we can obtain the representation 
\begin{align}\label{eq:tech inno bilinear form}
\begin{split}
&\bu^\top(\bL/\bLambda)^{-\alpha}\bhv_k\bhv_k^\top(\bL/\bLambda)^{-\alpha}\bv\\
=&-\frac{1}{2\pi i}\oint_{\cC_k}\bu^\top\left(\bG(z)-\bG(z)\bV\frac{1}{\bDelta^{-1}+\bV^\top\bG(z)\bV}\bV^\top\bG(z)\right)\bv\dd z,
\end{split}
\end{align} 
which expresses the bilinear form $\bu^\top(\bL/\bLambda)^{-\alpha}\bhv_k\bhv_k^\top(\bL/\bLambda)^{-\alpha}\bv$ for arbitrary deterministic unit vectors $\bu,\bv\in\R^n$ in terms of resolvent $\bG(z)$, enabling us to estimate the projection $\bu^\top(\bL/\bLambda)^{-\alpha}\bhv_k$. To deduce the asymptotic expansion of $\bu^\top(\bL/\bLambda)^{-\alpha}\bhv_k$, we replace all occurrences of $\bG(z)$ in \eqref{eq:tech inno bilinear form} with $\btUpsilon(z)$, which provides the relationship between the projection and its asymptotic limit $\bu^\top(\bL/\bLambda)^{-\alpha}\bv_k$.

To present the detailed formulas for the asymptotic expansions and error bounds, we examine the differences between \eqref{eq:tech inno eigenvalue} and \eqref{eq:tech inno tk}, as well as the error introduced when replacing $\bG(z)$ with $\btUpsilon(z)$ in \eqref{eq:tech inno bilinear form}. To determine the leading terms and the order of error terms, we need to characterize the asymptotic behavior of $\bu^\top(\bG(z)-\btUpsilon(z))\bv$ for some deterministic unit vectors $\bu$ and $\bv$. We expect to get some estimates of the form
\[
    |\bu^\top(\bG(z)-\btUpsilon(z))\bv|\leq\epsilon_n(z,\bu,\bv),
\]
where $\epsilon_n>0$ is a sequence of deterministic error control parameters and small enough compared to the leading terms as random matrix size $n$ increases. Such estimates are referred to as the \textit{anisotropic local laws} in the RMT literature; see, e.g., \cite{alex2014isotropic,knowles2013isotropic,knowles2017anisotropic}. In our context, the required local laws are stated in Section \ref{sec:prelim} as Theorems \ref{prop:ekG-UpsilonvLinfty}--\ref{prop:eiLWL(G-Upsilon)v}. These theorems provide the necessary tools to establish the high-order asymptotic expansions and error bounds for the empirical spiked eigenvalues, and the components and projections of the empirical spiked eigenvectors.

To prove the local laws in our paper, we first utilize the local laws of the intermediate matrix $\bR(z)$, which are established in \cite{SIMPLERC} for the specific case of $\alpha=0$ and $\btX$ is the adjacency matrix of a random graph. Combining the methods in \cite{SIMPLERC,erdHos2013spectral}, we obtain the corresponding local laws of $\bR$ under our more general setting of $\btX$, which are summarized in Theorem \ref{thm:local laws of R} (see Section \ref{Preliminary Estimates} of the Supplementary Material). Then we can derive the local laws of $\bG$ from those of $\bR$ by controlling the difference $\bG-\bR$. However, the presence of correlations between random matrices $\btX$ and $\bL$ poses a \textit{significant challenge} in extending the local laws of $\bR(z)$ to those of $\bG(z)$, particularly for Theorem \ref{thm:main thm}, which provides an estimate for $\be_i^\top\bG(z)\bv$. To overcome such a challenge, we define the resolvent 
\begin{equation}\label{eq:G[i] definition unrescaled}
	\bG_{[i]}(z)=\left(\bcW-z(\bL_{[i]}/\bLambda)^{2\alpha}\right)^{-1},
\end{equation}
where $\bL_{[i]}$ with $1 \leq i \leq n$ is a random diagonal matrix with diagonal entries $(L_{[i]})_i=L_i$ and 
\begin{equation}\label{eq:L[i] definition unrescaled}
	(L_{[i]})_j=\Lambda_j + \sum_{1 \leq s \leq n}^{(i)}W_{js}+\frac{\tau_j}{n}\sum_{1 \leq t,s \leq n}^{(i)}W_{ts} 
\end{equation}
for all $j\neq i$. We further obtain the local law for $\mathbf e_i^\top \bG \bv$ by first obtaining the corresponding local law for $\mathbf e_i^\top \bG_{[i]} \bv$, and then controlling the difference between $\bG$ and $\bG_{[i]}$. The main motivation for this approach is the observation that when we deal with the $i$th row and column of $\bG$, most correlations between $\bL$ and $\btX$ come from the entries in the $i$th row and column of $\btX$. Hence, by introducing $\bL_{[i]}$ we can reduce its correlation with $\btX$ greatly, which allows us to prove a sufficiently accurate local law for $\mathbf e_i^\top \bG_{[i]}\bv$. On the other hand, the difference between $\bL$ and $\bL_{[i]}$ is very small. This is because we have removed only a single entry from $\btX$ in each entry of $\bL_{[i]}$, leading to an asymptotically negligible difference. As a result, the difference between $\bG_{[i]}$ and $\bG$ is also asymptotically negligible. The details of this technical argument can be found in Lemmas \ref{lemma:L[i]} and \ref{lemma:G-G[i]} (see Section \ref{Local laws} of the Supplementary Material).

One of the major challenges in our paper is the \textit{insufficiency} of low-order expansions for the asymptotic expansions of the empirical spiked eigenvalue and the projection of the empirical spiked eigenvector $\bu^\top(\bL/\bLambda)^{-\al}\bhv_k$ for general deterministic vector $\bu \in \mathbb{R}^n$. During the manipulation of expressions in \eqref{eq:tech inno eigenvalue} and \eqref{eq:tech inno bilinear form}, we investigate $\bG(z)$ through series expansion
\begin{equation}\label{eq:tech inno series expansion}
    \bG(z)=\left({\bcW}-z(\bL/\bLambda)^{2\alpha}\right)^{-1}=-(\bL/\bLambda)^{-2\al}\sum_{l=0}^\infty z^{-(l+1)}\left(\bcW(\bL/\bLambda)^{-2\al}\right)^l.
\end{equation}
Upon detailed calculations, it is found that to derive the required CLTs under the extra assumption
\begin{equation}
    \|\bv_k\|_\infty\to 0,\quad \sqrt n\ll q^2,\quad \sqrt n\frac{\sdeg^{4\al}}{q^{2-8\al}}\ll |\delta_k|^2,
\end{equation}
we need to truncate the series expansion \eqref{eq:tech inno series expansion} at $l=3$. The inclusion of \textit{higher-order terms} in the expansion allows us to obtain the formulas presented in Theorems \ref{thm:eigenvalue2} and \ref{thm:projection2}. These formulas provide accurate enough approximations to ensure the validity of the CLTs stated in Corollaries \ref{cor:CLT_evalue} and \ref{cor:CLT_evector}. 

Furthermore, through the higher-order asymptotic expansions, we have also confirmed the interesting \textit{phase transition phenomenon} discussed in \cite{fan2020asymptotic}, where the limiting distribution of the projection $\bu^\top(\bL/\bLambda)^{-\alpha}\bhv_k$ depends on the proximity of the deterministic unit vector $\bu \in \mathbb{R}^n$ to $\bv_k$ (modulo the sign). Qualitatively speaking, if we denote the angle between $\bu$ and $\bv_k$ as $\gamma$, and the angle between $\bv_k$ and $\bhv_k$ as $\Delta\gamma$, then from the Taylor expansion we have that 
\[
    \bu^\top\bhv_k=\cos(\gamma+\Delta\gamma)=\bu^\top\bv_k-\sin(\gamma)\Delta\gamma-\frac{1}{2}\cos(\gamma)(\Delta\gamma)^2+O\left((\Delta\gamma)^3\right).
\]
When $\bu$ is far away from $\bv_k$, the leading term in the representation above is $\sin(\gamma)\Delta\gamma$, which yields an order-$1$ variation, as stated in part 1) of Theorem \ref{thm:projection2}. On the other hand, when $\bu$ is close to $\bv_k$, the order-$1$ term vanishes and the leading term becomes of order-$2$, as stated in part 2) of Theorem \ref{thm:projection2}. Such phase transition phenomenon provides valuable insights into the asymptotic behavior of the projection of the empirical spiked eigenvector in different regimes, and our higher-order asymptotic expansion confirms and quantifies this phenomenon.

We also want to highlight an interesting observation regarding the results of the empirical spiked eigenvectors. Instead of directly considering eigenvector $\bhv_k$ of random matrix $\bX=\bL^{-\al}\btX\bL^{-\al}$, we find it cleaner and more manageable to work with vector $\bL^{-\alpha}\bhv_k$ instead. This is equivalent to dealing with the eigenvector of random matrix $\bL^{-2\al}\btX$, which can be obtained by transforming $\bX$ using $\bL^{\al}$. Such consideration explains why we define the resolvent $\bG$ as in \eqref{eq:defGR unrescaled} instead of $(\bL^{-\al}\bW\bL^{-\al}-z)^{-1}$, giving rise to the appearance of term $\bu^\top(\bL/\bLambda)^{-\al}\bhv_k$ in Theorem \ref{thm:projection2}. This same approach can also be applied in other similar random matrix models, such as the sample correlation matrices.

\section{Applications of ATE-GL}
\label{new.Sec.appl}

In this section, we briefly discuss four applications of our newly established theoretical framework ATE-GL: 1) graph neural networks, 2) pure node confidence intervals in network inference, 3) confidence intervals for network parameters, and 4) uncertainty quantification for network community detection.

\subsection{Graph neural networks} \label{new.Sec.appl.GNN}

One natural application of ATE-GL is related to the idea of graph embedding in graph neural networks (GNNs). The key ingredient of GNN is to utilize the underlying graphical structure in the data (e.g., neighboring pixels in images and word patterns in text sequences) to improve the predictive power of deep neural networks (DNN). As reviewed in the survey paper \cite{GNN-2019}, an important class of graph neural networks is the spectral-based graph convolutional neural network, which we briefly review below. Considering the case of $\alpha=1/2$, matrix $\bI - \bX$ is the normalized Laplacian matrix. It is seen that the leading  $K$ eigenvalues and the associated eigenvectors of $\bX$ correspond to the smallest $K$ eigenvalues and the associated eigenvectors of the Laplacian matrix $\bI - \bX$. 

Let $\bF\in \mathbb R^{n\times d}$ be the feature matrix corresponding to the $n$ nodes in the network. Motivated by the classical convolutional neural network (CNN), \cite{GNN-spectral} proposed to construct the spectral convolutional layer that takes $\bF$ as the input and outputs a feature matrix $\bF_{\text{out}}$ of size $n\times d_{\text{out}}$ 
\begin{align}
    \bF_{\text{out}}(:,j) = \sigma\left(\sum_{i=1}^{d_{\text{out}}}\bV\text{diag}(\btheta_{i,j})\bV\bF(:,i) \right),
\end{align}
where $\btheta_{i,j}\in \mathbb R^{n}$ denotes a vector of network weights that can be learned in training the neural network, $\sigma(\cdot)$ is the activation function, $\bV$ is the eigenvector matrix of $\bX$, and $\bM(:,j)$ represents the $j$th column of a generic matrix $\bM$. When $n$ is large, calculating the eigenvector matrix $\bV$ of the normalized Laplacian matrix is computationally expensive. To overcome such difficulty, \cite{GNN-spectral} proposed to replace $\bV$ with its submatrix corresponding to the smallest $K_0$ eigenvalues of $\bI-\bX$. When the adjacency matrix $\widetilde{\bX}$ has the low-rank structure with rank $K$ as considered in this paper, the ideal choice is to choose $K_0=K$. This reduces to inferring the value of low rank $K$. 

\cite{han2023universal} recently proposed a universal test for testing and estimating the low rank $K$ of the adjacency matrix $\widetilde{\bX}$. The main idea of their test is to first estimate and remove the low rank structure in the adjacency matrix under the null hypothesis $H_0: K=K_0$. If $K_0$ is indeed the true value for the rank, the residual matrix should not exhibit any low-rank structure and be close to a centered Wigner matrix with independent entries modulo the symmetry. Then by subsampling the residual matrix entries, a test statistic can be constructed as the summation of the subsampled entries with self-normalization. They proved that under $H_0$, the test statistic is asymptotically standard normal, and the power depends on the signal strength, which can be roughly measured by the magnitude of the leading eigenvalues.  

Despite the generality and robustness properties of their test, their method works only under the assumption of mild degree heterogeneity. Yet, in practical applications, network data often exhibits  severe degree heterogeneity. It has been justified in the literature that the Laplacian matrix can help accommodate more severe degree heterogeneity. Motivated by this, we can replace the residual matrix in the test constructed in \cite{han2023universal} with $\widehat \bW$ defined at the end of Section \ref{new.sec.mainresu}, and construct a similar test for inferring the true rank $K$. The established theory in this paper can help establish the asymptotic null distribution. The same sequential testing procedure can also be exploited here to estimate the true rank. It is worth mentioning that this method is widely applicable to all data sets with the low-rank plus noise structure, and the application is much broader than the graph neural networks discussed in this subsection.

\subsection{Pure node confidence intervals in network inference} \label{new.Sec.appl.purenode}

Let us consider the DCMM model given in 
\eqref{intro_DCMM}. \cite{ke2024optimal} proposed to estimate the node memberships using the method of mixed-SCORE-Laplacian, which is a spectral method based on the generalized Laplacian matrix $\bX$ in \eqref{new.eq.FL.gLap}. Using our eigenvalue and eigenvector expansion results, we will discuss how to construct confidence intervals (CIs) for the pure nodes. For the completeness of the presentation, let us review how mixed-SCORE-Laplacian estimates the pure nodes in the noiseless scenario where $\widetilde{\bX} = \bTheta\bPi\bP\bPi^T\bTheta$. Define $\mathbf R(i,k) = v(i,k+1)/v(i,1)$ for each $1 \leq i \leq n$ and $1 \leq k \leq K-1$, and let $\br_1,\cdots, \br_n\in \mathbb R^{K-1}$ be the rows of $\mathbf R = (\mathbf R(i,k))$. In this case, it was shown in \cite{ke2024optimal} that vectors $\br_1,\cdots,\br_n$ fall on a $(K-1)$-dimensional simplex, where pure nodes have $\br_i$'s falling on the vertices. This suggests that the node with the largest $\|\br_i\|_2$ is a pure node. In particular, identifying pure nodes plays a key role in clustering for network data. 

In the noisy case of $\widetilde{\bX} = \bTheta\bPi\bP\bPi^T\bTheta + \bW$, vectors $\br_1,\cdots, \br_n$ can be estimated by using the empirical eigenvectors of $\bX$. Denote by $\hat \br_1,\cdots, \hat\br_n$ the corresponding noisy versions of $\br_1,\cdots,\br_n$. Naturally, we can estimate $\max_{1 \leq i \leq n} \|\br_i\|_2$ as $\max_{1 \leq i \leq n} \|\hat\br_i\|_2$. Thanks to the entrywise eigenvector expansion for $\hat\bv_i$'s, we can obtain the entrywise  expansion for $\|\hat\br_i\|_2^2$, based on which we can derive the asymptotic distribution of 
\begin{equation}\label{eq_teststat}
\max_{1 \leq i \leq n} \|\hat\br_i\|_2^2 - \max_{1 \leq i \leq n} \|\br_i\|_2^2,
\end{equation}
and hence derive a $(1-\alpha)$-CI for $\max_{1 \leq i \leq n} \|\br_i\|_2^2$. 
More precisely, given the asymptotic expansion \eqref{eq:main thm unrescaled}, the correlations between different entries $\hv_k(i)$ are precisely quantified. Drawing insights from \cite{fan2022simple,SIMPLERC}, we expect that the asymptotic distribution of the test statistic \eqref{eq_teststat} would be given by the maximum of several (asymptotically) independent $\chi^2$ distributions. This observation can provide a concrete expression for the CI. Then, nodes with $\|\br_i\|_2^2$ falling into the CI can be the candidate estimates for pure nodes.

\subsection{Confidence intervals for network parameters} \label{new.Sec.appl.netparaCI}

Recently, \cite{ke2024optimal} and \cite{jiang2024optimalestimationparametersdegree} proposed methods that can achieve the optimal estimation of various parameters in the DCMM model, where the former concerns the estimation of community membership matrix $\bPi$ and the latter concerns the estimation of the degree matrix $\bTheta$ and connectivity matrix $\bP$. Both methods are built on the constructed $\{\hat\br_i\}_{i=1}^n$ as reviewed in the last section. We first briefly review their proposed estimation methods and then discuss how our entrywise eigenvector and eigenvalue expansions can be exploited to construct confidence intervals for these estimated parameters. 

To simplify the presentation and gain better intuition, let us use the noiseless case of $\widetilde{\bX} = \bTheta\bPi\bP\bPi^T\bTheta$ to present the estimation idea. With the constructed vectors $\br_1,\cdots, \br_n$, a vertex hunting algorithm can be applied to estimate the simplex vertices, denoted as $\br_1^*,\cdots, \br_{K}^*$. The pure nodes will fall on one of the vertices, and each node will have a barycentric coordinate in the simplex with respect to the vertices. Denote by $\bw_i$ the barycentric coordinate for node $i$. It was shown in \cite{ke2024optimal} that there is an explicit relationship between $\bpi_i$ and $\bw_i$; that is,
$$
\bpi_i\propto [\text{diag}(\mathbf b_1)]^{-1}\bw_i $$
with $\mathbf b_1 \in \mathbb R^{K}$ and $\mathbf b_1(k) = [\lambda_1 + (\br^*_k)^T\text{diag}(\lambda_2,\cdots,\lambda_K)\br^*_k]^{-1/2}$ for each $1 \leq k \leq K$. Then the membership profile vector $\bpi_i$ can be estimated by normalizing $[\text{diag}(\mathbf b_1)]^{-1}\bw_i$ to have unit $L_1$-norm. 
Let $\bQ \in \mathbb R^{K\times K}$ be a matrix with the $k$th row being $(1, (\br^*_k)^T)$. \cite{jiang2024optimalestimationparametersdegree} proposed to estimate the connectivity matrix as $\bP = \mathbf b_1^T\bQ\bLambda\bQ^T\mathbf b_1$ and the degree matrix as $\bTheta(i,i)=\bv_1(i)\Lambda(i,i)^{1/2}(\bpi_1^T\mathbf b_1)^{-1}$.

When the adjacency matrix is observed with noise $\bW$, the population eigenvalues and eigenvectors are replaced with their empirical counterparts. It is seen that these estimates can all be written as functions of the eigenvalues and eigenvectors of the Laplacian matrix, thanks to which our entrywise expansions of eigenvalues and eigenvectors can be applied to construct confidence intervals for these network parameter estimates. As such, our theoretical framework ATE-GL enables various network inference tasks.

\subsection{Uncertainty quantification for network community detection} \label{new.Sec.appl.commdetUQ}

\cite{fan2022simple,
SIMPLERC} and \cite{bhattacharya2023inferences} studied the problem of testing a group of nodes under the DCMM sharing similar membership profiles, that is, their corresponding $\bpi_i$'s are close to each other. An important assumption in their study is that the degree heterogeneity should be mild, owing to the fact that their test statistics were constructed using the eigenvalues and eigenvectors of the adjacency matrix instead of the Laplacian matrix. Under severe degree heterogeneity, a new test will be needed for assessing the statistical uncertainty in the network clustering problem.   

It was shown in \cite{ke2024optimal} that under the DCMM, if two nodes $i$ and $j$ have the same membership profile, then their embedding locations on the simplex are also the same; that is, if $\bpi_i = \bpi_j$, it holds that $\br_i = \br_j$. Motivated by such observation, a new test for testing a group of modes in $\mathcal M \subset \{(i,j): i\neq j, \, 1 \leq i, j \leq n\}$ sharing similar membership profiles can be constructed based on the eigenvectors of the Laplacian matrix $\bX$.  

\section{Simulation study}
\label{new.Sec.simu}

In this section, we conduct a simulation study to verify the asymptotic distributions of the empirical spiked eigenvalues and spiked eigenvectors for the generalized Laplacian matrices built in Section \ref{sec:mainresult}.

\subsection{Simulation settings} \label{new.Sec.simu.setting}

Let us introduce the simulation design for the generalized (regularized) Laplacian matrix $\bX$. We first borrow the network setting of simulation example 1 in \cite{SIMPLERC} to generate the $n\times n$ symmetric random matrix $\btX$ with independent entries modulo the symmetry given in (\ref{eq:model}). Such setting considers the frequently used mixed membership (MM) model for the random adjacency matrix and was adopted in \cite{SIMPLERC} for the inference problem of group network testing under non-sharp nulls and weak signals; see Section 5 therein for more details. In particular, following \cite{SIMPLERC} we consider a network with size $n = 3000$ and $K = 5$ communities, where each community contains $n_0 = 300$ pure nodes. Note that each pure node in the $k$th community with $1 \leq k \leq K$ has a community membership probability vector $\bpi$ that is the $k$th basis vector $\be_k \in \mathbb{R}^K$. The remaining $n - K n_0$ nodes are divided into four groups of equal size. As in \cite{SIMPLERC}, we define the community membership probability vector $\bpi$ as $\ba_l$ for each mixed (i.e., non-pure) node from the $l$th group with $1 \leq l \leq 4$, where $\ba_1 = (0.1, 0.6, 0.1, 0.1, 0.1)^T$, $\ba_2 = (0.6, 0.1, 0.1, 0.1, 0.1)^T$, $\ba_3 = (0.1, 0.1, 0.6, 0.1, 0.1)^T$, and $\ba_4 = (1/K, \cdots, 1/K)^T$. We have fully specified the $n \times K$ matrix of community membership probability vectors $\bPi$ associated with the MM model; see (8) in \cite{SIMPLERC} for details. 

It remains to define the kernel matrix $\bP$ associated with the MM model (see (8) in \cite{SIMPLERC}), which is a $K \times K$ nonsingular matrix. Let the diagonal entries of $\bP$ be one and its $(j,k)$th entry $\rho/|j - k|$ for each $1 \leq j \neq k \leq K$ with $\rho=0.2$. We finally introduce the sparsity parameter $\theta$ for the mean matrix $\bH = \theta \bPi \bP \bPi^T$ in (\ref{eq:model}) and allow $\theta$ to vary in $\{0.1, 0.5, 0.9\}$, where a smaller value of $\theta$ represents a lower average node degree and consequently weaker signal strength. This completes the specification on the $n\times n$ symmetric random matrix $\btX$ given in (\ref{eq:model}).

We next define the generalized (regularized) Laplacian matrix $\bX=\bL^{-\alpha}\btX\bL^{-\alpha}$ given in (\ref{new.eq.FL.gLap}), where $\bL=\bL_{\tau,\lambda} = 
\diag\left(d_i+\tau_i\bar d+\lambda_i: i \in [n]\right)$ is a diagonal matrix given in (\ref{eq:defbL}) without the rescaling population parameters $q$ and $\sdeg$. For simplicity, we choose a pair of common regularization parameters $(\tau_i, \lambda_i) = (\tau, \lambda) = 10^{-4}$ with $1 \leq i \leq n$, ensuring that matrix $\bL$ is nonsingular almost surely. To investigate the finite-sample performance of the empirical spiked eigenvalues $\hdelta_k$'s and spiked eigenvectors $\bhv_k$'s of the generalized Laplacian matrix $\bX$, we generate $500$ data sets for each setting of $(\alpha, \theta)$, with $\alpha$ varying in $\{0.25, 1/2, 1, 2\}$ and $\theta$ varying in $\{0.1, 0.5, 0.9\}$.

\subsection{Simulation results} \label{new.Sec.simu.results}

\begin{figure}[tp]
\centering
\includegraphics[width=0.90\linewidth]{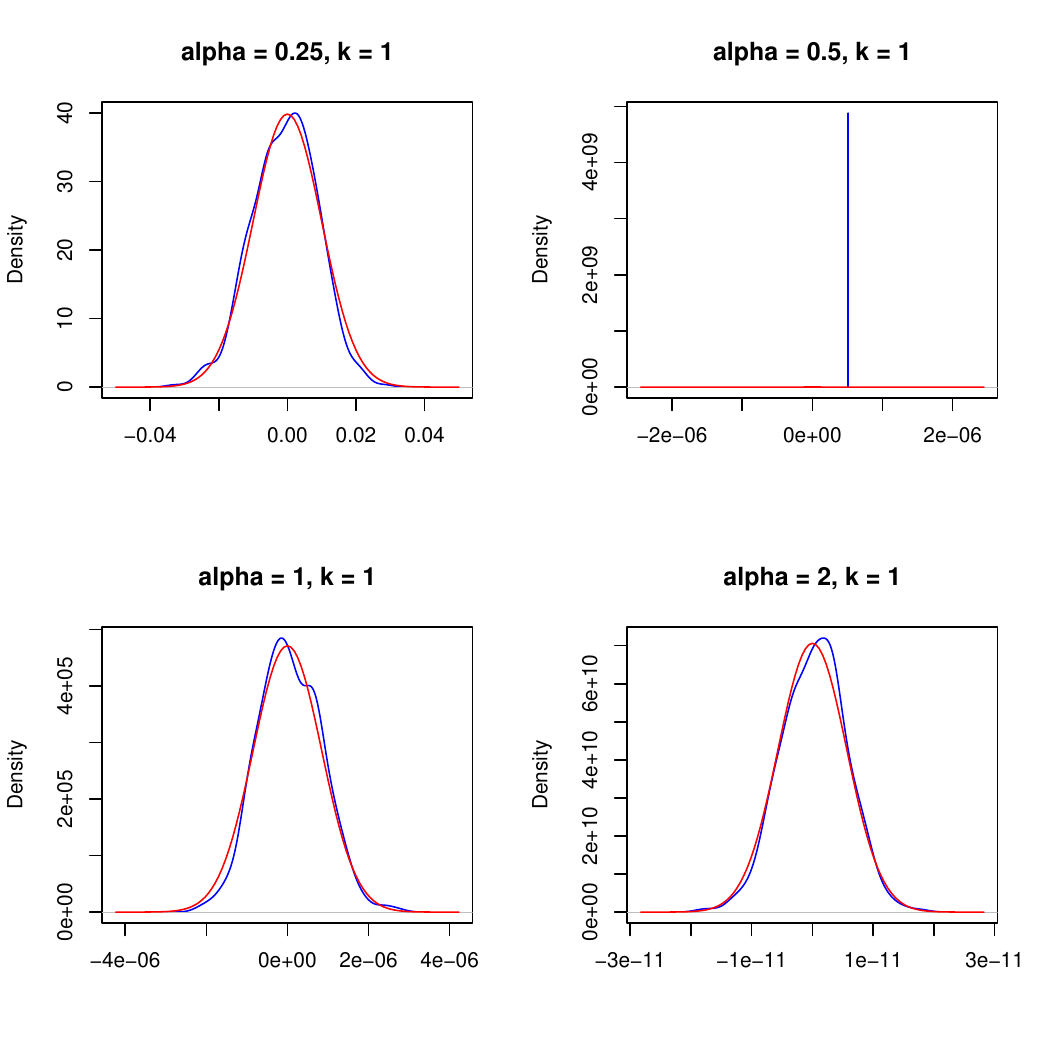}
\caption{The kernel density estimate (KDE) for the distribution of the empirical spiked eigenvalue $\hdelta_k$ corrected by $A_k$ for the generalized Laplacian matrix $\bX$  with $k = 1$ across different values of $\alpha$ based on $500$ replications for simulation example in Section \ref{new.Sec.simu} with $\theta = 0.9$. The  generalized (regularized) Laplacian matrix $\bX$ is as given in (\ref{new.eq.FL.gLap}) with $\bL=\bL_{\tau,\lambda} := 
\diag\left(d_i+\tau\bar d+\lambda: i \in [n]\right)$ without the rescaling population parameters $q$ and $\sdeg$. The blue curves represent the KDEs for the empirical spiked eigenvalue corrected by $A_k$, whereas the red curves stand for the target normal density. Both curves are centered with the asymptotic limit $t_k$. The top right plot is due to extremely small empirical standard deviations (as shown in Table \ref{tab1} with empirical SD = 7.75E-11 and asymptotic SD = 4.89E-07). This is associated with the fact that the normalized Laplacian matrix has a trivial largest eigenvalue at 1.
}
\label{fig1}
\end{figure}

\begin{figure}[h]
\centering
\includegraphics[width=0.90\linewidth]{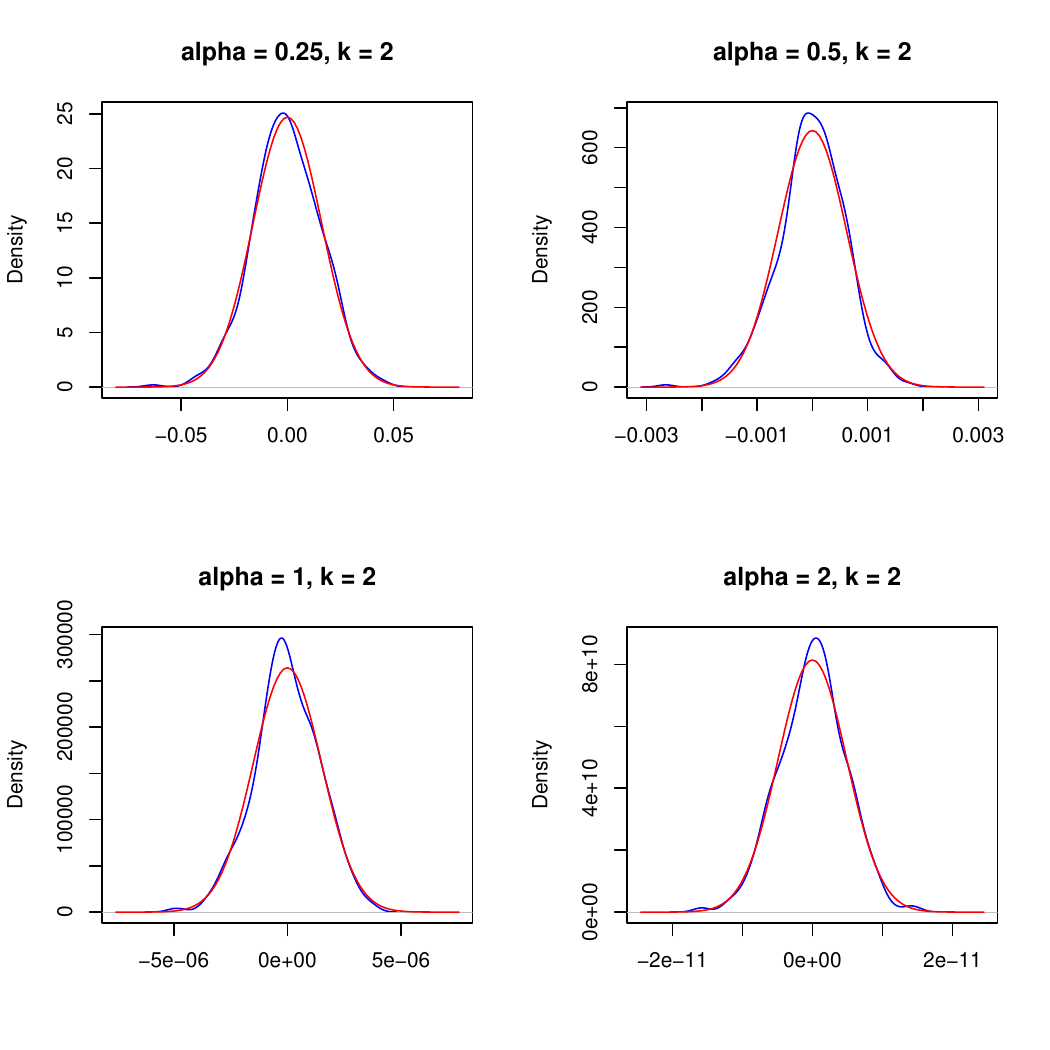}
\caption{The kernel density estimate (KDE) for the distribution of the empirical spiked eigenvalue $\hdelta_k$ corrected by $A_k$ for the generalized Laplacian matrix $\bX$  with $k = 2$ across different values of $\alpha$ based on $500$ replications for simulation example in Section \ref{new.Sec.simu} with $\theta = 0.9$. The  generalized (regularized) Laplacian matrix $\bX$ is as given in (\ref{new.eq.FL.gLap}) with $\bL=\bL_{\tau,\lambda} := 
\diag\left(d_i+\tau\bar d+\lambda: i \in [n]\right)$ without the rescaling population parameters $q$ and $\sdeg$. The blue curves represent the KDEs for the empirical spiked eigenvalue corrected by $A_k$, whereas the red curves stand for the target normal density. Both curves are centered with the asymptotic limit $t_k$.}
\label{fig2}
\end{figure}

\begin{figure}[h]
\centering
\includegraphics[width=0.90\linewidth]{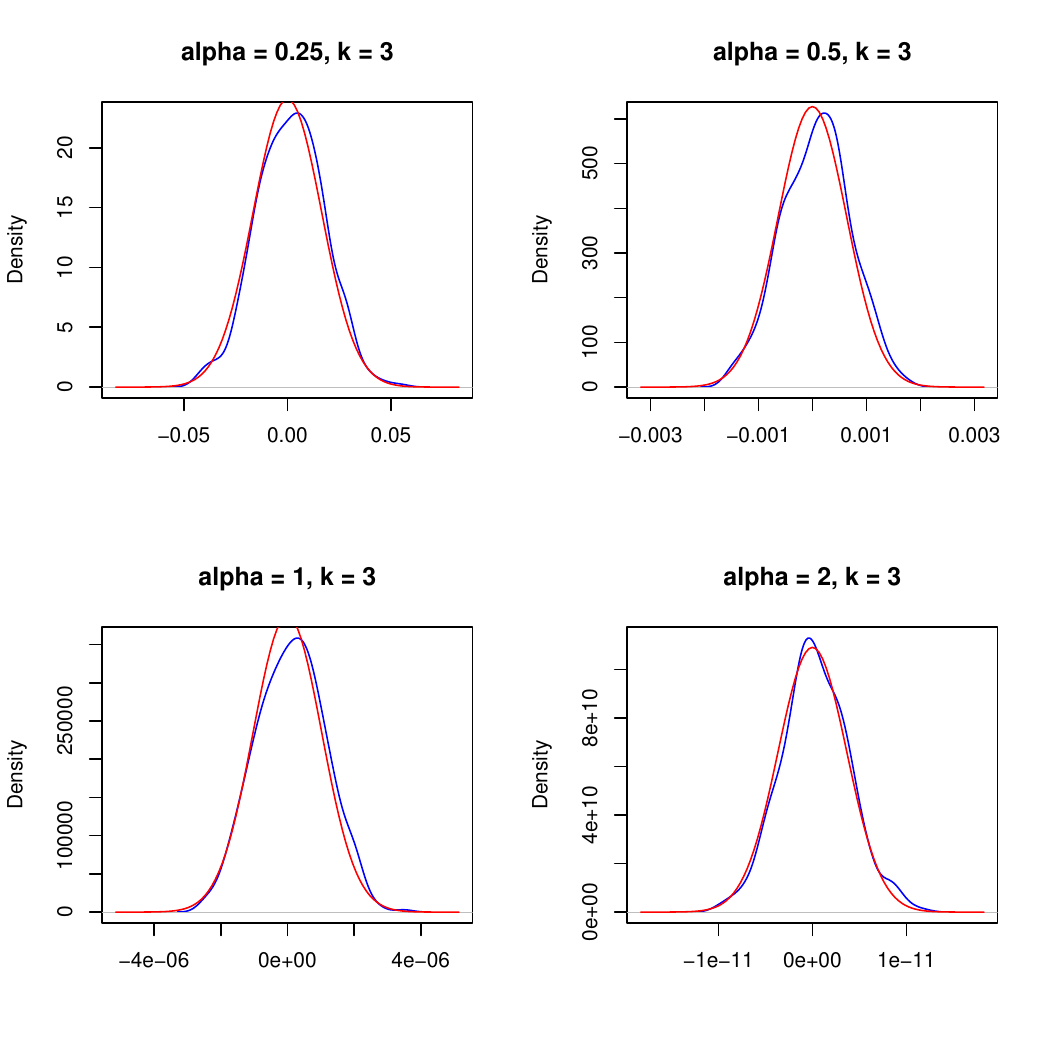}
\caption{The kernel density estimate (KDE) for the distribution of the empirical spiked eigenvalue $\hdelta_k$ corrected by $A_k$ for the generalized Laplacian matrix $\bX$  with $k = 3$ across different values of $\alpha$ based on $500$ replications for simulation example in Section \ref{new.Sec.simu} with $\theta = 0.9$. The  generalized (regularized) Laplacian matrix $\bX$ is as given in (\ref{new.eq.FL.gLap}) with $\bL=\bL_{\tau,\lambda} := 
\diag\left(d_i+\tau\bar d+\lambda: i \in [n]\right)$ without the rescaling population parameters $q$ and $\sdeg$. The blue curves represent the KDEs for the empirical spiked eigenvalue corrected by $A_k$, whereas the red curves stand for the target normal density. Both curves are centered with the asymptotic limit $t_k$.}
\label{fig3}
\end{figure}

\begin{figure}[h]
\centering
\includegraphics[width=0.90\linewidth]{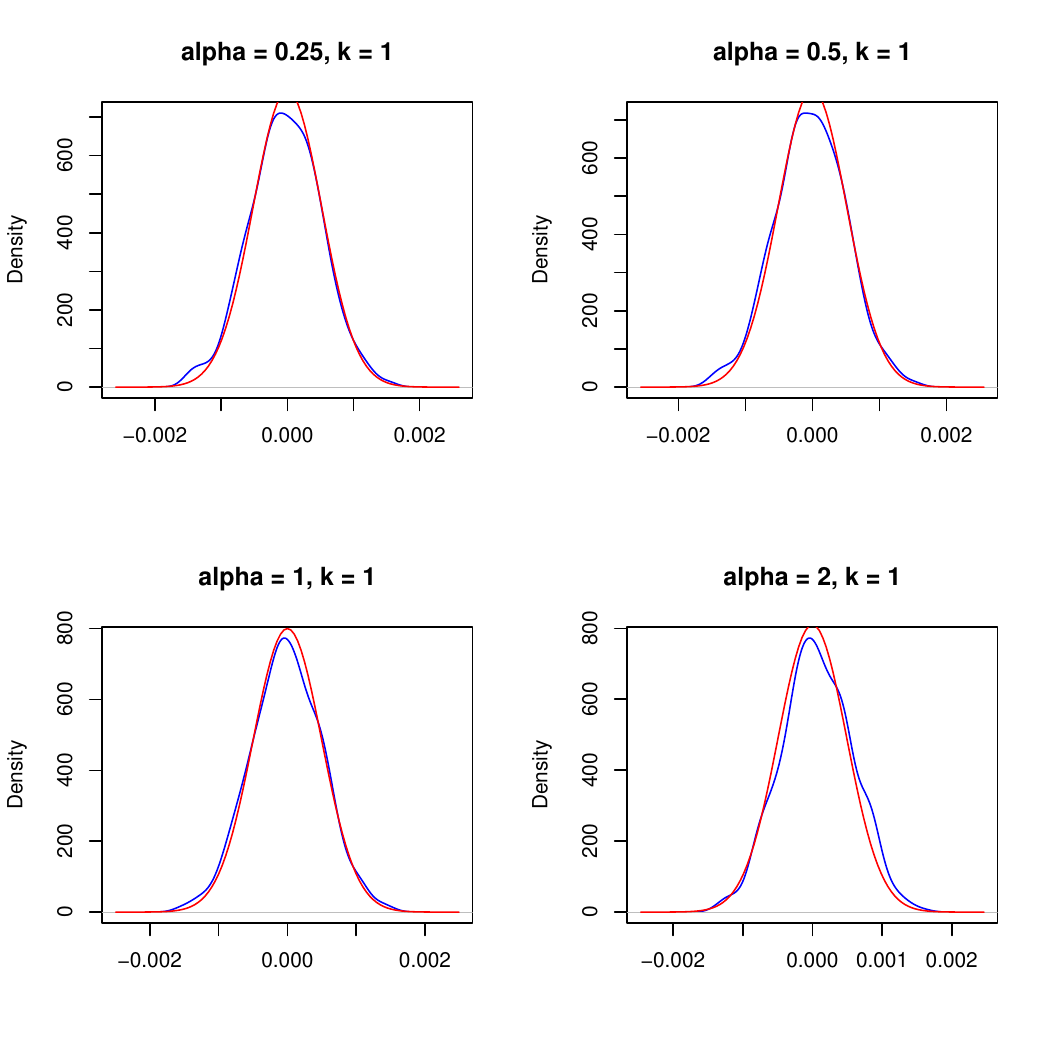}
\caption{The kernel density estimate (KDE) for the distribution of the empirical spiked eigenvector component $\bhv_k(i)$ (rescaled by $L_i^\al/\La_i^\al$) for the generalized Laplacian matrix $\bX$  with $k = 1$ and $i = 1$ across different values of $\alpha$ based on $500$ replications for simulation example in Section \ref{new.Sec.simu} with $\theta = 0.9$. The  generalized (regularized) Laplacian matrix $\bX$ is as given in (\ref{new.eq.FL.gLap}) with $\bL=\bL_{\tau,\lambda} := 
\diag\left(d_i+\tau\bar d+\lambda: i \in [n]\right)$ without the rescaling population parameters $q$ and $\sdeg$. The blue curves represent the KDEs for the rescaled empirical spiked eigenvector component, whereas the red curves stand for the target normal density. Both curves are centered with the asymptotic limit $\bv_k(i)$.}
\label{fig4}
\end{figure}

\begin{figure}[h]
\centering
\includegraphics[width=0.90\linewidth]{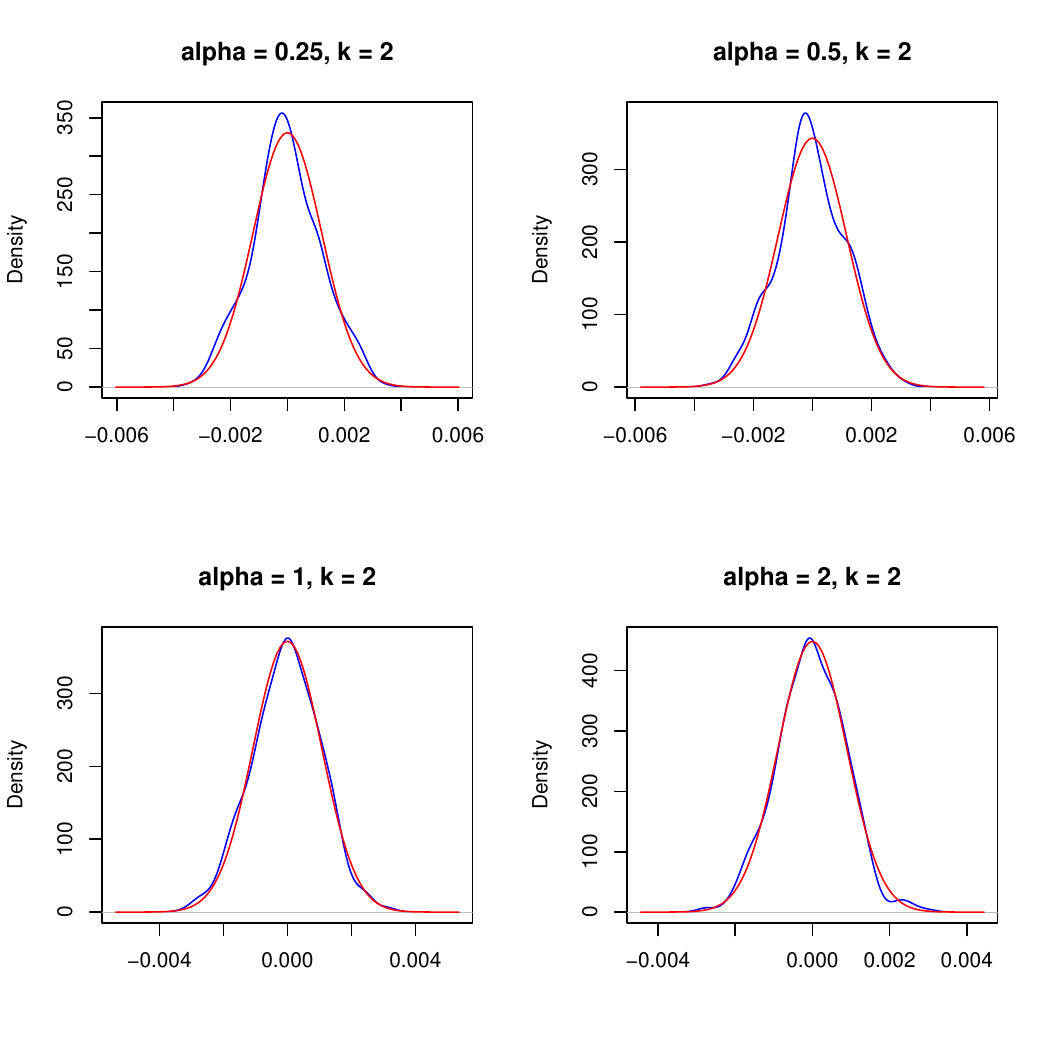}
\caption{The kernel density estimate (KDE) for the distribution of the empirical spiked eigenvector component $\bhv_k(i)$ (rescaled by $L_i^\al/\La_i^\al$) for the generalized Laplacian matrix $\bX$  with $k = 2$ and $i = 1$ across different values of $\alpha$ based on $500$ replications for simulation example in Section \ref{new.Sec.simu} with $\theta = 0.9$. The  generalized (regularized) Laplacian matrix $\bX$ is as given in (\ref{new.eq.FL.gLap}) with $\bL=\bL_{\tau,\lambda} := 
\diag\left(d_i+\tau\bar d+\lambda: i \in [n]\right)$ without the rescaling population parameters $q$ and $\sdeg$. The blue curves represent the KDEs for the rescaled empirical spiked eigenvector component, whereas the red curves stand for the target normal density. Both curves are centered with the asymptotic limit $\bv_k(i)$.}
\label{fig5}
\end{figure}

\begin{figure}[h]
\centering
\includegraphics[width=0.90\linewidth]{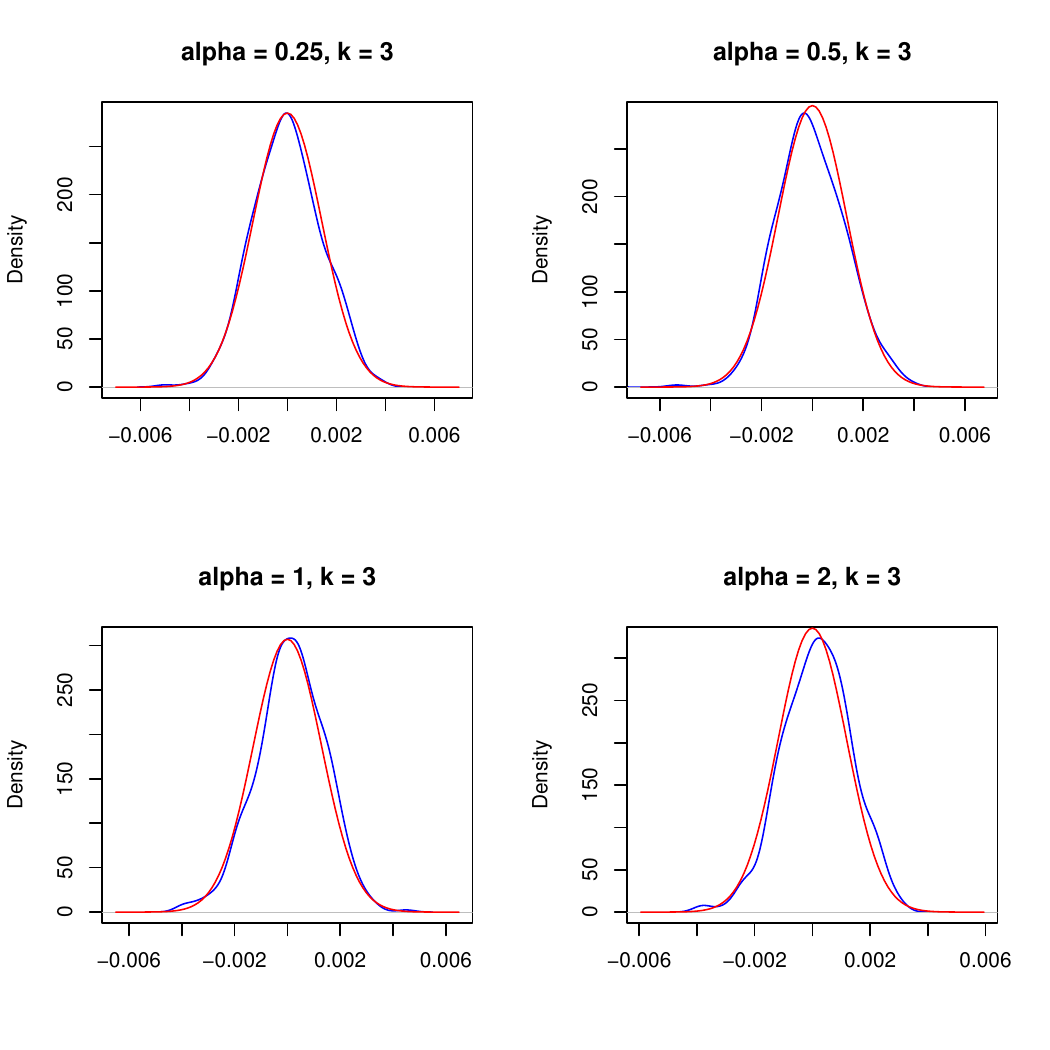}
\caption{The kernel density estimate (KDE) for the distribution of the empirical spiked eigenvector component $\bhv_k(i)$ (rescaled by $L_i^\al/\La_i^\al$) for the generalized Laplacian matrix $\bX$  with $k = 3$ and $i = 1$ across different values of $\alpha$ based on $500$ replications for simulation example in Section \ref{new.Sec.simu} with $\theta = 0.9$. The  generalized (regularized) Laplacian matrix $\bX$ is as given in (\ref{new.eq.FL.gLap}) with $\bL=\bL_{\tau,\lambda} := 
\diag\left(d_i+\tau\bar d+\lambda: i \in [n]\right)$ without the rescaling population parameters $q$ and $\sdeg$. The blue curves represent the KDEs for the rescaled empirical spiked eigenvector component, whereas the red curves stand for the target normal density. Both curves are centered with the asymptotic limit $\bv_k(i)$.}
\label{fig6}
\end{figure}

The simulation results associated with the empirical spiked eigenvalues $\hdelta_k$'s and spiked eigenvectors $\bhv_k$'s of the generalized Laplacian matrix $\bX$ are summarized in Figures \ref{fig1}--\ref{fig6} and Tables \ref{tab1}--\ref{tab6}. Specifically, Figures \ref{fig1}--\ref{fig3} depict the distributions of the empirical spiked eigenvalues $\hdelta_k$'s corrected by the theoretical values $A_k$'s across different values of $\alpha$ for the representative case of $\theta = 0.9$, with $1 \leq k \leq 3$, respectively, where each distribution curve is centered by the corresponding asymptotic limit $t_k$ given in Lemma \ref{lemma:tk}. It can be seen from Figures \ref{fig1}--\ref{fig3} that the distributions of the empirical spiked eigenvalues $\hdelta_k$'s corrected by $A_k$'s are indeed close to the target asymptotic distributions established in Corollary \ref{cor:CLT_evalue}. Indeed, we observed the bias issue for the original empirical spiked eigenvalues $\hdelta_k$'s (i.e., without any bias correction) even for the case of relatively dense networks (i.e., with a larger value of $\theta$). 
We have also implemented the bias correction idea for the empirical spiked eigenvalues $\hdelta_k$'s using estimates $\widehat{A}_k$'s instead of the theoretical values $A_k$'s, with the asymptotic limit $t_k$. The results are rather similar to those in Figures \ref{fig1}--\ref{fig3}; see Figures \ref{fig1_Ahat}--\ref{fig3_Ahat} in Section \ref{add.simu.results} of the Supplementary Material for details. Similarly, we have examined the idea of correction by estimate $\widehat{A}_k$ coupled with the empirical bias correction in Section \ref{new.sec.mainresu} for the empirical spiked eigenvalues $\hdelta_k$, with the asymptotic limit $\delta_k$ instead. Such idea also works well for the empirical spiked eigenvalue $\hdelta_k$ of the generalized Laplacian matrix $\bX$ across different settings; see Figures \ref{fig1_empbc_delta}--\ref{fig3_empbc_delta} in Section \ref{add.simu.results} of the Supplementary Material for details. These simulation results showcase the advantages of \textit{both bias-correction ideas} suggested in Section \ref{new.sec.mainresu}.

Figures \ref{fig4}--\ref{fig6} display the distributions of the empirical spiked eigenvector components $\bhv_k(i)$'s (rescaled by $L_i^\al/\La_i^\al$) across different values of $\alpha$ for the representative case of $\theta = 0.9$, with $1 \leq k \leq 3$, respectively, where each distribution curve is centered by the corresponding asymptotic limit $\bv_k(i)$ given in Corollary \ref{example:CLT entry}. For simplicity, we examine only the representative scenario of $i = 1$. It is interesting to observe from Figures \ref{fig4}--\ref{fig6} that the distributions of the empirical spiked eigenvectors $\hdelta_k$'s match rather closely the target asymptotic distributions established in Corollary \ref{example:CLT entry}.

We further provide in Tables \ref{tab1}--\ref{tab3} the means and standard deviations (SDs) of the empirical spiked eigenvalues $\hdelta_k$'s in comparison to their theoretical (i.e., asymptotic) counterparts given in Corollary \ref{cor:CLT_evalue}, and in Tables \ref{tab4}--\ref{tab6} the means and standard deviations (SDs) of the empirical spiked eigenvector components $\bhv_k(i)$'s (rescaled by $L_i^\al/\La_i^\al$) in comparison to their theoretical (i.e., asymptotic) counterparts given in Corollary \ref{example:CLT entry} across different settings of $(\alpha, \theta)$. From Tables \ref{tab1}--\ref{tab6}, we can see that our asymptotic theory established in Section \ref{sec:mainresult} on the empirical spiked eigenvalues and spiked eigenvectors for the generalized Laplacian matrices is still largely valid at the finite-sample level. In particular, it can be seen that the asymptotic theory becomes more accurate (in terms of both the mean and variance) as the network sparsity parameter $\theta$ increases, which is sensible since it contributes to the signal strength in the network model. An overall message is that the new asymptotic theory for the generalized Laplacian matrix built in our work is uniformly valid across different values of index $\alpha \in (0, \infty)$, empowering their practical utilities with flexibility.

\begin{table}[h]
\centering
\caption{The means and standard deviations (SDs) of the empirical spiked eigenvalue $\hdelta_k$ corrected by $A_k$ for the generalized Laplacian matrix $\bX$ with $k = 1$ as well as their asymptotic counterparts across different settings of $(\alpha, \theta)$ based on $500$ replications for simulation example in Section \ref{new.Sec.simu}. The  generalized (regularized) Laplacian matrix $\bX$ is as given in (\ref{new.eq.FL.gLap}) with $\bL=\bL_{\tau,\lambda} := 
	\diag\left(d_i+\tau\bar d+\lambda: i \in [n]\right)$ without the rescaling population parameters $q$ and $\sdeg$.}
\smallskip
\begin{tabular}{cc|cccc}
\hline
$\alpha$ & $\theta$ & Empirical Eigenvalue & Asymptotic Eigenvalue & Empirical SD & Asymptotic SD \\
\hline
0.25	& 0.1	& 9.7617	& 9.7592	& 0.0121	& 0.0127 \\
	& 0.5	& 21.6409	& 21.6406	& 0.0118	& 0.0115 \\
	& 0.9	& 29.0056	& 29.0063	& 0.0095	& 0.0100 \\
0.5	& 0.1	& 1.0101	& 1.0100	& 2.69E-09	& 2.65E-05 \\
& 0.5	& 1.0016	& 1.0016	& 2.37E-10	& 1.76E-06 \\
& 0.9	& 1.0006	& 1.0006	& 7.75E-11	& 4.89E-07 \\
1	& 0.1	& 0.0110	& 0.0110	& 2.74E-05	& 2.94E-05 \\
	& 0.5	& 0.0022	& 0.0022	& 2.43E-06	& 2.33E-06 \\
	& 0.9	& 0.0012	& 0.0012	& 7.86E-07	& 8.46E-07 \\
2	& 0.1	& 1.45E-06	& 1.45E-06	& 1.79E-08	& 1.61E-08 \\
	& 0.5	& 1.14E-08	& 1.14E-08	& 5.50E-11	& 5.08E-11 \\
	& 0.9	& 1.96E-09	& 1.96E-09	& 5.37E-12	& 5.65E-12 \\
\hline
\end{tabular}
\label{tab1}
\end{table}

\begin{table}[h]
\centering
\caption{The means and standard deviations (SDs) of the empirical spiked eigenvalue $\hdelta_k$ corrected by $A_k$ for the generalized Laplacian matrix $\bX$ with $k = 2$ as well as their asymptotic counterparts across different settings of $(\alpha, \theta)$ based on $500$ replications for simulation example in Section \ref{new.Sec.simu}. The  generalized (regularized) Laplacian matrix $\bX$ is as given in (\ref{new.eq.FL.gLap}) with $\bL=\bL_{\tau,\lambda} := 
\diag\left(d_i+\tau\bar d+\lambda: i \in [n]\right)$ without the rescaling population parameters $q$ and $\sdeg$.}
\smallskip
\begin{tabular}{cc|cccc}
\hline
$\alpha$ & $\theta$ & Empirical Eigenvalue & Asymptotic Eigenvalue & Empirical SD & Asymptotic SD \\
\hline
0.25	& 0.1	& 4.1352	& 4.1090	& 0.0230	& 0.0233 \\
& 0.5	& 8.7895	& 8.7882	& 0.0202	& 0.0201 \\
& 0.9	& 11.7246	& 11.7248	& 0.0160	& 0.0162 \\
0.5	& 0.1	& 0.4450	& 0.4427	& 0.0024	& 0.0025 \\
& 0.5	& 0.4238	& 0.4237	& 0.0010	& 0.0010 \\
& 0.9	& 0.4213	& 0.4214	& 0.0006	& 0.0006 \\
1	& 0.1	& 0.0055	& 0.0055	& 4.59E-05	& 4.66E-05 \\
& 0.5	& 0.0010	& 0.0010	& 3.92E-06	& 3.87E-06 \\
& 0.9	& 0.0006	& 0.0006	& 1.43E-06	& 1.51E-06 \\
2	& 0.1	& 8.43E-07	& 8.33E-07	& 1.40E-08	& 1.24E-08 \\
& 0.5	& 6.39E-09	& 6.39E-09	& 4.26E-11	& 4.07E-11 \\
& 0.9	& 1.09E-09	& 1.09E-09	& 4.74E-12	& 4.90E-12 \\
\hline
\end{tabular}
\label{tab2}
\end{table}

\begin{table}[h]
\centering
\caption{The means and standard deviations (SDs) of the empirical spiked eigenvalue $\hdelta_k$ corrected by $A_k$ for the generalized Laplacian matrix $\bX$ with $k = 3$ as well as their asymptotic counterparts across different settings of $(\alpha, \theta)$ based on $500$ replications for simulation example in Section \ref{new.Sec.simu}. The  generalized (regularized) Laplacian matrix $\bX$ is as given in (\ref{new.eq.FL.gLap}) with $\bL=\bL_{\tau,\lambda} := 
\diag\left(d_i+\tau\bar d+\lambda: i \in [n]\right)$ without the rescaling population parameters $q$ and $\sdeg$.}
\smallskip
\begin{tabular}{cc|cccc}
\hline
$\alpha$ & $\theta$ & Empirical Eigenvalue & Asymptotic Eigenvalue & Empirical SD & Asymptotic SD \\
\hline
0.25	& 0.1	& 3.6481	& 3.6124	& 0.0239	& 0.0246 \\
& 0.5	& 7.6343	& 7.6327	& 0.0208	& 0.0210 \\
& 0.9	& 10.1666	& 10.1650	& 0.0163	& 0.0166 \\
0.5	& 0.1	& 0.3876	& 0.3840	& 0.0024	& 0.0025 \\
& 0.5	& 0.3631	& 0.3631	& 0.0009	& 0.0010 \\
& 0.9	& 0.3605	& 0.3604	& 0.0006	& 0.0006 \\
1	& 0.1	& 0.0044	& 0.0043	& 3.16E-05	& 3.18E-05 \\
& 0.5	& 0.0008	& 0.0008	& 2.65E-06	& 2.65E-06 \\
& 0.9	& 0.0004	& 0.0004	& 1.03E-06	& 1.03E-06 \\
2	& 0.1	& 5.57E-07	& 5.76E-07	& 1.99E-07	& 9.53E-09 \\
& 0.5	& 4.35E-09	& 4.35E-09	& 3.31E-11	& 3.05E-11 \\
& 0.9	& 7.40E-10	& 7.40E-10	& 3.65E-12	& 3.65E-12 \\
\hline
\end{tabular}
\label{tab3}
\end{table}

\begin{table}[h]
\centering
\caption{The means and standard deviations (SDs) of the empirical spiked eigenvector component $\bhv_k(i)$ (rescaled by $L_i^\al/\La_i^\al$) for the generalized Laplacian matrix $\bX$  with $k = 1$ and $i = 1$ as well as their asymptotic counterparts across different settings of $(\alpha, \theta)$ based on $500$ replications for simulation example in Section \ref{new.Sec.simu}. The  generalized (regularized) Laplacian matrix $\bX$ is as given in (\ref{new.eq.FL.gLap}) with $\bL=\bL_{\tau,\lambda} := 
\diag\left(d_i+\tau\bar d+\lambda: i \in [n]\right)$ without the rescaling population parameters $q$ and $\sdeg$.}
\smallskip
\begin{tabular}{cc|cccc}
\hline
$\alpha$ & $\theta$ & Empirical Eigenvector & Asymptotic Eigenvector & Empirical SD & Asymptotic SD \\
\hline
0.25	& 0.1	& -0.01094	& -0.01907	& 0.00115	& 0.00188 \\
& 0.5	& -0.01647	& -0.01907	& 0.00066	& 0.00078 \\
& 0.9	& -0.01910	& -0.01907	& 0.00054	& 0.00052 \\
0.5	& 0.1	& -0.00621	& -0.01867	& 0.00078	& 0.00184 \\
& 0.5	& -0.01393	& -0.01867	& 0.00059	& 0.00076 \\
& 0.9	& -0.01870	& -0.01867	& 0.00053	& 0.00051 \\
1	& 0.1	& -0.00196	& -0.01755	& 0.00068	& 0.00176 \\
& 0.5	& -0.00977	& -0.01755	& 0.00053	& 0.00073 \\
& 0.9	& -0.01756	& -0.01755	& 0.00051	& 0.00050 \\
2	& 0.1	& -0.00018	& -0.01316	& 0.00065	& 0.00163 \\
& 0.5	& -0.00405	& -0.01316	& 0.00048	& 0.00070 \\
& 0.9	& -0.01310	& -0.01316	& 0.00051	& 0.00049 \\
\hline
\end{tabular}
\label{tab4}
\end{table}

\begin{table}[h]
\centering
\caption{The means and standard deviations (SDs) of the empirical spiked eigenvector component $\bhv_k(i)$ (rescaled by $L_i^\al/\La_i^\al$) for the generalized Laplacian matrix $\bX$  with $k = 2$ and $i = 1$ as well as their asymptotic counterparts across different settings of $(\alpha, \theta)$ based on $500$ replications for simulation example in Section \ref{new.Sec.simu}. The  generalized (regularized) Laplacian matrix $\bX$ is as given in (\ref{new.eq.FL.gLap}) with $\bL=\bL_{\tau,\lambda} := 
\diag\left(d_i+\tau\bar d+\lambda: i \in [n]\right)$ without the rescaling population parameters $q$ and $\sdeg$.}
\smallskip
\begin{tabular}{cc|cccc}
\hline
$\alpha$ & $\theta$ & Empirical Eigenvector & Asymptotic Eigenvector & Empirical SD & Asymptotic SD \\
\hline
0.25	& 0.1	& -0.00285	& -0.00548	& 0.00244	& 0.00396 \\
	& 0.5	& -0.00472	& -0.00548	& 0.00154	& 0.00174 \\
	& 0.9	& -0.00552	& -0.00548	& 0.00121	& 0.00121 \\
0.5	& 0.1	& -0.00194	& -0.00633	& 0.00134	& 0.00380 \\
	& 0.5	& -0.00467	& -0.00633	& 0.00128	& 0.00167 \\
	& 0.9	& -0.00637	& -0.00633	& 0.00116	& 0.00116 \\
1	& 0.1	& -0.00082	& -0.00777	& 0.00044	& 0.00345 \\
	& 0.5	& -0.00425	& -0.00777	& 0.00088	& 0.00153 \\
	& 0.9	& -0.00780	& -0.00777	& 0.00107	& 0.00107 \\
2	& 0.1	& -0.00013	& -0.01166	& 0.00018	& 0.00285 \\
	& 0.5	& -0.00358	& -0.01166	& 0.00052	& 0.00126 \\
	& 0.9	& -0.01169	& -0.01166	& 0.00088	& 0.00089 \\
\hline
\end{tabular}
\label{tab5}
\end{table}

\begin{table}[h]
\centering
\caption{The means and standard deviations (SDs) of the empirical spiked eigenvector component $\bhv_k(i)$ (rescaled by $L_i^\al/\La_i^\al$) for the generalized Laplacian matrix $\bX$  with $k = 3$ and $i = 1$ as well as their asymptotic counterparts across different settings of $(\alpha, \theta)$ based on $500$ replications for simulation example in Section \ref{new.Sec.simu}. The  generalized (regularized) Laplacian matrix $\bX$ is as given in (\ref{new.eq.FL.gLap}) with $\bL=\bL_{\tau,\lambda} := 
\diag\left(d_i+\tau\bar d+\lambda: i \in [n]\right)$ without the rescaling population parameters $q$ and $\sdeg$.}
\smallskip
\begin{tabular}{cc|cccc}
\hline
$\alpha$ & $\theta$ & Empirical Eigenvector & Asymptotic Eigenvector & Empirical SD & Asymptotic SD \\
\hline
0.25	& 0.1	& 0.00476	& 0.00848	& 0.00300	& 0.00468 \\ 
& 0.5	& 0.00718	& 0.00848	& 0.00187	& 0.00205 \\
& 0.9	& 0.00848	& 0.00848	& 0.00140	& 0.00140 \\
0.5	& 0.1	& 0.00057	& 0.00519	& 0.00216	& 0.00432 \\
& 0.5	& 0.00295	& 0.00519	& 0.00267	& 0.00192 \\
& 0.9	& 0.00496	& 0.00519	& 0.00190	& 0.00135 \\
1	& 0.1	& -0.00002	& -0.00083	& 0.00049	& 0.00405 \\
& 0.5	& -0.00029	& -0.00083	& 0.00102	& 0.00183 \\
& 0.9	& -0.00070	& -0.00083	& 0.00130	& 0.00130 \\
2	& 0.1	& 0.00006	& 0.00459	& 0.00008	& 0.00371 \\
& 0.5	& 0.00150	& 0.00459	& 0.00054	& 0.00167 \\
& 0.9	& 0.00475	& 0.00459	& 0.00120	& 0.00119 \\
\hline
\end{tabular}
\label{tab6}
\end{table}

\clearpage


\section{Discussions} \label{new.Sec.disc}


We have investigated in this paper the problem of extending latent embeddings with Laplacian matrices for graphs and manifolds by considering the generalized Laplacian matrices, a class of random matrices containing the Laplacian matrix and the random adjacency matrix as specific cases. Such class provides us flexibility for extracting the underlying latent structures in real applications while posing nontrivial challenges on the theoretical developments due to the intrinsic dependency associated with the random matrices. We have exploited the tools of generalized quadratic vector equations and local laws to unveil the asymptotic distributions for both empirical spiked eigenvectors and eigenvalues. The suggested ATE-GL framework for latent embeddings with generalized Laplacian matrices will enable us to conduct practical, flexible inference and uncertainty quantification.

To streamline the technical analysis, we have focused on the unnormalized random matrix with independent entries modulo symmetry. It would be interesting to consider such a random matrix with dependency, which in turn leads to stronger dependency for the corresponding generalized Laplacian matrices. Also, it is of practical merit to investigate the problem of rank inference under the ATE-GL framework. Another interesting problem is the eigenvector selection for downstream applications such as clustering and local manifold representation. For specific downstream applications of our ATE-GL theoretical framework, identifying the optimal parameter $\alpha \in (0, \infty)$ deserves further studies. These problems are beyond the scope of the current paper and will be interesting topics for future research.

\bibliographystyle{chicago}
\bibliography{references}

	
\newpage
\appendix
\setcounter{page}{1}
\setcounter{section}{0}
\renewcommand{\theequation}{A.\arabic{equation}}
\setcounter{equation}{0}
	
\begin{center}{\bf \Large Supplementary Material to ``Asymptotic Theory of Eigenvectors for Latent Embeddings with Generalized Laplacian Matrices''}
		
\bigskip
		
Jianqing Fan, Yingying Fan, Jinchi Lv, Fan Yang, and Diwen Yu
\end{center}
	
\noindent This Supplementary Material contains the proofs of Theorems \ref{thm:eigenvalue}--\ref{thm:K_0} and Corollaries \ref{example:CLT entry}--\ref{cor:CLT_evector}, as well as some propositions, key lemmas, additional technical details including some refined results under the network setting, and additional simulation results. All the notation used in the Supplementary Material is the same as defined in the main body of the paper, except that some of the notation will be redefined by rescaling as stated in Section \ref{sec.mainresu.rescaledmodel}.

\section{Main results for the rescaled model} \label{sec.mainresu.rescaledmodel}

To streamline the proofs of our major theoretical results presented in the main paper, this section aims to provide a clear understanding of the relationship between some key parameters and quantities mentioned in the main text of our paper and their rescaled counterparts through suitable rescalings. It is important to keep in mind that the sparsity parameters $\theta$, $\theta_i$, and $\bar{\theta}$, the rescaling parameters $q$ and $\sdeg$, the diverging parameter $\xi$, and the regularization parameters $\tau_i$ and $\lambda_i$ are \textit{always not rescaled} throughout our paper. These parameters are given as in Section \ref{modelsetting} and Definition \ref{def: model setting}. By maintaining the original scale of these parameters, we ensure consistency and coherence in our technical analyses.

To provide a detailed exposition of the rescaled model, let us introduce three key rescaled matrices
\begin{equation}\label{eq:rescaling tX,H,W}
    \btX\to\btX/q,\quad \bH\to\bH/q,\quad \bW\to\bW/q.
\end{equation}
It is important to reiterate that the values of the population parameters such as $\theta$, $\theta_i$, $\bar{\theta}$, $q$, and $\sdeg$ are \textit{not} affected by the rescaling procedure in \eqref{eq:rescaling tX,H,W} above. These values are determined by the original signal-plus-noise model \eqref{eq:model} as specified in Section \ref{modelsetting} and Definition \ref{def: model setting}. Throughout the rest of this section, the notation $\btX = (\tX_{ij})_{1 \leq i,j \leq n}$, $\bH = (H_{ij})_{1 \leq i,j \leq n}$, $\bW = (W_{ij})_{1 \leq i,j \leq n}$, and $s_{ij}:=\E|W_{ij}|^2$ should be understood implicitly as the rescaled versions rather than their original values. Correspondingly, we define the rescaled diagonal matrix $\bL$ as 
\begin{equation}\label{eq:defbL rescaled}
\bL\equiv\bL_{\tau,\lambda}:=\diag(L_1,\cdots,L_n)=\frac{1}{q\sdeg}\diag\left(d_i+\tau_i\bar d+\lambda_i/q: i \in [n]\right),
\end{equation}
where $d_i:=\sum_{j=1}^n\tX_{ij}$ and $\bar d:=n^{-1}\sum_{j=1}^nd_j$. Additionally, denote by 
\begin{equation} \label{new.eq.supp003}
\bLambda := \diag(\Lambda_1,\cdots,\Lambda_n) =\E \bL, 
\end{equation}
and $\bL_{[i]}$ with $1 \leq i \leq n$ the random diagonal matrix with diagonal entries $(L_{[i]})_i=L_i$ and 
\begin{equation}\label{eq:L[i] definition}
	(L_{[i]})_j=\Lambda_{j}+\frac{1}{q\sdeg}\left(\sum_{s}^{(i)}W_{js}+\frac{\tau_j}{n}\sum_{t,s}^{(i)}W_{ts}\right)
\end{equation}
for all $j\neq i$. 

We emphasize that the notation $\tX_{ij}$ used in the definitions of our rescaled $d_i$ and $\bL_{[i]}$ has been rescaled. Hence, the relationship between the original $\bL$ in \eqref{eq:defbL} and the rescaled $\bL$ in \eqref{eq:defbL rescaled}, the relationship between their expectations, and the relationship between the original $\bL_{[i]}$ in \eqref{eq:L[i] definition unrescaled} and the rescaled $\bL_{[i]}$ in \eqref{eq:L[i] definition} are given by
\begin{equation}\label{eq:rescaling L,Lambda}
    \bL\to \frac{\bL}{q^2\sdeg},\quad\bLambda\to\frac{\bLambda}{q^2\sdeg},\quad\bL_{[i]}\to\frac{\bL_{[i]}}{q^2\sdeg},
\end{equation}
respectively. From this point on, the notation $\bL$, $\bLambda$, and $\bL_{[i]}$ should be understood as referring to the rescaled matrices. Consequently, we define the generalized Laplacian matrix as 
\begin{equation} \label{new.eq.supp001}
\bX:=\bL^{-\alpha}\btX\bL^{-\alpha} 
\end{equation}
with $\bcW:=\bLambda^{-\al}\bW\bLambda^{-\al}$ for each $\alpha\in(0,\infty)$. Then we consider both empirical and population versions of the eigendecomposition given by 
\begin{equation} 
    \bX=\sum_{i\in[n]}\hdelta_i\bhv_i\bhv_i^\top \ \text{ and } \ \bLambda^{-\alpha}\bH\bLambda^{-\alpha}=\sum_{i\in[K]}\delta_i\bv_i\bv_i^\top,
\end{equation}
where we arrange the eigenvalues according to the descending order in magnitude with $|\hdelta_1|\geq\cdots\geq|\hdelta_n|$ and $|\delta_1|\geq\cdots\geq|\delta_K|>0$, and $\bhv_i$'s and $\bv_i$'s are the corresponding eigenvectors. Similarly, we define the diagonal matrices of spiked eigenvalues 
\begin{equation} 
\bhDelta:=\diag(\hdelta_1,\ldots, \hdelta_K) \ \text{ and } \ \bDelta:=\diag(\delta_1,\ldots, \delta_K), 
\end{equation}
as well as the corresponding spiked eigenvector matrices 
\begin{equation} 
\bhV=(\bhv_1,\cdots,\bhv_K) \ \text{ and } \ \bV=(\bv_1,\cdots,\bv_K). 
\end{equation}

Taking the above rescaling scheme into account, the relationships between matrices $\bX$, $\bcW$,  eigenvalues $\hdelta_k$, $\delta_k$, and their original values can be expressed as 
\begin{equation}\label{eq:rescaling X delta}
    \bX\to \frac{\sdeg^{2\al}\bX}{q^{1-4\al}},\quad \bcW\to\frac{\sdeg^{2\al}\bcW}{q^{1-4\al}},\quad \hdelta_k\to \frac{\sdeg^{2\al}\hdelta_k}{q^{1-4\al}},\quad\delta_k\to \frac{\sdeg^{2\al}\delta_k}{q^{1-4\al}},
\end{equation}
while the eigenvectors remain unchanged. Throughout the rest of this section, the notation $\bX$, $\bcW$, $\hdelta_k$, and $\delta_k$ should be understood as the rescaled versions as opposed to their original values. 

With the rescaled model introduced above, we are ready to restate the technical conditions correspondingly to ease the reading. Specifically, under the setting of the rescaled model, the assumptions given in \Cref{def: model setting} are restated below.

\begin{condition}\label{cond:rescaled model}
Assume some basic regularity conditions with a constant $C_0>0$ that 
\begin{enumerate}
\item The sparsity parameter $q$ satisfies that 
    \begin{equation}\label{eq: q condition}
   \xi^3\leq q \leq C_0n^{1/2}
   \end{equation}
   with $\xi$ given in (\ref{eq:xi condition}).
   
\item The regularization parameters satisfy that  $\tau_i\le C_0$ and $\lambda_i\le C_0q^2$ (allowing them to be zero or depend on $n$).

\item The entries of $\bW$ satisfy that
\begin{equation}\label{eq:moment conditions}
    \mathbb E W_{ij}=0,\quad s_{ij}=\mathbb E|W_{ij}|^2\leq\frac{C_0}{n}=C_0\theta,\quad \mathbb E|W_{ij}|^p\leq\frac{C_0^p}{nq^{p-2}}
\end{equation}
for all $ i,j\in [n]$ and $3\leq p\leq \xi $. 

\item The entries of $\bH$ are 
nonnegative and assume that 
\begin{equation}
\max_{i\in [n]} \theta_i \le C_0.  
\end{equation}

\item Matrix $\bL$ is positive definite almost surely.
\end{enumerate}
\end{condition}

Similarly, Assumption \ref{main_assm} can be restated below under the setting of the rescaled model.

\begin{assumption}\label{main_assm rescaled}
For a fixed $\alpha\in(0,\infty)$, assume that the following conditions hold for some $1\le K_0\le K$. 

\begin{enumerate}
\item  (Network sparsity) The sparsity parameter $q$ satisfies that $q \gg(\log n)^4$.

\item  (Spiked eigenvalues) It holds that $|\delta_k|\gg 1$ for all $1\leq k\leq K_0$.

\item (Eigengap) There exists some constant $\epsilon_0>0$ such that
\begin{equation}\label{eq:eigengap}
    \min_{1\leq k\leq K_0}\frac{|\delta_k|}{|\delta_{k+1}|}>1+\epsilon_0,
\end{equation}
where we do not require eigengaps for smaller eigenvalues $|\delta_k|$ with $K_0+1\leq k\leq K$.

\item (Low-rankness of signals)  The rank $K$ of $\bH$ satisfies that
\begin{equation}\label{eq:condK}
K\xi\left(\frac{1}{|\delta_{K_0}|\sdeg}+\frac{\xi}{q\sdeg^{2}}+\|\mathbf V\|_{\max}\right)\ll q.
\end{equation}
\end{enumerate}
\end{assumption}

We remark that by introducing a scaling factor $q^{-2}\sdeg^{-1}$ to the definition in \eqref{eq:defbL rescaled} in view of (\ref{eq:rescaling tX,H,W}) and compared to (\ref{eq:defbL}), with high probability the largest diagonal entries of the diagonal matrix $\bL$ are of order $\sdeg^{-1}$, while the smallest diagonal entries are of order $1$. Such normalization will be convenient for our technical analyses.

To present the main results under the setting of the rescaled model above, similarly, we provide the asymptotic limit $t_k$ of $\hdelta_k$. To define $t_k$, let us introduce the complex-valued vector $\bM\equiv \bM_n(z)=(M_1(z),\cdots,M_n(z))^\top\equiv(M_1,\cdots,M_n)^\top$ that is the $z$-dependent solution to the \textit{generalized quadratic vector equation (QVE)} given by

\begin{equation}\label{eq:M definition}
   \frac{1}{M_i}=-z-\Lambda_i^{-2\alpha}\sum_{j\in[n]}\Lambda_j^{-2\alpha}s_{ij}M_j
\end{equation}
with $\im M_{i}(z)\ge 0$ for all $i\in [n]$ and $z\in \C_+$, where $\C_+$ denotes the upper half of
the complex plane $\C$. It is well-known that 
\begin{itemize}
\item[1)] there exists a probability measure $\mu_c$ on $\R$ such that
\[
    \langle \bM\rangle:=\frac{1}{n}\sum_{i\in[n]}M_i(z)
\]
is the Stieltjes transform of $\mu_c$; 

\item[2)] probability measure $\mu_c$ is absolutely continuous with respect to the Lebesgue measure on $\R$, and its density $\rho_c$ is given by 
\[
    \rho_c(x)=\frac{1}{\pi}\lim_{\eta\to 0+}\Im \langle \bM(x+i\eta)\rangle
\]
for $x \in \R$;

\item[3)] measure $\mu_c$ is compactly supported on $\R$ with support $\supp(\mu_c)\subset [-2\sqrt{\fM},2\sqrt{\fM}]$, where $\fM:=\max_{i\in[n]}\sum_{j\in[n]}s_{ij}$; 

\item[4)] each $M_i(z)$  is the Stieltjes transform of some finite measure that has the same support as $\mu_c$ and is uniformly bounded, i.e., $\max_{z\in\C_+}|M_i(z)|\lesssim 1$. 
\end{itemize}
For more details, see, e.g., Corollary 1.3 of \cite{ajanki2017universality}.
Indeed, measure $\mu_c$ is known as the asymptotic empirical spectral distribution (ESD) of the noise random matrix $\bW$ \citep{ajanki2017universality}.

We next define the complex-valued deterministic diagonal matrix
\begin{equation}\label{eq:Upsilon definition}
    \bUpsilon(z):=\diag(M_1(z),\cdots,M_n(z))
\end{equation}
and the complex-valued deterministic matrix
\begin{equation}\label{eq:Upsilonk def}
	\bUpsilon_k(z):=\bUpsilon(z)-\bUpsilon(z)\bV_{-k}\frac{1}{\bDelta_{-k}^{-1}+\bV_{-k}^\top\bUpsilon(z)\bV_{-k}}\bV_{-k}^\top\bUpsilon(z)
\end{equation}
with $z\in \C_+$ and $1 \leq k \leq K$. For notational simplicity, we will drop the dependence on $z$ whenever there is no confusion. By comparing  \eqref{eq:tM definition} and \eqref{eq:M definition}, one can observe that the relationship between $\btUpsilon$ and $\bUpsilon$ and the relationship between $\btUpsilon_k$ and $\bUpsilon_k$ are given by 
\begin{equation}
    \btUpsilon(z)=\frac{\sdeg^{2\al}}{q^{1-4\al}}\bUpsilon\left(\frac{\sdeg^{2\al}}{q^{1-4\al}}z\right) \ \text{ and } \ \btUpsilon_k(z)=\frac{\sdeg^{2\al}}{q^{1-4\al}}\bUpsilon_k\left(\frac{\sdeg^{2\al}}{q^{1-4\al}}z\right),
\end{equation}
respectively. For each $1\le k\le K_0$, denote by  
\beq 
    \cI_k:=\biggl\{x\in\mathbb R:\frac{|\delta_k|}{1+\epsilon_0/2}\leq|x|\leq (1+\epsilon_0/2)|\delta_k|\biggr\},
\eeq
and let $t_k \in \R$ be the solution to the nonlinear equation
\begin{equation}\label{eq:t_k def}
    1+\delta_k\bv_k^\top\bUpsilon(x)\bv_k-\delta_k\bv_k^\top\bUpsilon(x)\bV_{-k}\frac{1}{\bDelta_{-k}^{-1}+\bV_{-k}^\top\bUpsilon(x)\bV_{-k}}\bV_{-k}^\top\bUpsilon(x)\bv_k=0
\end{equation}
over $x\in \cI_k$.

By resorting to similar arguments as in the proof of Lemma 3 in \cite{fan2020asymptotic} and Section A.2 of \cite{SIMPLERC}, we can establish the lemma below, which asserts the existence, uniqueness, and asymptotic properties of the population quantity $t_k$ defined in (\ref{eq:t_k def}).

\begin{lemma}\label{lemma:tk rescaled}
    Under parts (ii) and (iii) of Assumption \ref{main_assm rescaled}, for each $1 \leq k \leq K_0$, there exists a unique solution $x=t_k$ to equation \eqref{eq:t_k def} in the subset $\cI_k$, 
    and it holds that $t_k=\delta_k+O(|\delta_k|^{-1})$.
\end{lemma}

Lemma \ref{lemma:tk rescaled} above under the rescaled model corresponds to Lemma \ref{lemma:tk} in the main text. By comparing \eqref{eq:sft_k def} and \eqref{eq:t_k def}, we see the relationship between population quantity $\sft_k$ and its original value given by 
\begin{equation}\label{eq:rescaling t_k}
    \sft_k\to\frac{\sdeg^{2\al}t_k}{q^{1-4\al}},
\end{equation}
which is coherent with the rescaling in \eqref{eq:rescaling X delta}. Finally, we define the resolvents (i.e., the Green functions) of three random matrices
\begin{align}\label{eq:def resolvents}
	& \bG(z):=\left({\bcW}-z(\bL/\bLambda)^{2\alpha}\right)^{-1}, \ \ \bR(z):=\left({\bcW}-z\bI\right)^{-1}, \nonumber\\
 & \bG_{[i]}(z)=\left(\bcW-z(\bL_{[i]}/\bLambda)^{2\alpha}\right)^{-1}
\end{align}
with $z \in \C$ and $1 \leq i \leq n$. Comparing to the original definitions in \eqref{eq:defGR unrescaled} and \eqref{eq:G[i] definition unrescaled}, one can observe that the rescaling of the resolvents is given by
\begin{align} \label{eq:rescaling resolvents}
    & \bG(z)\to\frac{\sdeg^{2\al}}{q^{1-4\al}}\bG\left(\frac{\sdeg^{2\al}}{q^{1-4\al}}z\right), \ \ 
    \bR(z)\to\frac{\sdeg^{2\al}}{q^{1-4\al}}\bR\left(\frac{\sdeg^{2\al}}{q^{1-4\al}}z\right), \nonumber \\
    & \bG_{[i]}(z)\to\frac{\sdeg^{2\al}}{q^{1-4\al}}\bG_{[i]}\left(\frac{\sdeg^{2\al}}{q^{1-4\al}}z\right).
\end{align}
In summary, the specific rescaling scheme for the scaled model in this section is outlined in \eqref{eq:rescaling tX,H,W}, \eqref{eq:rescaling L,Lambda}, \eqref{eq:rescaling X delta}, \eqref{eq:rescaling t_k}, and \eqref{eq:rescaling resolvents} above. It is important to note that throughout the rest of this section and Sections \ref{sec:prelim}--\ref{addtech.details} later, 
the notations $\btX$, $\bH$, $\bW$, $s_{ij}$, $\bL$, $\bLambda$, $\bL_{[i]}$, $\bX$, $\bcW$, $\hdelta_k$, $\delta_k$, $t_k$, $\bG$, $\bR$, and $\bG_{[i]}$ should be interpreted as referring to the rescaled versions.

Now we present the main results of the paper under the setting of the rescaled model. It can be observed that Theorems \ref{thm:eigenvalue rescaled}--\ref{thm:K_0 rescaled} below are equivalent to Theorems \ref{thm:eigenvalue}--\ref{thm:K_0} in the main text, respectively. Therefore, to establish Theorems \ref{thm:eigenvalue}--\ref{thm:K_0}, it is sufficient to prove Theorems \ref{thm:eigenvalue rescaled}--\ref{thm:K_0 rescaled} in this section.

\begin{theorem}\label{thm:eigenvalue rescaled}
Under Condition \ref{cond:rescaled model} and Assumption \ref{main_assm rescaled}, it holds w.h.p.~that
\begin{equation}\label{eqn:t_k-gamma_k}
    |\hdelta_k-t_k|=O\biggl\{ |\delta_k| \frac{\xi\pnk}{q}\left(1+\frac{K }{ |\delta_k|^4}\right)\biggr\}
\end{equation}
for each $1\leq k\leq K_0$, where we introduce the notation
\begin{equation}\label{eq:Apsink}
\pnk:=\frac{1}{|\delta_k|\sdeg}+\frac{\xi}{q\sdeg^{2}}+\|\bV\|_{\max} 
 \end{equation}
for simplicity.
\end{theorem}

\begin{theorem}\label{thm:projection rescaled}
Under Condition \ref{cond:rescaled model} and Assumption \ref{main_assm rescaled}, for each $1\le k\le K_0$ it holds w.h.p. that 
\begin{equation}\label{eq:projectionvk}
\begin{split}
   \left|\bv_k^\top(\bL/\bLambda)^{-\alpha}\bhv_k- \frac{1}{\sqrt{\delta_k^2\bv_k^\top\bUpsilon_k'(t_k)\bv_k}}\right| \lesssim \frac{\xi\pnk}{q}\left(1+\frac{K }{ |\delta_k|^4}\right) ,    
\end{split}
\end{equation}
where we choose the direction of $\bhv_k$ such that $\bhv_k^\top\bv_k>0$. Further, for any deterministic unit vector $\bu$, it holds w.h.p. that 
\begin{equation}\label{eq:projection}
\begin{split}
    &\left|\bu^\top(\bL/\bLambda)^{-\alpha}\bhv_k+\frac{\delta_k\bu^\top\bUpsilon_k(t_k)\bv_k}{\sqrt{\delta_k^2\bv_k^\top\bUpsilon_k'(t_k)\bv_k}}\right| \\
    &\lesssim \frac{\xi\pnk}{q}\left[1+\frac{K }{ |\delta_k|^4} + {\|\bu^\top \bV_{-k}\|} \left(\sqrt K+\frac{K}{|\delta_k|^2}\right)\right].
\end{split}    
\end{equation}
Moreover, for the second terms on the left-hand side (LHS) of \eqref{eq:projectionvk} and \eqref{eq:projection}, we have that 
\begin{equation}\label{eq:projectionvk2}
\delta_k^2\bv_k^\top\bUpsilon_k'(t_k)\bv_k= 1+O(\delta_k^{-2}) \ \text{ and } \ \delta_k \bu^\top\bUpsilon_k(t_k)\bv_k = -  \bu^\top \bv_k +O(\delta_k^{-2}).
\end{equation}
\end{theorem}

\begin{theorem}\label{thm:main thm rescaled}
Assume that Condition \ref{cond:rescaled model} and Assumption \ref{main_assm rescaled} are satisfied, and 
\begin{equation}\label{eq:main thm assumption}
 K\pnk\sdeg\lesssim 1,\quad \|\bV\|_{\max}\ll \frac{1}{|\delta_k|\sdeg}+\frac{\xi}{q\sdeg^{2}} 
\end{equation}
for $1\le k\le K_0$. Then for each $i\in [n]$, it holds w.h.p.~that
\begin{align}\label{eq:main thm}
	\begin{split}
	 \hv_k(i)&=(\Lambda_i/L_i)^{\al} v_k(i)+\frac{1}{t_kL_i^{\alpha}}\sum_{j\in[n]}{W}_{ij}{\Lambda}_j^{-\alpha}v_k(j)\\
	&\quad+O\left(\|\bV\|_{\max}\left(\frac{\sqrt K}{|\delta_k|}+\frac{ K\xi}{q}\right)\left(\frac{1}{|\delta_k|\sdeg}+\frac{ \xi}{q\sdeg^{2}}\right)\right)\\
 &\quad+O\left(\frac{\xi}{\sqrt n |\delta_k|}\left(\frac{1}{|\delta_k|}+\frac{\xi}{q\sdeg}\right)\right),
	\end{split}
\end{align}
where we choose the direction of $\bhv_k$ such that $\bhv_k^\top\bv_k>0$. Consequently, we obtain that 
\begin{align}\label{eq:main thm expand}
\begin{split}
   	 \hv_k(i)&= v_k(i)- \frac{\al}{\Lambda_i\sdeg}\bigg( \frac{1}{q} \sum_{j\in[n]} W_{ij} +\frac{\tau_i}{nq} \sum_{j,l\in[n]} W_{jl}\bigg)v_k(i) +\frac{1}{t_k}\sum_{j\in[n]}\Lambda_i^{-\al}{W}_{ij}{\Lambda}_j^{-\alpha}v_k(j)\\
	&\quad+O\left(\|\bV\|_{\max}\left(\frac{\sqrt K}{|\delta_k|}+\frac{ K\xi}{q}\right)\left(\frac{1}{|\delta_k|\sdeg}+\frac{ \xi}{q\sdeg^{2}}\right)\right)\\
 &\quad+O\left(\frac{\xi}{\sqrt n |\delta_k|}\left(\frac{1}{|\delta_k|}+\frac{\xi}{q\sdeg}\right)\right). 
\end{split}
\end{align}
\end{theorem}

\begin{theorem}\label{thm:eigenvalue2 rescaled}
    Under Condition \ref{cond:rescaled model} and Assumption \ref{main_assm rescaled}, it holds w.h.p. that
\begin{align}\label{eqn:t_k-gamma_k2}
\begin{split}
    &\hdelta_k-t_k - A_k= -2\alpha t_k\bv_k^\top\frac{\bL-\bLambda}{\bLambda}\bv_k+\bv_k^\top\bcW\bv_k+B_k\\
    &\quad+O\left(\frac{1}{|\delta_k|^2}+\frac{\xi^3|\delta_k|}{q^3\sdeg^3}+\frac{\sqrt K\xi|\delta_k|\pnk}{q}\left(\frac{1}{|\delta_k|^2}+\frac{\sqrt K\xi\pnk}{q}\right)\right),
\end{split}
\end{align}
where $A_k$ is a deterministic term given by  
\begin{align*}
A_k& =\alpha(2\alpha+1)t_k\E\bv_k^\top\frac{(\bL-\bLambda)^2}{\bLambda^2}\bv_k-4\alpha\E\bv_k^\top\frac{\bL-\bLambda}{\bLambda}\bcW\bv_k,
\end{align*}
and $B_k$ is a centered random error satisfying  
\[
    \var (B_k)\lesssim \frac{|\delta_k|^2\|\bv_k\|_\infty^2}{q^4\sdeg^4}+\frac{|\delta_k|^2}{q^4n^2\sdeg^4}+\frac{\|\bv_k\|_\infty^2}{q^2\sdeg^2}+\frac{1}{q^2n\sdeg^2}+\frac{1}{\sqrt n q|\delta_k|^2}.
\]
\end{theorem}

\begin{theorem}\label{thm:projection2 rescaled}
Assume that Condition \ref{cond:rescaled model} and Assumption \ref{main_assm rescaled} are satisfied. Then we have that 

1) For each $1\leq k\leq K_0$ and any deterministic unit vector $\bu$ such that $\bu^\top\bv_k=0$, it holds w.h.p. that
\begin{align}\label{eq:projection2}
\begin{split}
    &\bu^\top(\bL/\bLambda)^{-\alpha}\bhv_k -\cA_k= \sft_k\bu^\top\bV_{-k}\frac{1}{\sft_k-\bDelta_{-k}}\bV_{-k}^\top\left(-2\alpha\frac{\bL-\bLambda}{\bLambda}+\sft_k^{-1}\bcW\right)\bv_k\\
    +&\bw^\top\left(-2\alpha\frac{\bL-\bLambda}{\bLambda}+t_k^{-1}\bcW\right)\bv_k+\sum_{l\in[K]\setminus\{k\}}\frac{\sft_k\bu^\top\bv_l}{\sft_k-\delta_l}\cB_{k,l}+\cB_k^{\bw}\\
    +&O\left(K\left(\frac{1}{|\delta_k|^2}+\frac{\xi\pnk}{q}\right)\left(\frac{1}{|\delta_k|}+\frac{\xi}{q\beta_n}\right)+\frac{K^{3/2}\xi\pnk}{q}\left(\frac{\xi\pnk}{q}+\frac{1}{|\delta_k|^2}\right)\right),
\end{split}
\end{align}
where we choose the sign of $\bhv_k$ such that $\bhv_k^\top\bv_k>0$, $\bw=(\bI-\bV\bV^\top)\bu$, $\cA_k$ is a deterministic term given by 
\begin{align*}
\cA_k & =\bw^\top\left(\alpha(2\alpha+1)\frac{(\bL-\bLambda)^2}{\bLambda^2}- \frac{2\alpha}{ t_k}\left(\frac{\bL-\bLambda}{\bLambda}\bcW+\bcW\frac{\bL-\bLambda}{\bLambda}\right)+\frac{\bcW^2}{t_k^{2}}\right)\bv_k\\
&\quad+t_k\bu^\top\bV_{-k}\frac{1}{t_k-\bDelta_{-k}}\E \bV_{-k}^\top\left(\alpha(2\alpha+1)\frac{(\bL-\bLambda)^2}{\bLambda^2} \right.\\
& \quad \left.- \frac{2\alpha}{ t_k}\left(\frac{\bL-\bLambda}{\bLambda}\bcW+\bcW\frac{\bL-\bLambda}{\bLambda}\right)+\frac{\bcW^2}{t_k^{2}}\right)\bv_k,
\end{align*}
$\cB_{k}^{\bw}$ is a centered random variable satisfying
\[
\var(\cB_{k}^{\bw}) \lesssim\frac{\|\bv_k\|_\infty\|\bw\|_\infty}{q^4\sdeg^4}+\frac{|\bw|}{q^4n^2\sdeg^4}+\frac{1}{|\delta_k|^2\sdeg^2}\left(\frac{\|\bv_k\|_\infty\|\bw\|_\infty}{q^2}+\frac{|\bw|}{q^2n}\right)+\frac{|\bw|}{q\sqrt n|\delta_k|^4},
\]
and for each $l\in[K]\setminus\{k\}$, $\cB_{k,l}$ is a centered random variable satisfying
\[
\var(\cB_{k,l}) \lesssim\frac{\|\bv_k\|_\infty\|\bv_l\|_\infty}{q^4\sdeg^4}+\frac{1}{q^4n^2\sdeg^4}+\frac{1}{|\delta_k|^2\sdeg^2}\left(\frac{\|\bv_k\|_\infty\|\bv_l\|_\infty}{q^2}+\frac{1}{q^2n}\right)+\frac{1}{q\sqrt n|\delta_k|^4}.
\]

2) For the case of $\bu=\bv_k$ and each $1 \leq k \leq K_0$, it holds w.h.p. that
\begin{align}\label{eq:projection3}
\begin{split}
     & \bv_k^\top(\bL/\bLambda)^{-\alpha}\bhv_k-\bv_k^\top(\bL/\bLambda)^{-\alpha}\bv_k -\fA_k =\frac{\alpha^2}{2}\bv_k^\top\left(\frac{\bL-\bLambda}{\bLambda}\right)^2\bv_k\\
    &\quad-\frac{1}{2t_k^{2}}\bv_k^\top\bcW^2\bv_k+\fB_k+O\left(\frac{K}{|\delta_k|^4}+\frac{K\xi^2\psi_n^2}{q^2}\right),
\end{split}
\end{align}
where $\fA_k$ is a deterministic term given by 
\[
\fA_k:=(\delta_k^2\bv_k^\top\bUpsilon'_k(t_k)\bv_k)^{-1/2}-1+\frac{1}{2}\bv_k^\top(t_k^2\bUpsilon'(t_k)+2t_k\bUpsilon(t_k)+\bI)\bv_k\]
and $\fB_k$ is a random variable satisfying
\[
\E\fB_k^2\lesssim\frac{n^2\|\bv_k\|_\infty^4}{q^8\sdeg^6}+\frac{n^2\|\bv_k\|_\infty^4}{q^2|\delta_k|^6}.
\]
\end{theorem}

\begin{theorem}\label{thm:K_0 rescaled}
Assume that Condition \ref{cond:rescaled model} and Assumption \ref{main_assm rescaled} are satisfied, 
\begin{equation}\label{eq:K_0+1 assumption}
|\delta_{K_0+1}|\gg 1,\quad \left|\frac{\delta_{K_0+1}}{\delta_{K_0+2}}\right|\geq 1+\epsilon_0,\quad K\xi\psi_n(\delta_{K_0+1})\ll q, 
\end{equation}
$K_0$ can be represented as
\begin{equation}
    K_0=\max\left\{k\in[K]:|\delta_k|\geq a_n\right\}
\end{equation}
with some deterministic sequence $a_n\to\infty$, and there exists some deterministic sequence $a_n'\to\infty$ such that
\begin{equation}\label{eq:a_n assumption}
    \limsup_{n\to\infty}\left|\frac{a_n'}{a_n}\right|<1,\quad \limsup_{n\to\infty}\frac{|\delta_{K_0+1}|}{a_n'}<1.
\end{equation}
Then the estimate of the latent embedding dimensionality defined as 
\begin{equation}
    \hK_0:=\max\{k\in[K]:|\hdelta_k|\geq a_n'\}
\end{equation}
is a consistent estimator of $K_0$, i.e., $\P\{\hK_0 = K_0\} \rightarrow 1$ as $n \rightarrow \infty$.
\end{theorem}

\section{Preliminary estimates and local laws} \label{sec:prelim}
Throughout this section, we continue to examine the rescaled setting of our model as introduced in Section \ref{sec.mainresu.rescaledmodel}, which includes rescalings \eqref{eq:rescaling tX,H,W}, \eqref{eq:rescaling L,Lambda}, \eqref{eq:rescaling X delta}, and \eqref{eq:rescaling t_k}.
\subsection{Some preliminary estimates} \label{Preliminary Estimates}

In this subsection, we provide some preliminary estimates that will be used in our technical analyses. 
Note that we record in \Cref{lemma:estimation} in Subsection \ref{add.tech.lem} 
some large deviation estimates for random variables that satisfy \eqref{eq:moment conditions}. 
We begin with giving a concentration estimate for the rescaled diagonal matrix $\bL$ in (\ref{eq:defbL rescaled}). 
To this end, we define the diagonal random error matrix $\bcE$ as
\begin{equation} \label{new.eq.supp002}
\bcE:=\diag(\cE_1,\cdots,\cE_n)=\bL-\bLambda.
\end{equation}

With the aid of \Cref{lemma:estimation}, we can readily obtain the results in the two lemmas below.

\begin{lemma}\label{prop:Lambda and E}
Under 
Condition \ref{cond:rescaled model}, there exist some constants $C_1,c_1>0$ (depending on $\tau$, $\lambda$, and $C_0$) such that for all $i\in[n]$, 
		\begin{equation}\label{eq:Lambda}
		C_1^{-1} \le {\Lambda}_i =  \sdeg^{-1}(\theta_i + \tau_i \bar \theta+ \lambda_i/q^2)  \le C_1\sdeg^{-1},
		\end{equation}
		and with $(c_1,\xi)$-high probability,
		\begin{equation}\label{eq:E}
			\max_{i\in[n]}|\cE_i|\le C_1\sdeg^{-1}\xi/q .
		\end{equation}
		Consequently, we have that with $(c_1,\xi)$-high probability,
\begin{equation}\label{eq:LasympLambda}
    \|\bLambda\|,\, \|\bL\|\lesssim \sdeg^{-1},\ \ \|\bLambda^{-1}\|,\, \|\bL^{-1}\|\lesssim 1,\ \ \|\bL/\bLambda\|\sim 1.
\end{equation}
\end{lemma}
 
\begin{lemma}\label{prop:L-Lambda}
Under 
Condition \ref{cond:rescaled model},  for each fixed $\alpha\in 
(0, \infty)$, there exist some constants $C_2 ,c_2>0$ (depending on $\tau$, $\lambda$, $C_0$, and $\al$) such that  
\begin{equation}\label{eq:L-Lambda}
    \left|\frac{L_i^{\alpha}-\Lambda_i^{\alpha}}{\Lambda_i^\alpha}\right|\le {C_2}\frac{\xi}{q\sdeg}
\end{equation}
with $i\in[n]$ holds with $(c_2,\xi)$-high probability. 
\end{lemma}

We next introduce the technical notion of \textit{minors} of matrices as given in the definition below. 

\begin{definition}[Minors]
Given an $n\times n$ matrix $\bA=\bW,\ \bcW,\ \bUpsilon,\ \bLambda$, or $\bL$, and a subset $\mathbb T\subset[n]$, we define the minor $\bA^{(\mathbb T)}:=(A_{ij}:i,j\notin \mathbb T)$ as a matrix of size $(n-|\mathbb T|)\times(n-|\mathbb T|)$ defined by removing all rows and columns of $\bA$ with indices belonging to $\mathbb T$. We keep the names of indices for $\bA^{(\mathbb T)}$, i.e., $A^{(\mathbb T)}_{ij}=A_{ij}$ for $i,j\notin \mathbb T$.   
Then we define the resolvent minors as
\begin{align*}
    \bG^{(\mathbb T)}(z)&:=\big[\bcW^{(\mathbb T)} -z(\bL^{(\mathbb T)}/\bLambda^{(\mathbb T)})^{2\alpha}\big]^{-1}, \ \text{ and } \ 
    \bR^{(\mathbb T)}(z):=\big(\bcW^{(\mathbb T)} -z\big)^{-1}.
\end{align*}
For simplicity of notation, we will abbreviate $(\{i\})=(i)$, $(\{i,j\}):=(ij)$, and $\sum_{i\in[n]}^{(\mathbb T)}=\sum_{i\in[n]\setminus\mathbb T}$. As a convention, we define $A^{(\mathbb T)}_{ij}=R^{(\mathbb T)}_{ij}=G_{ij}^{(\mathbb T)}=0$ whenever $i$ or $j$ belongs to $\mathbb T$.
\end{definition}
 
Let us define a parameter of order $1$ as 
\begin{equation} \label{new.eq.supp004}
\fM:=(\max_{i\in[n]}\Lambda_i^{-\al}\sum_{j\in[n]}s_{ij} \Lambda_j^{-\al}) \vee 1,
\end{equation}
where $\vee$ denotes the maximum of two given numbers. With an application of arguments as in \cite{erdHos2013spectral} and \Cref{prop:L-Lambda}, we can prove the bounds on the operator norms of $\bcW$, $\bG$, $\bR$, and their minors in the proposition below.
 

\begin{proposition}\label{prop:W,R,G}
Under 
Condition \ref{cond:rescaled model}, there exist some constants $C_3,c_3>0$ such that with $(c_3,\xi)$-high probability,
    \begin{equation}\label{eq:W}
        \max\left\{\|\bcW\|,\ \max_{i\in[n]}\|\bcW^{(i)}\|,\ \max_{i,j\in[n]}\|\bcW^{(ij)}\|\right\}\leq 2\sqrt{\fM}+\xi/\sqrt{q}.
    \end{equation}
    Consequently, given any $\fC>2\sqrt{\fM} +\kappa$ with some constant $\kappa>0$, it holds that with $(c_3,\xi)$-high probability,
    \begin{align}
        \sup_{z\in S(\fC)}\left(|z|-2\sqrt{\fM}\right)\max\left\{\|\bR(z)\|,\ \max_{i\in[n]}\|\bR^{(i)}(z)\|,\ \max_{i,j\in[n]}\|\bR^{(ij)}(z)\|\right\}\le C_3,\label{eq:R}\\
        \sup_{z\in S(\fC)}\left(|z|-2\sqrt{\fM}\right)\max\left\{\|\bG(z)\|,\ \max_{i\in[n]}\|\bG^{(i)}(z)\|,\ \max_{i,j\in[n]}\|\bG^{(ij)}(z)\|\right\}\le C_3,\label{eq:G}
    \end{align}
    where we define the spectral domain $S(\fC):=\{z=E+i\eta: \fC\le |E|\le n^{\fC},\ \eta\geq 0\}$. 
\end{proposition}

\begin{remark}
  Note that condition \eqref{eq:signal conditions} implies the trivial bound on the signals 
\beq\label{eq:bounddeltak}
|\delta_1|^2 \lesssim \|\bH\|^2 \le   \sum_{i,j \in [n]}H_{ij}^2 \le \sum_{i \in [n]} \bigg(\sum_{j \in [n]} H_{ij}\bigg)^2 \le n q^2 \lesssim n^2.
\eeq
Hence, since $\fC>2$ in view of (\ref{new.eq.supp004}), we see that the spectral domain $S(\fC)$ contains all the subsets $\cal I_k$ defined in \eqref{eq:defnIk}.
\end{remark}

We proceed with stating some fine estimates on $\bR(z)$, called the \emph{local laws}, which show that $\bUpsilon$ defined in \eqref{eq:Upsilon definition} is the asymptotic limit of $\bG(z)$ in various senses (together with some precise rates of convergence). These local laws have been proved (under slightly different assumptions) in \cite{SIMPLERC}.   

\begin{theorem}[Local laws of $\bR$]\label{thm:local laws of R}
Assume that 
Condition \ref{cond:rescaled model} is satisfied, $q\gg(\log n)^4$, and $\fC>2\sqrt{\fM}+\kappa$ with some constant $\kappa>0$. Then there exist some constants $c_4,C_4>0$ such that the events
\begin{align}
    &\bigcap_{z\in S(\fC)}\biggl\{\max_{i\in[n]}|{R}_{ii}(z)-{M}_i(z)|\leq\frac{C_4}{|z|^2}\bigg(\frac{1}{q}+\frac{\xi}{q|z|}+\frac{\xi^{2}}{\sqrt n|z|}\bigg)\bigg\},\label{eq:entry law of R}\\
	&\bigcap_{z\in S(\fC)}\bigg\{\max_{i\neq j\in[n]}|R_{ij}(z)|\leq\frac{C_4}{|z|^2}\bigg(\frac{1}{q}+\frac{\xi^{2}}{\sqrt n|z|}\bigg)\biggr\}
\end{align}
hold with $(c_4,\xi)$-high probability. Moreover, for any deterministic unit vectors $\bu,\bv$ and constant $D>0$, the events
\begin{equation}\label{eq:anisotropic law of R}
	\bigcap_{z\in S(\fC)}\biggl\{|\bu^\top(\bR(z)-\bUpsilon(z))\bv|\leq \frac{C_4\log n}{q|z|^2}\biggr\},
\end{equation}
\begin{equation}\label{eq:prop 1 of R}
	\bigcap_{z\in S(\fC)}\biggl\{\max_{i\in[n]}|\be_i^\top(\bR(z)-\bUpsilon(z))\bv|\leq \frac{C_4}{|z|^2}\biggl({\frac{\xi}{\sqrt n}}+\frac{\xi}{q}\|\bv\|_{\infty}\biggr)\biggr\},
\end{equation}
\begin{equation}\label{eq:prop 2 of R}
	\bigcap_{z\in S(\fC)}\biggl\{\max_{i\in[n]}|\be_i^\top\bLambda^{-\alpha}\bW\bLambda^{-\alpha}(\bR(z)-\bUpsilon(z))\bv|\leq \frac{C_4}{|z|^2}\biggl({\frac{\xi}{\sqrt n}}+\|\bv\|_{\infty}\biggr)\biggr\}
\end{equation}
hold with probability at least $1-n^{-D}$.
\end{theorem}

Using the \textit{generalized QVE} in \eqref{eq:M definition} and the definition in \eqref{eq:Upsilon definition}, we can easily show the estimates on $\bUpsilon(z)$ and its first and second derivatives in the lemma below.

\begin{lemma}\label{lemma:Upsilon}
 For $z\in S(\fC)$ with $\fC>2\sqrt{\fM}+\kappa$ and some constant $\kappa>0$, we have the estimates
\begin{equation}\label{eq:Upsilon}
    \bUpsilon(z)=-z^{-1}+\bcE_1(z),\ \bUpsilon'(z)=z^{-2}+\bcE_2(z),\ \bUpsilon''(z)=-2z^{-3}+\bcE_3(z),
\end{equation}
where $\bcE_1(z)$, $\bcE_2(z)$, and $\bcE_3(z)$ are deterministic diagonal matrices satisfying $\|\bcE_1(z)\|=O(|z|^{-3})$, $\|\bcE_2(z)\|=O(|z|^{-4})$, and $\|\bcE_3(z)\|=O(|z|^{-5})$.
\end{lemma}


\subsection{Local laws of $\bG$}\label{Local laws}

In this subsection, using the preliminary estimates presented in Subsection \ref{Preliminary Estimates}, we will establish some local laws (see Theorems \ref{prop:ekG-UpsilonvLinfty}--\ref{prop:eiLWL(G-Upsilon)v}) on the resolvent $\bG(z)$, which are the core technical RMT results used in the proofs of our main results. We first present a proposition below. 

\begin{proposition}\label{prop:G-R}
Under 
Condition \ref{cond:rescaled model}, given any $\fC>2\sqrt{\fM}+\kappa$ with some constant $\kappa>0$, 
there exist some constants $C_5,c_5>0$ such that with $(c_5,\xi)$-high probability,  
\begin{align}
   & \sup_{z\in S(\fC)}\|\bG(z)-\bR(z)\|\le \frac{C_5\xi}{q|z|\sdeg}, \label{eq:G-R}\\
   & \sup_{z\in S(\fC)}\|\bG(z)-\bUpsilon(z)\|\le  C_5\left(\frac{1}{|z|^2}+\frac{\xi}{q|z|\sdeg}\right).\label{eq:G-Y0}
\end{align}
\end{proposition}

Combining the local laws of $\bR$ in \Cref{thm:local laws of R} with \Cref{prop:G-R} above, we can immediately derive some local laws on $\bG$. However, they are not sharp enough for our purpose. For the rest of this section, we will derive some 
\textit{more refined} local laws on $\bG$ that give almost sharp error estimates. 
One main difficulty in proving the locals on $\bG$ is the issue that random matrices $\bL$ and $\bW$ are \textit{not} independent of each other. A useful observation is that the $i$th diagonal entry $L_i$ depends mainly on the entries in the $i$th row and column of random noise matrix $\bW$. To decouple such dependence, we introduce the intermediate resolvent as defined in \eqref{eq:G[i] definition unrescaled}.
For defining the $j$th diagonal entry of the random diagonal matrix $\bL$, we remove the contributions from the entries in the $i$th row and column of matrix $\bW$. As a consequence, $\bL^{(i)}_{[i]}$ and $\bG^{(i)}_{[i]}$ are independent of the entries in the $i$th row and column of matrix $\bW$. Using Lemma \ref{lemma:estimation}, we can easily control the difference between $\bL$ and $\bL_{[i]}$ in the lemma below.

\begin{lemma}\label{lemma:L[i]}
Under 
Condition \ref{cond:rescaled model}, for each fixed $\alpha\in (0, \infty)$, there exist some constants $C_6,c_6>0$ such that with $(c_6,\xi)$-high probability,  
    \begin{equation}\label{eq:L-L[i]}
        \left\|\bL-\bL_{[i]}\right\|_F\le \frac{C_6}{q\sdeg},\quad  \left\|\frac{\bL^{\alpha}-\bL_{[i]}^{\alpha}}{\bL^\alpha}\right\|_F\leq\frac{C_6}{q\sdeg}, \quad \left\|\frac{\bL_{[i]}^{\alpha}-\bLambda^{\alpha}}{\bLambda^\alpha}\right\|\leq\frac{C_6\xi}{q\sdeg}.
    \end{equation}
    Consequently, given any $\fC>2\sqrt{\fM}+\kappa$ with some constant $\kappa>0$, there exist some constants $C_7,c_7>0$ such that for all $z\in S(\fC)$, 
    \begin{align}
       & \sup_{z\in S(\fC)} \max_{i\in [n]} \left\|\bG_{[i]}-\bR(z)\right\|  \le \frac{C_7\xi}{q|z|\sdeg},\label{eq:G[i]-R}\\
        & \sup_{z\in S(\fC)} \max_{i\in [n]} \left\|\bG_{[i]}^{(i)}-\bUpsilon^{(i)}(z)\right\|  \le C_7\left(\frac{1}{|z|^2}+\frac{\xi}{q|z|\sdeg}\right),\label{eq:G[i]-R_add}\\
    & \max\left\{\max_{i\in [n]}\left\|\bG_{[i]}(z)\right\|,\ \max_{i,j\in[n]}\left\|\bG_{[i]}^{(i)}(z)\right\|,\ \max_{i,j,k\in[n]}\left\|\bG_{[i]}^{(jk)}(z)\right\|\right\}\le \frac{C_7}{|z|}\label{eq:G[i]-R2}
    \end{align}
   hold with $(c_7,\xi)$-high probability.
\end{lemma}

With the aid of Lemma \ref{lemma:L[i]} above, we can easily bound the difference between bilinear forms $\bu^\top \bG(z)\bv$ and $\bu^\top \bG_{[i]}(z)\bv$. We remark that in the lemma below, vectors $\bu$ and $\bv$ are \textit{not} necessarily deterministic (in contrast to some other results in this paper). 


 \begin{lemma}\label{lemma:G-G[i]}
Assume that Condition \ref{cond:rescaled model} is satisfied and $\fC>2\sqrt{\fM}+\kappa$ with some constant $\kappa>0$. Then for any vectors $\bu,\bv\in \C^n$ (which could be random) and all $z\in S(\fC)$, we have that 
 \begin{align}
    |\bu^\top(\bG(z)-\bG_{[i]}(z))\bv|&\lesssim\frac{1}{q\sdeg}|\bu| \left(\|\bG\bv\|_{\infty}\wedge \|\bG_{[i]}\bv\|_{\infty}\right) \label{eq:uG-G[i]v}
 \end{align}
 with high probability. Such estimate also holds for $\bG^{(i)}$, i.e., 
  \begin{align}
    \left|\bu^\top\left(\bG^{(i)}(z)-\bG^{(i)}_{[i]}(z)\right)\bv\right|&\lesssim\frac{1}{q\sdeg}|\bu| \left(\|\bG^{(i)}\bv\|_{\infty}\wedge \|\bG^{(i)}_{[i]}\bv\|_{\infty}\right).\label{eq:uG-G[i]v2}
 \end{align}
 \end{lemma}

By Schur's complement formula, we have the resolvent identities collected in the lemma below. The reader can also refer to Lemma 3.4 in \cite{erdHos2013spectral} for proof.
\begin{lemma}[Resolvent identities]\label{lemma:resolvent identities}
The following resolvent identities hold for $\bG(z)$. 
\begin{enumerate}
    \item  For each $i\in[n]$, we have
    \begin{equation}\label{eq:resolvent identity 1 of G}
    \frac{1}{G_{ii}}=-z(L_i/\Lambda_i)^{2\al}-\ocW_{ii}-\sum_{k,l\in[n]}^{(i)}\ocW_{ik}\ocW_{il}G_{kl}^{(i)}.
\end{equation}

\item For each $i\neq j\in[n]$, we have
\begin{align}\label{eq:resolvent identity 2 of G}
    G_{ij}&=-G_{ii} \sum_{k\in[n]}^{(i)}\ocW_{ik}G_{kj}^{(i)}=G_{ii}G_{jj}^{(i)} \bigg(-\ocW_{ij}+\sum_{k,l\in[n]}^{(i,j)}\ocW_{ik}W_{jl}G_{kl}^{(ij)}\bigg).
\end{align}

\item For each $k\in[n]\setminus\{i,j\}$, we have
\begin{align}\label{eq:resolvent identity 3 of G}
    G_{ij}^{(k)}=G_{ij}-\frac{G_{ik}G_{kj}}{G_{kk}}.
\end{align}
\end{enumerate}
Same identities also hold for $\bG_{[i]}$ and $\bR$ by replacing $\bL$ with $\bL_{[i]}$ and $\bLambda$, respectively.
\end{lemma}

Before stating and proving the local laws of $\bG(z)$, we first provide the (almost) sharp estimates on $\be_i^\top\bG(z)\bv$ and $\be_i^\top\bG_{[j]}(z)\bv$ for any $i\in [n]$ and deterministic vector $\bv$ in the lemma below. 

 \begin{lemma}\label{lemma:Linfty norms}
 Under the conditions of \Cref{thm:local laws of R}, for any deterministic unit vector $\bv\in \C^n$ and any $z\in S(\fC)$,  we have that
     \begin{align}
         \max_{i\in[n]}|\be_i^\top\bG(z)\bv|&\lesssim{\frac{\xi}{\sqrt n|z|^2}}+\frac{\|\bv\|_\infty}{|z|},\label{eq:eiGv}\\
        \max_{i,j\in[n]}|\be_i^\top\bG_{[j]}(z)\bv|&\lesssim{\frac{\xi}{\sqrt n|z|^2}}+\frac{\|\bv\|_\infty}{|z|},\label{eq:eiG[i]v}\\
        \max_{i\in[n]}\left|\frac{\be_i^\top\bG_{[i]}(z)\bv}{(G_{[i]})_{ii}}\right| &\lesssim {\frac{\xi}{\sqrt n|z|}}+\|\bv\|_\infty \label{eq:eiG[i]v_div}
     \end{align}
     with high probability.
 \end{lemma}
 
We are now ready to state and prove the \textit{three local laws} for $\bG$ in Theorems \ref{prop:ekG-UpsilonvLinfty}--\ref{prop:eiLWL(G-Upsilon)v} below. These refined local law results serve as the key tools for the proofs of our main results presented in \Cref{sec:mainresult}. 


\begin{theorem}\label{prop:ekG-UpsilonvLinfty}
Under the conditions of \Cref{thm:local laws of R}, for each constant $D>0$, there exists some constant $C_8>0$ such that for any deterministic unit vector $\bv\in \C^n$,  the events
     \begin{align}
		&\bigcap_{z\in S(\fC)}\biggl\{\max_{i\in[n]}|\be_i^\top(\bG(z)-\bUpsilon(z))\bv|\leq C_8\frac{\xi}{|z|}\left(\frac{1}{\sqrt n|z|}+\frac{\|\bv\|_{\infty}}{q\sdeg}\right)\biggr\},\label{eq:eiG-UpsilonvLinfty}\\
		&\bigcap_{z\in S(\fC)}\biggl\{\max_{i,j\in[n]}|\be_i^\top(\bG_{[j]}(z)-\bUpsilon(z))\bv|\leq C_8\frac{\xi}{|z|}\left(\frac{1}{\sqrt n|z|}+\frac{\|\bv\|_{\infty}}{q\sdeg}\right)\biggr\},\label{eq:eiGj-UpsilonvLinfty}\\
	&\bigcap_{z\in S(C_0)}\biggl\{\max_{i\ne j\in[n]}\left|\be_j^\top\big(\bG_{[i]}^{(i)}(z)-\bUpsilon^{(i)}(z)\big)\bv\right|\leq C_8\frac{\xi}{|z|}\left(\frac{1}{\sqrt n|z|}+\frac{\|\bv\|_{\infty}}{q\sdeg}\right)\biggr\} \label{eq:eiG(i)-UpsilonvLinfty}
    \end{align}
    hold with probability at least $1-n^{-D}$.
 \end{theorem}


\begin{theorem}\label{prop:uG-Upsilonv anisotropic}
Under the conditions of \Cref{thm:local laws of R}, for each constant $D>0$, there exists some constant $C_9>0$ such that for any deterministic unit vectors $\bu,\bv\in \C^n$,  the event
\begin{equation}\label{eq:uG-Upsilonv anisotropic}
    \bigcap_{z\in S({\fC})}\left\{|\bu^\top(\bG(z)-\bUpsilon(z))\bv|\le  C_9\frac{\xi}{q|z|}\left( \frac{1}{|z|\sdeg}+\frac{\xi}{q\sdeg^{2}} +\|\bu\|_\infty\wedge \|\bv\|_\infty\right)\right\}
\end{equation}
 holds with probability at least $1-n^{-D}$.
 \end{theorem}

\begin{theorem}\label{prop:eiLWL(G-Upsilon)v}
Under the conditions of \Cref{thm:local laws of R}, for each constant $D>0$, there exists some constant $C_{10}>0$ such that for any deterministic unit vector $\bv\in \C^n$,  the event
\begin{equation}\label{eq:eiLWL(G-Upsilon)v}
\begin{split}
    \bigcap_{z\in S(\fC)} & \bigg\{\max_{i\in[n]}\left|\be^\top_i\bcW (\bG-\bUpsilon)\bv\right| \leq {C_{10}} \left(\left(\frac{1}{|z|}+\frac{\xi}{q\sdeg}\right)\frac{\xi}{\sqrt n |z|}\right. \\
    &\quad \left. +\left(\frac{1}{|z|}+\frac{1}{q\sdeg}\right)\frac{\|\bv\|_\infty}{|z|}\right)\bigg\}
    \end{split}
\end{equation}
holds with probability at least $1-n^{-D}$.
\end{theorem}

\section{Proofs of Theorems \ref{thm:eigenvalue rescaled}--\ref{prop:eiLWL(G-Upsilon)v} and Corollaries \ref{example:CLT entry}--\ref{cor:CLT_evector}} \label{Sec.newA}

In this section, we will provide the complete proofs for our main results in Theorems \ref{thm:eigenvalue}--\ref{thm:K_0}. As mentioned in Section \ref{sec.mainresu.rescaledmodel}, Theorems \ref{thm:eigenvalue}--\ref{thm:K_0} in the main text are equivalent to Theorems \ref{thm:eigenvalue rescaled}--\ref{thm:K_0 rescaled}, respectively, that are stated under the setting of the rescaled model. Thus to this end, it remains to prove Theorems \ref{thm:eigenvalue rescaled}--\ref{prop:eiLWL(G-Upsilon)v}.

We start by providing a sketch of the main ideas of the technical analyses that will be exploited in the proofs. Our proofs will be mainly based on the key estimates provided in \eqref{eq:vv0}--\eqref{eq:vV22}
below, which follow from \Cref{lemma:Upsilon} and the local laws established in Theorems \ref{prop:ekG-UpsilonvLinfty} and \ref{prop:uG-Upsilonv anisotropic}. 
Specifically, let us consider the spectral domain $S(\fC)$ with $\fC\gg 1$. Then in light of \Cref{lemma:Upsilon} and the fact of $\bv_k^\top \bV_{-k}=0$, we have that for all $z\in S(\fC)$, 
\begin{equation}\label{eq:vv0}
\begin{split}
    \bv_k^{\top}\bUpsilon(z)\bv_k=-z^{-1}+ O &(|z|^{-3}),\quad \bV_{-k}^\top\bUpsilon(z)\bV_{-k}=-z^{-1}\bI+O(|z|^{-3}),\\
    & \bv_{k}^\top\bUpsilon(z)\bV_{-k}=O(|z|^{-3}),
\end{split}      
\end{equation}
where in the second and third expressions above, $O(|z|^{-3})$ denotes a matrix $\bcE$ and a vector $\boldsymbol{\varepsilon}$ satisfying $\|\bcE\|=O(|z|^{-3})$ and $|\boldsymbol{\varepsilon}|=O(|z|^{-3})$, respectively. Further, it follows from Theorems \ref{prop:ekG-UpsilonvLinfty} and \ref{prop:uG-Upsilonv anisotropic} that the estimates 
\begin{align}
		& \max_{i\in[n]}|\be_i^\top(\bG(z)-\bUpsilon(z))\bv_k|\lesssim\frac{\xi}{|z|}\bigg(\frac{1}{\sqrt n|z|}+\frac{\|\bv_k\|_\infty}{q\sdeg}\bigg), \label{eq:eiG-Upsilonuk}\\
	& \max_{i\in[n]}|\be_i^\top(\bG(z)-\bUpsilon(z))\bV_{-k}|\lesssim\frac{\sqrt K\xi}{|z|}\bigg(\frac{1}{\sqrt n|z|}+\frac{\|\bV_{-k}\|_{\max}}{q\sdeg}\bigg), \label{eq:eiG-UpsilonU-k}\\
 & |\bv_k^\top(\bG(z)-\bUpsilon(z))\bv_k|\lesssim \frac{\xi}{q|z|}\bigg(\frac{1}{|z|\sdeg}+\frac{\xi}{q\sdeg^{2}}+\|\bv_k\|_\infty \bigg), \label{eq:vv}\\
& \|\bV_{-k}^\top(\bG(z)-\bUpsilon(z))\bV_{-k}\|\lesssim \frac{K\xi}{q|z|}\bigg(\frac{1}{|z|\sdeg}+\frac{\xi}{q\sdeg^{2}}+\|\bV_{-k}\|_{\max}\bigg), \label{eq:VV}\\
& \|\bv_k^\top(\bG(z)-\bUpsilon(z))\bV_{-k}\|\lesssim \frac{\sqrt K\xi}{q|z|}\bigg(\frac{1}{|z|\sdeg}+\frac{\xi}{q\sdeg^{2}}+\|\bV\|_{\max}\bigg) \label{eq:vV}
\end{align}
hold w.h.p. uniformly over $z\in S(\fC)$. Combining \eqref{eq:vv0} with \eqref{eq:vv}--\eqref{eq:vV} and using \eqref{eq:condK}, we can obtain that for all $|z|\gtrsim |\delta_{K_0}| $,
\begin{equation}\label{eq:vv22}
 |\bv_k^\top \bG(z) \bv_k|\lesssim  |z|^{-1},\quad 
 \|\bV_{-k}^\top \bG(z) \bV_{-k}\|\lesssim |z|^{-1},
\end{equation}
\begin{equation}\label{eq:vV22}
 \|\bv_k^\top \bG(z) \bV_{-k}\|\lesssim \frac{1}{|z|^3}+\frac{\sqrt K\xi}{q|z|}\bigg(\frac{1}{|z|\sdeg}+\frac{\xi}{q\sdeg^{2}}+\|\bV\|_{\max}\bigg)
\end{equation}
with high probability.

\subsection{Proof of \Cref{thm:eigenvalue rescaled}}
\label{The estimation of the spiked eigenvalues}


The key ingredient of the proof is to show that $\hdelta_k$ satisfies the same equation as in (\ref{eq:t_k def}) but with $\bUpsilon(x)$ replaced by $\bG(x)$; see  \eqref{eq:equdeltak} below. Then taking a subtraction of (\ref{eq:t_k def}) and \eqref{eq:equdeltak} and applying the local laws in \eqref{eq:vv}--\eqref{eq:vV}, we can derive the desired conclusion. 
Specifically, combining the eigengap condition \eqref{eq:eigengap} in Assumption \ref{main_assm rescaled} with \eqref{eq:vv0}
, we see that there exists a constant $C>0$ such that 
\begin{equation}\label{eq:denominator}
 \left\|\delta_k^{-1}(\bDelta_{-k}^{-1}+\bV_{-k}^\top\bUpsilon(z)\bV_{-k})^{-1}\right\|\leq C
\end{equation}
for all $z\in\cI_k$. 
From (\ref{eq:t_k def}), (\ref{eq:vv0}), and (\ref{eq:denominator}), it holds that 
\begin{equation}\label{eq:tkdeltak}
    t_k=\delta_k+O(|\delta_k|^{-1}).
\end{equation}
Moreover, with the aid of (\ref{eq:VV}) and \eqref{eq:denominator}, we can deduce that w.h.p.,
\begin{align*}
	\left\|\frac{1}{{\bDelta}_{-k}^{-1}+\bV_{-k}^\top\bG(z)\bV_{-k}}-\frac{1}{{\bDelta}_{-k}^{-1}+\bV_{-k}^\top\bUpsilon(z)\bV_{-k}}\right\|\lesssim  |\delta_k|\frac{K\xi\pnk}{q} .
\end{align*}
In view of \eqref{eq:condK}, it follows that for all $z\in\cI_k$,
\begin{equation}\label{eq:denominator G}
	\left\|\frac{1}{\bDelta_{-k}^{-1}+\bV_{-k}^\top\bG(z)\bV_{-k}}\right\|\lesssim|\delta_k|
\end{equation}
w.h.p.

With an application of Weyl's inequality \citep{Weyl} and Proposition \ref{prop:W,R,G}, it holds that w.h.p., $\wh \delta_k\in \cal I_k$ for each $1\le k \le K_{0}$. We will make a useful claim that w.h.p., $\wh \delta_k$ satisfies the nonlinear equation
\begin{equation}\label{eq:equdeltak}
1+\delta_k\bv_k^\top\bG(\hdelta_k)\bv_k-\delta_k\bv_k^\top\bG(\hdelta_k)\bV_{-k}\frac{1}{\bDelta_{-k}^{-1}+\bV_{-k}^\top\bG(\hdelta_k)\bV_{-k}}\bV_{-k}^\top\bG(\hdelta_k)\bv_k=0.
\end{equation}
In fact, $\hdelta_k$ is a solution to equation $\det(\bX-z \bI)=0$ over $z\in \cal I_k$. Moreover, for all $|z|\gg 1$, $\bG(z)$ exists and is nonsingular with high probability by Proposition \ref{prop:W,R,G}. 
Hence, with the spectral decomposition $\bLambda^{-\alpha}\bH\bLambda^{-\alpha}=\bV\bDelta\bV^\top$ (recall \eqref{spectral decompositions}) and the identity $\det(\bI+\bA\bB)=\det(\bI+\bB\bA)$ for any conformable matrices $\bA$ and $\bB$, we observe that equation $\det(\bX-z \bI)=0$ is equivalent to
$$\det(\bG(z)^{-1}+\bLambda^{-\alpha}\bH\bLambda^{-\alpha})=0 \iff\det(\bDelta^{-1}+\bV^\top\bG(z)\bV)=0.$$
Let us write the second equation above as 
\begin{align*}
    \det \begin{pmatrix}
        \delta_k^{-1}+\bv_k^\top\bG(z)\bv_k & \bv_k^\top\bG(z)\bV_{-k}\\
        \bV_{-k}^\top\bG(z)\bv_k & \bDelta^{-1}_{-k}+\bV_{-k}^\top\bG(z)\bV_{-k}
    \end{pmatrix} = 0.
\end{align*}
Using Schur's formula for the determinant, this equation is further equivalent to $\det(\bDelta^{-1}_{-k}+\bV_{-k}^\top\bG(z)\bV_{-k})=0$ or
\begin{align*}
1+\delta_k\bv_k^\top\bG(z)\bv_k-\delta_k\bv_k^\top\bG(z)\bV_{-k}\frac{1}{\bDelta_{-k}^{-1}+\bV_{-k}^\top\bG(z)\bV_{-k}}\bV_{-k}^\top\bG(z)\bv_k=0.
\end{align*}
In light of \eqref{eq:denominator G}, we see that matrix $\bDelta_{-k}^{-1}+\bV_{-k}^{\top}\bG({\hdelta}_k)\bV_{-k}$ is nonsingular w.h.p., which entails that equation \eqref{eq:equdeltak} indeed holds w.h.p. 

We are now ready to establish \eqref{eqn:t_k-gamma_k}. Subtracting (\ref{eq:t_k def})
from (\ref{eq:equdeltak}), 
we obtain that w.h.p.,
\begin{align}\label{eqn:t_k-gamma_k 1}
\begin{split}
    \bv_k^\top\big[\bG(\hdelta_k)-\bUpsilon(t_k)\big]\bv_k &= \bv_k^\top\bG(\hdelta_k)\bV_{-k}\frac{1}{\bDelta^{-1}_{-k}+\bV_{-k}^\top\bG(\hdelta_k)\bV_{-k}}\bV_{-k}^\top\bG(\hdelta_k)\bv_k\\
    &-\bv_k^\top\bUpsilon(t_k)\bV_{-k}\frac{1}{\bDelta_{-k}^{-1}+\bV_{-k}^\top\bUpsilon(t_k)\bV_{-k}}\bV_{-k}^\top\bUpsilon(t_k)\bv_k.
\end{split}
\end{align}
From (\ref{eq:vv}), it holds that w.h.p.,
\begin{align}\label{eqn:t_k-gamma_k 1.1}
\begin{split}
    \bv_k^\top\big[\bG(\hdelta_k)-\bUpsilon(t_k)\big]\bv_k=&\bv_k^\top\big[\bUpsilon(\hdelta_k)-\bUpsilon(t_k)\big]\bv_k+O\bigg(\frac{\xi\pnk}{q|\delta_k|}\bigg).
\end{split}
\end{align}
Using (\ref{eq:VV})--(\ref{eq:denominator}) and (\ref{eq:denominator G}), we can deduce that w.h.p.,
\begin{align}\label{eqn:t_k-gamma_k 1.2}
\begin{split}
    &~\bv_k^\top\bG(z)\bV_{-k}\frac{1}{\bDelta^{-1}_{-k}+\bV_{-k}^\top\bG(z)\bV_{-k}}\bV_{-k}^\top\bG(z)\bv_k\\
    &-~\bv_k^\top\bUpsilon(z)\bV_{-k}\frac{1}{\bDelta^{-1}_{-k}+\bV_{-k}^\top\bUpsilon(z)\bV_{-k}}\bV_{-k}^\top\bUpsilon(z)\bv_k\\
     &\lesssim ~ \frac{\sqrt K\xi}{q}\pn(z)\left(\frac{1}{|z|^3}+\frac{\sqrt K}{|z|^5}+\frac{\sqrt{K}\xi}{q|z|}\pn(z)\right),
\end{split}
\end{align}
where we have used the notation in \eqref{eq:Apsink} with $\delta_k$ replaced by a general $z$, and the asymptotic bound above is understood \textit{implicitly} for the absolute value of the quantity involved (for notational simplicity). 
For the deterministic term, with the aid of \eqref{eq:Upsilon} and \eqref{eq:vv0}, we can rewrite it as
\begin{align}\label{eqn:t_k-gamma_k 2}
&~\bv_k^\top\bUpsilon(\hdelta_k)\bV_{-k}\frac{1}{\bDelta^{-1}_{-k}+\bV_{-k}^\top\bUpsilon(\hdelta_k)\bV_{-k}}\bV_{-k}^\top\bUpsilon(\hdelta_k)\bv_k\\
    &=~\bv_k^\top\bUpsilon(t_k)\bV_{-k}\frac{1}{\bDelta^{-1}_{-k}+\bV_{-k}^\top\bUpsilon(t_k)\bV_{-k}}\bV_{-k}^\top\bUpsilon(t_k)\bv_k+O\bigg(\frac{|\hdelta_k-t_k|}{|\delta_k|^6}\bigg).\nonumber
\end{align}

Plugging \eqref{eqn:t_k-gamma_k 1.1}, \eqref{eqn:t_k-gamma_k 1.2} (with $z=\hdelta_k$), and (\ref{eqn:t_k-gamma_k 2}) into \eqref{eqn:t_k-gamma_k 1} and using \eqref{eq:condK}, we can obtain that
\begin{equation}\label{eqn:t_k-gamma_k 3}
    \left|\bv_k^\top\big[\bUpsilon(\hdelta_k)-\bUpsilon(t_k)\big]\bv_k \right| \lesssim\frac{\xi}{q|\delta_k|}\pnk\left(1+\frac{K }{ |\delta_k|^4}\right)+\frac{|\hdelta_k-t_k|}{|\delta_k|^6}
\end{equation}
w.h.p. By Corollary 3.4 in \cite{ajanki2017universality}, $M_i(z)$ is the Stieltjes transform of a finite measure $\mu_i$ on $\mathbb R$ given by 
\begin{equation}
    M_i(z)=\int_{\mathbb R}\frac{\mu_i(\dd x)}{x-z},
\end{equation}
where the support of $\mu_i$ satisfies $\supp\{\mu_i\}\subset[-2\sqrt{\fM},2\sqrt{\fM}]$. Thus, $M_i(x)$ is strictly increasing in $x$ on $(-\infty,-2\sqrt{\fM}]$ and $[2\sqrt{\fM},+\infty)$, respectively. Such property implies that
\[
    \left|\bv_k^\top(\bUpsilon(\hdelta_k)-\bUpsilon(t_k))\bv_k\right|\gtrsim \frac{|\hdelta_k-t_k|}{|\delta_k|^2}.
\]
Therefore, plugging this result into (\ref{eqn:t_k-gamma_k 3}) and solving for $|\hdelta_k-t_k|$ yield the desired conclusion, which completes the proof of \Cref{thm:eigenvalue rescaled}.

\subsection{Proof of \Cref{thm:projection rescaled}} \label{sec:The estimations about the spiked eigenvectors}

We start by describing the main ideas of the proof. To study the asymptotic behavior of the spiked eigenvectors, let us define the contour
\begin{equation}\label{eq:C_k definition}
	\cC_k:=\{z\in\mathbb C:|z-t_k|=ct_k\},
\end{equation}
where $c=c(\epsilon_0)>0$ is small enough such that $(1\pm c)t_k\in\cI_k$. Under part (iii) of Assumption \ref{main_assm rescaled} and Theorem \ref{thm:eigenvalue rescaled}, contour $\cC_k$ encloses $\hdelta_k$ and no other eigenvalues of $\bX$ w.h.p. 
Then using Cauchy's integral formula, we can estimate the projections of $\bhv_k$ by evaluating the loop integral $\int_{\cC_k}\bu^\top(\bX-z \bI)^{-1}\bv \dd z$ for any deterministic vectors $\bu$ and $\bv$ in $\R^n$. 
In particular, by taking $\bv=\bv_k$, we will obtain an estimate of the quadratic form
\beq\label{eq:quadform}
\bu^\top(\bL/\bLambda)^{-\alpha}\bhv_k\bhv_k^\top(\bL/\bLambda)^{-\alpha}\bv_k.
\eeq
If we further take $\bu=\bv_k$, we can get an estimate of $\bv_k^\top(\bL/\bLambda)^{-\alpha}\bhv_k$. Then dividing \eqref{eq:quadform} by $\bv_k^\top(\bL/\bLambda)^{-\alpha}\bhv_k$ will conclude the proof.


Specifically, we first establish a contour integral representation for the quadratic form \eqref{eq:quadform} above. 
To this end, let us define a new resolvent
\begin{align}
	\bG_k(z)&:=\left(\bLambda^{-\alpha}\btX\bLambda^{-\alpha}-\delta_k\bv_k\bv_k^\top-z(\bL/\bLambda)^{2\alpha}\right)^{-1} \nonumber\\
 &=\left( \bcW +\bV_{-k}\bDelta_{-k}\bV_{-k}^\top-z(\bL/\bLambda)^{2\alpha}\right)^{-1} \nonumber\\
	&=\bG(z)-\bG(z)\bV_{-k}\frac{1}{\bDelta^{-1}_{-k}+\bV_{-k}^\top\bG(z)\bV_{-k}}\bV_{-k}^\top\bG(z),\label{eq:Gk def}
\end{align}
where in the last step above, we have used the Woodbury matrix identity
\begin{equation}\label{eq:Woodbury's identity}
    (\bA+\bU\bB\bV)^{-1}=\bA^{-1}-\bA^{-1}\bU(\bB^{-1}+\bV\bA^{-1}\bU)^{-1}\bV\bA^{-1}
\end{equation}
for any nonsingular matrices $\bA, \bB$ and any matrices $\bU, \bV$. Then applying \eqref{eq:Woodbury's identity} again, we can write that
\begin{align*}
	(\bX-z \bI)^{-1}&=(\bL/\bLambda)^\alpha(\bG_k^{-1}(z)+\delta_k\bv_k\bv_k^\top)^{-1}(\bL/\bLambda)^\alpha\\
    &=(\bL/\bLambda)^\alpha\bigg(\bG_k(z)-\bG_k(z)\bv_k\frac{1}{\delta_k^{-1}+\bv_k^\top\bG_k(z)\bv_k}\bv_k^\top\bG_k(z)\bigg)(\bL/\bLambda)^\alpha.
\end{align*}

With an application of Cauchy's integral formula, we can deduce that 
\begin{align*}
    &\bu^\top(\bL/\bLambda)^{-\alpha}\bhv_k\bhv_k^\top(\bL/\bLambda)^{-\alpha}\bv=-\frac{1}{2\pi i}\oint_{\cC_k}\bu^\top(\bL/\bLambda)^{-\alpha}(\bX-z)^{-1}(\bL/\bLambda)^{-\alpha}\bv \dd z\\
&=-\frac{1}{2\pi i}\oint_{\cC_k}\bu^\top \bigg(\bG_k(z)-\bG_k(z)\bv_k\frac{1}{\delta_k^{-1}+\bv_k^\top\bG_k(z)\bv_k}\bv_k^\top\bG_k(z)\bigg) \bv \dd z
\end{align*}
for any deterministic vectors $\bu$ and $\bv$. Using Weyl's inequality and Proposition \ref{prop:W,R,G}, we obtain that w.h.p.,
\[
	\left|\lambda_l( \bcW +\bV_{-k}\bDelta_{-k}\bV^\top_{-k})-\lambda_l(\bV_{-k}\bDelta_{-k}\bV^\top_{-k})\right|\leq\|\bcW\|=O(1),
\]
where $\lambda_l(\cdot)$ denotes the $l$th eigenvalue of a given symmetric matrix. Then due to the eigengap condition in \eqref{eq:eigengap}, contour $\cC_k$ does not enclose any eigenvalue of $\bcW +\bV_{-k}\bDelta_{-k}\bV^\top_{-k}$, i.e., $\bG_k(z)$ is nonsingular in the regime enclosed by $\cC_k$. Thus, it holds that w.h.p.,
\[
	\oint_{\cC_k}\bx^\top   \bG_k(z)   \by \dd z=0,
\]
which in turn leads to 
\begin{align}\label{eq:oint}
\begin{split}
	\bu^\top(\bL/\bLambda)^{-\alpha}\bhv_k\bhv_k^\top(\bL/\bLambda)^{-\alpha}\bv=\frac{1}{2\pi i}\oint_{\cC_k}\frac{\bu^\top\bG_k(z)\bv_k\bv_k^\top\bG_k(z)\bv}{\delta_k^{-1}+\bv_k^{\top}\bG_k(z)\bv_k}\dd z.
\end{split}
\end{align}

It remains to estimate the right-hand side (RHS) of \eqref{eq:oint} above. Since $\bUpsilon$ is a deterministic approximation of $\bG(z)$ due to the local laws, $\bUpsilon_k$ in \eqref{eq:Upsilonk def} is the corresponding deterministic approximation of $\bG_k(z)$. 
Let us now control the differences between some bilinear forms of $\bG_k(z)$ and $\bUpsilon_k(z)$ using the local law established in \Cref{prop:uG-Upsilonv anisotropic}.
Applying \eqref{eq:vv} and \eqref{eqn:t_k-gamma_k 1.2}, it holds that w.h.p., 
\begin{align}\label{eq:ukGk-Upsilonkuk}
\begin{split}
	 &\bv_k^\top\left(\bG_k(z)-\bUpsilon_k(z)\right)\bv_k\\
	&\lesssim \frac{\xi}{q|\delta_k|}\pnk + \frac{\sqrt K\xi}{q}\pnk\left(\frac{1}{|\delta_k|^3}+\frac{\sqrt K}{|\delta_k|^5}+\frac{\sqrt{K}\xi}{q|\delta_k|}\pnk\right)\\
 &\lesssim \frac{\xi}{q|\delta_k|}\pnk\left(1+\frac{K }{ |\delta_k|^4}\right) 
 \end{split}
\end{align}
uniformly in $z\in \cC_k$, where we have used \eqref{eq:condK} in the second step. 
With the aid of \eqref{eq:Upsilon} and \eqref{eq:denominator}, we can deduce that
\begin{equation}\label{eq:vvx0}
 \bu^\top \bUpsilon(z) \bv_k =-{\bu^\top \bv_k}/{ z} + \OO(|z|^{-3}), \quad \|\bu^\top \bUpsilon(z) \bV_{-k}\|\lesssim {\|\bu^\top \bV_{-k}\|}/{ |z|} + |z|^{-3},
\end{equation}
\begin{equation}\label{eq:Upsilonk}
 \bu^\top \bUpsilon_k(z) \bv_k = - \bu^\top \bv_k/z + \OO(|z|^{-3}),
 \quad \bv_k^\top\bUpsilon_k'(z)\bv_k = z^{-2} +\OO(|z|^{-4}).
\end{equation}
From \eqref{eq:uG-Upsilonv anisotropic}, it follows that with high probability,
\begin{equation}\label{eq:vvx}
 |\bu^\top(\bG(z)-\bUpsilon(z))\bv_k|\lesssim \frac{\xi}{q|z|}\left(\frac{1}{|z|\sdeg}+\frac{\xi}{q\sdeg^{2}}+\|\bv_k\|_\infty \right),
\end{equation}
\begin{equation}\label{eq:vVx}
 \|\bu^\top(\bG(z)-\bUpsilon(z))\bV_{-k}\|\lesssim \frac{\sqrt K\xi}{q|z|}\left(\frac{1}{|z|\sdeg}+\frac{\xi}{q\sdeg^{2}}+\|\bV_{-k}\|_{\max}\right)
\end{equation}
 uniformly in $z \in S(\fC)$. 

Combining the estimates \eqref{eq:vvx0}--\eqref{eq:vVx} with \eqref{eq:vv0}, (\ref{eq:vv})--\eqref{eq:denominator}, and \eqref{eq:denominator G}, we can obtain that for any deterministic unit vector $\bu$, 
\begin{align}\label{eq:x(G_k-Upsilon_k)v_k}
\begin{split}
     \bu^\top(\bG_k(z)-\bUpsilon_k(z))\bv_k &\lesssim \frac{\xi}{q|\delta_k|}\pnk +  \frac{ K\xi^2}{q^2 |\delta_k|}\pnk^2\\
    &+  \frac{\sqrt K\xi}{q|\delta_k|}\pnk \left( {\|\bu^\top \bV_{-k}\|} +\frac{1}{|\delta_k|^2}\right)\left(1+\frac{\sqrt K}{|\delta_k|^2}\right)  \\
    &\lesssim \frac{\xi}{q|\delta_k|}\pnk \left[1+\left( {\|\bu^\top \bV_{-k}\|} +\frac{1}{|\delta_k|^2}\right)\left(\sqrt K+\frac{K}{|\delta_k|^2}\right) \right]
\end{split}
\end{align}
uniformly in $z\in \cC_k$ w.h.p. Then using \eqref{eq:Upsilonk} and \eqref{eq:x(G_k-Upsilon_k)v_k}, we can immediately get that w.h.p.,
\begin{align}\label{eq:xUpsilon_kv_k}
    \bu^\top\bG_k(z)\bv_k\lesssim  \frac{|\bu^\top \bv_k|}{|\delta_k|}+\frac{1}{|\delta_k|^3}+\frac{\xi}{q|\delta_k|}\pnk \left(1+\sqrt K {\|\bu^\top \bV_{-k}\|}  \right)
\end{align}
uniformly in $z\in \cal C_k$. 

We now estimate \eqref{eq:oint} for the case of $\bu=\bv=\bv_k$. By \eqref{eq:Upsilonk}, we see that for all $z\in\cC_k$,
\begin{equation}\label{eq:v_kUpsilon_kv_k}
    \bv_k^\top\bUpsilon_k(z)\bv_k=-z^{-1}+O(|z|^{-3}),
\end{equation}
which entails that 
\begin{equation}\label{eq:deno}
	\min_{z\in\cC_k}\left|1+\delta_k\bv_k^\top\bUpsilon_k(z)\bv_k\right|=\frac{c}{1+c}+o(1).
\end{equation} 
With (\ref{eq:ukGk-Upsilonkuk}) and (\ref{eq:deno}), we can deduce that w.h.p.,  
\begin{align}
&	\bv_k^\top(\bL/\bLambda)^{-\alpha}\bhv_k\bhv_k^\top(\bL/\bLambda)^{-\alpha}\bv_k  \nonumber\\
&=\frac{1}{2\pi i}\oint_{\cC_k}\frac{(\bv_k^\top\bG_k(z)\bv_k)^2}{\delta_k^{-1}+\bv_k^\top\bG_k(z)\bv_k}\dd z =\frac{1}{2\pi i\delta_k}\oint_{\cC_k}\frac{1}{1+\delta_k\bv_k^\top\bG_k(z)\bv_k}\dd z \nonumber\\
	&=\frac{1}{2\pi i\delta_k}\oint_{\cC_k}\frac{1}{1+\delta_k\bv_k^\top\bUpsilon_k(z)\bv_k}\dd z+\OO\left(\frac{\xi}{q}\pnk\left(1+\frac{K }{ |\delta_k|^4}\right)\right) \nonumber\\
 &=\frac{1}{\delta_k^2\bv_k^\top\bUpsilon_k'(t_k)\bv_k}+\OO\left(\frac{\xi}{q}\pnk\left(1+\frac{K }{ |\delta_k|^4}\right)\right),\label{eq:vkvkhat}
\end{align}
where we have used the residue theorem from complex analysis at the pole $z = t_k$ in the last step above.
Moreover, it follows from \eqref{eq:tkdeltak} and \eqref{eq:Upsilonk} that 
\[
	\delta_k^2\bv_k^\top\bUpsilon_k'(t_k)\bv_k=\delta_k^2/t_k^2+O\left( {\delta_k^2}/{t_k^4}\right)=1+O(\delta_k^{-2}),
\]
which yields the first estimate in \eqref{eq:projectionvk2}. Hence, taking the square root of \eqref{eq:vkvkhat}, we can obtain \eqref{eq:projectionvk}.



We next take $\bv=\bv_k$ in (\ref{eq:oint}). With the aid of (\ref{eq:xUpsilon_kv_k}), \eqref{eq:ukGk-Upsilonkuk}, and \eqref{eq:deno}, we can show that w.h.p.,
\begin{align}
    &\bu^\top(\bL/\bLambda)^{-\alpha}\bhv_k\bhv_k^\top(\bL/\bLambda)^{-\alpha}\bv_k 
    =-\frac{1}{2\pi i}\oint_{\cC_k}\frac{\bu^\top\bG_k(z)\bv_k}{1+\delta_k\bv_k^{\top}\bG_k(z)\bv_k}\dd z \nonumber\\
   & = -\frac{1}{2\pi i}\oint_{\cC_k}\frac{\bu^\top\bG_k(z)\bv_k}{1+\delta_k\bv_k^{\top}\bUpsilon_k(z)\bv_k}\dd z+\cal E_{\bu} = -\frac{\bu^\top\bG_k(t_k)\bv_k}{\delta_k\bv_k^\top\bUpsilon_k'(t_k)\bv_k}+\cal E_{\bu} ,\label{eq:xv_khatv_khatv_k}
\end{align}
where $\cal E_{\bu}$ is a random error that can be bounded w.h.p. as
\begin{align*}
    \cal E_{\bu} \lesssim &~ \frac{\xi}{q}\pnk\left(1+\frac{K }{ |\delta_k|^4}\right)  \left[ |\bu^\top \bv_k| +\frac{1}{|\delta_k|^2}+\frac{\xi}{q}\pnk \left(1+\sqrt K {\|\bu^\top \bV_{-k}\|}  \right)\right]  ,
\end{align*}
and we have used the residue theorem at the pole $z = t_k$ in the last step above. Then an application of  \eqref{eq:x(G_k-Upsilon_k)v_k} and \eqref{eq:Upsilonk} yields that w.h.p.,
\begin{align}
    &\bu^\top(\bL/\bLambda)^{-\alpha}\bhv_k\bhv_k^\top(\bL/\bLambda)^{-\alpha}\bv_k   = -\frac{\bu^\top\bUpsilon_k(t_k)\bv_k}{\delta_k\bv_k^\top\bUpsilon_k'(t_k)\bv_k}+\cal E_{\bu} \label{eq:xv_khatv_khatv_k2}\\
    &\qquad +\OO\left\{\frac{\xi}{q}\pnk \left[1+\left( {\|\bu^\top \bV_{-k}\|} +\frac{1}{|\delta_k|^2}\right)\left(\sqrt K+\frac{K}{|\delta_k|^2}\right) \right]\right\}\nonumber\\
    &=-\frac{\bu^\top\bUpsilon_k(t_k)\bv_k}{\delta_k\bv_k^\top\bUpsilon_k'(t_k)\bv_k}+\OO\left\{\frac{\xi}{q}\pnk\left[1+\frac{K }{ |\delta_k|^4} + {\|\bu^\top \bV_{-k}\|} \left(\sqrt K+\frac{K}{|\delta_k|^2}\right)\right] \right\}, \nonumber
\end{align}
where we have used \eqref{eq:condK} to simplify the error term.
Dividing \eqref{eq:xv_khatv_khatv_k2} by \eqref{eq:projectionvk} gives \eqref{eq:projection}.
Finally, plugging \eqref{eq:tkdeltak} into the first expression in \eqref{eq:Upsilonk} results in the second estimate in \eqref{eq:projectionvk2}. This concludes the proof of \Cref{thm:projection rescaled}.


\subsection{Proof of \Cref{thm:main thm rescaled}} \label{Sec.proof.thm3}

We first observe that the estimate in \eqref{eq:main thm} is an immediate consequence of the proposition below under the extra assumption (\ref{eq:main thm assumption}). 

\begin{proposition}\label{prop:main_strong}
Under Condition \ref{cond:rescaled model} and Assumption \ref{main_assm rescaled}
, for each $1\le k\le K_0$ and $i\in [n]$, it holds that w.h.p.,   
    \begin{align}\label{eq:main thm without assumption}
	\begin{split}
	\hv_k(i) =&~(\Lambda_i/L_i)^{\alpha}v_k(i)+\frac{1}{t_kL_i^\alpha}\sum_{j\in[n]}{W}_{ij}{\Lambda}_j^{-\alpha}v_k(j) \\
    &+O\left(\|\bv_k\|_\infty \left(\frac{1}{|\delta_k|^2}+\frac{\xi \pnk}{q }\right) + \|\bV_{-k}\|_{\max}  \left(\frac{\sqrt K}{|\delta_k|^2}+\frac{K\xi}{q}\pnk\right)\right)\\
    &+ O\left(\frac{\xi}{\sqrt n |\delta_k|}\left(\frac{1+\sqrt{K}|\delta_k|^{-1}}{|\delta_k|}  +\frac{\xi\sdeg^{-1}  +K\xi \pnk}{q}\right) \right).
	\end{split}
\end{align}

\end{proposition}

We next aim to prove \eqref{eq:main thm expand}. With an application of \Cref{prop:Lambda and E} and the Taylor expansion, it holds that w.h.p.,
$$ \left(L_i/\Lambda_i\right)^{-\alpha} = 1 - \frac{\al}{\Lambda_i\sdeg}\bigg( \frac{1}{q} \sum_j W_{ij} +\frac{\tau_i}{nq} \sum_{j,l} W_{jl}\bigg) + \OO\left(\frac{\xi^2}{q^2\sdeg^{2}} \right). $$
From \eqref{eq:large_dev1}, it follows that w.h.p., 
$$ \sum_{l\in[n]}{W}_{il}{\Lambda}_l^{-\alpha}v_k(l) \lesssim \frac{\xi}{q}\|\bv_k\|_\infty + \frac{\xi}{\sqrt{n}}.$$
Then using \Cref{prop:Lambda and E} and the above two estimates, we can obtain that  
\begin{align*}
	 \hv_k(i)&= v_k(i)- \frac{\al}{\Lambda_i\sdeg}\bigg( \frac{1}{q} \sum_{j\in[n]} W_{ij} +\frac{\tau_i}{nq} \sum_{j,l\in[n]} W_{jl}\bigg)v_k(i) +\frac{1}{t_k}\sum_{j\in[n]}\Lambda_i^{-\al}{W}_{ij}{\Lambda}_j^{-\alpha}v_k(j)\\
	&+O\left(\|\bV\|_{\max}\left(\frac{\sqrt K}{|\delta_k|}+\frac{ K\xi}{q}\right)\left(\frac{1}{|\delta_k|\sdeg}+\frac{ \xi}{q\sdeg^{2}}\right)\right)\\
 &+O\left(\frac{\xi}{\sqrt n |\delta_k|}\left(\frac{1}{|\delta_k|}+\frac{\xi}{q\sdeg}\right)\right).
\end{align*}
Hence, reorganizing the terms above yields \eqref{eq:main thm expand}.

It remains to establish Proposition \ref{prop:main_strong} above. To this end, let us take $\bx=\be_i^{\top}$ and $\by=\bv_k$ in \eqref{eq:oint}. 
First, with the aid of \Cref{lemma:Upsilon} and \eqref{eq:denominator}, we see that that for all $z\in \cC_k$,
\begin{equation}\label{eq:YYmax}
    \be_i^\top\bUpsilon\bv_k\lesssim |\delta_k|^{-1}|\bv_k(i)|,\quad |\be_i^\top\bUpsilon\bV_{-k}| \lesssim  \sqrt{K}|\delta_k|^{-1}\|\bV_{-k}(i)\|_{\max} , 
\end{equation}
\begin{equation}\label{eq:YYmax2}
    \be_i^\top\bUpsilon_k\bv_k\lesssim {|\delta_k|}^{-1}{|\bv_k(i)|}+{\sqrt{K}|\delta_k|^{-3}\|\bV_{-k}(i)\|_{\max}}. 
\end{equation}
Then an application of \eqref{eq:eiG-Upsilonuk}--\eqref{eq:denominator}, \eqref{eq:denominator G}, and \eqref{eq:YYmax} gives that w.h.p., 
\begin{align}
&|\be_i^\top({\bG}_k-\bUpsilon_k)\bv_k|\leq|\be_i^\top(\bG-\bUpsilon)\bv_k|\nonumber\\
 &\quad+\left|\be_i^\top\bUpsilon\bV_{-k}\biggl(\frac{1}{\bDelta_{-k}^{-1}+\bV_{-k}^\top\bG\bV_{-k}}-\frac{1}{\bDelta_{-k}^{-1}+\bV_{-k}^\top\bUpsilon\bV_{-k}}\biggr)\bV_{-k}^\top\bUpsilon\bv_k\right|\nonumber\\
	&\quad+\left|\be_i^\top \bUpsilon\bV_{-k}\frac{1}{\bDelta_{-k}^{-1}+\bV_{-k}^\top\bG\bV_{-k}}\bV_{-k}^\top(\bUpsilon-\bG)\bv_k\right|\nonumber\\
    &\quad+\left|\be_i^\top(\bG-\bUpsilon)\bV_{-k}\frac{1}{\bDelta_{-k}^{-1}+\bV_{-k}^\top\bG\bV_{-k}}\bV_{-k}^\top\bG\bv_k\right|\nonumber\\
    &\lesssim \frac{\xi}{|\delta_k|}\left(\frac{1}{\sqrt n|\delta_k|}+\frac{\|\bv_k\|_\infty}{q\sdeg}\right) +  {\|\bV_{-k}\|_{\max}} \frac{K^{3/2}\xi}{q|\delta_k|^3}\pnk + \|\bV_{-k}\|_{\max} \frac{K\xi}{q|\delta_k|}\pnk  \nonumber\\
    & \quad+ {\sqrt K\xi} \left(\frac{1}{\sqrt n|\delta_k|}+\frac{\|\bV_{-k}\|_{\max}}{q\sdeg}\right) \left(\frac{1}{|\delta_k|^3}+\frac{\sqrt K\xi}{q|\delta_k|}\pnk\right)\nonumber\\
    &\lesssim  \left( 1+\frac{\sqrt{K}}{|\delta_k|^2}\right) \left(\frac{ \xi}{\sqrt n|\delta_k|^2}+  {\|\bV_{-k}\|_{\max}} \frac{K\xi}{q|\delta_k|} \pnk\right) +\frac{\xi \|\bv_k\|_\infty}{q|\delta_k|\sdeg}\nonumber\\
    &\quad+\|\bV_{-k}\|_{\max}\frac{K\xi}{q|\delta_k|\sdeg}\left(\frac{1}{|\delta_k|^2}+\frac{\xi\pnk}{q}\right) \label{eq:eiGktilde-Upsilonkuk}
\end{align}
uniformly in $z\in \cC_k\cup \cI_k$,  where we have used \eqref{eq:condK} in the last step above to simplify the estimate. 
In light of \eqref{eq:condK} and \eqref{eq:YYmax2}--(\ref{eq:eiGktilde-Upsilonkuk}), we can deduce that  w.h.p., 
\begin{align}
	 & \be_i^\top\bG_k(z)\bv_k = \be_i^\top\bUpsilon_k(z)\bv_k+ \be_i^\top\left[\bG_k(z)-\bUpsilon_k(z)\right]\bv_k \nonumber\\
 &\lesssim \frac{\|\bv_k\|_\infty}{|\delta_k|}+\left( 1+\frac{\sqrt{K}}{|\delta_k|^2}\right) \frac{\xi}{\sqrt n|\delta_k|^2}+  {\|\bV_{-k}\|_{\max}} \left(  \frac{\sqrt{K}}{|\delta_k|^{3}}+\frac{K\xi}{q|\delta_k|} \pnk\right) \label{eq:eiGktilde-Upsilonkuk2}
 \end{align} 
uniformly in $z\in \cC_k\cup \cI_k$.


We are now ready to establish the asymptotic expansion of $\bhv_k(i)$. 
Taking $\bx=\be_i$ and $\by=\bv_k$ in (\ref{eq:oint}) and applying (\ref{eq:eiGktilde-Upsilonkuk2}), \eqref{eq:ukGk-Upsilonkuk}, and \eqref{eq:deno}, we can show that w.h.p., 
\begin{align}
	&\be^\top_i(\bL/\bLambda)^{-\alpha}\bhv_k\bhv_k^\top(\bL/\bLambda)^{-\alpha}\bv_k
	=-\frac{1}{2\pi i}\oint_{\cC_k}\frac{\be_i^\top\bG_k(z)\bv_k}{1+\delta_k\bv_k^\top\bG_k(z)\bv_k}\dd z \nonumber\\
	&\quad=-\frac{1}{2\pi i}\oint_{\cC_k}\frac{\be_i^\top\bG_k(z)\bv_k}{1+\delta_k\bv_k^\top\bUpsilon_k(z)\bv_k}\dd z+\cal E_{i} = -\frac{\be_i^\top\bG_k(t_k)\bv_k}{\delta_k\bv_k^\top\bUpsilon'_k(t_k)\bv_k} + \cal E_i,\label{eq:eivkhatvkhatuk}
\end{align}
where $\cal E_{i}$ is a random error that can be bounded w.h.p. as
\begin{align*}
    \cal E_{i}&\lesssim |\delta_k|\frac{\xi \pnk}{q }\left(1+\frac{K }{ |\delta_k|^4}\right) \\
    &\quad\times \left[\frac{\|\bv_k\|_\infty}{|\delta_k|}+\left( 1+\frac{\sqrt{K}}{|\delta_k|^2}\right) \frac{\xi}{\sqrt n|\delta_k|^2}+  {\|\bV_{-k}\|_{\max}} \left(  \frac{\sqrt{K}}{|\delta_k|^{3}}+\frac{K\xi}{q|\delta_k|} \pnk\right)\right].
\end{align*}
Then dividing (\ref{eq:eivkhatvkhatuk}) by (\ref{eq:vkvkhat}) and using \eqref{eq:projectionvk2} and \eqref{eq:tkdeltak}, it holds that w.h.p.,
\begin{align}\label{eq:L/Lambda vkhati}
	(L_i/\Lambda_i)^{-\alpha}\hv_k(i)
 =- t_k  {\be_i^\top\bG_k(t_k)\bv_k}  +\cal E_i',
\end{align}
where $\cal E_{i}'$ is a random error satisfying that w.h.p.,
\begin{align*}
    \cal E_{i}'&\lesssim \left(\frac{1}{|\delta_k|}+\frac{\xi|\delta_k| \pnk}{q }\right)  \left[\frac{\|\bv_k\|_\infty}{|\delta_k|}+\left( 1+\frac{\sqrt{K}}{|\delta_k|^2}\right) \frac{\xi}{\sqrt n|\delta_k|^2}\right.\\
    &\quad\left. +  {\|\bV_{-k}\|_{\max}} \left(  \frac{\sqrt{K}}{|\delta_k|^{3}}+\frac{K\xi\pnk}{q|\delta_k|} \right)\right].
\end{align*}

Further, from the definition of $\bG_k$, it follows that w.h.p.,
\begin{align*}
	\be_i^\top(\bG(t_k)-\boldsymbol{G}_k(t_k))\bv_k&=\be_i^\top\bG(t_k)\bV_{-k}\frac{1}{\bDelta_{-k}^{-1}+\bV_{-k}^\top\bG(t_k)\bV_{-k}}\bV_{-k}^\top\bG(t_k)\bv_k\\
 &\lesssim \left(\sqrt{K}\|\bV_{-k}\|_{\max} + \frac{\sqrt K\xi}{\sqrt n|\delta_k|} \right)\left(\frac{1}{|\delta_k|^3}+\frac{\sqrt K\xi}{q|\delta_k|}\pnk\right), 
\end{align*} 
where we have used \eqref{eq:eiG-UpsilonU-k} and \eqref{eq:YYmax} to bound $\be_i^\top\bG(t_k)\bV_{-k}$, \eqref{eq:vV22} to bound $\bV_{-k}^\top\bG(t_k)\bv_k$, and \eqref{eq:denominator G} to bound the denominator.
Plugging the above estimate into \eqref{eq:L/Lambda vkhati} and using \eqref{eq:condK}, we can obtain that w.h.p.,
\begin{align}\label{eq:L/Lambda vkhati2}
	 (L_i/\Lambda_i)^{-\alpha}\hv_k(i)
 &=-t_k \be_i^\top\bG(t_k)\bv_k +O\left(\left(\frac{1}{|\delta_k|^2}+\frac{\xi \pnk}{q }\left(1+\frac{K}{|\delta_k|^4}\right)\right) {\|\bv_k\|_\infty}\right)\\
 &\quad+O\left(\sqrt{K}\|\bV_{-k}\|_{\max} + \frac{\sqrt K\xi}{\sqrt n|\delta_k|} \right)\left(\frac{1}{|\delta_k|^2}+\frac{\sqrt K\xi}{q}\pnk\right).\nonumber
\end{align} 

We next handle the first term on the RHS of \eqref{eq:L/Lambda vkhati2} 
\begin{align}\label{eq:tkeiGtildevk}
\begin{split}
	-t_k\be_i^\top\bG(t_k)\bv_k&=(L_i/\Lambda_i)^{-2\alpha} \be_i^\top t_k(\bL/\bLambda)^{2\alpha}\frac{1}{t_k(\bL/\bLambda)^{2\alpha}- \bcW }\bv_k\\
 &=(L_i/\Lambda_i)^{-2\alpha}v_k(i)- (L_i/\Lambda_i)^{-2\alpha} \be_i^\top \bcW \bG(t_k)\bv_k.
 \end{split}
\end{align}
Together with (\ref{eq:L/Lambda vkhati2}), it yields that w.h.p.,
\begin{align}\label{eq:vkhati}
\begin{split}
\hv_k(i) =&~(\Lambda_i/L_i)^{\alpha}v_k(i)-(\Lambda_i/L_i)^{\alpha} \be_i^\top\bcW \bG(t_k)\bv_k+O\left(\left(\frac{1}{|\delta_k|^2}+\frac{\xi \pnk}{q }\right)\left(1+\frac{K}{|\delta_k|^4}\right) {\|\bv_k\|_\infty}\right)\\
    &+O\left( \left(\sqrt{K}\|\bV_{-k}\|_{\max} + \frac{\sqrt K\xi}{\sqrt n|\delta_k|} \right)\left(\frac{1}{|\delta_k|^2}+\frac{\sqrt K\xi}{q}\pnk\right)\right)\\
=&~(\Lambda_i/L_i)^{\alpha}v_k(i)-(\Lambda_i/L_i)^{\alpha} \be_i^\top\bcW \bUpsilon(t_k)\bv_k \\
    &+O\left(\|\bv_k\|_\infty \left(\frac{1}{|\delta_k|^2}+\frac{\xi \pnk}{q }\right) + \|\bV_{-k}\|_{\max}  \left(\frac{\sqrt K}{|\delta_k|^2}+\frac{K\xi}{q}\pnk\right)\right)\\
    &+ O\left(\frac{\xi}{\sqrt n |\delta_k|}\left(\frac{1+\sqrt{K}|\delta_k|^{-1}}{|\delta_k|}  +\frac{\xi\sdeg^{-1}  +K\xi \pnk}{q}\right) \right),
\end{split}
\end{align}
where we have used \Cref{prop:eiLWL(G-Upsilon)v} in the second step above.

Finally, recalling that $\bUpsilon(z)+z^{-1}=\bcE_1=O(|z|^{-3})$ by \eqref{eq:Upsilon} 
and using \eqref{eq:large_dev1}, we can deduce that
\begin{align}\label{eq:error term Upsilon+1/z}
\begin{split}
& \Lambda_i^{\alpha} \be_i^\top\bcW \left(\bUpsilon(t_k)+t_k^{-1}\right)\bv_k =   \be_i^\top\bW\bLambda^{-\alpha}\bcE_1(t_k)\bv_k\\
& \quad =\sum_{j\in[n]} W_{ij}\Lambda_i^{-\al}(\cE_1(t_k))_{jj}v_k(j) 
    \lesssim\frac{\xi}{|\delta_k|^3}\bigg(\frac{1}{\sqrt n}+\frac{1}{q}\|\bv_k\|_\infty\bigg) 
\end{split}
\end{align}
 with high probability. Therefore, a combination of \eqref{eq:LasympLambda}, \eqref{eq:vkhati}, and  (\ref{eq:error term Upsilon+1/z}) leads to  \eqref{eq:main thm without assumption}, which completes the proof of \Cref{thm:main thm rescaled}.


\subsection{Proof of \Cref{thm:eigenvalue2 rescaled}} \label{Sec.proof.thm4}

We now aim to derive the asymptotic expansion of the spiked eigenvalue $\hdelta_k$. To accomplish this, we utilize \eqref{eqn:t_k-gamma_k 1} while employing a more accurate estimate of $\bG$ through the Taylor expansion. We begin by applying \eqref{eq:E}, \eqref{eq:LasympLambda}, \eqref{eq:W}, \eqref{eq:G}, and the Taylor expansion to obtain that w.h.p., 
\begin{align}\label{eqn:eval G expansion}
\begin{split}
    \bG(z)& =-\frac{1}{(\bL/\bLambda)^{2\alpha}z}\frac{1}{\bI-\bcW(\bL/\bLambda)^{-2\alpha}z^{-1}}\\
&=-(\bL/\bLambda)^{-2\alpha}z^{-1}-(\bL/\bLambda)^{-2\alpha}\bcW(\bL/\bLambda)^{-2\alpha}z^{-2}\\
&\quad-\bcW^2z^{-3}+O\left(|\delta_k|^{-4}+\frac{\xi}{q|\delta_k|^3\sdeg}\right)
\end{split}
\end{align}
uniformly in $z\in \cC_k\cup\cI_k$. Combining \eqref{eqn:t_k-gamma_k 1}, \eqref{eqn:t_k-gamma_k 1.2}, \eqref{eqn:t_k-gamma_k 2}, and \eqref{eqn:eval G expansion}, it holds that w.h.p., 

\begin{align}
\begin{split}
    t_k^{-1}-\hdelta_k^{-1}&=\hdelta_k^{-1}\bv_k^\top((\bL/\bLambda)^{-2\alpha}-\bI+\hdelta_k^{-1}(\bL/\bLambda)^{-2\alpha}\bcW(\bL/\bLambda)^{-2\alpha}+\hdelta_k^{-2}\bcW^2)\bv_k\\
    &\quad+\bv_k^\top(t_k^{-1}+\bUpsilon(t_k))\bv_k+O\left(\frac{1}{|\delta_k|^4}+\frac{\xi}{q|\delta_k|^3\sdeg}\right)\\
    &\quad+O\left(\frac{\sqrt K\xi\pnk}{q}\left(\frac{1}{|\delta_k|^3}+\frac{\sqrt K\xi\pnk}{q|\delta_k|}\right)+\frac{|\hdelta_k-t_k|}{|\delta_k|^6}\right).
\end{split}
\end{align}
This immediately yields a rough estimate of $\hdelta_k$ that w.h.p., 
\begin{align}\label{eq:eval expansion 1}
\begin{split}
    \hdelta_k-t_k&=t_k\bv_k^\top((\bL/\bLambda)^{-2\alpha}-\bI+t_k^{-1}(\bL/\bLambda)^{-2\alpha}\bcW(\bL/\bLambda)^{-2\alpha}+t_k^{-2}\bcW^2)\bv_k\\
    &\quad+t_k^2\bv_k^\top(t_k^{-1}+\bUpsilon(t_k))\bv_k+O\left(\frac{1}{|\delta_k|^2}+\frac{\xi}{q|\delta_k|\sdeg}\right.\\
    &\quad\left.+\frac{\sqrt K\xi|\delta_k|\pnk}{q}\left(\frac{1}{|\delta_k|^2}+\frac{\sqrt K\xi\pnk}{q}\right)\right),
\end{split}
\end{align}
where we have utilized \eqref{eq:LasympLambda}, \eqref{eq:W}, and the assumption of $|\delta_k|\gg1$. 

By applying the Taylor expansion once again, we can further rewrite \eqref{eq:eval expansion 1} in the form of \eqref{eqn:t_k-gamma_k2}, where the centered random error $B_k$ is defined as
\begin{align}\label{eq:eval B_k def}
\begin{split}
    B_k&:=\alpha(2\alpha+1)t_k\bv_k^\top\frac{(\bL-\bLambda)^2}{\bLambda^2}\bv_k-4\alpha\bv_k^\top\frac{\bL-\bLambda}{\bLambda}\bcW\bv_k\\
    &\quad+\frac{1}{t_k}\bv_k^\top\bcW^2\bv_k+t_k^2\bv_k^\top(t_k^{-1}+\bUpsilon(t_k))\bv_k-A_k.
\end{split}
\end{align}
It is crucial to demonstrate that the variances of the random quadratic terms in \eqref{eq:eval B_k def} above satisfy the inequality stated in \Cref{thm:eigenvalue2}. We now provide the bounds for the variances of
\[
    \bu^\top\frac{(\bL-\bLambda)^2}{\bLambda^2}\bv_k,\quad \bu^\top\frac{\bL-\bLambda}{\bLambda}\bcW\bv_k,\quad\bu^\top\bcW\frac{\bL-\bLambda}{\bLambda}\bv_k,\quad\bu^\top\bcW^2\bv_k
\]
for any deterministic unit vector $\bu$. To control the variance of $\bu^\top({(\bL-\bLambda)}/{\bLambda})^2\bv_k$, we start with controlling each term that appears in the variance. Specifically, we calculate the value of
\[
\E W_{ij}W_{ls}\bu^\top\frac{(\bL-\bLambda)^2}{\bLambda^2}\bv_k
\]
for each $i,j,l,s\in[n]$. Using \eqref{eq:network moment}, \eqref{eq:LasympLambda}, and some direct calculations, it can be shown that
\[
    \E W_{ij}W_{ls}\bu^\top\frac{(\bL-\bLambda)^2}{\bLambda^2}\bv_k\lesssim\frac{\|\bu\|_\infty\|\bv_k\|_\infty}{q^2n^3\sdeg^2} \quad \text{ if }\{i,j\}\cap\{l,s\}=\emptyset,
\]
\[
    \E W_{ij}W_{is}\bu^\top\frac{(\bL-\bLambda)^2}{\bLambda^2}\bv_k\lesssim\frac{|u(i)v_k(i)|}{q^2n^2\sdeg^2}+\frac{\|\bu\|_\infty\|\bv_k\|_\infty}{q^2n^3\sdeg^2} \quad \text{ if } j\neq s,
\]
\begin{align*}
    & \E W_{ij}^2\bu^\top\frac{(\bL-\bLambda)^2}{\bLambda^2}\bv_k=s_{ij}\mathbb E\bv_k^\top\frac{(\bL-\bLambda)^2}{\bLambda^2}\bv_k\\
&\quad+O\left(\frac{|u(i)v_k(i)|+|u(j)v_k(j)|+n^{-1}\|\bu\|_\infty\|\bv_k\|_\infty}{q^4n\sdeg^2}\right).
\end{align*}

By counting the number of appearances of the mentioned cases in $\E(\bu^\top({(\bL-\bLambda)}/{\bLambda})^2\bv_k)^2$ and summing them up, we can deduce that 
\begin{align}\label{eq:(L-La/La)^2 var}
    \var\left(\bu^\top\frac{(\bL-\bLambda)^2}{\bLambda^2}\bv_k\right)\lesssim\frac{\|\bu\|_\infty\|\bv_k\|_\infty}{q^4\sdeg^4}+\frac{1}{q^4n^2\sdeg^4}.
\end{align}
Using similar arguments, one can establish the bounds for the variances of the other quadratic terms. We provide the results below and omit the technical details for simplicity
\begin{align}\label{eq:L-La/La W var}
\var\left(\bu^\top\frac{\bL-\bLambda}{\bLambda}\bcW\bv_k\right)+\var\left(\bu^\top\bcW\frac{\bL-\bLambda}{\bLambda}\bv_k\right)\lesssim\frac{\|\bu\|_\infty\|\bv_k\|_\infty}{q^2\sdeg^2}+\frac{1}{q^2n\sdeg^2},
\end{align}
\begin{align}\label{eq:W^2 var}
    \var\left(\bu^\top\bcW^2\bv_k\right)\lesssim\frac{1}{n}+\frac{1}{q\sqrt n}+\frac{\|\bu\|_\infty\|\bv_k\|_\infty}{q^2}.
\end{align}
Finally, in view of \eqref{eq:Y_l recursive formula}, setting $\bu=\bv_k$ in \eqref{eq:(L-La/La)^2 var}--\eqref{eq:W^2 var} and exploiting \eqref{eq:eval B_k def}, we conclude the proof of \Cref{thm:eigenvalue2 rescaled}.

\subsection{Proof of \Cref{thm:projection2 rescaled}} \label{Sec.proof.thm5}

We proceed with deriving the asymptotic expansions for the spiked eigenvectors $\bu^\top(\bL/\bLambda)^{-\al}\bhv_k$. The procedure is still based on \eqref{eq:oint} while we estimate $\bG_k$ using \eqref{eqn:eval G expansion} to get more accurate results. It turns out that as suggested in \cite{fan2020asymptotic} for the specific case of $\alpha=0$, the asymptotic variance of $\bv_k^\top(\bL/\bLambda)^{-\al}\bhv_k$ is much smaller than that of $\bu^\top(\bL/\bLambda)^{-\al}\bhv_k$ when $\bu$ is \textit{not} parallel to $\bv_k$. As a result, we will need to analyze these two scenarios separately in this proof.

We first aim to prove part 2) of \Cref{thm:projection2 rescaled}.
Taking $\bu=\bv_k$ in \eqref{eq:oint}, it follows from \eqref{eq:E}, \eqref{eq:LasympLambda}, \eqref{eq:W}, \eqref{eq:G}, \eqref{eq:Upsilon}, and \eqref{eq:vv}--\eqref{eq:vV22} that 
\begin{align}\label{eq:parallel expansion 1}
\begin{split}
    &\bv_k^\top(\bL/\bLambda)^{-\alpha}\bhv_k\bhv_k^\top(\bL/\bLambda)^{-\alpha}\bv_k\\
    &=\delta_k^{-2}(\bv_k^\top\bUpsilon'_k(t_k)\bv_k)^{-1}+\frac{1}{2\pi i}\oint_{\cC_k}\frac{\bv_k^\top(\bUpsilon(z)+z^{-1}\cA(z))\bv_k}{(1+\delta_k\bv_k^\top\bUpsilon_k(z)\bv_k)^2}\dd z\\
    &\quad+O\left(\frac{K}{|\delta_k|^4}+\frac{K\xi^2\pnk^2}{q^2}\right)\\
    &=\bv_k^\top\Big(t_k^2\bUpsilon'(t_k)+2t_k\bUpsilon(t_k)+(\bL/\bLambda)^{-2\alpha}-t_k^{-2}(\bL/\bLambda)^{-2\alpha}\bcW(\bL/\bLambda)^{-2\alpha}\\
    &\quad \times\bcW(\bL/\bLambda)^{-2\alpha}-2t_k^{-3}\bcW^3\Big)\bv_k+\delta_k^{-2}(\bv_k^\top\bUpsilon'_k(t_k)\bv_k)^{-1}\\
    &\quad+O\left(\frac{K}{|\delta_k|^4}+\frac{K\xi^2\pnk^2}{q^2}\right),
\end{split}
\end{align}
where we denote by 
\begin{align*}
    \cA(z)&:=(\bL/\bLambda)^{-2\alpha}+(\bL/\bLambda)^{-2\alpha}\bbW (\bL/\bLambda)^{-2\alpha}z^{-1}\\
    &\quad+(\bL/\bLambda)^{-2\alpha}(\bcW (\bL/\bLambda)^{-2\alpha})^2z^{-2}+\bcW^3z^{-3}
\end{align*}
and have resorted to the Cauchy residue theorem in the second step above. 

By taking the square root of the expression in  \eqref{eq:parallel expansion 1} and applying the Taylor expansion on the right-hand side, we can deduce that w.h.p., 
\begin{align}\label{eq:parallel expansion 2}
\begin{split}
    &\bv_k^\top(\bL/\bLambda)^{-\alpha}\bhv_k-\bv_k^\top(\bL/\bLambda)^{-\alpha}\bv_k\\
    &=\frac{\alpha^2}{2}\bv_k^\top\left(\frac{\bL-\bLambda}{\bLambda}\right)^2\bv_k-\frac{\alpha^2(\alpha+1)}{6}\bv_k^\top\left(\frac{\bL-\bLambda}{\bLambda}\right)^3\bv_k\\
    &\quad-\frac{t_k^{-2}}{2}\bv_k^\top (\bL/\bLambda)^{-2\alpha}\bcW(\bL/\bLambda)^{-2\alpha}\bcW(\bL/\bLambda)^{-2\alpha}\bv_k-t_k^{-3}\bv_k^\top\bcW^3\bv_k\\
    &\quad+(\delta_k^2\bv_k^\top\bUpsilon'_k(t_k)\bv_k)^{-1/2}-1+\frac{1}{2}\bv_k^\top(t_k^2\bUpsilon'(t_k)+2t_k\bUpsilon(t_k)+\bI)\bv_k\\
    &\quad+O\left(\frac{K}{|\delta_k|^4}+\frac{K\xi^2\pnk^2}{q^2}\right).
\end{split}
\end{align}
In view of the asymptotic expansion
\begin{align}\label{eq:parallel expansion 3}
\begin{split}
    &\bv_k^\top (\bL/\bLambda)^{-2\alpha}\bcW(\bL/\bLambda)^{-2\alpha}\bcW(\bL/\bLambda)^{-2\alpha}\bv_k\\
    & =\bv_k^\top\bcW^2\bv_k-4\alpha\bv_k^\top\left(\frac{\bL-\bLambda}{\bLambda}\right)\bcW^2\bv_k\\
    &\quad-2\alpha\bv_k^\top\bcW\left(\frac{\bL-\bLambda}{\bLambda}\right)\bcW\bv_k+O\left(\frac{\xi^2}{q^2\sdeg^2}\right),
\end{split}
\end{align}
it remains to bound the second moments of the cubic terms in \eqref{eq:parallel expansion 2} and \eqref{eq:parallel expansion 3}. Using similar arguments as in the proof of \eqref{eq:(L-La/La)^2 var}, it holds that (we omit the complicated details here for simplicity)
\begin{equation}\label{eq:L-La^3 L^2}
    \E\left(\bv_k^\top\left(\frac{\bL-\bLambda}{\bLambda}\right)^3\bv_k\right)^2\lesssim\frac{n^2\|\bv_k\|_\infty^4}{q^8\sdeg^6},
\end{equation}
\begin{equation}\label{eq:L-La^2W L^2}
    \E\left(\bv_k^\top\left(\frac{\bL-\bLambda}{\bLambda}\right)\bcW^2\bv_k\right)^2+\mathbb E\left(\bv_k^\top\bcW\left(\frac{\bL-\bLambda}{\bLambda}\right)\bcW\bv_k\right)^2\lesssim\frac{n^2\|\bv_k\|_\infty^4}{q^4\sdeg^4},
\end{equation}
\begin{equation}\label{eq:W^3 L^2}
    \E(\bv_k^\top\bcW^3\bv_k)^2\lesssim\frac{n^2\|\bv_k\|_\infty^4}{q^2}.
\end{equation}
Thus, combining \eqref{eq:parallel expansion 2}--\eqref{eq:W^3 L^2} yields the conclusion in part 2) of \Cref{thm:projection2 rescaled}.

We next move on to proving part 1) of \Cref{thm:projection2 rescaled}. We start with estimating the integral term in \eqref{eq:oint} for a general $\bu$. With slight abuse of notation, denoted by 
\[
    \bA(z):=(\bL/\bLambda)^{-2\alpha}+z^{-1}(\bL/\bLambda)^{-2\alpha}\bcW(\bL/\bLambda)^{-2\alpha}+z^{-2}\bcW^2.
\]
From \eqref{eq:E}, \eqref{eq:LasympLambda}, \eqref{eq:W}, \eqref{eq:G}, and \eqref{eq:VV}--\eqref{eq:vV22}, it follows that w.h.p.,
\begin{align}\label{eq:projection integral term}
\begin{split}
    &\frac{\bu^\top\bG_k(z)\bv_k}{1+\delta_k\bv_k^{\top}\bG_k(z)\bv_k}-\frac{\bu^\top\bG_k(z)\bv_k}{1+\delta_k\bv_k^{\top}\bUpsilon_k(z)\bv_k}\\
    &=-\delta_k\frac{z^{-1}\bu^\top\bA(z)\bv_k\bv_k^\top(z^{-1}\bA(z)+\bUpsilon(z))\bv_k}{(1+\delta_k\bv_k^\top\bUpsilon_k(z)\bv_k)^2}\\
    &\quad+O\left(K\left(\frac{1}{|\delta_k|^2}+\frac{\xi\pnk}{q}\right)\left(\frac{1}{|\delta_k|}+\frac{\xi}{q\beta_n}\right)\right)
\end{split}
\end{align}
uniformly in $z\in\cC_k\cup\cI_k$. 
An application of the Cauchy residue theorem gives 
\begin{align}\label{eq:projection integral 1}
\begin{split}
    &\frac{1}{2\pi i}\oint_{\cC_k}\frac{z^{-1}\bu^\top\bA(z)\bv_k\bv_k^\top(z^{-1}\bA(z)+\bUpsilon(z))\bv_k}{(1+\delta_k\bv_k^\top\bUpsilon_k(z)\bv_k)^2}\dd z\\
    &=\frac{1}{(\delta_k\bv_k^\top\bUpsilon_k'(t_k)\bv_k)^2}\frac{\partial (t_k^{-1}\bu^\top\bA(t_k)\bv_k\bv_k^\top(t_k^{-1}\bA(t_k)+\bUpsilon(t_k))\bv_k)}{\partial t_k}\\
    &\quad-t_k^{-1}\bu^\top\bA(t_k)\bv_k\bv_k^\top(t_k^{-1}\bA(t_k)+\bUpsilon(t_k))\bv_k\frac{\bv_k^\top\bUpsilon_k''(t_k)\bv_k}{\delta_k^2(\bv_k^\top\bUpsilon_k'(t_k)\bv_k)^3}.
\end{split}
\end{align}

Using the Taylor expansion, we can deduce for the two terms in \eqref{eq:projection integral 1} that w.h.p., 
\begin{align}\label{eq:projection integral 2}
\begin{split}
    &\frac{1}{(\delta_k\bv_k^\top\bUpsilon_k'(t_k)\bv_k)^2}\frac{\partial t_k^{-1}\bu^\top\bA(t_k)\bv_k\bv_k^\top(t_k^{-1}\bA(t_k)+\bUpsilon(t_k))\bv_k}{\partial t_k}\\
    &=t_k^{-1}\bu^\top\bv_k\bv_k^\top\Big(4\alpha\frac{\bL-\bLambda}{\bLambda}-2\alpha(2\alpha+1)\left(\frac{\bL-\bLambda}{\bLambda}\right)^2\\
    &\quad-3t_k^{-1}(\bL/\bLambda)^{-2\alpha}\bcW(\bL/\bLambda)^{-2\alpha}-4t_k^{-2}\bcW^2\Big)\bv_k\\
    &\quad-2\alpha\bu^\top\frac{\bL-\bLambda}{\bLambda}\bv_k\bv_k^\top\left(4\alpha\frac{\bL-\bLambda}{\bLambda}t_k^{-1}-3t_k^{-2}\bcW\right)\bv_k\\
&\quad+t_k^{-2}\bu^\top\bcW\bv_k\bv_k^\top\left(6\alpha\frac{\bL-\bLambda}{\bLambda}-4t_k^{-1}\bcW\right)\bv_k\\
    &\quad+\bu^\top\bv_k\bv_k^\top(t_k\bUpsilon'(t_k)-\bUpsilon(t_k)-2t_k^{-1})\bv_k+O\left(\frac{\xi^3}{q^3|\delta_k|\sdeg^3}+\frac{1}{|\delta_k|^4}\right)
\end{split}
\end{align}
and
\begin{align}\label{eq:projection integral 3}
\begin{split}
    &-t_k^{-1}\bu^\top\bA(t_k)\bv_k\bv_k^\top(t_k^{-1}\bA(t_k)+\bUpsilon(t_k))\bv_k\frac{\bv_k^\top\bUpsilon_k''(t_k)\bv_k}{\delta_k^2(\bv_k^\top\bUpsilon_k'(t_k)\bv_k)^3}\\
    &=2\bu^\top\bv_k\bv_k\Big(-2\alpha t_k^{-1}\frac{\bL-\bLambda}{\bLambda}+\alpha(2\alpha+1)t_k^{-1}\left(\frac{\bL-\bLambda}{\bLambda}\right)^2\\
    &\quad+t_k^{-2}(\bL/\bLambda)^{-2\alpha}\bcW(\bL/\bLambda)^{-2\alpha}+t_k^{-3}\bcW^2\Big)\bv_k\\
    &\quad-4\alpha\bu^\top\frac{\bL-\bLambda}{\bLambda}\bv_k\bv_k^\top\left(-2\alpha t_k^{-1}\frac{\bL-\bLambda}{\bLambda}+t_k^{-2}\bcW\right)\bv_k\\
&\quad+2t_k^{-1}\bu^\top\bcW\bv_k\bv_k^\top\left(-2\alpha t_k^{-1}\frac{\bL-\bLambda}{\bLambda}+t_k^{-2}\bcW\right)\bv_k\\
&\quad+2\bu^\top\bv_k\bv_k^\top(t_k^{-1}+\bUpsilon(t_k))\bv_k+O\left(\frac{\xi^3}{q^3|\delta_k|\sdeg^3}+\frac{1}{|\delta_k|^4}\right),
\end{split}
\end{align}
where we have used \eqref{eq:Upsilon} in the second estimate \eqref{eq:projection integral 3} above. 

By resorting to the Cauchy integral formula, the Taylor expansion, \eqref{eq:E}, \eqref{eq:LasympLambda}, \eqref{eq:W}, \eqref{eq:G}, and \eqref{eq:VV}--\eqref{eq:vV22}, we can show that w.h.p., 
\begin{align}\label{eq:projection integral 4}
\begin{split}
    &\frac{1}{2\pi i}\oint_{\cC_k}\frac{\bu^\top\bG_k(z)\bv_k}{1+\delta_k\bv_k^{\top}\bUpsilon_k(z)\bv_k}\dd z=\frac{\bu^\top\bG_k(t_k)\bv_k}{\delta_k\bv_k^\top\bUpsilon_k'(t_k)\bv_k}\\
    &=\frac{-t_k^{-1}}{\delta_k\bv_k^\top\bUpsilon_k'(t_k)\bv_k}\bu^\top\Big(\bI-2\alpha\frac{\bL-\bLambda}{\bLambda}+\alpha(2\alpha+1)\left(\frac{\bL-\bLambda}{\bLambda}\right)^2\\
    &\quad+t_k^{-1}(\bL/\bLambda)^{-2\alpha}\bcW(\bL/\bLambda)^{-2\alpha}+t_k^{-2}\bcW^2\Big)\bv_k\\
    &\quad+\frac{-t_k^{-2}}{\delta_k\bv_k^\top\bUpsilon_k'(t_k)\bv_k}\bu^\top\bV_{-k}\frac{1}{\bDelta_{-k}^{-1}-t_k^{-1}}\\
    &\quad\times\bV_{-k}^\top\Big(-2\alpha\frac{\bL-\bLambda}{\bLambda}+\alpha(2\alpha+1)\left(\frac{\bL-\bLambda}{\bLambda}\right)^2\\
    &\quad+t_k^{-1}(\bL/\bLambda)^{-2\alpha}\bcW(\bL/\bLambda)^{-2\alpha}+t_k^{-2}\bcW\Big)\bv_k\\
    &\quad+\frac{-t_k^{-2}}{\delta_k\bv_k^\top\bUpsilon_k'(t_k)\bv_k}\bu^\top\left(-2\alpha\frac{\bL-\bLambda}{\bLambda}+t_k^{-1}\bcW\right)\bV_{-k}\frac{1}{\bDelta_{-k}^{-1}-t_k^{-1}}\bV_{-k}^\top\\
    &\quad\times \left(-2\alpha\frac{\bL-\bLambda}{\bLambda}+t_k^{-1}\bcW\right)\bv_k\\
    &\quad+O\left(\frac{K}{|\delta_k|^3}+\frac{K\xi^3}{q^3\sdeg^3}+\frac{K^{3/2}\xi\pnk}{q}\left(\frac{\xi\pnk}{q}+\frac{1}{|\delta_k|^2}\right)\right).
\end{split}
\end{align}

Moreover, with the aid of \eqref{eq:large_dev1}
, we have that w.h.p., 
\begin{align}\label{eq:projection integral 5}
\begin{split}
    \bv_k^\top\frac{\bL-\bLambda}{\bLambda}\bv_k&=q^{-1}\sdeg^{-1}\sum_{i\in[n]}\left(\sum_{j\in[n]}W_{ij}+\frac{\tau}{n}\sum_{l,j\in[n]}W_{lj}\right)\Lambda_i^{-1}v_k(i)^2\\
    &\lesssim\frac{\xi}{q\sdeg}\left(\frac{\|\bv_k\|_\infty^2}{q}+\left(\frac{1}{n}\sum_{i,j\in[n]}|v_k(i)|^4\right)^{1/2}\right)\\
    &\lesssim\frac{\xi\|\bv_k\|_\infty}{q\sdeg}
\end{split}
\end{align}
and
\begin{align}\label{eq:projection integral 6}
\begin{split}
    \bv_k^\top\bcW\bv_k&=\sum_{i,j\in[n]}\overline{W}_{ij}v_k(i)v_k(j)\\
    &\lesssim\xi\left(\frac{\|\bv_k\|_\infty^2}{q}+\left(\frac{1}{n}\sum_{i,j\in[n]}v_k(i)^2v_k(j)^2\right)^{1/2}\right)\\
    & \lesssim\frac{\xi}{\sqrt n} \lesssim\frac{\xi}{q}.
\end{split}
\end{align}

Combining \eqref{eq:oint} and \eqref{eq:projection integral term}--\eqref{eq:projection integral 6}, it holds that w.h.p., 
\begin{align}\label{eq:projection expansion perp 1}
\begin{split}
    &\bu^\top(\bL/\bLambda)^{-\alpha}\bhv_k\bhv_k^\top(\bL/\bLambda)^{-\alpha}\bv_k=-\frac{1}{2\pi i}\oint_{\cC_k}\frac{\bu^\top\bG_k(z)\bv_k}{1+\delta_k\bv_k^{\top}\bG_k(z)\bv_k}\dd z\\
    &=\frac{t_k^{-1}}{\delta_k\bv_k^\top\bUpsilon_k'(t_k)\bv_k}\bu^\top\Big(\bI-2\alpha\frac{\bL-\bLambda}{\bLambda}+\alpha(2\alpha+1)\left(\frac{\bL-\bLambda}{\bLambda}\right)^2\\
    &\quad+t_k^{-1}(\bL/\bLambda)^{-2\alpha}\bcW(\bL/\bLambda)^{-2\alpha}+t_k^{-2}\bcW^2\Big)\bv_k\\
    &\quad+\frac{t_k^{-2}}{\delta_k\bv_k^\top\bUpsilon_k'(t_k)\bv_k}\bu^\top\bV_{-k}\frac{1}{\bDelta_{-k}^{-1}-t_k^{-1}} \bV_{-k}^\top\Big(-2\alpha\frac{\bL-\bLambda}{\bLambda}\\
    &\quad+\alpha(2\alpha+1)\left(\frac{\bL-\bLambda}{\bLambda}\right)^2+t_k^{-1}(\bL/\bLambda)^{-2\alpha}\bcW(\bL/\bLambda)^{-2\alpha}+t_k^{-2}\bcW^2\Big)\bv_k\\
    &\quad+\frac{t_k^{-2}}{\delta_k\bv_k^\top\bUpsilon_k'(t_k)\bv_k}\bu^\top\left(-2\alpha\frac{\bL-\bLambda}{\bLambda}+t_k^{-1}\bcW\right)\bV_{-k}\frac{1}{\bDelta_{-k}^{-1}-t_k^{-1}} \\
    &\quad \times \bV_{-k}^\top\left(-2\alpha\frac{\bL-\bLambda}{\bLambda}+t_k^{-1}\bcW\right)\bv_k\\
    &\quad-\bu^\top\bv_k\bv_k^\top(t_k^{-1}(\bL/\bLambda)^{-2\alpha}\bcW(\bL/\bLambda)^{-2\alpha}+2t_k^{-2}\bcW^2)\bv_k\\
    &\quad-t_k\bu^\top\bv_k\bv_k^\top(t_k\bUpsilon'(t_k)+\bUpsilon(t_k))\bv_k\\
    &\quad+O\left(K\left(\frac{1}{|\delta_k|^2}+\frac{\xi\pnk}{q}\right)\left(\frac{1}{|\delta_k|}+\frac{\xi}{q\beta_n}\right)\right.\\
    &\quad\left.+\frac{K^{3/2}\xi\pnk}{q}\left(\frac{\xi\pnk}{q}+\frac{1}{|\delta_k|^2}\right)\right).
\end{split}
\end{align}
In addition, note that with the aid of \eqref{eq:parallel expansion 2} and \eqref{eq:projection integral 5}, an application of the Taylor expansion gives that w.h.p., 
\begin{align}\label{eq:projection expansion perp 2}
\begin{split}
    (\bv_k^\top(\bL/\bLambda)^{-\alpha}\bhv_k)^{-1}&=|\delta_k|(\bv_k^\top\bUpsilon'_k(t_k)\bv_k)^{1/2}\left(1-\frac{1}{2}\bv_k^\top(t_k^2\bUpsilon'(t_k)+2t_k\bUpsilon(t_k)\right.\\
    &\quad\left.+(\bL/\bLambda)^{-2\alpha}-t_k^{-2}\bcW^2)\bv_k\right)\\
    &\quad+O\left(\frac{1}{|\delta_k|^4}+\frac{\xi^4}{q^4\sdeg^4}+\frac{\xi^2\|\bv_k\|_\infty}{q^2\sdeg^2}\right).
\end{split}
\end{align}

Multiplying \eqref{eq:projection expansion perp 1} and \eqref{eq:projection expansion perp 2} above, we can obtain the asymptotic expansion of the projection 
\begin{align}\label{eq:projection expansion perp 3}
\begin{split}
    &\bu^\top(\bL/\bLambda)^{-\alpha}\bhv_k-\frac{\bu^\top\bv_k}{(t_k^2\bv_k^\top\bUpsilon_k(t_k)\bv_k)^{1/2}}\\
    &=-\frac{1}{2}\bu^\top\bv_k\bv_k^\top(t_k^2\bUpsilon'(t_k)+2t_k\bUpsilon(t_k)+(\bL/\bLambda)^{-2\alpha}-t_k^{-2}\bcW^2)\bv_k\\
    &\quad+\bu^\top\left(-2\alpha\frac{\bL-\bLambda}{\bLambda}+\alpha(2\alpha+1)\left(\frac{\bL-\bLambda}{\bLambda}\right)^2\right.\\
    &\quad\left.+t_k^{-1}(\bL/\bLambda)^{-2\alpha}\bcW(\bL/\bLambda)^{-2\alpha}+t_k^{-2}\bcW^2\right)\bv_k\\
    &\quad+t_k^{-1}\bu^\top\bV_{-k}\frac{1}{\bDelta_{-k}^{-1}-t_k^{-1}} \bV_{-k}^\top\left(-2\alpha\frac{\bL-\bLambda}{\bLambda}+\alpha(2\alpha+1)\left(\frac{\bL-\bLambda}{\bLambda}\right)^2\right.\\
    &\quad\left.+t_k^{-1}(\bL/\bLambda)^{-2\alpha}\bcW(\bL/\bLambda)^{-2\alpha}+t_k^{-2}\bcW^2\right)\bv_k\\
    &\quad+t_k^{-1}\bu^\top\left(-2\alpha\frac{\bL-\bLambda}{\bLambda}+t_k^{-1}\bcW\right)\bV_{-k}\frac{1}{\bDelta_{-k}^{-1}-t_k^{-1}}\\
    &\quad\times\bV_{-k}^\top\left(-2\alpha\frac{\bL-\bLambda}{\bLambda}+t_k^{-1}\bcW\right)\bv_k\\
    &\quad-\bu^\top\bv_k\bv_k^\top(t_k^{-1}(\bL/\bLambda)^{-2\alpha}\bcW(\bL/\bLambda)^{-2\alpha}+2t_k^{-2}\bcW^2)\bv_k\\
    &\quad-t_k\bu^\top\bv_k\bv_k^\top(t_k\bUpsilon'(t_k)+\bUpsilon(t_k))\bv_k\\
    &\quad+O\left(K\left(\frac{1}{|\delta_k|^2}+\frac{\xi\pnk}{q}\right)\left(\frac{1}{|\delta_k|}+\frac{\xi}{q\beta_n}\right)\right.\\
    &\quad\left.+\frac{K^{3/2}\xi\pnk}{q}\left(\frac{\xi\pnk}{q}+\frac{1}{|\delta_k|^2}\right)\right).
\end{split}
\end{align}

Finally, when $\bu^\top\bv_k=0$, by writing $\bu=\bV_{-k}\bV_{-k}^\top\bu+\bw$ in \eqref{eq:projection expansion perp 3} and using \eqref{eq:(L-La/La)^2 var}--\eqref{eq:W^2 var}, we can derive the desired conclusion in part 1) of \Cref{thm:projection2 rescaled}. This completes the proof of \Cref{thm:projection2 rescaled}.

\subsection{Proof of \Cref{thm:K_0 rescaled}}

From the proof of \Cref{thm:eigenvalue rescaled} in Section \ref{The estimation of the spiked eigenvalues}, we see that \eqref{eq:tkdeltak} also holds for $k=K_0+1$. Let us consider the probability
\begin{align} \label{new.eq.supp.A.144}
\begin{split}
    \P[\hK_0\neq K_0]&=\P[|\hdelta_{K_0}|<a_n']+\P[|\hdelta_{K_0+1}|\geq a_n']\\
    &=\P\left[1<\frac{a'_n}{|\delta_{K_0}|}+\frac{|\hdelta_{K_0}-t_{K_0}|}{|\delta_{K_0}|}+O(|\delta_{K_0}|^{-2})\right]\\
    &\quad+\P\left[\frac{a'_n}{|\delta_{K_0+1}|}\leq 1+\frac{|\hdelta_{K_0+1}-t_{K_0+1}|}{|\delta_{K_0+1}|}+O(|\delta_{K_0+1}|^{-2})\right]\\
    &\rightarrow 0,
\end{split}
\end{align}
where in the second step above, we have used \eqref{eq:tkdeltak}, and in the last step above, we have used $|\delta_{K_0}|\geq a_n$, \eqref{eq:condK}, \eqref{eqn:t_k-gamma_k}, \eqref{eq:K_0+1 assumption}, and \eqref{eq:a_n assumption}. Thus, an application of (\ref{new.eq.supp.A.144}) above concludes the proof of \Cref{thm:K_0 rescaled}.

\subsection{Proof of Corollary \ref{example:CLT entry}} \label{Sec.proof.cor1}

An application of the classical Lindeberg--Feller central limit theorem (CLT) (see, e.g., \cite{chung2001course}) gives that
\begin{equation*}
 \frac{1}{\sigma_{k,i} t_k}\sum_{l\in[n]}{W}_{il}{\Lambda}_l^{-\alpha}v_k(l) \to \mathcal N(0,1)
\end{equation*}
in law provided that $\|\bv_k\|_\infty\to 0$. Together with \eqref{eq:entry_cond_CLT}, this leads to the desired CLT for $(L_i^{\alpha}\hv_k(i)-\Lambda_i^\alpha v_k(i))/{\sigma_{k,i}}$, which completes the proof of Corollary \ref{example:CLT entry}.

\subsection{Proof of Corollary \ref{cor:CLT_evalue}} \label{Sec.proof.cor2}
Applying the classical Lindeberg--Feller CLT gives that
\begin{equation*}
 \frac{1}{\varsigma_k}\left(-2\alpha t_k\bv_k^\top\frac{\bL-\bLambda}{\bLambda}\bv_k+\bv_k^\top\bcW\bv_k\right) \to \mathcal N(0,1)
\end{equation*}
in law provided that $\|\bv_k\|_\infty\to 0$. This along with \eqref{eq:CLT_evalue var assumption} yields the desired CLT for ${(\hdelta_k-t_k-A_k)}/{\varsigma_k}$, which concludes the proof of Corollary \ref{cor:CLT_evalue}.

\subsection{Proof of Corollary \ref{cor:CLT_evector}} \label{Sec.proof.cor3}

The proof for part 1) of Corollary \ref{cor:CLT_evector} is still a simple application of the classical Lindeberg--Feller CLT, and thus, we omit the details there. It remains to prove part 2) of Corollary \ref{cor:CLT_evector}. Clearly, we need only to establish the CLT for 
\[
\frac{\alpha^2}{2}\bv_k^\top\left(\frac{\bL-\bLambda}{\bLambda}\right)^2\bv_k-\frac{1}{2t_k^{2}}\bv_k^\top\bcW^2\bv_k.
\]
Such a term can be written as
\begin{align}\label{eq: parallel CLT decomposition}
    &\frac{\alpha^2}{2}\bv_k^\top\left(\frac{\bL-\bLambda}{\bLambda}\right)^2\bv_k-\frac{1}{2t_k^{2}}\bv_k^\top\bcW^2\bv_k \nonumber\\
    &=\frac{\alpha^2}{2q^2\sdeg^2}\sum_{i\in[n]}\frac{v_k(i)^2}{\Lambda_i^{2}}\left(\sum_{j_1,j_2\in[n]}W_{ij_1}W_{ij_2}+\frac{\tau_i}{n}\sum_{j_1,j_2,l\in[n]}W_{ij_1}W_{lj_2}\right. \nonumber\\
    &\quad\left.+\frac{\tau_i^2}{n^2}\sum_{j_1,j_2,l_1,l_2\in[n]}W_{j_1l_1}W_{j_2l_2}\right) -\frac{1}{2t_k^2}\sum_{i,j,l\in[n]}v_k(i)v_k(l)\ocW_{ij}\ocW_{jl} \nonumber\\
    &=\frac 1 2\sum_{1\leq i\leq j\leq n}(W_{ij}b_{ij}+W_{ij}^2c_{ij}),
\end{align}
where we denote by 
\begin{align}
\begin{split}
    b_{ij}:=&\sum_{1\leq l<j}W_{il}f_k(i,j,l)+\sum_{1\leq l<i}W_{jl}f_k(j,i,l)\\
    &+\left(\sum_{1\leq l_1\leq l_2<j}+\sum_{1\leq l_1<i,\,l_2=j}\right)W_{l_1l_2}g_k(i,j,l_1,l_2)
\end{split}
\end{align}
and
\begin{align}
\begin{split}
    c_{ij}:=&\frac{\alpha^2}{q^2\sdeg^2}(1+\delta_i^j)^{-1}\left(\frac{v_k(i)^2}{\Lambda_i^{2}}+\frac{v_k(j)^2}{\Lambda_j^{2}}\right)+g_k(i,j,i,j)\\
    &-\frac{1}{t_k^2}(1+\delta_i^j)^{-1}\Lambda_i^{-2\alpha}\Lambda_j^{-2\alpha}(v_k(i)^2+v_k(j)^2)
\end{split}
\end{align}
with \begin{equation}
    f_k(i,j,l):=(1+\delta_i^j)^{-1}\left(\frac{\alpha^2}{q^2\sdeg^2}\Lambda_i^{-2}v_k(i)^2-\frac{2}{t_k^2}\Lambda_i^{-2\alpha}\Lambda_j^{-\alpha}\Lambda_l^{-\alpha}v_k(j)v_k(l)\right)
\end{equation}
and 
\begin{align}
\begin{split}
    g_k(i,j,l_1,l_2):=&\frac{\alpha^2}{nq^2\sdeg^2}\left(\frac{2-\delta_{l_1}^{l_2}}{1+\delta_i^j}\left(\frac{\tau_i v_k(i)^2}{\Lambda_i^2}+\frac{\tau_j v_k(j)^2}{\Lambda_j^2}\right)\right.\\
    &+\left.\frac{2-\delta_i^j}{1+\delta_{l_1}^{l_2}}\left(\frac{\tau_{l_1} v_k(l_1)^2}{\Lambda_{l_1}^2}+\frac{\tau_{l_2} v_k(l_2)^2}{\Lambda_{l_2}^2}\right)\right)\\
    &+\frac{\alpha^2}{n^2q^2\sdeg^2}(2-\delta_i^j)(2-\delta_{l_1}^{l_2})\sum_{m\in[n]}\frac{\tau_m^2v_k(m)^2}{\Lambda_m^2}.
\end{split}
\end{align}

Based on \eqref{eq: parallel CLT decomposition}, we can calculate the mean
\begin{equation}
    \E\left[\frac{\alpha^2} 2\bv_k^\top\left(\frac{\bL-\bLambda}{\bLambda}\right)^2\bv_k-\frac{1}{2t_k^{2}}\bv_k^\top\bcW^2\bv_k\right]=\frac 1 2\sum_{1\leq i\leq j\leq n}s_{ij}c_{ij}.
\end{equation}
Observe that for each integer $t\in[2^{-1}n(n+1)]$, there exist unique $i,j\in[n]$ such that
\[
    t=i+2^{-1}j(j-1).
\]
With such property, we can define the $\sigma$-algebras
\begin{equation}
    \cF_t:=\sigma\{W_{l,s}:1\leq l\leq s<j \text{ or }1\leq s\leq i\leq l=j\}.
\end{equation}
In light of such representation, we see that 
\[
    \frac{\alpha^2} 2\bv_k^\top\left(\frac{\bL-\bLambda}{\bLambda}\right)^2\bv_k-\frac{1}{2t_k^{2}}\bv_k^\top\bcW^2\bv_k-\E\left[\frac{\alpha^2} 2\bv_k^\top\left(\frac{\bL-\bLambda}{\bLambda}\right)^2\bv_k-\frac{1}{2t_k^{2}}\bv_k^\top\bcW^2\bv_k\right]
\]
is in fact a sum of martingale differences with respect of the filtration $\{\cF_{i+2^{-1}j(j-1)}\}$, since for each $1\leq i\leq j\leq n$ we have that 
\begin{equation}
    \E[W_{ij}b_{ij}-(W_{ij}^2-s_{ij})c_{ij}|\cF_{i+2^{-1}j(j-1)-1}]=0.
\end{equation}

Let us define the sum of the conditional variances as
\begin{align}
\begin{split}
    P_k&=P_k(n):=\frac 1 4\sum_{1\leq i\leq j\leq n}\E[(W_{ij}b_{ij}-(W_{ij}^2-s_{ij})c_{ij})^2|\cF_{i+2^{-1}j(j-1)-1}]\\
    &=\frac 1 4\sum_{1\leq i\leq j\leq n}(s_{ij}b_{ij}^2+2\gamma_{ij}b_{ij}c_{ij}+\kappa_{ij}c_{ij}^2),
\end{split} 
\end{align}
where $\gamma_{ij}:=\E W_{ij}^3$ and $\kappa_{ij}:=\E(W_{ij}^2-s_{ij})^2$. In particular, the mean of $P_k$ is given by 
\begin{align}\label{eq:variance para def}
\begin{split}
    &\mathfrak s_{\bv_k,k}^2:=\E P_k\\
    &=\frac 1 4\sum_{1\leq i\leq j\leq n}s_{ij}\left(\sum_{1\leq l<j}s_{il}{(2-\delta_i^l)}\left({f_k(i,j,l)}+g_k(i,j,i,l)\right)^2\right.\\
    &\quad+\left.\sum_{1\leq l<i}s_{jl}{(2-\delta_j^l)}(f_{k}(j,i,l)+g_k(i,j,j,l))^2\right.\\
    &\quad+\left.\sum_{1\leq l_1\leq l_2<j,l_1,l_2\notin\{i,j\}}s_{l_1l_2}{(2-\delta_{l_1}^{l_2})}g_k(i,j,l_1,l_2)^2\right)\\
    &\quad+\frac 1 4\sum_{1\leq i\leq j\leq n}\kappa_{ij}c_{ij}^2,
\end{split} 
\end{align}
and the variance of $P_k$ can be calculated as
\begin{align}\label{eq:variance Pk def}
\begin{split}
    & \kappa_{\bv_k}: =\var(P_k) \\
    & =\frac{1}{16}\sum_{i_1,i_2,j_1,j_2\in[n],i_1\leq j_1,i_2\leq j_2}\E\left((s_{i_1j_1}(b_{i_1j_1}^2-\E b_{i_1j_1}^2)+2\gamma_{i_1j_1}b_{i_1j_1}c_{i_1j_1})\right.\\
    &\quad\times\left.(s_{i_2j_2}(b_{i_2j_2}^2-\E b_{i_2j_2}^2)+2\gamma_{i_2j_2}b_{i_2j_2}c_{i_2j_2})\right).
\end{split}
\end{align}

Let us recall the classical martingale CLT; see, e.g., Lemma 9.12 of \cite{bai2006spectral}. If a martingale difference sequence $\{Y_t\}$ with respect to a filtration $\{\cF_t\}$ satisfies the conditions 
\begin{itemize}
    \item[a)] $\frac{\sum_{t\in[T]}\E[Y_t^2|\cF_{t-1}]}{\sum_{t\in[T]}\E Y_t^2} 
    {\to} 1$ in probability,

\item[b)] $\frac{\sum_{t\in[T]}\E[Y_t^2I(|Y_t|/\sqrt{\sum_{t\in[T]}\E Y_t^2}\geq\epsilon)]}{\sum_{t\in[T]}\E Y_t^2}\leq\frac{\sum_{t\in[T]}\E Y_t^4}{(\sum_{t\in[T]}\E Y_t^2)^2}\to 0$ for any $\epsilon>0$,
\end{itemize}
then we have $\frac{\sum_{t\in[T]}Y_t}{\sqrt{\sum_{t\in[T]}\E Y_t^2}}\to \cN(0,1)$ in distribution as $T\to\infty$, where $I(\cdot)$ stands for the indicator function. It follows from the assumption of $\kappa_{\bv_k}^{1/4}\ll \mathfrak s_{\bv_k,k}$ that $P_k/\E P_k\to 1$ in probability, which shows that condition a) above is satisfied. It remains to verify condition b) above in order to invoke the classical martingale CLT.

From some simple calculations and \eqref{eq:moment conditions}, we can deduce that 
\begin{equation}\label{eq:fk cij}
\max_{i,j,l\in[n]}|f_k(i,j,l)|\lesssim\left(\frac{1}{q^2\sdeg^2}+\frac{1}{|\delta_k|^2}\right)\|\bv_k\|_\infty^2, \ \max_{i,j,l_1,l_2\in[n]}|g_k(i,j,l_1,l_2)|\lesssim\frac{\|\bv_k\|_\infty^2}{nq^2\sdeg^2}
\end{equation}
\begin{equation}
    \max_{i,j\in[n]}|c_{ij}|\lesssim\left(\frac{1}{q^2\sdeg^2}+\frac{1}{|\delta_k|^2}\right)\|\bv_k\|_\infty^2,
\end{equation}
\begin{equation}
    \E b_{ij}^2\lesssim\left(\frac{1}{q^2\sdeg^2}+\frac{1}{|\delta_k|^2}\right)^2\|\bv_k\|_\infty^4\frac{i+j}{n},
\end{equation}
\begin{equation}
    \E b_{ij}^3\lesssim\left(\frac{1}{q^2\sdeg^2}+\frac{1}{|\delta_k|^2}\right)^3\|\bv_k\|_\infty^6\frac{i+j}{nq},
\end{equation}
\begin{equation}\label{eq:bij fourth moment}
    \E b_{ij}^4\lesssim\left(\frac{1}{q^2\sdeg^2}+\frac{1}{|\delta_k|^2}\right)^4\|\bv_k\|_\infty^8\left(\frac{i+j}{nq^2}+\frac{i^2+j^2}{n^2}\right).
\end{equation}

Finally, with an application of \eqref{eq:fk cij}--\eqref{eq:bij fourth moment}, we can bound the sum of the fourth moments as 
\begin{align}
\begin{split}
    &\sum_{1\leq i\leq j\leq n}\E (W_{ij}b_{ij}-(W_{ij}^2-s_{ij})c_{ij})^4\\
    &=\sum_{1\leq i\leq j\leq n}(\E W_{ij}^4\E b_{ij}^4+4c_{ij}\E W_{ij}^3(W_{ij}^2-s_{ij})\E b_{ij}^3\\
    &\quad+6c_{ij}^2\E W_{ij}^2(W_{ij}^2-s_{ij})^2\E b_{ij}^2+c_{ij}^4\E(W_{ij}^2-s_{ij})^4)\\
    &\lesssim\left(\frac{1}{q^8\sdeg^8}+\frac{1}{|\delta_k|^8}\right)\|\bv_k\|_\infty^8\sum_{1\leq i\leq j\leq n}\left(\frac{i^2+j^2}{n^3q^2}+\frac{i+j}{n^2q^4}+\frac{1}{nq^6}\right)\\
    &\lesssim\frac{n}{q^2}\left(\frac{1}{q^8\sdeg^8}+\frac{1}{|\delta_k|^8}\right)\|\bv_k\|_\infty^8\\
    &\ll \mathfrak s_{\bv_k,k}^4,
\end{split}
\end{align}
where in the last step above, we have used \eqref{eq:CLT_evector2 var assumption}. This shows that condition b) above is also satisfied.
Therefore, an application of the classical martingale CLT yields the desired conclusion in part 2) of \Cref{cor:CLT_evector}. This completes the proof of Corollary \ref{cor:CLT_evector}.

\subsection{Proof of \Cref{thm:local laws of R}} \label{Sec.proof.thm6} 

The local laws in \Cref{thm:local laws of R} can be proved using similar arguments as those in \cite{SIMPLERC}. The only difference is that there are several places in \cite{SIMPLERC} where Bernstein's inequality was applied and its role can be replaced with Lemma \ref{lemma:estimation} in Section \ref{add.tech.lem} correspondingly. For simplicity, we omit the details here.

\subsection{Proof of \Cref{prop:ekG-UpsilonvLinfty}} \label{Sec.proof.thm7}

For any $z\in S(\fC)$, applying (\ref{eq:resolvent identity 2 of G}) to $\bG_{[i]}(z)$ gives that w.h.p.,
\begin{align}\label{eq:eiG[i]-Upsilonv deco}
\begin{split}
\mathbf{e}_{i}^\top(\bG_{[i]}-\bUpsilon)\mathbf{v}&=-(G_{[i]})_{ii}\sum_{l \in [n]\setminus\{i\}}\ocW_{il} (G_{[i]}^{(i)})_{l\bv^{(i)}} +v(i)\left((G_{[i]})_{ii}-M_i\right)\\
	&\lesssim \frac{\xi}{|z|}\bigg(\frac{\max_{1 \leq l\neq i \leq n}|(G_{[i]}^{(i)})_{l\bv^{(i)}}|}{q}+\frac{1}{\sqrt n |z|}\bigg)+\frac{\xi}{q|z|\sdeg}\|\bv\|_\infty,
 \end{split}
\end{align}
where we have used a similar argument as in \eqref{eq:boundG[i]0} with the aid of  \eqref{eq:large_dev1} and \eqref{eq:G[i]-R2}, and have applied \eqref{eq:entry law of R} and \eqref{eq:G[i]-R} to control $(G_{[i]})_{ii}-M_i$. In view of  \eqref{eq:boundG[i]} and (\ref{eq:eiG[i]v}), it holds that w.h.p.,
\begin{align}\label{eq:etG[i](i)v(i)}
	|(G_{[i]}^{(i)})_{l\bv^{(i)}}| 
	&\lesssim\left(1+\frac{1}{q\sdeg}\right)\left({\frac{\xi}{\sqrt n|z|^2}}+\frac{\|\bv\|_\infty}{|z|}\right).
\end{align}
Plugging it into \eqref{eq:eiG[i]-Upsilonv deco}, we can obtain that for any $z\in S(\fC)$, the event 
\begin{equation}\label{eq:eiGj-UpsilonvLinfty_weak}
 \left\{\left|\mathbf{e}_{i}^\top(\bG_{[i]}-\bUpsilon)\mathbf{v}\right| \lesssim \frac{\xi}{|z|}\bigg(\frac{1}{\sqrt n|z|}+\frac{\|\bv\|_\infty}{q\sdeg}\bigg)\right\}
\end{equation}
holds with high probability. 

We next make use of a standard $\epsilon$-net argument with respect to $z\in S(\fC)$. More specifically, using a union bound, we first get a uniform estimate for all $z$ in an $(n|z|)^{-3}$-net $S(\fC)\cap \{(n|z|)^{-3}\mathbb Z^2\}$. Second, by the Lipschitz continuity of $\max_{i\in[n]}|\be^\top_i(\bG_{[i]}(z)-\bUpsilon(z))\bv|$ (with Lipschitz constant $\OO(1)$ due to \eqref{eq:W}), the inequality can be extended uniformly to all $S(\fC)$. Then it follows that the event 
\begin{equation}\bigcap_{z\in S(\fC)}\biggl\{\max_{i\in[n]}|\be^\top_i(\bG_{[i]}(z)-\bUpsilon(z))\bv|\leq C\frac{\xi}{|z|}\bigg(\frac{1}{\sqrt n|z|}+\frac{\|\bv\|_\infty}{q\sdeg}\bigg)\biggr\} \label{eq:eiGi-UpsilonvLinfty}
\end{equation}
holds with high probability for a constant $C>0$. In light of \eqref{eq:eiGi-UpsilonvLinfty}, estimate (\ref{eq:eiG-UpsilonvLinfty}) for each fixed $z\in S(\fC)$ follows from a simple application of Lemmas \ref{lemma:G-G[i]} and \ref{lemma:Linfty norms}. Then with the aid of (\ref{eq:eiG-UpsilonvLinfty}) for each fixed $z\in S(\fC)$,  estimate (\ref{eq:eiGj-UpsilonvLinfty})  for such $z$ also holds with an application of Lemmas \ref{lemma:G-G[i]} and \ref{lemma:Linfty norms}. Using the $\epsilon$-net argument again, we can obtain (\ref{eq:eiG-UpsilonvLinfty}) and (\ref{eq:eiGj-UpsilonvLinfty}). 

To establish \eqref{eq:eiG(i)-UpsilonvLinfty}, let us first observe that 
$$\be_j^\top\big(\bG_{[i]}^{(i)}(z)-\bUpsilon^{(i)}(z)\big)\bv= \mathbf 1_{i\ne j}\be_j^\top \big(\bG_{[i]}^{(i)}(z)-\bUpsilon(z)\big)\bv^{(i)}. $$
Applying (\ref{eq:resolvent identity 3 of G}) to $\bG_{[i]}$, we have that for $j\ne i$, 
\begin{align}
& \left|\be_j^\top(\bG_{[i]}(z)-\bG_{[i]}^{(i)}(z) )\bv^{(i)}\right| =\left|({G}_{[i]})_{ji}\right|  \left|\frac{({G}_{[i]})_{i\bv^{(i)}}}{({G}_{[i]})_{ii}} \right| \nonumber\\
&\quad \lesssim \frac{\xi}{|z|}\bigg(\frac{1}{\sqrt n|z|}+\frac{1}{q}\bigg)\left({\frac{\xi}{\sqrt n|z|}}+\|\bv\|_\infty\right) \label{eq:ejG[i](i)} 
\end{align}
w.h.p., where, in the second step above, we have used  (\ref{eq:eiG[i]v_div}) to control $\left|{({G}_{[i]})_{i\bv^{(i)}}}/{({G}_{[i]})_{ii}} \right|$
and \eqref{eq:eiGj-UpsilonvLinfty_weak} with $\bv=\be_j$ to control $\left|({G}_{[i]})_{ji}\right|$.
Therefore, a combination of \eqref{eq:eiGj-UpsilonvLinfty} and \eqref{eq:ejG[i](i)} leads to  \eqref{eq:eiG(i)-UpsilonvLinfty}, which concludes the proof of \Cref{prop:ekG-UpsilonvLinfty}.

\subsection{Proof of \Cref{prop:uG-Upsilonv anisotropic}} \label{Sec.proof.thm8}

Denote by 
$
     \bcE':=(\bL/\bLambda)^{2\alpha}-\bI.
$
By \Cref{prop:L-Lambda}, we have that 
$$\|\bcE'\|\lesssim \sdeg^{-1}\xi/q$$ with high probability. Then it follows from Theorem \ref{thm:local laws of R} and \Cref{prop:G-R} that w.h.p.,
\begin{align}
    \bu^\top(\bG(z)-\bUpsilon(z))\bv&=\bu^\top(\bG-\bR)\bv+O\left(\frac{\log n}{q|z|^2}\right) \nonumber\\
    &= z\bu^\top\bG\bcE'\bR\bv+O\left(\frac{\log n}{q|z|^2}\right) \nonumber\\
    &=z\bu^\top\bUpsilon\bcE'\bUpsilon\bv+O\bigg(\frac{\xi}{q|z|^2\sdeg}+\frac{\xi^2}{q^2|z|\sdeg^{2}}\bigg).\label{eq:uG-Upsilonv 1} 
 \end{align}  
Let us estimate the first term on the RHS of \eqref{eq:uG-Upsilonv 1} above. With the Taylor expansion of $\bcE'$, we can write that w.h.p.,
\begin{align}
z\bu^\top\bUpsilon\bcE'\bUpsilon\bv&=z\sum_{i\in[n]}u(i)v(i)\Upsilon_i^2\cE'_i\nonumber\\
    &=\frac{z}{q}\sum_{i\in[n]}u(i)v(i)\Upsilon_i^2\cdot \frac{2\al}{\Lambda_i}\bigg(\sum_{j\in[n]}W_{ij}+\frac{1}{n}\sum_{k,l\in[n]}W_{kl}\bigg)\nonumber\\
    &\quad+O\left(\frac{\xi^2}{q^2|z|\sdeg^{2}} \right).\label{eq:uG-Upsilonv 2}
\end{align}

Applying \eqref{eq:large_dev1} to the first term on the RHS above shows that w.h.p.,
\begin{align*}
    \frac{1}{n}\sum_{k,l\in[n]}W_{kl} &\lesssim\frac{\xi}{n}\left(\frac{1}{q}+\sqrt{n}\right)\lesssim\frac{\xi}{\sqrt{n}},\\
\sum_{i,j\in[n]}u(i)v(i)\Upsilon_i^2\frac{2\al}{\Lambda_i} W_{ij} 
    &\lesssim   \frac{\xi}{|z|^2} \bigg[\frac{\|\bu\|_\infty\|\bv\|_\infty}{q}+ \bigg(\frac{1}{n}\sum_{i,j\in[n]}|u(i)|^2|v(i)|^2\bigg)^{1/2}\bigg] 
    \\
    &\lesssim \frac{\xi}{|z|^2}\|\bu\|_\infty\wedge \|\bv\|_\infty.
\end{align*}
Plugging these two estimates into \eqref{eq:uG-Upsilonv 2}, we can deduce that w.h.p., 
\begin{align}
\left|z\bu^\top\bUpsilon\bcE'\bUpsilon\bv\right|\lesssim \frac{\xi}{q|z|}\|\bu\|_\infty\wedge \|\bv\|_\infty +\frac{\xi}{q\sqrt{n}|z|} + \frac{\xi^2}{q^2|z|\sdeg^{2}},
\end{align}
where $\wedge$ represents the minimum of two given numbers. Together with \eqref{eq:uG-Upsilonv 1}, this  yields (\ref{eq:uG-Upsilonv anisotropic}) for each fixed $z\in S(\fC)$. Finally, an application of the $\epsilon$-net argument results in the desired conclusion in (\ref{eq:uG-Upsilonv anisotropic}). This completes the proof of \Cref{prop:uG-Upsilonv anisotropic}.

\subsection{Proof of \Cref{prop:eiLWL(G-Upsilon)v}} \label{Sec.proof.thm9}

From \eqref{eq:uG-G[i]v}, we can deduce that w.h.p.,
\begin{align}\label{eq:Lambda W Lambda deco0}
\begin{split}
     \be^\top_i  \bcW (\bG-\bUpsilon)\bv &=   \be^\top_i  \bcW (\bG_{[i]}-\bUpsilon)\bv+\OO\left[\frac{1}{q\sdeg}\left({\frac{\xi}{\sqrt n|z|^2}}+\frac{\|\bv\|_\infty}{|z|}\right)\right],
\end{split}
\end{align}
where we have used \eqref{eq:W} to bound $\|\bcW\be_i\|$ and \eqref{eq:eiGv} to bound $\|\bG\bv\|_{\max}$. In view of \eqref{eq:resolvent identity 3 of G}, it holds that 
\begin{align}
     \be^\top_i  \bcW (\bG_{[i]}-\bUpsilon)\bv  = &  \sum_{j\in[n]\setminus\{i\}} \ocW_{ij}\left(\bG_{[i]}^{(i)}-\bUpsilon^{(i)}\right)_{j\bv}+ \sum_{j\in[n]\setminus\{i\}} \ocW_{ij}\frac{(G_{[i]})_{ji}(G_{[i]})_{i\bv}}{(G_{[i]})_{ii}} \nonumber\\
     &+\ocW_{ii} \left((G_{[i]})_{i\bv}-\Upsilon_{i\bv}\right).\label{eq:Lambda W Lambda deco}
\end{align}
With condition \eqref{eq:moment conditions}, a simple application of the Markov inequality gives that 
\begin{equation}
\max_{i,j\in [n]} |\ocW_{ij}|\lesssim q^{-1}  
\end{equation}
with high probability. This along with \eqref{eq:eiGj-UpsilonvLinfty} yields that w.h.p.,
\begin{equation}\label{eq:localthird}
    \ocW_{ii} \left((G_{[i]})_{i\bv}-\Upsilon_{i\bv}\right)\lesssim \frac{\xi}{q|z|}\left(\left(1+\frac{1}{q\sdeg}\right)\frac{1}{\sqrt n|z|}+\frac{\|\bv\|_{\infty}}{q\sdeg}\right).
\end{equation}

With the aid of \eqref{eq:large_dev1}, we can bound the first term on the RHS of \eqref{eq:Lambda W Lambda deco} above as
\begin{align}\label{eq:Lambda W Lambda 1}
\begin{split}
    \sum_{j \in[n]\setminus\{i\}} \ocW_{ij}\left(\bG_{[i]}^{(i)}-\bUpsilon^{(i)}\right)_{j\bv} \lesssim&  \frac{\xi^2}{q|z|}\left(\left(1+\frac{1}{q\sdeg}\right)\frac{1}{\sqrt n|z|}+\frac{\|\bv\|_{\infty}}{q\sdeg}\right) \\
    &+  \frac{\xi}{\sqrt{n}}\left(\frac{1}{|z|^2}+\frac{\xi}{q|z|\sdeg}\right) ,
\end{split}
\end{align}
where we have used (\ref{eq:eiG(i)-UpsilonvLinfty}) to bound $\|(\bG_{[i]}^{(i)}-\bUpsilon^{(i)})\bv\|_{\infty}$ and used \eqref{eq:G[i]-R_add} to bound $\|(\bG_{[i]}^{(i)}-\bUpsilon^{(i)})\bv\|_2$.  
In view of (\ref{eq:resolvent identity 2 of G}), we can write the second term on the RHS of \eqref{eq:Lambda W Lambda deco} above as
\begin{equation}\label{eq:Lambda W Lambda 1.5}
    \sum_{j \in [n]\setminus\{i\}} \ocW_{ij}\frac{(G_{[i]})_{ji}(G_{[i]})_{i\bv}}{(G_{[i]})_{ii}}=- (G_{[i]})_{i\bv}\sum_{j,k \in[n]\setminus\{i\}} \ocW_{ij}\ocW_{ki}(G_{[i]}^{(i)})_{jk}.
\end{equation}
Then applying \eqref{eq:large_dev2} and \eqref{eq:large_dev3}, we can deduce that w.h.p., 
\begin{align*}
    &\sum_{j,k\in[n]\setminus\{i\}} \ocW_{ij}\ocW_{ki}(G_{[i]}^{(i)})_{jk} - \sum_{j\in[n]\setminus\{i\}}\Lambda_i^{-4\alpha}\Lambda_j^{-4\alpha}s_{ij} (G_{[i]}^{(i)})_{jj} \\
    &\lesssim   \xi\frac{\max_{j\in[n]}|(G_{[i]}^{(i)})_{jj}|}{q}+{\xi^2}\frac{\max_{1 \leq j\ne k\leq n}|(G_{[i]}^{(i)})_{jk}|}{q} + \frac{{\xi^2}}{n}\bigg( \sum_{j,k\in[n]\setminus\{i\}} |(G_{[i]}^{(i)})_{jk}|^2\bigg)^{1/2} \\
    &\lesssim \frac{\xi}{q|z|} +  \frac{\xi^3}{q|z|}\left(\frac{1}{\sqrt n|z|}+\frac{1}{q\sdeg}\right)  + \frac{\xi^2}{\sqrt{n}|z|} \\
    &\lesssim \frac{\xi}{q|z|} + \frac{\xi^3}{q^2|z|\sdeg}   + \frac{\xi^2}{\sqrt{n}|z|}, 
\end{align*}
where in the second step above, we have used (\ref{eq:G[i]-R2}) to bound $\max_{j\in[n]}|(G_{[i]}^{(i)})_{jj}|$, used (\ref{eq:eiG(i)-UpsilonvLinfty}) with $\bv=\be_k$  to bound $\max_{1\leq j\ne k\leq n}|(G_{[i]}^{(i)})_{jk}|$, and used \eqref{eq:G[i]-R2} to bound $\sum_{j,k\in[n]\setminus \{i\}} |(G_{[i]}^{(i)})_{jk}|^2=\tr [\bG_{[i]}^{(i)}(\bG_{[i]}^{(i)} )^*]$. 

The above estimate along with (\ref{eq:G[i]-R2}) 
shows that w.h.p., 
$$\sum_{j,k\in[n]\setminus\{i\}} \ocW_{ij}\ocW_{ki}(G_{[i]}^{(i)})_{jk} \lesssim \left(1+\frac{\xi^3}{q^2\sdeg}\right)\frac{1}{|z|}. $$
Plugging it into \eqref{eq:Lambda W Lambda 1.5} and using \eqref{eq:eiG[i]v}, it holds that  
\begin{align}\label{eq:Lambda W Lambda 2}
\begin{split}
    \sum_{j\in[n]\setminus\{i\}} \ocW_{ij}\frac{(G_{[i]})_{ji}(G_{[i]})_{i\bv}}{(G_{[i]})_{ii}}&\lesssim \left(1+\frac{1}{q\sdeg}\right)\left(1+\frac{\xi^3}{q^2\sdeg}\right)\left({\frac{\xi}{\sqrt n|z|^3}}+\frac{\|\bv\|_\infty}{|z|^2}\right).
\end{split}
\end{align}
Then combining \eqref{eq:localthird}, \eqref{eq:Lambda W Lambda 1}, and \eqref{eq:Lambda W Lambda 2}, we can obtain that w.h.p.,
$$\be^\top_i  \bcW (\bG_{[i]}-\bUpsilon)\bv\lesssim    \left(\frac{1}{|z|}+\frac{\xi}{q\sdeg}\right)  \frac{\xi}{\sqrt{n}|z|}+ \left( \frac{\xi^2}{q^2\sdeg}+\frac{1}{|z|}\right)\frac{\|\bv\|_\infty}{|z|}. $$
Therefore, plugging this into \eqref{eq:Lambda W Lambda deco0} yields estimate \eqref{eq:eiLWL(G-Upsilon)v} for each fixed $z\in S(\fC)$. This concludes the proof of \Cref{prop:eiLWL(G-Upsilon)v}.


\section{Proofs of propositions and key lemmas} \label{proof.keylems}

In this section, we still investigate the rescaled setting of our model as initially introduced in Section \ref{sec.mainresu.rescaledmodel}. This rescaled setting involves the rescalings given by \eqref{eq:rescaling tX,H,W}, \eqref{eq:rescaling L,Lambda}, \eqref{eq:rescaling X delta}, and \eqref{eq:rescaling t_k}.
\subsection{Proof of Proposition \ref{prop:W,R,G}} \label{Sec.proof.prop1}

The estimate \eqref{eq:W} can be shown using the same arguments as in the proof of Lemma 4.3 in \cite{erdHos2013spectral}. Note that \eqref{eq:R} is a simple consequence of \eqref{eq:W} by definition. Then in light of \eqref{eq:LasympLambda} and \eqref{eq:L-Lambda}, we have that for $\mathbb T=\emptyset,\ \{i\}$, or $\{i,j\}$,
$$
    \|(\bL^{(\mathbb T)}/\bLambda^{(\mathbb T)})^{2\alpha}\|\sim 1
$$
with $(c,\xi)$-high probability for some constant $c>0$. This together with \eqref{eq:W} leads to \eqref{eq:G}, which completes the proof of Proposition \ref{prop:W,R,G}.


\subsection{Proof of Proposition \ref{prop:G-R}} \label{Sec.proof.prop2}

Recall the simple matrix identity $$\bA^{-1}=\bB^{-1}+\bA^{-1}(\bB-\bA)\bB^{-1}$$ for any nonsingular matrices $\bA$ and $\bB$. An application of (\ref{eq:L-Lambda}) and Proposition \ref{prop:W,R,G} yields that with $(c_5,\xi)$-high probability, 
	\begin{eqnarray*}
		\|\bG(z)-\bR(z)\|=\|z\bG((\bL/\bLambda)^{2\alpha}-\bI)\bR\|
		\lesssim\frac{\xi}{q|z|\sdeg}.
	\end{eqnarray*}
Furthermore, using Proposition \ref{prop:W,R,G} and \eqref{eq:Upsilon}, we can deduce that 
 \begin{align*}
    \|\bG(z)-\bUpsilon(z)\|&\leq\left\|z^{-1}\bG(z)\left[ \bcW +(\bI-(\bL/\bLambda)^{2\alpha})z\right] \right\|+\left\|\bUpsilon(z)+z^{-1}\bI\right\|\\
    &\lesssim \frac{1}{|z|^2}+\frac{\xi}{q|z|\sdeg}
\end{align*}
with $(c_5,\xi)$-high probability. This concludes the proof of Proposition \ref{prop:G-R}.

\subsection{Proof of Lemma \ref{prop:Lambda and E}} \label{Sec.proof.lem1}

The inequality \eqref{eq:Lambda} follows directly from the definition. From \eqref{eq:large_dev1}, we see that with $(c_1,\xi)$-high probability,
	\begin{align*}
		|{\cE}_i|&\lesssim \frac{|d_i-\mathbb Ed_i|+|\bar d-\mathbb E{\bar d}|}{q\sdeg} \\
  &= \frac{1}{q\sdeg} \bigg|\sum_{j\in[n]} W_{ij}\bigg|+\frac{1}{nq\sdeg}\bigg|\sum_{i,j\in[n]} W_{ij}\bigg| \\
  &\lesssim\frac {\xi} {q\sdeg} \left(\frac{1}{q}+1\right)+\frac {\xi} {nq\sdeg}  \left(\frac{1}{q}+\sqrt{n}\right) \\
  &\lesssim \frac \xi {q\sdeg},
	\end{align*}
which leads to (\ref{eq:E}). An application of the Taylor expansion shows that \eqref{eq:LasympLambda} is a simple consequence of \eqref{eq:condK} and   \eqref{eq:Lambda}--\eqref{eq:E}, which completes the proof of Lemma \ref{prop:Lambda and E}.

\subsection{Proof of Lemma \ref{prop:L-Lambda}} \label{Sec.proof.lem2}

With the Taylor expansion, we have that  
\begin{equation*}
    \left|\frac{L_{i}^{\alpha}-\Lambda_{i}^{\alpha}}{\Lambda_i^\alpha}\right|=\alpha t^{\alpha-1}\frac{\cE_{i}}{\Lambda_i},
\end{equation*}
where $t$ is some number between $L_{j}/\Lambda_{j}$ and $1$. 
Then we see that (\ref{eq:L-Lambda}) follows immediately from \Cref{prop:Lambda and E}. This concludes the proof of Lemma \ref{prop:L-Lambda}.

\subsection{Proof of Lemma \ref{lemma:Upsilon}} \label{Sec.proof.lem3}
Lemma \ref{lemma:Upsilon} is directly from the Laurent series of $\btUpsilon(z)$ in \eqref{eq:Upsilon Laurent}.

\subsection{Proof of Lemma \ref{lemma:L[i]}} \label{Sec.proof.lem4}

The third estimate in \eqref{eq:L-L[i]} can be proved in the same way as for \Cref{prop:L-Lambda}, which also entails that $$1\le\|\bL_{[i]}\|\le \sdeg^{-1}$$ with $(c,\xi)$-high probability for some constant $c>0$. Together with \eqref{eq:LasympLambda} and a simple application of the mean-value theorem, it gives that  
$$\|(\bL^{\alpha}-\bL_{[i]}^{\alpha})\bL^{-\alpha}\|_F \lesssim \|\bL-\bL_{[i]}\|_F$$
with $(c,\xi)$-high probability. Hence, the second estimate in \eqref{eq:L-L[i]} follows directly from the first estimate in \eqref{eq:L-L[i]}, which we aim to establish next.

By definition, it holds that for $j\ne i,$
$$L_j-(L_{[i]})_j=\frac{1}{q\sdeg}W_{ij} + \frac{\tau_j}{nq\sdeg} W_{ii} + \frac{2\tau_j}{nq\sdeg}\sum_{l\in[n]\setminus\{i\}}W_{il}.$$ 
With the aid of \eqref{eq:large_dev1} and \eqref{eq:large_dev2}, we can deduce that with $(c,\xi)$-high probability,
$$W_{ii} + 2\sum_{l\in[n]\setminus\{i\}}W_{il} \lesssim {\xi} \ \text{ and } \ \sum_{j\in[n]\setminus\{i\}}|W_{ij}|^2 - \sum_{j\in[n]\setminus\{i\}}s_{ij} \lesssim \frac{\xi}{q}.$$
This gives that with $(c,\xi)$-high probability,
\[
    \left\|\bL-\bL_{[i]}\right\|_F^2\lesssim \frac{1}{q^2\sdeg^2}\sum_{j\in[n]\setminus\{i\}}\left(|W_{ij}|^2 + \frac{\xi^2}{n^2}\right)  \lesssim \frac{1}{q^2\sdeg^2},
 \] 
which yields the first estimate in (\ref{eq:L-L[i]}). 
Further, combining (\ref{eq:L-L[i]}) with Proposition \ref{prop:W,R,G}, we can easily derive (\ref{eq:G[i]-R}) and (\ref{eq:G[i]-R2}).
Finally, estimate \eqref{eq:G[i]-R_add} can be established in the same way as for \eqref{eq:G-Y0}, which completes the proof of Lemma \ref{lemma:L[i]}.


\subsection{Proof of Lemma \ref{lemma:G-G[i]}} \label{Sec.proof.lem5}

It follows from the definition that 
\begin{align*}
	\bu^\top \left(\bG(z)-\bG_{[i]}(z)\right)\bv&=z\bu^\top\bG(z)\bLambda^{-2\alpha}\left(\bL^{2\alpha}-\bL^{2\alpha}_{[i]}\right)\bG_{[i]}(z)\bv\\
	&\lesssim |\bu|\left|\bLambda^{-2\alpha}\left(\bL^{2\alpha}-\bL^{2\alpha}_{[i]}\right)\bG_{[i]}(z)\bv\right|\\
 &  \lesssim |\bu|\left\|\bLambda^{-2\alpha}(\bL^{2\alpha}-\bL_{[i]}^{2\alpha})\right\|_F\left\|\bG_{[i]}(z)\bv\right\|_{\infty}\\
 &\lesssim\frac{1}{q\sdeg}|\bu|\left\|\bG_{[i]}(z)\bv\right\|_{\infty},
\end{align*}
where we have used \eqref{eq:G} in the second step above and (\ref{eq:L-L[i]}) in the last step. 
The term $\|\bG_{[i]}(z)\bv\|_{\infty}$ above can also be replaced with $\|\bG(z)\bv\|_{\infty}$ by writing the first step as $$z\bu^\top \bG_{[i]}(z)\bLambda^{-2\alpha}(\bL^{2\alpha}-\bL^{2\alpha}_{[i]})\bG(z)\bv$$ and using \eqref{eq:G[i]-R}. Thus, we obtain \eqref{eq:uG-G[i]v}. The estimate in \eqref{eq:uG-G[i]v2} can be proved in a similar fashion, which concludes the proof of Lemma \ref{lemma:G-G[i]}.


\subsection{Proof of Lemma \ref{lemma:resolvent identities}} \label{Sec.proof.lem6}

We will focus on proving the conclusion for $\bG$,  since the proof for $\bG_{[i]}$ follows a similar approach. Additionally, the proof for $\bR$ can be derived directly from Lemma 3.4 in \cite{erdHos2013spectral}. Denote by 
 \[
 \bQ(z):=(\bL^{-\alpha}\bW\bL^{-\alpha}-z)^{-1}=(\bL/\bLambda)^{\alpha}\bG(z)(\bL/\bLambda)^{\alpha},
 \]
 \[
    \bQ^{(\mathbb T)}(z):=((\bL^{(\mathbb T)})^{-\alpha}\bW^{(\mathbb T)}(\bL^{(\mathbb T)})^{-\alpha}-z)^{-1}=(\bL^{(\mathbb T)}/\bLambda^{(\mathbb T)})^{\alpha}\bG^{(\mathbb T)}(z)(\bL^{(\mathbb T)}/\bLambda^{(\mathbb T)})^{\alpha},
 \]
and
\[
    \bG_{[i]}^{(\mathbb T)}(z):=((\bLambda^{(\mathbb T)})^{-\alpha}\bW^{(\mathbb T)}(\bLambda^{(\mathbb T)})^{-\alpha}-z(\bL_{[i]}^{(\mathbb T)}/\bLambda^{(\mathbb T)})^{2\alpha})^{-1}
\]
for any $\mathbb T\subset [n]$. An application of Lemma 3.4 in \cite{erdHos2013spectral} gives the following resolvent identities for $\bQ(z)$ (which can also be shown using the Schur complement formula): 
\begin{itemize}
\item[(i)] For each $i\in[n]$, we have that
\begin{equation}
    \frac{1}{Q_{ii}}=-z-\Lambda_{i}^{-2\alpha}W_{ii}-\sum_{k,l\in[n]\setminus\{i\}}(\Lambda_i^2\Lambda_k\Lambda_l)^{-\alpha}Q_{kl}^{(i)}.
\end{equation}

\item[(ii)] For each $i\neq j\in[n]$, we have that 
\begin{equation}
\begin{split}
    Q_{ij}&=-Q_{ii}\Lambda_i^{-\alpha}\sum_{k\in[n]\setminus\{i\}}W_{ik}\Lambda_k^{-\alpha}Q_{kj}^{(i)} \\
    &=Q_{ii}Q_{jj}^{(ij)}(\Lambda_i\Lambda_j)^{-\alpha}\bigg(-W_{ij}+\sum_{k,l\in[n]\setminus\{i,j\}}(\Lambda_k\Lambda_l)^{-\alpha}W_{ik}W_{jl}Q_{kl}^{(ij)}\bigg).
\end{split} 
\end{equation}

\item[(iii)] For each $k\in[n]\setminus\{i,j\}$, we have that 
\begin{equation}
    Q_{ij}^{(k)}=Q_{ij}-\frac{Q_{ik}Q_{kj}}{Q_{kk}}.
\end{equation}
\end{itemize}

Using the resolvent identities above, we can derive some further resolvent identities for $\bG(z)$ and $\bG_{[i]}$. Specifically, it holds that for all $l\neq s$, 
\begin{align}
\begin{split}
    G_{ls}(z)&=(L_l/\Lambda_l)^{-\alpha}Q_{ls}(z)(L_s/\Lambda_s)^{-\alpha}\\
    &=-(L_l/\Lambda_l)^{-\alpha}(Q_{ll}(z)\sum_{1\leq t\neq l\leq n}(\bL^{-\alpha}\bW\bL^{-\alpha})_{lt}Q_{ts}^{(l)}(z))(L_s/\Lambda_s)^{-\alpha}\\
    &=-G_{ll}(z)\Lambda_l^{-\alpha}\sum_{t\in [n]\setminus\{l\}}\Lambda_t^{-\alpha} W_{lt}G_{ts}^{(l)}(z)
\end{split}
\end{align}
and similarly,  
\begin{equation}
    (G_{[i]})_{ls}(z)=-(G_{[i]})_{ll}(z)\Lambda_l^{-\alpha}\sum_{t\in[n]\setminus\{l\}}\Lambda_t^{-\alpha} W_{lt}(G_{[i]}^{(l)})_{ts}(z).
\end{equation}
Using similar arguments, we can deduce that for each $k\in[n]\setminus\{i,j\}$,
\begin{align}
\begin{split}
    G_{ij}^{(k)}&=(L_i/\Lambda_i)^{-\alpha}Q_{ij}^{(k)}(L_j/\Lambda_j)^{-\alpha}\\
    &=(L_i/\Lambda_i)^{-\alpha}\bigg(Q_{ij}-\frac{Q_{ik}Q_{kj}}{Q_{kk}}\bigg)(L_j/\Lambda_j)^{-\alpha}\\
    &=G_{ij}-\frac{G_{ik}G_{kj}}{G_{kk}}
\end{split}
\end{align}
and
\begin{equation}
    (G_{[k]}^{(k)})_{ij}=(G_{[k]})_{ij}-\frac{(G_{[k]})_{ik}(G_{[k]})_{kj}}{(G_{[k]})_{kk}}.
\end{equation}
This completes the proof of Lemma \ref{lemma:resolvent identities}.

\subsection{Proof of Lemma \ref{lemma:Linfty norms}} \label{Sec.proof.lem7}


Denote by $\bv^{(i)}$ the vector with components $\bv^{(i)}(j)=\mathbf 1_{j\ne i}v(j)$, i.e., $\bv^{(i)}$ is obtained by setting the $i$th component of $\bv$ as zero. 
Using (\ref{eq:resolvent identity 2 of G}) for $\bG_{[i]}$ and recalling the notation in \eqref{eq:gen_entries}, it holds that w.h.p.,
\begin{equation}\label{eq:eiGv-exp}
    \be_i^\top\bG_{[i]}\bv=-(G_{[i]})_{ii} \sum_{l\in[n]\setminus\{i\}}\ocW_{il} (G_{[i]}^{(i)})_{l\bv^{(i)}}+(G_{[i]})_{ii}v(i).
\end{equation}
Note that by the definition in \eqref{eq:G[i] definition unrescaled}, $(G_{[i]}^{(i)})$ is independent of the entries $\ocW_{il}$. 
Hence, we can apply \eqref{eq:large_dev1} to $\sum_{l\in[n]\setminus\{i\}}\ocW_{il} (G_{[i]}^{(i)})_{l\bv^{(i)}}$ and obtain that w.h.p.,
\begin{align}
\left|\frac{\be_i^\top\bG_{[i]}(z)\bv}{(G_{[i]})_{ii}}\right|&\lesssim {\xi}  \frac{\max_{1\leq l\neq i\leq n}|(G_{[i]}^{(i)})_{l\bv^{(i)}}|}{q}+\xi\bigg(\frac{1}{n}\sum_{l\in[n]\setminus\{i\}} |(G_{[i]}^{(i)})_{l\bv^{(i)}}|^2\bigg)^{1/2} + {\|\bv\|_\infty}  \nonumber\\
	&\lesssim \frac{\xi}{q}\max_{1\leq l\neq i\leq n}|({G}_{[i]}^{(i)})_{l\bv^{(i)}}|+{\frac{\xi}{\sqrt n |z|}}+ {\|\bv\|_\infty} ,\label{eq:boundG[i]00}
\end{align}
where we have used \eqref{eq:G[i]-R2} to bound $\left(\sum_{l}^{(i)} |(G_{[i]}^{(i)})_{l\bv^{(i)}}|^2\right)^{1/2}$ by $\OO(|z|^{-1})$.  

Plugging \eqref{eq:boundG[i]00} into \eqref{eq:eiGv-exp} and using again \eqref{eq:G[i]-R2} to bound $(G_{[i]})_{ii}$, we can deduce that 
\begin{align}
\left|\be_i^\top\bG_{[i]}\bv\right|&\lesssim\frac{\xi}{|z|}\left(\frac{\max_{1\leq l\neq i\leq n}|(G_{[i]}^{(i)})_{l\bv^{(i)}}|}{q}+\bigg(\frac{1}{n}\sum_{l\in[n]\setminus\{i\}} |(G_{[i]}^{(i)})_{l\bv^{(i)}}|^2\bigg)^{1/2}\right)+\frac{\|\bv\|_\infty}{|z|} \nonumber\\
	&\lesssim \frac{\xi}{q|z|}\max_{1\leq l\neq i\leq n}|({G}_{[i]}^{(i)})_{l\bv^{(i)}}|+{\frac{\xi}{\sqrt n |z|^2}}+\frac{\|\bv\|_\infty}{|z|},\label{eq:boundG[i]0}
\end{align}
where we have used \eqref{eq:G[i]-R2} to bound $(G_{[i]})_{ii}$ and $\left(\sum_{l}^{(i)} |(G_{[i]}^{(i)})_{l\bv^{(i)}}|^2\right)^{1/2}$ by $\OO(|z|^{-1})$. Moreover, applying (\ref{eq:resolvent identity 3 of G}) and (\ref{eq:resolvent identity 2 of G}) to $\bG_{[i]}$, it follows that w.h.p.,
\begin{align}
    ({G}_{[i]}^{(i)})_{l\bv^{(i)}}&=({G}_{[i]})_{l\bv^{(i)}}-\frac{({G}_{[i]})_{li}({G}_{[i]})_{i\bv^{(i)}}}{({G}_{[i]})_{ii}} \nonumber\\
    &=({G}_{[i]})_{l\bv^{(i)}}+({G}_{[i]})_{i\bv^{(i)}} \cdot \sum_{k\in [n]\setminus\{i\}}(G_{[i]}^{(i)})_{lk}\ocW_{ki} \nonumber\\
    &\lesssim \max_{l\in[n]}|(G_{[i]})_{l\bv^{(i)}}| \lesssim \max_{l\in[n]}|(G_{[i]})_{l\bv}| + \frac{|v(i)|}{|z|},\label{eq:boundG[i]}
\end{align}
where in the third step above, we have again applied \eqref{eq:large_dev1} and \eqref{eq:G[i]-R2} to get that 
$$\sum_{k\in[n]\setminus\{i\}}(G_{[i]}^{(i)})_{lk}\ocW_{ki} \lesssim   {\xi}\bigg(\frac{1}{q|z|}+\bigg(\frac{1}{n}\sum_{k\in[n]\setminus\{i\}} |(G_{[i]}^{(i)})_{lk}|^2\bigg)^{1/2}\bigg)\lesssim \frac{\xi}{q|z|}. $$ 
Then combining \eqref{eq:boundG[i]0} and \eqref{eq:boundG[i]} yields that w.h.p.,
\begin{align}\label{eq:boundG[i]1}
	\left|\be_i^\top\bG_{[i]}\bv\right|&\lesssim \frac{\xi}{q|z|} \max_{l \in [n]}|(G_{[i]})_{l\bv}| +{\frac{\xi}{\sqrt n |z|^2}}+\frac{\|\mathbf{v}\|_\infty}{|z|}.
\end{align} 

On the other hand, an application of \Cref{lemma:G-G[i]} shows that w.h.p.,
$$|(G_{[i]})_{l\bv}-G_{l\bv}| \lesssim  \frac{1}{q\sdeg}\max_{l \in [n]}|G_{l\bv}|.$$
Plugging it into \eqref{eq:boundG[i]1}, we have that  w.h.p.,
\[
	|\be_i^\top\bG\bv|\lesssim \frac{\xi}{q|z|}\max_{l\in[n]}|G_{l\bv}|+\frac{\xi}{q^2|z|\sdeg}\max_{l\in[n]}|G_{l\bv}|+{\frac{\xi}{\sqrt n |z|^2}}+\frac{\|\mathbf{v}\|_\infty}{|z|}.
\]
Notice that with a simple union bound argument, we see that such an estimate holds uniformly in $i\in [n]$ w.h.p. Then taking the maximum of the left-hand side above over $i \in [n]$ gives that w.h.p.,
\[
	\max_{i\in[n]} |\be_i^\top\bG\bv|\lesssim \left(\frac{\xi}{q|z|} +\frac{\xi}{q^2|z|\sdeg}\right)\max_{i\in[n]}|G_{i\bv}|+{\frac{\xi}{\sqrt n |z|^2}}+\frac{\|\mathbf{v}\|_\infty}{|z|},
\]
which together with the assumption of $\xi\ll q^2|z|\sdeg$ yields (\ref{eq:eiGv}). Thus, an application of \Cref{lemma:G-G[i]} and (\ref{eq:eiGv}) leads to (\ref{eq:eiG[i]v}). 
Finally, applying \eqref{eq:boundG[i]} and (\ref{eq:eiG[i]v}) to \eqref{eq:boundG[i]00}, we can derive \eqref{eq:eiG[i]v_div}, which concludes the proof of Lemma \ref{lemma:Linfty norms}.

\section{Additional technical details and additional simulation results} \label{addtech.details}

\subsection{Refined results under the network setting} \label{app:network}

Throughout this subsection, we consider the rescaled setting of our model as introduced in Section \ref{sec.mainresu.rescaledmodel}, which involves the rescalings specified in \eqref{eq:rescaling tX,H,W}, \eqref{eq:rescaling L,Lambda}, \eqref{eq:rescaling X delta}, and \eqref{eq:rescaling t_k}. 
Specifically, we aim to present some more refined RMT results for the generalized Laplacian matrices under the network setting. Specifically, we will state the main results and some key steps of the technical analyses when $\btX$ represents the adjacency matrix of an undirected random graph. For such a case, the entries of $\btX$ have Bernoulli distributions before rescaling. For the rescaled $\bW$, it holds that 
\begin{equation}\label{eq:network case}
		\max_{i,j\in[n]}|{W}_{ij}|\leq \frac{1}{q} \ \text{ and } \ \max_{i,j\in[n]}s_{ij}\lesssim\frac{1}{n}
\end{equation}
almost surely (instead of  with $(c_0,\xi)$-high probability). With these properties, we can replace Lemma \ref{lemma:estimation} used in the proofs of our main results earlier with the lemma below.

\begin{lemma}[Bernstein's inequality \citep{vershynin2018high}]\label{lemma:Bernstein's inequality}
    Let $(x_i)_{i\in[n]}$ be a family of centered independent random variables satisfying that $\max_{i\in[n]}|x_i|\leq \phi_n$ for some ($n$-dependent) parameter $\phi_n>0$. Then it holds that for each $t>0$, 
\[
    \mathbb P\bigg(\sum_{i\in[n]}x_i>t\bigg)\leq2\exp\bigg(-\frac{ct^2}{\sum_{i\in[n]}\mathbb E x_i^2+\phi_n t}\bigg)
\]
with $c>0$ some absolute constant.
\end{lemma}

With Bernstein's inequality in Lemma \ref{lemma:Bernstein's inequality} above, we have that for some absolute constant $a>0$,
\begin{equation}
    \bigg|\sum_{i\in[n]}x_i\bigg|\leq\bigg(\sum_{i\in[n]}\mathbb E x_i^2\bigg)^{1/2}\xi^{1/2}+\phi_n\xi
\end{equation}
with $(a,\xi)$-high probability. Then for each constant $D>0$, there exists some constant $C>0$ such that
\begin{equation}
\begin{split}
    & \mathbb P\bigg\{\bigg|\sum_{i\in[n]}x_i\bigg|\leq C\bigg[\bigg(\sum_{i\in[n]}\mathbb E x_i^2\bigg)^{1/2}(\log n)^{1/2}+\phi_n\log n\bigg]\bigg\}\\
    &\geq 1-n^{-D}.
    \end{split}
\end{equation}
Hence, it follows that 
\begin{equation}
    \bigg|\sum_{i\in[n]}x_i\bigg|\lesssim \bigg(\sum_{i\in[n]}\mathbb E x_i^2\bigg)^{1/2}(\log n)^{1/2}+\phi_n\log n
\end{equation}
holds w.h.p.

For the bilinear forms of centered independent random variables with $O(n^{-1})$ variances, we have the lemma below.

\begin{lemma}[Lemma 3.8 of \cite{erdHos2013spectral}]\label{lemma:Bernstein}
    Let $(x_i)_{i\in[n]}$ and $(y_i)_{i\in[n]}$ be independent families of centered independent complex-valued random variables, and $(B_{ij})_{i,j\in[n]}$ a family of deterministic complex numbers. Assume that all components $x_i$ and $y_i$ have variances at most $n^{-1}$ and satisfying that $\max_{i\in[n]}|x_i|\leq\phi_n$ and $\max_{i\in[n]}|y_i|\leq\phi_n$ for some ($n$-dependent) parameter $\phi_n\geq n^{1/2}$. Then it holds with $(a,\xi)$-high probability that
\begin{align}
    &\bigg|\sum_{i,j\in[n]}x_i B_{ij}y_j\bigg|\leq\xi^2\bigg[\phi_n^2B_d+\phi_nB_o+\frac{1}{n}\bigg(\sum_{i\neq j\in[n]}|B_{ij}|^2\bigg)^{1/2}\bigg],\\
    &\bigg|\sum_{i\in[n]}\bar{x}_iB_{ii}x_i-\sum_{i\in[n]}(\mathbb E|x_i|^2)B_{ii}\bigg|\leq\bigg(\xi^{1/2}\phi_n+\xi\phi_n^2\bigg)B_d,\\
    &\bigg|\sum_{i\neq j\in[n]}\bar{x}_iB_{ij}x_j\bigg|\leq\xi^2\bigg[\phi_nB_o+\frac{1}{n}\bigg(\sum_{i\neq j\in[n]}|B_{ij}|^2\bigg)^{1/2}\bigg],
\end{align}
where $a>0$ is an absolute constant.
\end{lemma}

We emphasize that the bounds given in Lemmas \ref{lemma:Bernstein's inequality} and \ref{lemma:Bernstein} above are \textit{not} obtained by simply replacing $\xi$ with $\log n$ in Lemma \ref{lemma:estimation}. The parameter $\xi$ is replaced with $(\log n)^{1/2}$ at some places, which would yield sharper results. We now state the corresponding \textit{local laws} under the properties in (\ref{eq:network case}) for the network setting.

\begin{proposition}[Corresponding to \Cref{prop:ekG-UpsilonvLinfty}] \label{new.eq.supp.prop4}
     Under the conditions of \Cref{thm:local laws of R} and \eqref{eq:network case}, for each constant $D>0$, there exists some constant $C_7>0$ such that for any deterministic unit vector $\bv$, all the events
     \begin{equation}
		\bigcap_{z\in S(C_0)}\biggl\{\max_{i\in[n]}|\be^\top_i(\bG_{[i]}(z)-\bUpsilon(z))\bv|\leq C_7\frac{(\log n)^{1/2}}{|z|}\bigg(\frac{1}{\sqrt n|z|}+\frac{1}{q\sdeg}\|\bv\|_\infty\bigg)\biggr\},
	\end{equation}
    \begin{equation}
		\bigcap_{z\in S(C_0)}\biggl\{\max_{i\in[n]}|\be_i^\top(\bG(z)-\bUpsilon(z))\bv|\leq C_7\frac{(\log n)^{1/2}}{|z|}\bigg(\frac{1}{\sqrt n|z|}+\frac{1}{q\sdeg}\|\bv\|_\infty\bigg)\biggr\},
    \end{equation}
    \begin{equation}
		\bigcap_{z\in S(C_0)}\biggl\{\max_{i,j\in[n]}|\be_i^\top(\bG_{[j]}(z)-\bUpsilon(z))\bv|\leq C_7\frac{(\log n)^{1/2}}{|z|}\bigg(\frac{1}{\sqrt n|z|}+\frac{1}{q\sdeg}\|\bv\|_\infty\bigg)\biggr\}
    \end{equation}
    hold with probability at least $1-n^{-D}$.
\end{proposition}

\begin{proposition}[Corresponding to \Cref{prop:uG-Upsilonv anisotropic}] \label{new.eq.supp.prop5}
 Under the conditions of \Cref{thm:local laws of R} and \eqref{eq:network case}, for each constant $D>0$, there exists some constant $C_8>0$ such that for any deterministic unit vectors $\bu$ and $\bv$, the event
\begin{equation}
\begin{split}
    \bigcap_{z\in S({C_0})} & \left\{|\bu^\top(\bG(z)-\bUpsilon(z))\bv|\lesssim C_8\frac{{(\log n)^{1/2}}}{q|z|}\right. \\
    &\left.\times \left(\frac{1}{|z|\sdeg}+\frac{{(\log n)^{1/2}}}{q\sdeg^{2}}+\|\bu\|_\infty\wedge\|\bv\|_\infty\right)\right\}
\end{split}
\end{equation}
 holds with probability at least $1-n^{-D}$.
 \end{proposition}
 
\begin{proposition}[Corresponding to \Cref{prop:eiLWL(G-Upsilon)v}] \label{new.eq.supp.prop6}
    Under the conditions of \Cref{thm:local laws of R} and \eqref{eq:network case}, for each constant $D>0$, there exists some constant $C_9>0$ such that for any deterministic unit vector $\bv$, the event
\begin{equation}
\begin{split}
    \bigcap_{z\in S({C}_0)}&\biggl\{\max_{i\in[n]}|\be^\top_i\bLambda^{-\alpha}\bW\bLambda^{-\alpha}(\bG-\bUpsilon)\bv|\leq\frac{C_9}{|z|}\\
    &\times \bigg(\bigg(\frac{1}{|z|}+\frac{(\log n)^{1/2}}{q\sdeg}\bigg)\sqrt{\frac{\log n}{n}}+\bigg(\frac{1}{|z|}+\frac{1}{q\sdeg}\bigg)\|\bv\|_\infty\bigg)\biggr\}
    \end{split}
\end{equation}
holds with probability at least $1-n^{-D}$.
\end{proposition}

With the \textit{sharper} local laws given in Propositions \ref{new.eq.supp.prop4}--\ref{new.eq.supp.prop6} above, we can improve our main results in Theorems \ref{thm:eigenvalue rescaled}--\ref{thm:projection2 rescaled} 
to the improved ones below. 

\begin{theorem}[Corresponding to \Cref{thm:eigenvalue rescaled}]
Assume that Condition \ref{cond:rescaled model}, Assumption \ref{main_assm rescaled}, and \eqref{eq:network case} are satisfied. Then it holds for each $1\leq k\leq K_0$ that w.h.p.,
\begin{equation}
    |\hdelta_k-t_k|=O\biggl\{|\delta_k|\frac{\sqrt{\log n}}{q}\omega_n(\delta_k)\left(1+\frac{K}{|\delta_k|^4}\right)\biggr\},
\end{equation}
where for simplicity we have introduced the notation
\begin{equation}\label{eq:Apsink network}
\omega_n(\delta_k):=\frac{1}{|\delta_k|\sdeg}+\frac{\sqrt{\log n}}{q\sdeg^{2}}+\|\bV\|_{\max}.
 \end{equation}
\end{theorem}

\begin{theorem}[Corresponding to \Cref{thm:projection rescaled}]
Assume that Condition \ref{cond:rescaled model}, Assumption \ref{main_assm rescaled}, and \eqref{eq:network case} are satisfied. Then it holds for each $1\leq k\leq K_0$ that w.h.p., 
\begin{equation}
\begin{split}
   \left|\bv_k^\top(\bL/\bLambda)^{-\alpha}\bhv_k- \frac{1}{\sqrt{\delta_k^2\bv_k^\top\bUpsilon_k'(t_k)\bv_k}}\right| \lesssim \frac{\sqrt{\log n}}{q}\omega_n(\delta_k)\left(1+\frac{K }{ |\delta_k|^4}\right) ,    
\end{split}
\end{equation}
where we choose the direction of $\bhv_k$ such that $\bhv_k^\top\bv_k>0$.
Moreover, for any deterministic unit vector $\bu$, it holds that w.h.p., 
\begin{equation}
\begin{split}
    &\left|\bu^\top(\bL/\bLambda)^{-\alpha}\bhv_k+\frac{\delta_k\bu^\top\bUpsilon_k(t_k)\bv_k}{\sqrt{\delta_k^2\bv_k^\top\bUpsilon_k'(t_k)\bv_k}}\right| \\
    &\lesssim \frac{\sqrt{\log n}}{q}\omega_n(\delta_k)\left[1+\frac{K }{ |\delta_k|^4} + {\|\bu^\top \bV_{-k}\|} \left(\sqrt K+\frac{K}{|\delta_k|^2}\right)\right].
\end{split}    
\end{equation} 
\end{theorem}

\begin{theorem}[Corresponding to \Cref{thm:main thm rescaled}]
Assume that Condition \ref{cond:rescaled model}, Assumption \ref{main_assm rescaled}, and \eqref{eq:network case} are satisfied, and 
\begin{equation}
K\omega_n(\delta_k)\sdeg\ll 1,\quad \|\bV\|_{\max}\ll \frac{1}{|\delta_k|\sdeg}+\frac{(\log n)^{1/2}}{q\sdeg^{2}}
\end{equation}
for each $1\leq K_0\leq K$. Then for each $1\leq K_0\leq K$ and $i\in [n]$, it holds w.h.p.~that
\begin{align}
	\begin{split}
	\hv_k(i)&=(\Lambda_i/L_i)^\alpha v_k(i)+\frac{1}{t_kL_i^{\alpha}}\sum_{j\in[n]}{W}_{ij}{\Lambda}_j^{-\alpha}v_k(j)\\
	&+O\left(\|\bV\|_{\max}\left(\frac{\sqrt K}{|\delta_k|}+\frac{K(\log n)^{1/2}}{q}\right)\left(\frac{1}{|\delta_k|\sdeg}+\frac{(\log n)^{1/2}}{q\sdeg^{2}}\right)\right)\\
 &+O\left(\sqrt{\frac{\log n}{n}}\frac{1}{|\delta_k|}\left(\frac{1}{|\delta_k|}+\frac{(\log n)^{1/2}}{q\sdeg}\right)\right),
	\end{split}
\end{align}
where we choose the direction of $\bhv_k$ such that $\bhv_k^\top\bv_k>0$. Consequently, we have that w.h.p.,
\begin{align}
\begin{split}
    \hv_k(i)&= v_k(i)- \frac{\al}{\Lambda_i\sdeg}\bigg( \frac{1}{q} \sum_{j\in[n]} W_{ij} +\frac{\tau_i}{nq} \sum_{j,l\in[n]} W_{jl}\bigg)v_k(i) +\frac{1}{t_k}\sum_{j\in[n]}\Lambda_i^{-\al}{W}_{ij}{\Lambda}_j^{-\alpha}v_k(j)\\
    &+O\left(\|\bV\|_{\max}\left(\frac{\sqrt K}{|\delta_k|}+\frac{K(\log n)^{1/2}}{q}\right)\left(\frac{1}{|\delta_k|\sdeg}+\frac{(\log n)^{1/2}}{q\sdeg^{2}}\right)\right)\\
 &+O\left(\sqrt{\frac{\log n}{n}}\frac{1}{|\delta_k|}\left(\frac{1}{|\delta_k|}+\frac{(\log n)^{1/2}}{q\sdeg}\right)+\frac{(\log n)^{3/2}}{q^2|\delta_k|\sdeg}\|\bv_k\|_\infty\right)
\end{split}
\end{align}
for each $1\leq K_0\leq K$ and $i\in [n]$.
\end{theorem}

\begin{proposition}[Corresponding to \Cref{prop:main_strong}]
Assume that Condition \ref{cond:rescaled model}, Assumption \ref{main_assm rescaled}, and \eqref{eq:network case} are satisfied. Then for each $1\le k\le K_0$ and $i\in [n]$, we have that w.h.p.,
    \begin{align}
    \begin{split}
        \hv_k(i) =&~(\Lambda_i/L_i)^{\alpha}v_k(i)+\frac{1}{t_kL_i^\alpha}\sum_{j\in[n]}{W}_{ij}{\Lambda}_j^{-\alpha}v_k(j) \\
        &+O\left(\|\bv_k\|_\infty\left(\frac{1}{|\delta_k|^2}+\frac{(\log n)^{1/2}\omega_n(\delta_k)}{q}\left(1+\frac{K}{|\delta_k|^4}\right)\right)\right)\\
        &+O\left(\|\bV_{-k}\|_{\max}\left(\frac{\sqrt K}{|\delta_k|^2}+\frac{K(\log n)^{1/2}\omega_n(\delta_k)}{q}\right)\right)\\
        &+O\left(\sqrt{\frac{\log n}{n}}\frac{1}{|\delta_k|}\left(\frac{\sqrt K}{|\delta_k|^2}+\frac{1}{|\delta_k|}+\frac{K(\log n)^{1/2}\omega_n(\delta_k)}{q}+\frac{(\log n)^{1/2}}{q\sdeg}\right)\right).
    \end{split}
    \end{align}
\end{proposition}

\begin{theorem}[Corresponding to \Cref{thm:eigenvalue2 rescaled}]
Assume that Condition \ref{cond:rescaled model}, Assumption \ref{main_assm rescaled}, and \eqref{eq:network case} are satisfied. Then it holds w.h.p. that
\begin{align}
\begin{split}
    &\hdelta_k-t_k - A_k= -2\alpha t_k\bv_k^\top\frac{\bL-\bLambda}{\bLambda}\bv_k+\bv_k^\top\bcW\bv_k+B_k\\
    &\quad+O\left(\frac{1}{|\delta_k|^2}+\frac{(\log n)^{3/2}|\delta_k|}{q^3\sdeg^3}+\frac{\sqrt K(\log n)^{1/2}|\delta_k|\omega_n(\delta_k)}{q} \right. \\
    & \quad\left. \times\left(\frac{1}{|\delta_k|^2}+\frac{\sqrt K(\log n)^{1/2}\omega_n(\delta_k)}{q}\right)\right),
\end{split}
\end{align}
where $A_k$ is a deterministic term given by 
\begin{align*}
A_k & =\alpha(2\alpha+1)t_k\E\bv_k^\top\frac{(\bL-\bLambda)^2}{\bLambda^2}\bv_k-2\alpha\E\bv_k^\top\frac{\bL-\bLambda}{\bLambda}\bcW\bv_k,
\end{align*}
and $B_k$ is a centered random error satisfying
\[
    \var(B_k)\lesssim \frac{|\delta_k|^2\|\bv_k\|_\infty^2}{q^4\sdeg^4}+\frac{|\delta_k|^2}{q^4n^2\sdeg^4}+\frac{\|\bv_k\|_\infty^2}{q^2\sdeg^2}+\frac{1}{q^2n\sdeg^2}+\frac{1}{\sqrt n q|\delta_k|^2}.
\]
\end{theorem}

\begin{theorem}[Corresponding to \Cref{thm:projection2 rescaled}]
Assume that Condition \ref{cond:rescaled model}, Assumption \ref{main_assm rescaled}, and \eqref{eq:network case} are satisfied. Then we have that 

    1) For each $1\leq k\leq K_0$ and any deterministic unit vector $\bu$ such that $\bu^\top\bv_k=0$, it holds w.h.p. that
\begin{align}
\begin{split}
    &\bu^\top(\bL/\bLambda)^{-\alpha}\bhv_k -\cA_k= \sft_k\bu^\top\bV_{-k}\frac{1}{\sft_k-\bDelta_{-k}}\bV_{-k}^\top\left(-2\alpha\frac{\bL-\bLambda}{\bLambda}+\sft_k^{-1}\bcW\right)\bv_k\\
    &+\bw^\top\left(-2\alpha\frac{\bL-\bLambda}{\bLambda}+t_k^{-1}\bcW\right)\bv_k+\sum_{l\in[K]\setminus\{k\}}\frac{\sft_k\bu^\top\bv_l}{\sft_k-\delta_l}\cB_{k,l}+\cB_k^{\bw}\\
    &+O\left(K\left(\frac{1}{|\delta_k|^2}+\frac{(\log n)^{1/2}}{q}\omega_n(\delta_k)\right)\left(\frac{1}{|\delta_k|}+\frac{(\log n)^{1/2}}{q\beta_n}\right)\right)\\
    &+O\left(\frac{K^{3/2}(\log n)^{1/2}}{q}\omega_n(\delta_k)\left(\frac{(\log n)^{1/2}}{q}\omega_n(\delta_k)+\frac{1}{|\delta_k|^2}\right)\right),
\end{split}
\end{align}
where we choose the sign of $\bhv_k$ such that $\bhv_k^\top\bv_k>0$, $\bw=(\bI-\bV\bV^\top)\bu$, $\cA_k$ is a deterministic term given by 
\begin{align*}
\cA_k & =\bw^\top\left(\alpha(2\alpha+1)\frac{(\bL-\bLambda)^2}{\bLambda^2}- \frac{2\alpha}{ t_k}\left(\frac{\bL-\bLambda}{\bLambda}\bcW+\bcW\frac{\bL-\bLambda}{\bLambda}\right)+\frac{\bcW^2}{t_k^{2}}\right)\bv_k\\
&\quad+t_k\bu^\top\bV_{-k}\frac{1}{t_k-\bDelta_{-k}}\E \bV_{-k}^\top\left(\alpha(2\alpha+1)\frac{(\bL-\bLambda)^2}{\bLambda^2} \right.\\
& \quad \left.- \frac{2\alpha}{ t_k}\left(\frac{\bL-\bLambda}{\bLambda}\bcW+\bcW\frac{\bL-\bLambda}{\bLambda}\right)+\frac{\bcW^2}{t_k^{2}}\right)\bv_k,
\end{align*}
$\cB_{k}^{\bw}$ is a centered random variable satisfying
\[
\var(\cB_{k}^{\bw}) \lesssim\frac{\|\bv_k\|_\infty\|\bw\|_\infty}{q^4\sdeg^4}+\frac{|\bw|}{q^4n^2\sdeg^4}+\frac{1}{|\delta_k|^2\sdeg^2}\left(\frac{\|\bv_k\|_\infty\|\bw\|_\infty}{q^2}+\frac{|\bw|}{q^2n}\right)+\frac{|\bw|}{q\sqrt n|\delta_k|^4},
\]
and for each $l\in[K]\setminus\{k\}$, $\cB_{k,l}$ is a centered random variable satisfying
\[
\var\,  \cB_{k,l}\lesssim\frac{\|\bv_k\|_\infty\|\bv_l\|_\infty}{q^4\sdeg^4}+\frac{1}{q^4n^2\sdeg^4}+\frac{1}{|\delta_k|^2\sdeg^2}\left(\frac{\|\bv_k\|_\infty\|\bv_l\|_\infty}{q^2}+\frac{1}{q^2n}\right)+\frac{1}{q\sqrt n|\delta_k|^4}.
\]

2) For the case of $\bu=\bv_k$ and each $1 \leq k \leq K_0$, it holds w.h.p. that
\begin{align}
\begin{split}
     & \bv_k^\top(\bL/\bLambda)^{-\alpha}\bhv_k-\bv_k^\top(\bL/\bLambda)^{-\alpha}\bv_k -\fA_k\\
     &=\frac{\alpha^2}{2}\bv_k^\top\left(\frac{\bL-\bLambda}{\bLambda}\right)^2\bv_k-\frac{1}{2t_k^{2}}\bv_k^\top\bcW^2\bv_k\\
    &\quad+\fB_k+O\left(\frac{K}{|\delta_k|^4}+\frac{K\log n}{q^2}\omega_n(\delta_k)\right),
\end{split}
\end{align}
where $\fA_k$ is a deterministic term given by 
\[
\fA_k:=(\delta_k^2\bv_k^\top\bUpsilon'_k(t_k)\bv_k)^{-1/2}-1+\frac{1}{2}\bv_k^\top(t_k^2\bUpsilon'(t_k)+2t_k\bUpsilon(t_k)+\bI)\bv_k\]
and $\fB_k$ is a random variable satisfying
\[
\E\fB_k^2\lesssim\frac{n^2\|\bv_k\|_\infty^4}{q^8\sdeg^6}+\frac{n^2\|\bv_k\|_\infty^4}{q^2|\delta_k|^6}.
\]
\end{theorem}

\subsection{Additional technical lemma} \label{add.tech.lem}

\begin{lemma}[Lemma 3.8 of \cite{erdHos2013spectral}]\label{lemma:estimation}
Let $a_1,\ldots,a_n,b_1,\ldots, b_n$ be centered and independent (complex-valued) random variables satisfying that
\begin{equation}
    \mathbb E\mathbb|a_i|^p\leq\frac{C^p}{nq^{p-2}}, \ \ \mathbb E\mathbb|b_i|^p\leq\frac{C^p}{nq^{p-2}}
\end{equation}
with $i\in [n]$ for some $2\leq p\leq(\log n)^{A_0\log\log n}$. Then there exists some constant $\upsilon=\upsilon(C)>0$ such that for all $\xi$ satisfying (\ref{eq:xi condition}) and any deterministic values $A_i,B_{ij}\in\mathbb C$, we have that with $(\upsilon,\xi)$-high probability,
\begin{align}
    & \bigg|\sum_{i\in[n]}A_ia_i\bigg| \leq \xi\bigg[\frac{\max_{i\in[n]}|A_i|}{q}+\bigg(\frac{1}{n}\sum_{i\in[n]}|A_i|^2\bigg)^{1/2}\bigg],\label{eq:large_dev1}\\
    & \bigg|\sum_{i\in[n]}\bar a_iB_{ii}a_i-\sum_{i\in[n]}\sigma_i^2B_{ii}\bigg| \leq \xi\frac{B_d}{q},\label{eq:large_dev2}\\
    & \bigg|\sum_{i\neq j\in[n]}\bar a_iB_{ij}a_j\bigg| \leq\xi^{2}\bigg[\frac{B_o}{q}+\bigg(\frac{1}{n^2}\sum_{i\neq j\in[n]}|B_{ij}|^2\bigg)^{1/2}\bigg],\label{eq:large_dev3}\\
&\left|\sum_{i,j\in[n]}a_iB_{ij}b_j\right| \leq\xi^{2}\bigg[\frac{B_d}{q^2}+\frac{B_o}{q}+\bigg(\frac{1}{n^2}\sum_{i\neq j}|B_{ij}|^2\bigg)^{1/2}\bigg],\label{eq:large_dev4}
\end{align}
where $\sigma_i^2$ denotes the variance of $a_i$, and
\[
    B_d:=\max_{i\in[n]} |B_{ii}|, \ B_o:=\max_{i\neq j\in[n]}|B_{ij}|.
\]
\end{lemma}

\subsection{Additional simulation results} \label{add.simu.results}

\begin{figure}[tp]
\centering
\includegraphics[width=0.90\linewidth]{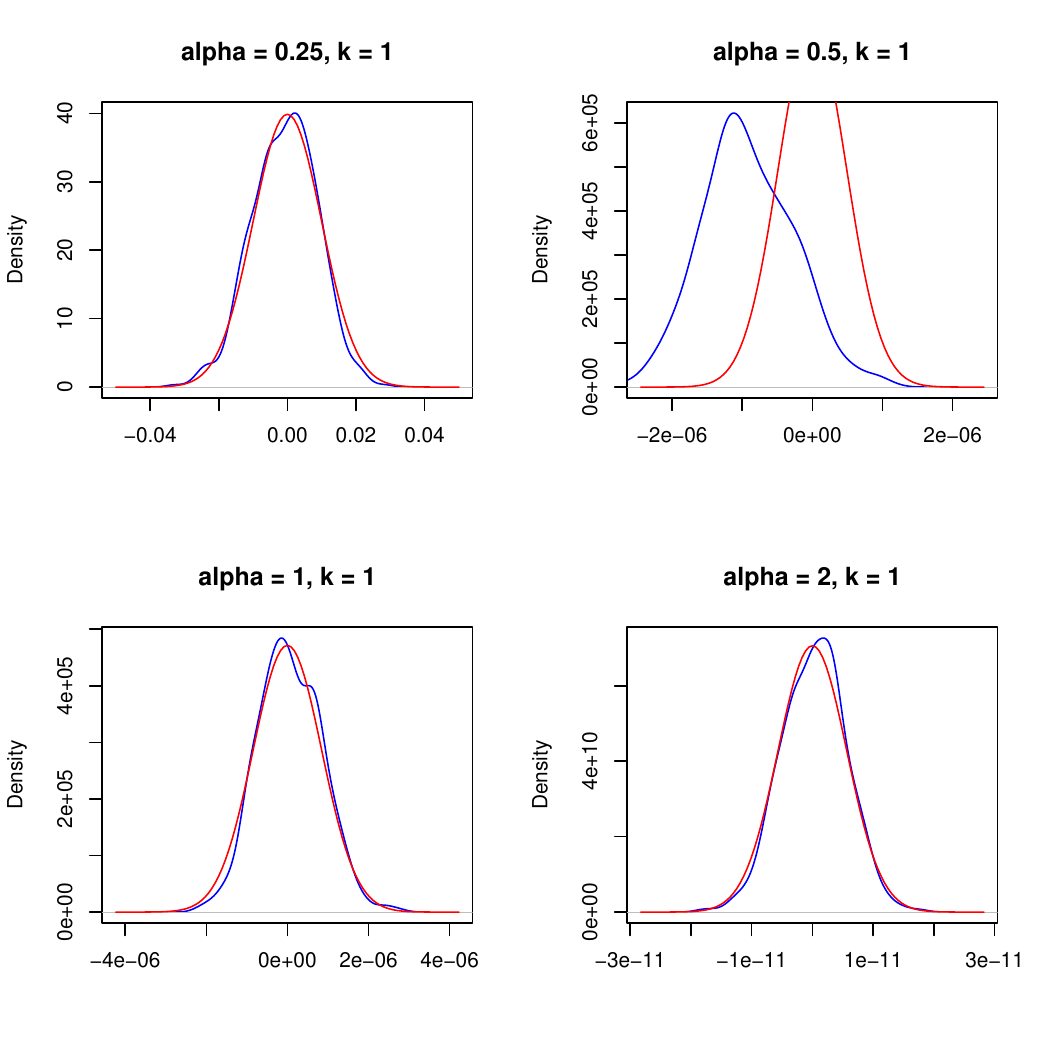}
\caption{The kernel density estimate (KDE) for the distribution of the empirical spiked eigenvalue $\hdelta_k$ corrected by estimate $\widehat{A}_k$ for the generalized Laplacian matrix $\bX$  with $k = 1$ across different values of $\alpha$ based on $500$ replications for simulation example in Section \ref{new.Sec.simu} with $\theta = 0.9$. The  generalized (regularized) Laplacian matrix $\bX$ is as given in (\ref{new.eq.FL.gLap}) with $\bL=\bL_{\tau,\lambda} := 
\diag\left(d_i+\tau\bar d+\lambda: i \in [n]\right)$ without the rescaling population parameters $q$ and $\sdeg$. The blue curves represent the KDEs for the empirical spiked eigenvalue corrected by estimate $\widehat{A}_k$, whereas the red curves stand for the target normal density. Both curves are centered with the asymptotic limit $t_k$. The top right plot is due to relatively small empirical standard deviations. This is associated with the fact that the normalized Laplacian matrix has a trivial largest eigenvalue at 1.}
\label{fig1_Ahat}
\end{figure}

\begin{figure}[h]
\centering
\includegraphics[width=0.90\linewidth]{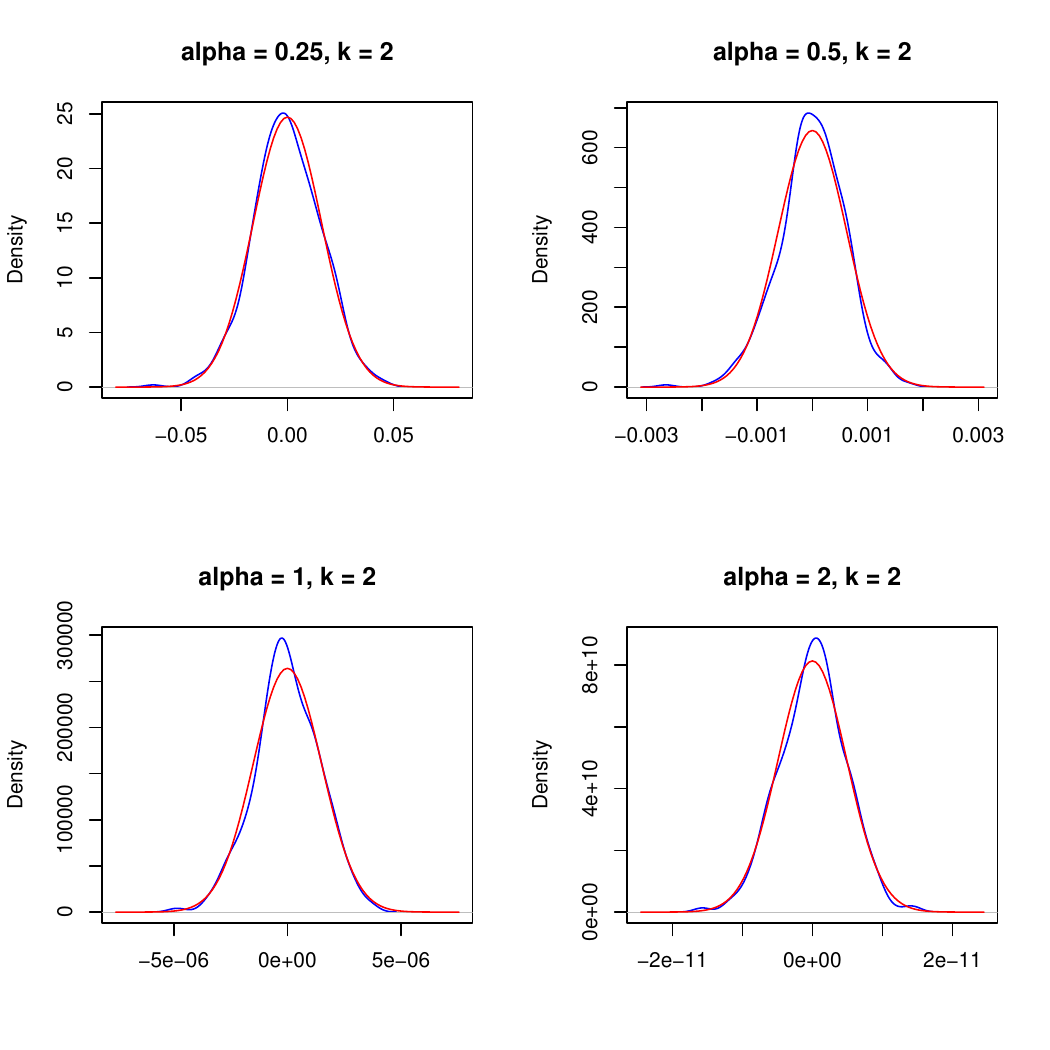}
\caption{The kernel density estimate (KDE) for the distribution of the empirical spiked eigenvalue $\hdelta_k$ corrected by estimate $\widehat{A}_k$ for the generalized Laplacian matrix $\bX$  with $k = 2$ across different values of $\alpha$ based on $500$ replications for simulation example in Section \ref{new.Sec.simu} with $\theta = 0.9$. The  generalized (regularized) Laplacian matrix $\bX$ is as given in (\ref{new.eq.FL.gLap}) with $\bL=\bL_{\tau,\lambda} := 
\diag\left(d_i+\tau\bar d+\lambda: i \in [n]\right)$ without the rescaling population parameters $q$ and $\sdeg$. The blue curves represent the KDEs for the empirical spiked eigenvalue corrected by estimate $\widehat{A}_k$, whereas the red curves stand for the target normal density. Both curves are centered with the asymptotic limit $t_k$.}
\label{fig2_Ahat}
\end{figure}

\begin{figure}[h]
\centering
\includegraphics[width=0.90\linewidth]{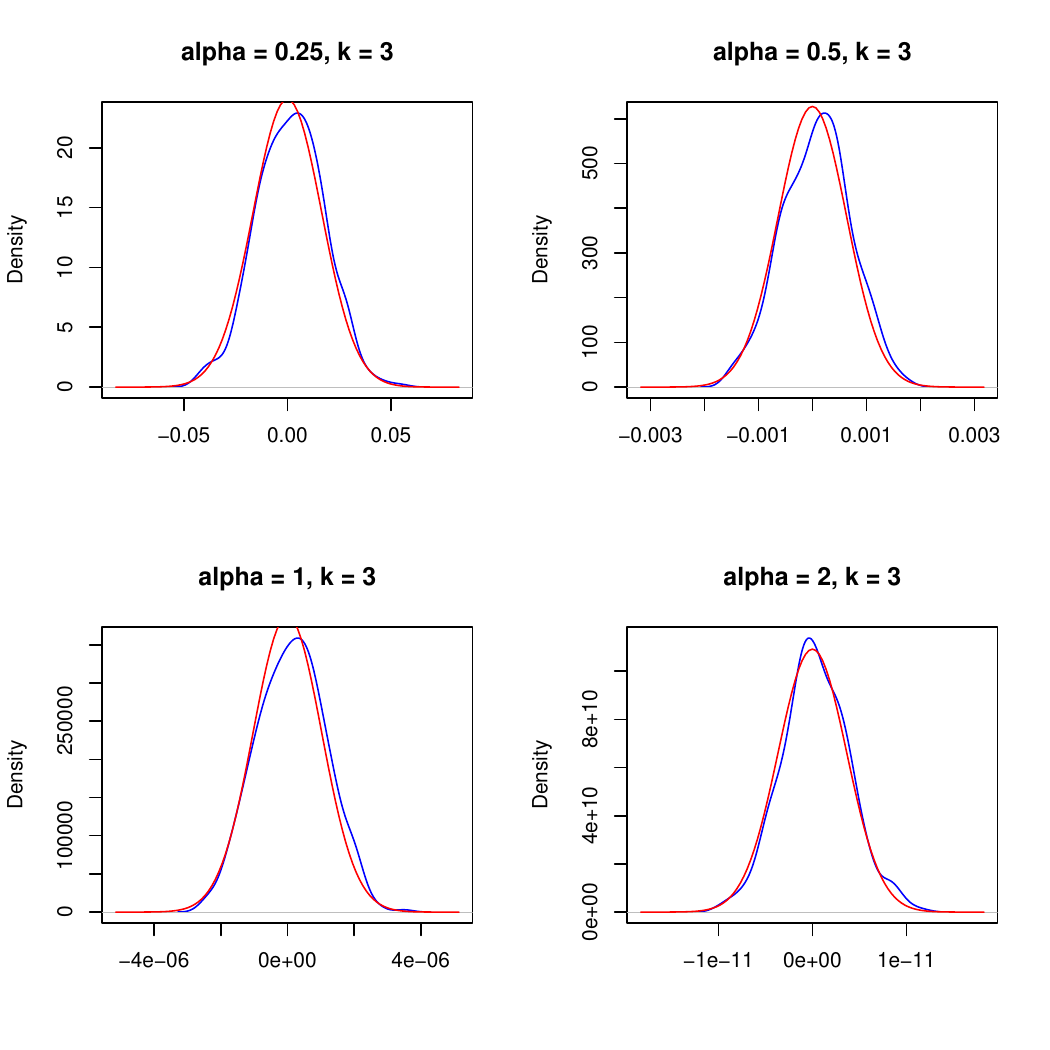}
\caption{The kernel density estimate (KDE) for the distribution of the empirical spiked eigenvalue $\hdelta_k$ corrected by estimate $\widehat{A}_k$ for the generalized Laplacian matrix $\bX$  with $k = 3$ across different values of $\alpha$ based on $500$ replications for simulation example in Section \ref{new.Sec.simu} with $\theta = 0.9$. The  generalized (regularized) Laplacian matrix $\bX$ is as given in (\ref{new.eq.FL.gLap}) with $\bL=\bL_{\tau,\lambda} := 
\diag\left(d_i+\tau\bar d+\lambda: i \in [n]\right)$ without the rescaling population parameters $q$ and $\sdeg$. The blue curves represent the KDEs for the empirical spiked eigenvalue corrected by estimate $\widehat{A}_k$, whereas the red curves stand for the target normal density. Both curves are centered with the asymptotic limit $t_k$.}
\label{fig3_Ahat}
\end{figure}

\begin{figure}[tp]
\centering
\includegraphics[width=0.90\linewidth]{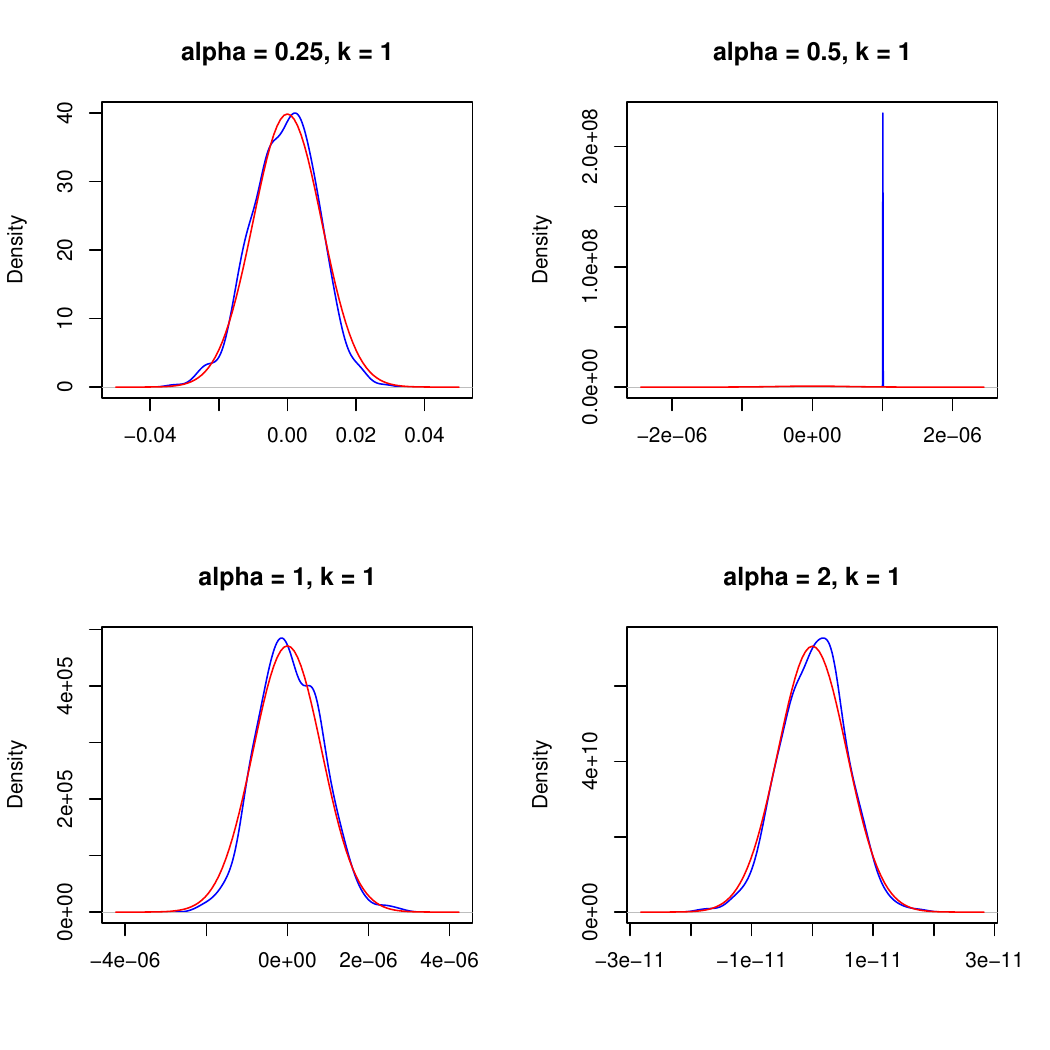}
\caption{The kernel density estimate (KDE) for the distribution of the empirical spiked eigenvalue $\hdelta_k$ corrected by estimate $\widehat{A}_k$ coupled with the empirical bias correction in Section \ref{new.sec.mainresu} for the generalized Laplacian matrix $\bX$  with $k = 1$ across different values of $\alpha$ based on $500$ replications for simulation example in Section \ref{new.Sec.simu} with $\theta = 0.9$. The  generalized (regularized) Laplacian matrix $\bX$ is as given in (\ref{new.eq.FL.gLap}) with $\bL=\bL_{\tau,\lambda} := 
\diag\left(d_i+\tau\bar d+\lambda: i \in [n]\right)$ without the rescaling population parameters $q$ and $\sdeg$. The blue curves represent the KDEs for the empirical spiked eigenvalue corrected by estimate $\widehat{A}_k$ coupled with the empirical bias correction in Section \ref{new.sec.mainresu}, whereas the red curves stand for the target normal density. Both curves are centered with the asymptotic limit $\delta_k$. The top right plot is due to extremely small empirical standard deviations (similar to Figure \ref{fig1}). This is associated with the fact that the normalized Laplacian matrix has a trivial largest eigenvalue at 1.}

\label{fig1_empbc_delta}
\end{figure}

\begin{figure}[h]
\centering
\includegraphics[width=0.90\linewidth]{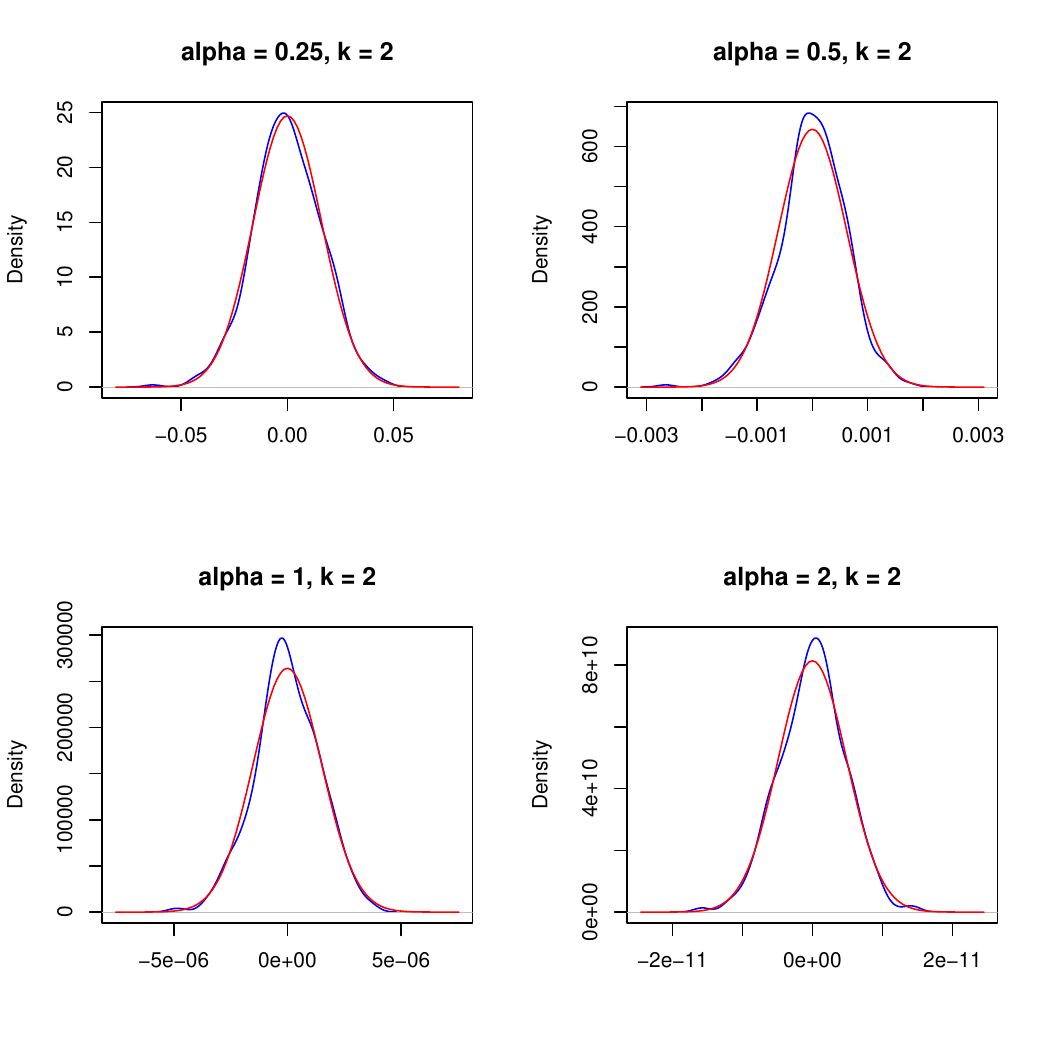}
\caption{The kernel density estimate (KDE) for the distribution of the empirical spiked eigenvalue $\hdelta_k$ corrected by estimate $\widehat{A}_k$ coupled with the empirical bias correction in Section \ref{new.sec.mainresu} for the generalized Laplacian matrix $\bX$  with $k = 2$ across different values of $\alpha$ based on $500$ replications for simulation example in Section \ref{new.Sec.simu} with $\theta = 0.9$. The  generalized (regularized) Laplacian matrix $\bX$ is as given in (\ref{new.eq.FL.gLap}) with $\bL=\bL_{\tau,\lambda} := 
\diag\left(d_i+\tau\bar d+\lambda: i \in [n]\right)$ without the rescaling population parameters $q$ and $\sdeg$. The blue curves represent the KDEs for the empirical spiked eigenvalue corrected by estimate $\widehat{A}_k$ coupled with the empirical bias correction in Section \ref{new.sec.mainresu}, whereas the red curves stand for the target normal density. Both curves are centered with the asymptotic limit $\delta_k$.}
\label{fig2_empbc_delta}
\end{figure}

\begin{figure}[h]
\centering
\includegraphics[width=0.90\linewidth]{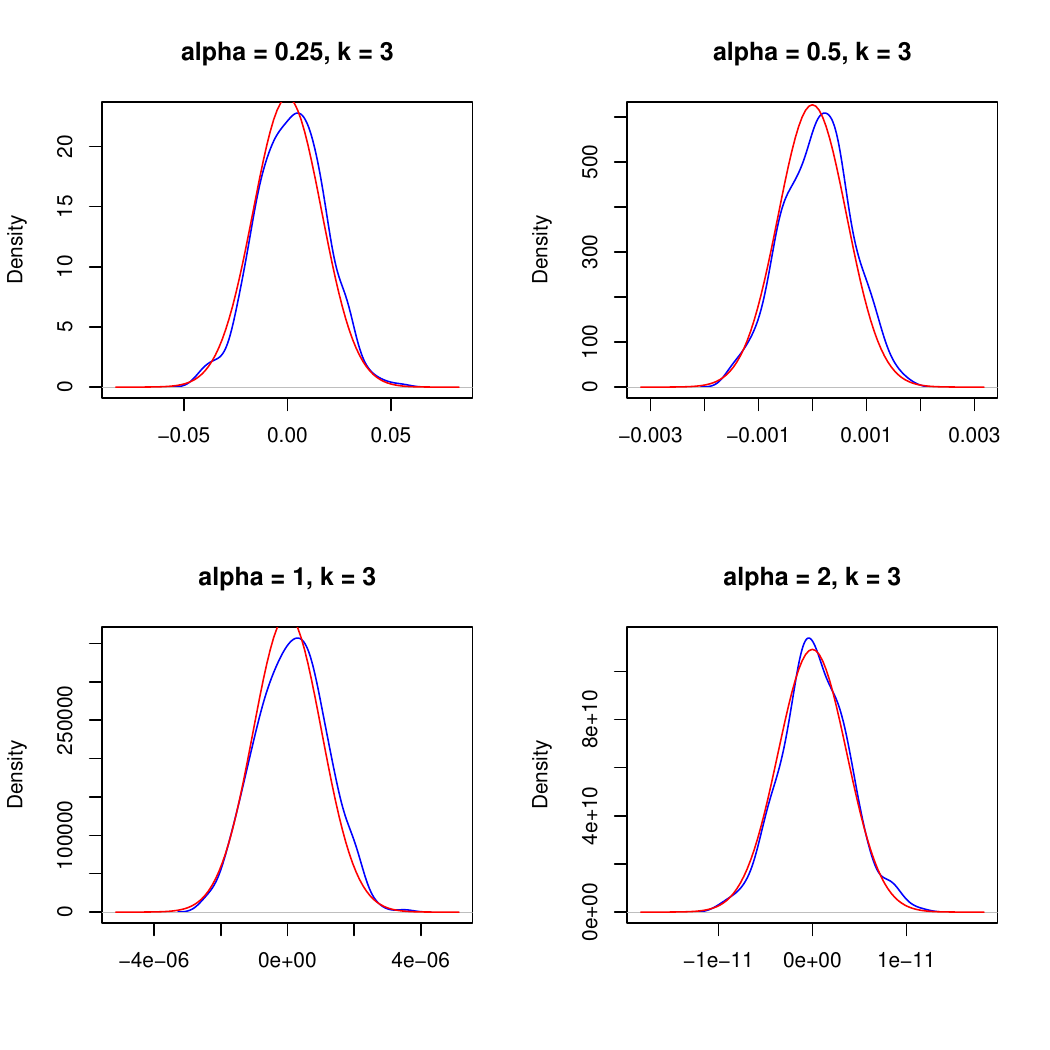}
\caption{The kernel density estimate (KDE) for the distribution of the empirical spiked eigenvalue $\hdelta_k$ corrected by estimate $\widehat{A}_k$ coupled with the empirical bias correction in Section \ref{new.sec.mainresu} for the generalized Laplacian matrix $\bX$  with $k = 3$ across different values of $\alpha$ based on $500$ replications for simulation example in Section \ref{new.Sec.simu} with $\theta = 0.9$. The  generalized (regularized) Laplacian matrix $\bX$ is as given in (\ref{new.eq.FL.gLap}) with $\bL=\bL_{\tau,\lambda} := 
\diag\left(d_i+\tau\bar d+\lambda: i \in [n]\right)$ without the rescaling population parameters $q$ and $\sdeg$. The blue curves represent the KDEs for the empirical spiked eigenvalue corrected by estimate $\widehat{A}_k$ coupled with the empirical bias correction in Section \ref{new.sec.mainresu}, whereas the red curves stand for the target normal density. Both curves are centered with the asymptotic limit $\delta_k$.}
\label{fig3_empbc_delta}
\end{figure}

In this section, we will present some additional simulation results. In particular, Figures \ref{fig1_Ahat}--\ref{fig3_Ahat} are the counterparts of Figures \ref{fig1}--\ref{fig3}, respectively, where the empirical spiked eigenvalue $\hdelta_k$ is now corrected by estimate $\widehat{A}_k$ instead of being corrected by the theoretical value $A_k$ with the asymptotic limit $t_k$. Similarly, Figures \ref{fig1_empbc_delta}--\ref{fig3_empbc_delta} correspond to Figures \ref{fig1}--\ref{fig3}, respectively, where the empirical spiked eigenvalue $\hdelta_k$ is now corrected by estimate $\widehat{A}_k$ coupled with the empirical bias correction in Section \ref{new.sec.mainresu} with the asymptotic limit $\delta_k$. Indeed, from Figures \ref{fig1_Ahat}--\ref{fig3_empbc_delta} we can see that \textit{both} ideas of bias correction using estimate $\widehat{A}_k$ toward the population quantity $t_k$, and correction by estimate $\widehat{A}_k$ coupled with the empirical bias correction in Section \ref{new.sec.mainresu} toward the population quantity $\delta_k$ instead work well for the empirical spiked eigenvalue $\hdelta_k$ of the generalized Laplacian matrix $\bX$ across different settings.

\end{document}